\documentclass{article} 

\usepackage[margin=2.5cm]{geometry}

\usepackage{graphicx}%
\usepackage{multirow}%
\usepackage{amsmath,amssymb,amsfonts}%
\usepackage{amsthm}%
\usepackage{mathrsfs}%
\usepackage[title]{appendix}%
\usepackage{xcolor}%
\usepackage{textcomp}%
\usepackage{manyfoot}%
\usepackage{booktabs}%
\usepackage{algorithm}%
\usepackage{algorithmicx}%
\usepackage{algpseudocode}%
\usepackage{listings}%

\usepackage[utf8]{inputenc}%
\usepackage[T1]{fontenc}%
\usepackage{hyperref}%
\usepackage{url}%
\usepackage{nicefrac}%
\usepackage{microtype}%
\usepackage{subcaption}%
\usepackage{wrapfig}%
\usepackage{enumitem}%
\usepackage{ragged2e}%
\usepackage{tabularx}%

\DeclareUnicodeCharacter{207A}{\ensuremath{^+}}  
\DeclareUnicodeCharacter{207B}{\ensuremath{^-}}  
\DeclareUnicodeCharacter{2082}{\ensuremath{_2}}  
\DeclareUnicodeCharacter{2084}{\ensuremath{_4}}  
\DeclareUnicodeCharacter{03C0}{$\pi$}            

\newcommand{\figref}[1]{Fig.~\ref{#1}}

\newcommand{\appref}[1]{Appendix~\ref{#1}}

\usepackage{xcolor}
\usepackage[textsize=footnotesize]{todonotes}
\usepackage{xspace}
\definecolor{Maroon}{RGB}{128,0,0} 

\newif\iffinal

\iffinal
    
    \newcommand{\YC}[1]{}
    \newcommand{\yuxin}[1]{}
\else
    
    \newcommand{\YC}[1]{\todo[fancyline,color=blue!40]{YC: #1}\xspace}
    \newcommand{\yuxin}[1]{\textcolor{Maroon}{[YC: #1]}}
\fi

\theoremstyle{thmstyleone}%

\theoremstyle{thmstyletwo}%

\theoremstyle{thmstylethree}%
\newcommand{\keywords}[1]{%
  \par\addvspace{10pt}%
  \noindent\textbf{Keywords:} #1\par%
}

\makeatletter
\newcommand{\mytableofcontents}{%
  \begingroup
    \parskip=0pt
    \parindent=0pt
    \@starttoc{toc}%
  \endgroup
}
\makeatother

\usepackage{tocloft}

\newcommand{\appendixtableofcontents}{%
  \begingroup
  \renewcommand{\cftsecnumwidth}{4em}%
  \renewcommand{\cftsecpresnum}{}
  \renewcommand{\cftsecaftersnum}{}
  \renewcommand{\cftsecfont}{%
    \ifnum\pdfstrcmp{\thesection}{\thesection}=0\fi
  }%
  \tableofcontents
  \endgroup
}

\begin{document}


\title{Hierarchical Deep Research with Local–Web RAG: Toward Automated System-Level Materials Discovery}


\author{%
Rui Ding\textsuperscript{1,2,\footnotemark[1]} \and
Rodrigo Pires Ferreira\textsuperscript{1,2,\footnotemark[1]} \and
Yuxin Chen\textsuperscript{3,\footnotemark[2]} \and
Junhong Chen\textsuperscript{1,2,\footnotemark[2]}\\[0.5em]
\textsuperscript{1}Pritzker School of Molecular Engineering, University of Chicago
\\
\textsuperscript{2}Chemical Sciences and Engineering Division, Argonne National Laboratory 
\\
\textsuperscript{3}Department of Computer Science, University of Chicago 
}

\footnotetext[1]{These authors contributed equally to this work.}
\footnotetext[2]{Corresponding authors: chenyuxin@uchicago.edu, junhongchen@uchicago.edu}

\date{}

\maketitle


\abstract{We present a long-horizon, hierarchical deep research (DR) agent
designed for complex materials and device discovery problems that exceed the
scope of existing Machine Learning (ML) surrogates and closed-source commercial agents. Our
framework instantiates a locally deployable DR instance that integrates local
retrieval-augmented generation with large language model reasoners, enhanced
by a Deep Tree of Research (DToR) mechanism that adaptively expands and prunes
research branches to maximize coverage, depth, and coherence. We systematically
evaluate across 27 nanomaterials/device topics using a large language model (LLM)-as-judge
rubric with five web-enabled state-of-the-art models as jurors. In addition,
we conduct dry-lab validations on five representative tasks, where human
experts use domain simulations (e.g., density functional theory, DFT) to verify whether DR-agent proposals
are actionable. Results show that our DR agent produces reports with quality
comparable to— and often exceeding— those of commercial systems
(ChatGPT-5-thinking/o3/o4-mini-high Deep Research) at a substantially lower cost,
while enabling on-prem integration with local data and tools.}


\keywords{deep research, hierarchical reasoning, materials discovery,
          retrieval-augmented generation, large language models,
          computational materials science, autonomous agents}

\section{Introduction}\label{sec:intro}

Machine learning (ML) for nanomaterials and device discovery has advanced from tractable single-entity predictions to increasingly complex, system-level challenges.
At the molecular or crystal level (S1), modern surrogates routinely predict
intrinsic properties~\cite{xie2018cgcnn}.
Small assemblies (S2)---interfaces, heterostructures, or simple 2--3-component
devices---introduce coupling effects yet remain manageable.
Real-world nano-devices, however, operate at system scope (S3): multi-layer
perovskite stacks with interdependent layers, core--shell--doped catalysts with engineered
interfaces, or nano-architected battery electrodes where transport across length scales
governs performance~\cite{severson2019cyclelife}.
The ultimate challenge lies in cross-domain integration (S4): platforms that span
material families, recognition chemistries, and transduction modalities---for example,
engineering 2D-material field-effect transistor (FET) sensors with aptamer functionalization
for environmental detection~\cite{szymanski2023alab,wang2022aptamer_gfet}.

Meeting this escalation in scope requires a matching escalation in depth.
We frame depth from single-pass prediction/forward screening (D1), to
multi-stage pipelines with structured outputs (D2), to long-horizon
synthesis with iterative planning (D3), and ultimately hierarchical, breadth–depth
exploration (D4).
While recent agentic frameworks have successfully demonstrated hierarchical orchestration for complex scientific workflows~\cite{gottweis2025towards,ghafarollahi2025sciagents,zou2025elagente}, ensuring robust execution and simulation management, device/system-level optimization and design (S3--S4) presents a distinct challenge.
The governing laws for device integration are not fully encoded in standard simulation packages, requiring long-horizon hierarchical autonomous discovery (D3--D4) with potential to navigate the combinatorial explosion of processing parameters sparsely distributed across heterogeneous knowledge sources.
Existing approaches typically face limitations in this unstructured knowledge regime~\cite{montoya2020camd,merchant2023gnome,chanussot2021oc20,levine2025omol25} (Detailed in \appref{app:related}).

To address this gap, we introduce a hierarchical deep-research (DR) agent\footnote{We use \emph{Deep Research (DR)} to denote a single-threaded, iterative retrieval--reasoning--writing loop run by one agent. \emph{Deep Tree of Research (DToR)} denotes a tree-structured orchestration that spawns and selects among multiple DR branches with hierarchical planning and merge. Throughout, "hierarchical deep research" is synonymous with DToR; we henceforth use only DR and DToR for clarity.} purpose-built for long-horizon scientific inquiry. Given a natural-language inquiry, the system instantiates a locally deployable DR instance that iterates between local and web retrieval-augmented generation (RAG) and large language model (LLM) reasoners (summarizers, reflectors). We further expand the DR instance into a
Deep Tree of Research (DToR) where each instance is treated as a node in a tree-structured workflow. DToR adaptively expands/prunes branches to improve coverage, depth, and coherence. The agent is self-directed/navigated (plans its own research path), runs locally for controllability and privacy, and targets S3–S4 hypotheses.
At a glance, our goal is to measure synthesis quality by a programmatic evaluation harness (anchored rubric, A/B preferences) plus dry-lab checks. We then show that orchestration via DToR consistently improves synthesis quality over single-instance DR across model backbones and local-corpus budgets at comparable compute.

\noindent\textbf{Contributions.} (1) A democratized, locally deployable framework for scientific deep research, enabling materials researchers to exercise fine-grained control over cost and preferences using open-source LLMs while achieving performance that surpasses commercial state-of-the-art solutions in most cases (\S\ref{sec:experiments}).
(2) A hierarchical orchestration (DToR) that yields robust gains over single-instance DR across LLM backbones and local data budgets, supported by comprehensive factorial and targeted component ablations.
(3) An evaluation program combining anchored rubric scores, repeated A/B duels, and dry-lab validation to assess synthesis quality.

\subsection{Related work}
Recent work on structured reasoning for LLMs has moved beyond linear prompting toward explicit search structures.
Tree-of-Thoughts (ToT) frames inference as a branching search over intermediate `thoughts',
while Graph-of-Thoughts (GoT) generalizes this to arbitrary graphs to model non-linear
dependencies and subproblem coordination
\cite{yao2023tree,besta2024graph,yao2024effectivegot}.
In parallel, reason–act and self-improvement paradigms—ReAct, Reflexion, and Self-Refine—provide
strong baselines for multi-step decision-making and revision
(\cite{yao2022react,shinn2023reflexion,madaan2023selfrefine}).
Foundational web-agent environments/benchmarks—BrowseComp, WebLINX, WebArena, and BrowserGym—formalize
browsing/navigation competencies and support reproducible evaluation
\cite{wei2025browsecomp,pan2024weblinx,zhou2023webarena,lesellier2024browsergym}.
Work on retrieval is also moving toward more adaptive strategies:
Self-RAG \cite{asai2024selfrag} learns when to retrieve and critique outputs, CRAG \cite{yan2024crag} evaluates retrieval quality and escalates to web search when evidence is weak, and GraphRAG \cite{peng2024graphragsurvey} uses graph-structured knowledge for non-flat retrieval.
Despite this progress, there is no framework that unifies structured branching and adaptive retrieval in a way that scales under practical compute and data limits for domain-specific research settings.

\section{Method}\label{sec:method}

D4 requires structured and budget-aware planning beyond short-term queries. We introduce DToR, a breadth-then-depth controller for DR that starts from multiple parallel perspectives, selectively expands or prunes based on explicit gap checks, runs evidence-first research steps (local RAG with adapted web search), and produces source-grounded reports.

\begin{figure}[t]
  \centering
  \includegraphics[width=0.9\textwidth]{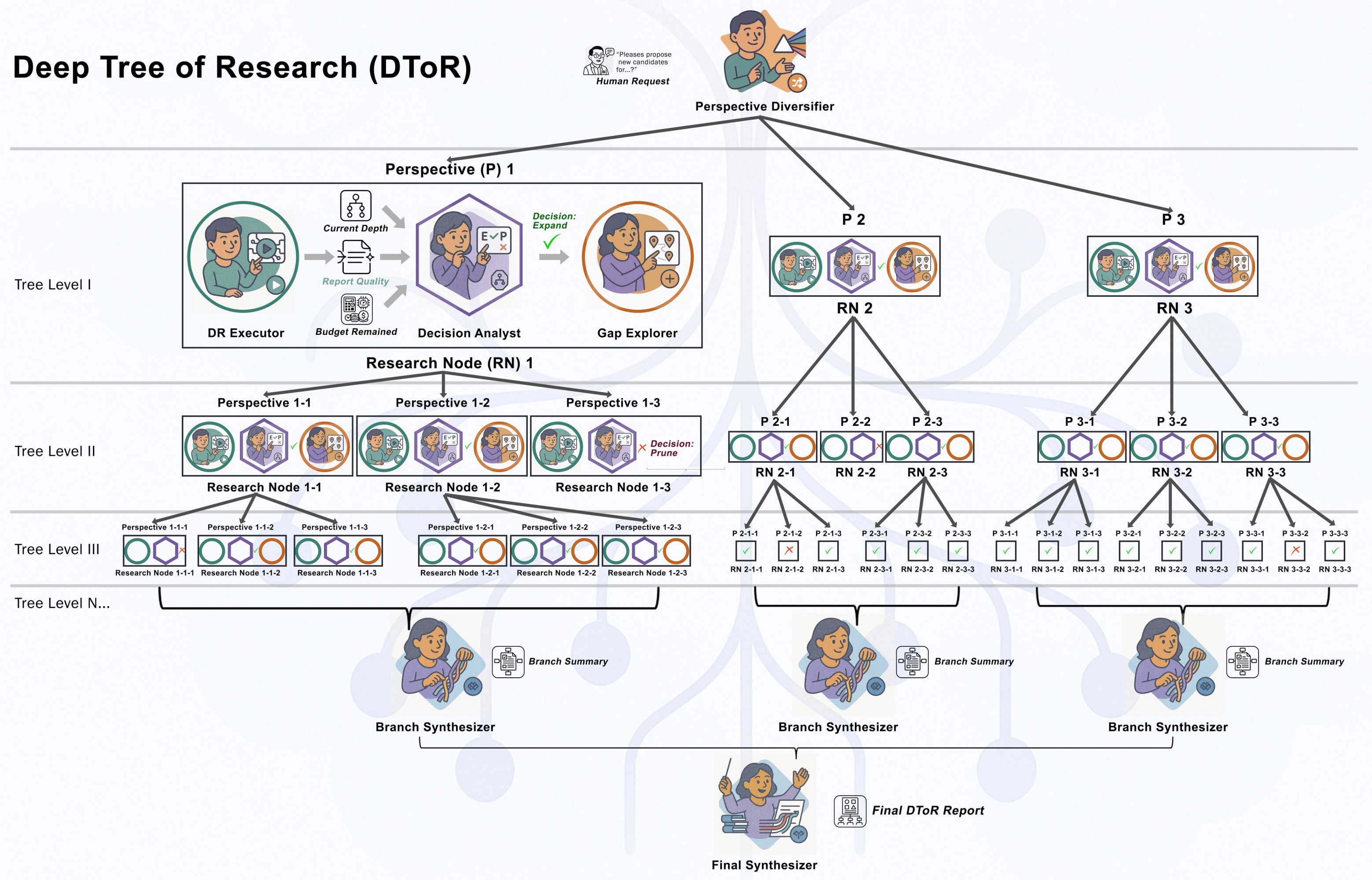}
  \caption{The DToR Depth-Breadth Workflow}
  \label{fig:DR_DToR_overview}
\end{figure}

\subsection{Single DR Instance}
As shown in \figref{fig:supp_1_single_DR_schematic}\footnote{Supporting figures (S1--S13) illustrating detailed experimental results and supplementary analyses are presented in \appref{app:supp_figures}.} (with comparison against common standard
DR agent \cite{langchain2025localdeepresearcher}), each DR instance is an evidence-first loop that
(1) generates a search query; (2) performs local RAG; (3) summarizes local evidence;
(4) generates a complementary query to induce topical diversity; (5) conducts web research for
both queries; (6) integrates/updates previous retrieved segments in a running summary;
(7) reflects to propose follow-up query.
The process would loop until a set maximum round before (8) finalizing.
Three design choices distinguish this instance from common local DR
agents\cite{langchain2025localdeepresearcher} which only conduct (1)-(5)-(6)-(7):
local-first retrieval (before web) to reduce drift, hallucination, and surface domain priors; diversity-aware query generation (local summary derived complementary query) to guarantee comprehensiveness; and robust I/O discipline to prevent controller stalls on local-hosted LLMs. Compared with classic ReAct/Reflexion-style linear loops, this instance adds an explicit breadth mechanism and
escalation policy akin to adaptive/corrective-RAG, while remaining compatible with the drop-in alternative RAG methods.
\subsection{DToR: from perspectives to gap-driven expansion}
Based on the single DR instance, we treat each DR instance as a Research Node (RN) inside a branch-and-bound controller (\figref{fig:DR_DToR_overview}
). Starting from a user query, a diversifier proposes several orthogonal Perspectives (P1, P2...)—each seeds one branch with a budget:
maximum depth, nodes per branch, and total branches. For the active branch, the router (i) runs the next pending RN (the single DR instance), producing a RN summary; (ii) invokes analyst based on remaining depth/budget/current RN report quality to give a decision {EXPAND, PRUNE}; and, if EXPAND with budget remaining, (iii) invokes knowledge gap explorer to materializes new RNs from this RN
targeted to those gaps (depth+1) and updates remaining budget. Branches that stagnate are pruned; (iv) branches that reach depth or converge are synthesized into a perspective report. When all branches reach synthesis, a final synthesizer reconciles cross-branch evidence, resolves conflicts, and outputs a provenance-rich, consolidated report. Conceptually, this is a tree-of-search over (query,
evidence, summary) states rather than over purely symbolic "thoughts," inheriting the benefits of structured exploration/backtracking from tree/graph-based reasoning while remaining tailored to information-seeking and citation fidelity under explicit resource budgets.
\section{Experiments}\label{sec:experiments}

\subsection{Experimental setup}
We evaluate across 27 expert-crafted topics spanning materials and devices (e.g., per- and polyfluoroalkyl substances (PFAS) FET probes, CO$_2$ sensing in 2D materials, Li/Na-selective membranes, CO$_2$-reduction catalysts); full topic prompts are provided in \appref{app:topics}.
For each topic, we consider 41 agents---11 commercial and 30 local (split evenly between single DR-instance and DToR-mode)---on consumer-level hardware.
We evaluate local agents across LLM backbones (gpt-oss120B, gpt-oss20B, QwQ32B) and local RAG budgets (local0/local100/local500), holding orchestration fixed and varying only backbone and corpus size; full agent configurations in \appref{app:exp_details_DR}. Using an LLM-as-judge protocol (\appref{app:scoring_template}), we generate 41 distinct report variants per topic and have five web-enabled SOTA judges---Claude 4 Opus (thinking), Gemini 2.5 Pro, Grok-3 (thinking),
ChatGPT-o3, and ChatGPT-o4-mini-high---score each report under a fixed, double-blind rubric (judges are blinded to agent identities) with five equally weighted dimensions---(1) relevance, (2) depth, (3) clarity, (4) applicability, and (5) novelty---whose concise definitions are provided in Appendix~\ref{app:scoring_template}.

Judge view and anonymization: LLM judges see only the final report text---no agent/model identifiers, tool logs, or token/count metadata---and assign numeric scores on five anchored dimensions (relevance, depth/rigor, clarity/reproducibility, applicability, novelty) under a fixed instruction to ``act as an experienced materials scientist'' (\appref{app:scoring_template}). The rubric provides positive/negative exemplars per dimension; unresolved contradictions or failure to reconcile conflicts between local and web evidence are penalized under Depth/Rigor and Clarity/Reproducibility, while reports that explicitly compare sources and justify a resolution are rewarded.
Access to local vs.\ web contexts is controlled at generation time (agents run in local-only or web-enabled modes), whereas judges never see those contexts---only the synthesized report---so the evaluation targets synthesis/reasoning quality rather than retrieval exposure.
Judges assign 0–10 per dimension with one-sentence justifications; numeric ratings are extracted by a deterministic parser. We repeat the evaluation three times per topic–report–judge combination, yielding $27\times 41 \times  5 \times 3 = 16,605$ rubric-based judgments. Inter-judge agreement (Pearson correlation on scores and Spearman rank correlation) remains high across all dimensions.

We then run A/B dueling preferences using the same setup. For each topic, we form three pools: commercial, local gpt-oss, and local non-gpt-oss. Within each pool we rank reports by rubric average and select the top-3; the resulting 9 reports then enter all pairwise duels ($\binom{9}{2}=36$ pairs per topic).
Duels are double-blind and order-randomized, guided by a concise ``experienced materials scientist'' prompt emphasizing clarity, gap prioritization, and actionable next steps. Each duel is repeated three times, yielding $27 \times 36 \times 3 = 2916 A/B$ trials and $14,580$ individual judge decisions.

Finally, we conduct dry-lab validation on five representative tasks---PFAS sensor probe, PFAS degradation catalysts, battery binder selection, oxygen evolution reaction (OER) catalyst stability, and CO\texorpdfstring{$_2$}{2} sensor probe---by sending anonymized best-local and best-commercial reports to domain experts who extract candidates and build simulation environments
(\appref{app:dry_lab_details}).
We score ten simulation metrics per task with domain-standard protocols (e.g., explicit-solvent ab initio molecular dynamics, AIMD, when necessary) and aggregate per-task and overall performance; full simulation setups and metric definitions are in \appref{app:dry_lab_details}.
We also run a full-factorial ablation across Method $\times$ LLM backbone $\times$ local-corpus budget (27 topics, 41 agents) to isolate main effects and interactions (\figref{fig:main_rubric_results}b), and conduct targeted component ablations on the best-performing configuration across three representative topics to further dissect orchestration contributions (\appref{app:additional_ablation}).
To characterize the benchmark itself, we compute per-topic difficulty (mean performance across agents) and discrimination power (standard deviation of agent scores), revealing which topics best separate strong from weak systems; full topic-level analysis is in \appref{app:topic_analysis}.
All runs were executed on a workstation with 4$\times$NVIDIA RTX A5500 (Ubuntu); completion-time statistics and energy estimates are reported in \appref{app:power_estimation}.
The DToR framework source code is publicly available at \url{https://github.com/ruiding-uchicago/DToR_deep_research}.
\subsection{Experimental results}

\begin{figure}
  \centering
  \includegraphics[width=0.9\textwidth]{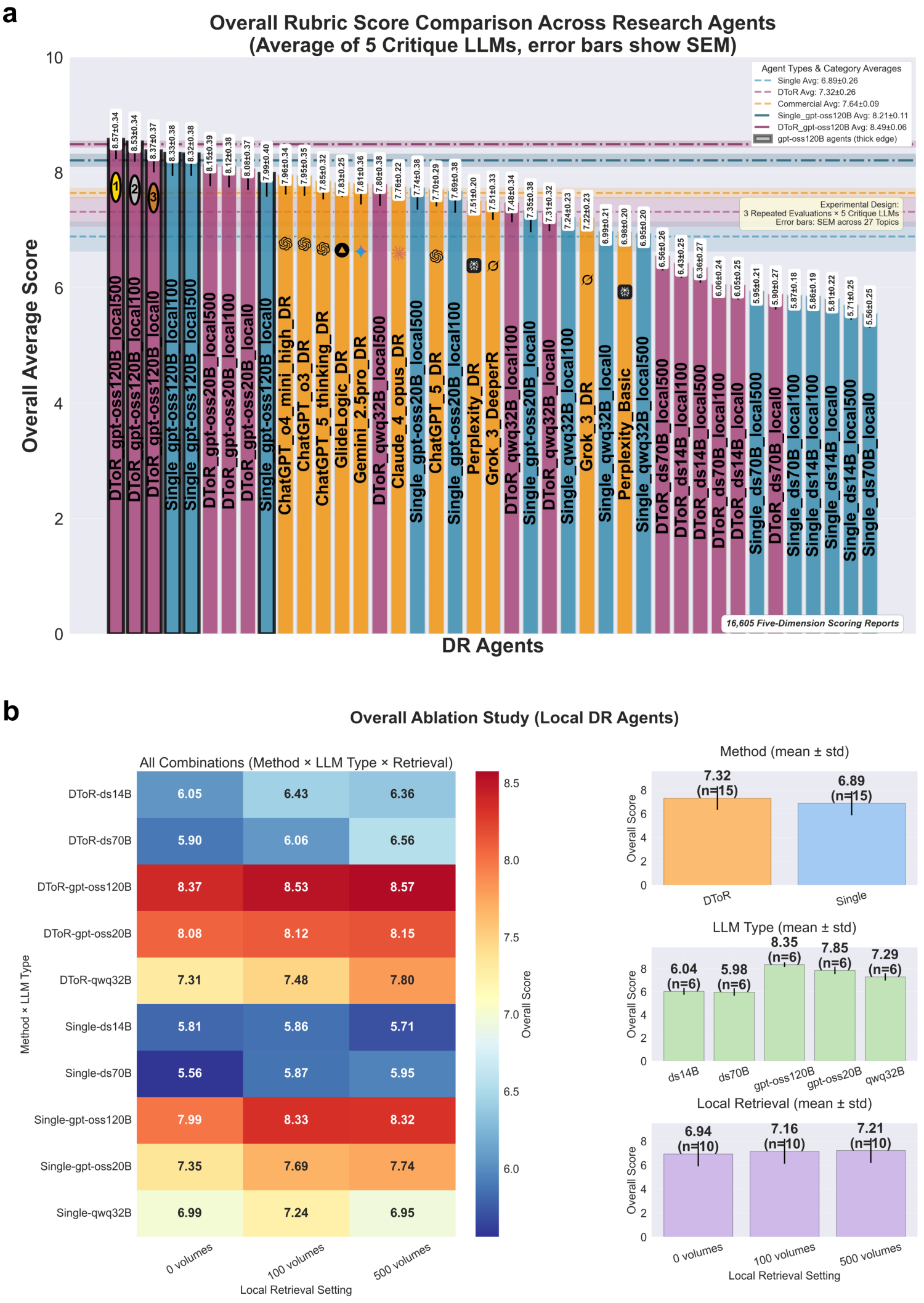}
  \caption{(a) Overall five-dimension rubric scores across 41 agents (27 topics; 5 judges; 3 trials).
(b) Full-factorial ablation (Method $\times$ LLM $\times$ Local retrieval); heatmap shows overall means; side bars show each factor impact.
}
  \label{fig:main_rubric_results}
\end{figure}

Across the five LLM-as-judge dimensions on 27 topics $\times$ 41 agents (\figref{fig:main_rubric_results}a), DToR\_gpt-oss120B\_local500 (running DToR with gpt-oss120B and all 500 volumes of local corpus for RAG; see details in \appref{app:exp_details_DR})
achieves an average score of 8.57/10, ranking 1st overall. When the local RAG budget is reduced to local100 and local0, the scores remain 8.53/10 and 8.37/10, both still top-3. Even without the DToR controller (i.e., a single DR instance), gpt-oss120B reaches 8.33/10 (local100) and 8.32/10 (local500), ranking 4th and 5th. Under the same setup with gpt-oss20B, DToR yields 8.15/10, 8.12/10, and
8.08/10 for local500/local100/local0, respectively. All of the above surpass commercial baselines including ChatGPT-o4-mini-high (7.96/10), ChatGPT-o3 (7.95/10), and ChatGPT-5-thinking (7.85/10), as well as other widely used systems such as Gemini 2.5 Pro (7.81/10), Claude Opus 4 (7.78/10), Grok 3 (7.51/10), and Perplexity (7.51/10). Notably, DToR\_QwQ32B\_local500 driven by laptop compatible
lightweight LLM also attains 7.80/10, outperforming roughly half of the commercial systems. 

A closer look at the five dimensions (\figref{fig:supp_3_each_dim_score}) shows that clarity and depth are the most challenging, with the lowest means and the largest dispersion. Even so, DToR yields the largest gains over a single-DR instance precisely on these two dimensions, with average improvements of +0.69 and +0.72, respectively. Excluding relevance—which exhibits a ceiling effect and thus limited separability—DToR delivers clear improvements on the remaining four dimensions. For example, on novelty and applicability, DToR improves Single\_gpt-oss120B\_local500 by +0.43 and +0.42 points on average, pushing it above the leading commercial
baseline. Inter-rater consistency across the five LLM critics was high: average Pearson correlation $\mathrm{p}=0.97$ on raw scores and Spearman $\rho=0.97$ on rank orders (\figref{fig:supp_4_corr_score_rank}).

Reliability is further supported by agent-specific and inter-trial consistency (\figref{fig:supp_5_agent_specific_consistency}, \figref{fig:supp_5_dim_trial_consistency}).
From an ablation study perspective (\figref{fig:main_rubric_results}b), enabling local RAG (7.21/10) and especially DToR mode (7.32/10) increases the average score.
Across all 27 topics and 41 agents, Method $\times$ Backbone $\times$ Local-corpus budget factorial analysis shows that the dominant effect is orchestration: DToR consistently outperforms single-instance DR once a non-trivial local corpus is present (local100/local500), whereas at local0 the gap narrows and a strong backbone (gpt-oss120B) can close it.
To further isolate orchestration components, we conducted targeted component ablations on the best-performing configuration (DToR\_gpt-oss120B\_local500) across three representative topics: systematically degrading tree depth (3$\rightarrow$2 perspectives), reflection loops (3$\rightarrow$0 iterations), and retrieval counts (5$\rightarrow$1 papers per query) all reduce performance, with reflection proving most critical; removing web search (local-only or LLM-only modes) similarly degrades quality, confirming gap-triggered web complement adds value; driving the same orchestration with API models (gpt-5-mini/nano) maintains relative orderings but at materially higher cost (\figref{fig:supp_11_add_ablation_5dim}, \figref{fig:supp_12_add_ablation_duel}, \figref{fig:supp_13_add_ablation_consistency}; full details in \appref{app:additional_ablation}).
Per-topic analysis reveals a clear difficulty spectrum: sensing/characterization tasks (in-situ TEM, printed FET arrays) are easiest, while battery electrochemistry topics (lithium ion battery: LIB binders, fluoroether anion receptors) are hardest; discrimination power shows environmental sensing topics (microplastics, antibiotics detection) separate agents weakly ($\sigma \approx 0.7$), whereas battery/materials synthesis challenges (LIB fluoroether receptor, ambient-pressure diamond growth) sharply distinguish strong from weak systems ($\sigma > 1.1$).
Notably, topic difficulty and discrimination are decoupled: LIB\_Fluoroether\_Anion\_Receptor exhibits both the highest difficulty ($\mu = 6.93$) and the largest discrimination ($\sigma = 1.24$), sharply separating strong and weak agents; conversely, Microplastics\_Sensing\_2D shows the lowest discrimination ($\sigma = 0.69$) at mid-pack difficulty ($\mu = 7.19$); Catalytic\_Mineralization\_of\_Microplastics is relatively easy ($\mu = 7.49$) yet still discriminates well ($\sigma = 1.03$), while Wastewater\_Resource\_Recovery\_Cats is hard ($\mu = 7.04$) but offers weak separation ($\sigma = 0.93$).
Critically, DToR agents maintain consistent performance advantages across both high-$\sigma$ and low-$\sigma$ topics, confirming that orchestration benefits are robust to topic characteristics rather than prompt-specific artifacts; full per-topic difficulty rankings and discrimination analysis are in \appref{app:topic_analysis}. 
\begin{figure}[h!]
  \centering
  \includegraphics[width=0.9\textwidth]{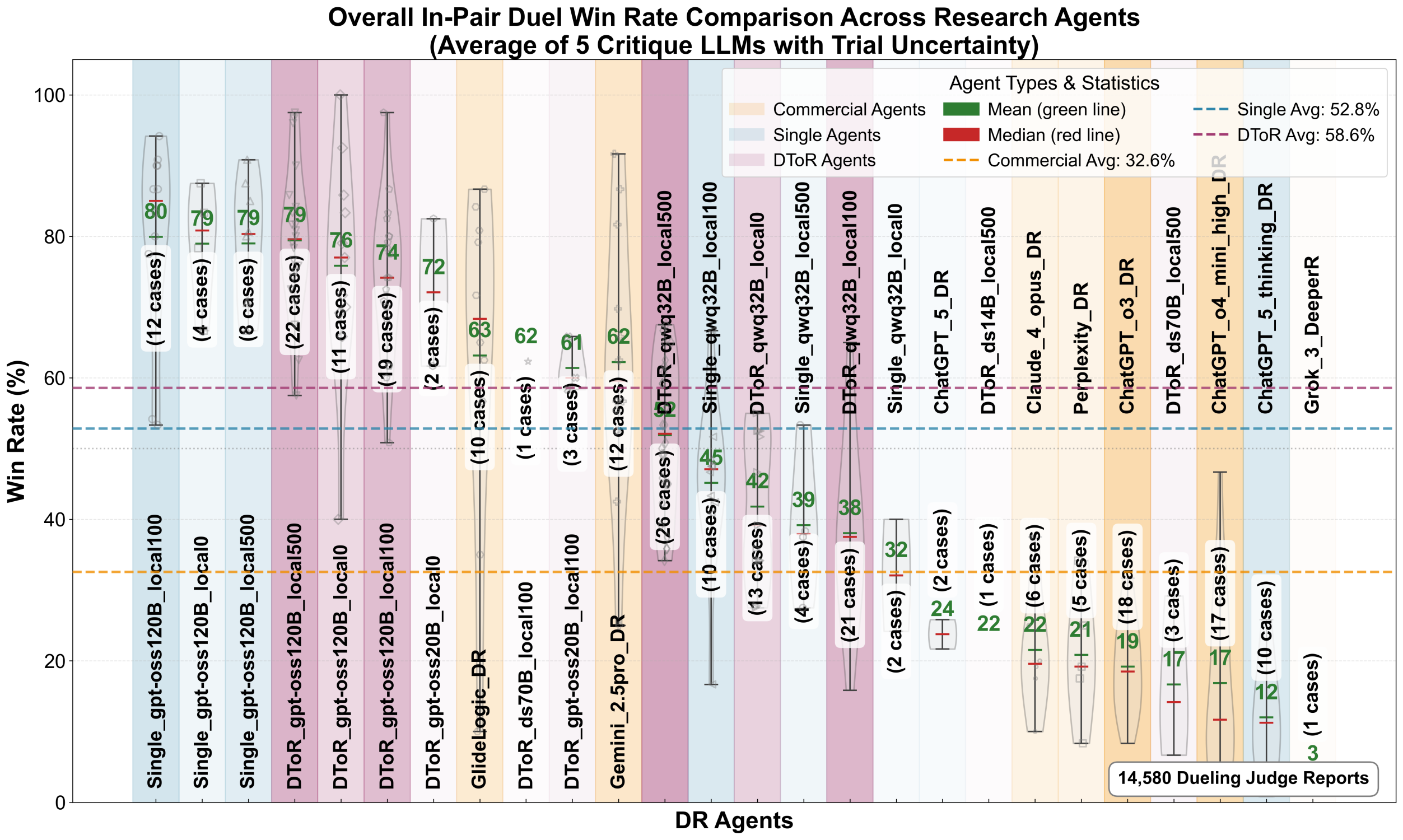}
  \caption{Win Rate Violin Plot - Shows the distribution of win rates across
different agents and topics with detailed statistical information and performance
variations}
  \label{fig:main_duel_violin}
\end{figure}

To mitigate potential bias inherent in numerical rubric scoring, we further complement these results with pairwise dueling—a side-by-side comparison that directly elicits preferences from LLM judges without imposing a rigid numerical scale. Pairwise dueling (\figref{fig:main_duel_violin}) confirms these findings across all 27 topics (\figref{fig:supp_6_duel_perf}) with strong experimental consistency (\figref{fig:supp_3_duel_consistency}): (1) DToR agents show a higher average win rate (58.6\%) than Single DR agents (52.8\%) in head-to-head comparisons; (2) DToR\_gpt-oss120B\_local500 achieves an impressive 79\% mean win rate; (3) gpt-oss-driven DToR and Single agents both exhibit similarly high win rates above 74\%, demonstrating a clear advantage. These outcomes further support the previous rubric scoring results, confirming that our locally hosted DToR agents generate superior DR reports.

\begin{figure}[h!]
  \centering
  \includegraphics[width=0.9\textwidth]{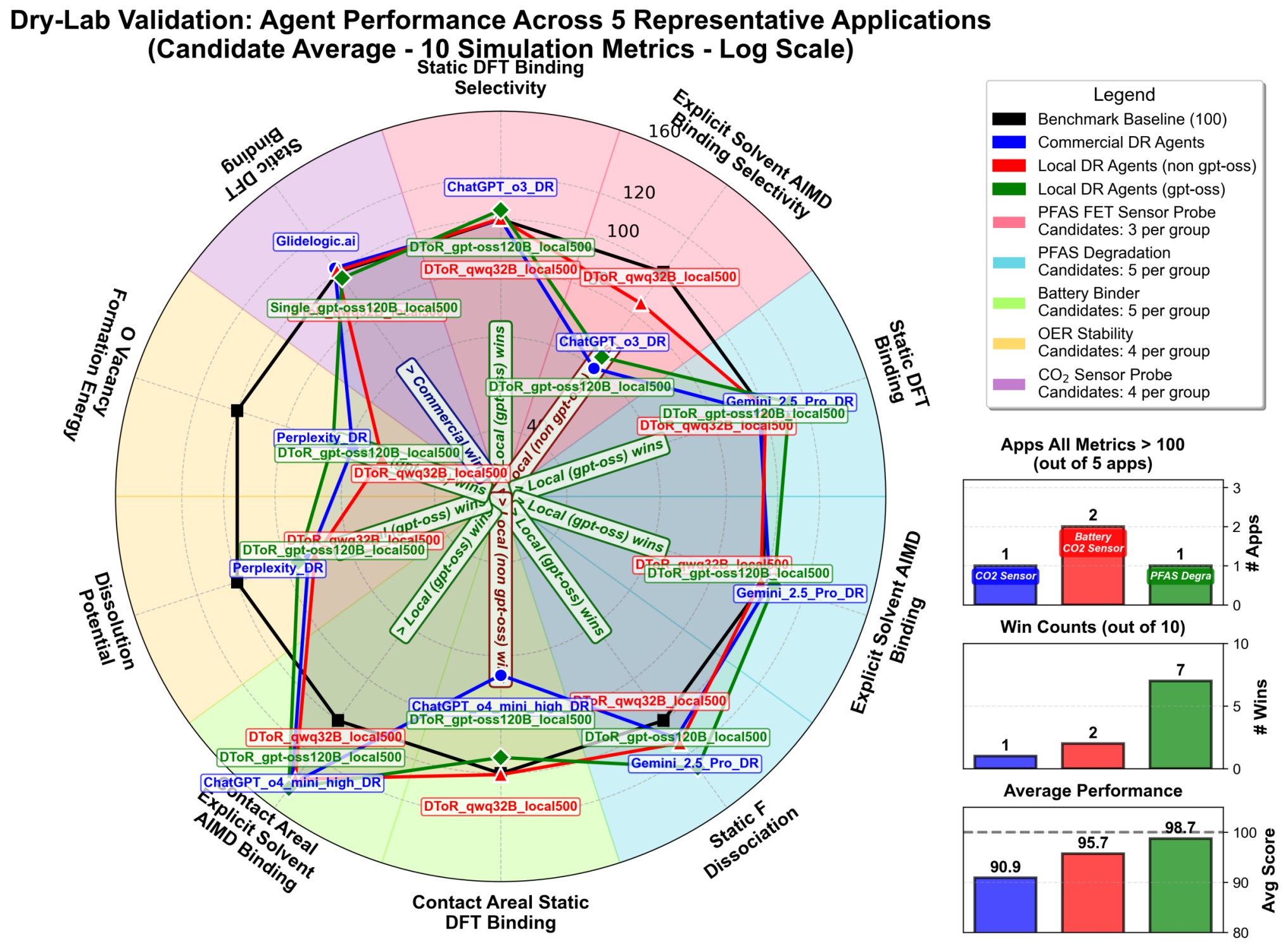}
  \caption{Dry-lab validation on five tasks: radar of 10 simulation metrics comparing best local vs. best commercial DR Agents. The bar plots located at bottom right indicate respectively: the time of agent proposed candidates obtain score surpassed domain prior (>100); the time of agent proposed candidates obtain the highest score among three groups; the overall 10 metrics average score.}
  \label{fig:main_dry_lab}
\end{figure}

In dry-lab validation spanning 10 metrics over five application scenarios (\figref{fig:main_dry_lab}), we comprehensively validate across domain benchmarks with DFT and explicit-solvent AIMD, scoring each task against a well-recognized baseline normalized to 100.
The tasks and metrics are:
(i) PFAS sensor binding---static-DFT binding energies $\Delta E_{\text{bind}}$ for PFAS vs.\ interferents and explicit-solvent end-state potential-energy gaps from short AIMD, with selectivity scored as $\Delta\Delta E$ (PFAS minus interferent) and anchored to $\beta$-cyclodextrin\cite{wang2024reversible} (Score=100);
(ii) PFAS degradation catalysis---C--F bond dissociation energies from constrained scans and solvated binding stability proxies, normalized to Ti$_4$O$_7$ (Magnéli)\cite{lin2018ti4o7} as the domain benchmark (100);
(iii) battery binder adhesion (on NCM811)---areal-normalized $\Delta E_{\text{bind}}/A$ (static DFT, contact area via convex-hull atoms within 3.5\,\AA) and explicit-solvent end-state gap (AIMD) to capture electrolyte competition, anchored to polyvinylidene fluoride (PVDF)\cite{zhong2021pvdf} (100);
(iv) OER catalyst stability---dissolution potential $U_{\text{diss}}$ (tendency for surface metal leaching) and oxygen-vacancy formation energy $V_{\mathrm{O}}$ (lattice robustness), normalized to IrO$_2$\cite{ding2025oermulti} (100); and
(v) CO$_2$ sensor binding---grid-sampled static-DFT binding energies on candidate surfaces, linearly transformed and anchored to g-C$_3$N$_4$\cite{karthik2023gc3n4co2} (100).
Across these five applications, we observe four all-metrics $>$ 100 events (i.e., every metric for a task exceeds its baseline): commercial once on CO$_2$ sensor; local DR (non-gpt-oss) twice on CO$_2$ sensor and battery binder; and local DR (gpt-oss) once on PFAS degradation.
In these cases, the DR-proposed candidates point to concrete, testable improvements over their domain baselines in terms of qualitative dry lab trials: for CO$_2$ sensing, stronger and more uniform adsorption than g-C$_3$N$_4$ (reduced site-to-site dispersion); for battery binders, higher areal-normalized adhesion than PVDF under dry interfaces and competitive performance under explicit electrolyte; and for PFAS degradation, lower C--F dissociation energies relative to Ti$_4$O$_7$ while maintaining adequate adsorption stability.
Notably, explicit-solvent AIMD often reorders static-DFT rankings, revealing candidates that look promising in vacuum but lose ground when solvation/competition is included---an effect we quantify in the AIMD score panels.
The gpt-oss-based local DR agent leads on 7 out of 10 metrics when judged by per-metric averages and also achieves a higher overall score aggregating across metrics (98.7). Related details are available in \appref{app:dry_lab_details}. 

\begin{figure}[t]
  \centering
  \includegraphics[width=1\textwidth]{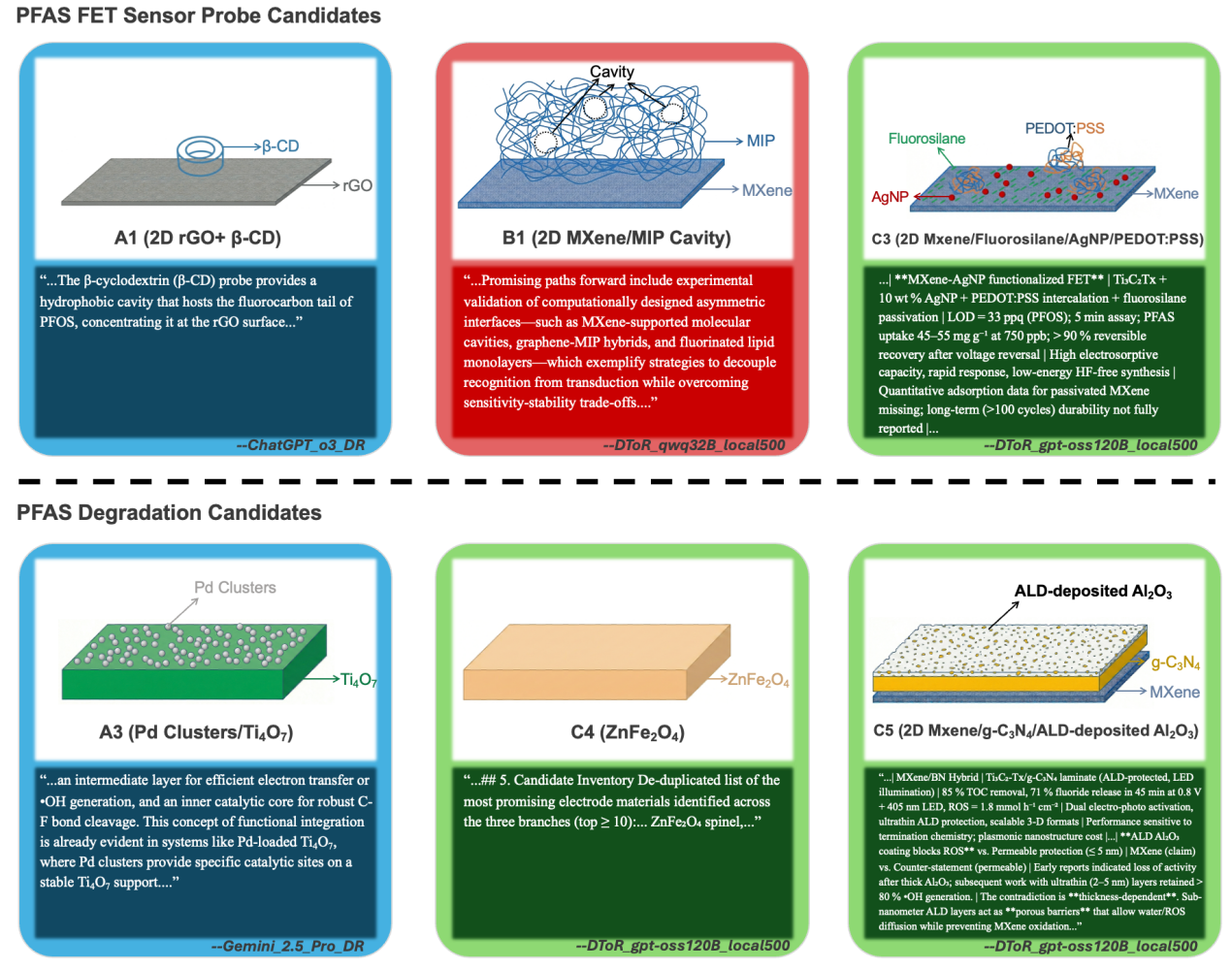}
\caption{Visual representations of selected candidates drawn based on specific descriptions from the DR reports. The top panel illustrates candidates for PFAS sensing (A1, B1, C3), while the bottom panel shows candidates for PFAS degradation (A3, C4, C5), accompanied by the original text excerpts generated by the agents.Dry lab qualitative studies are available in \appref{dry_lab_PFAS_sens} and \appref{dry_lab_PFAS_degra}}
  \label{fig:good_bad_examples}
\end{figure}

While dry-lab simulations provide an initial filter, looking more closely at the candidates proposed by the DR agents reveals a feasibility gap in current LLM–RAG orchestration(\figref{fig:good_bad_examples}). In PFAS sensing, the agents successfully identified rational designs like B1 and A1, yet DToR agent also produced “kitchen-sink” failures like C3, which stacks four mutually incompatible phases (hydrophobic fluorosilane, hydrophilic MXene, Ag nanoparticles, acidic PEDOT:PSS). In practice, synthesis of such a complex system would be infeasible, because the agents ignored likely colloidal incompatibilities (e.g., phase separation and acid-induced etching) that would render the device chemically unstable\cite{cameron2020pedotpss, kumar2023mxenestability}. Similarly, for PFAS degradation, cost-effective and chemically robust candidates like C4 and A3 stand in contrast to C5, a triple heterostructure (MXene/g-C$_3$N$_4$/Al$_2$O$_3$) that neglects potential challenges in interfacial charge-transfer resistance and fabrication complexity\cite{mandari2025gcn_mxene_review, bagnall2025al2o3_stability}. These over-engineered failures stem from a fundamental limitation in current agent architectures: without intrinsic wet-lab feasibility priors, cost models, or external physics validators, LLMs are prone to “inverse-design hallucinations” \cite{miret2025llms_materials_discovery, kalai2025hallucinate} that assume functional additivity where physical incompatibility exists. This underscores that future AI4Science frameworks must integrate dedicated domain-validator modules (e.g., synthesis-aware ReAct loops) to ground theoretical generation in experimental reality\footnote{Extended discussion with further citations is available in \appref{candidate_feasibility_discussions}}..

\begin{figure}[h!]
  \centering
  \includegraphics[width=0.9\textwidth]{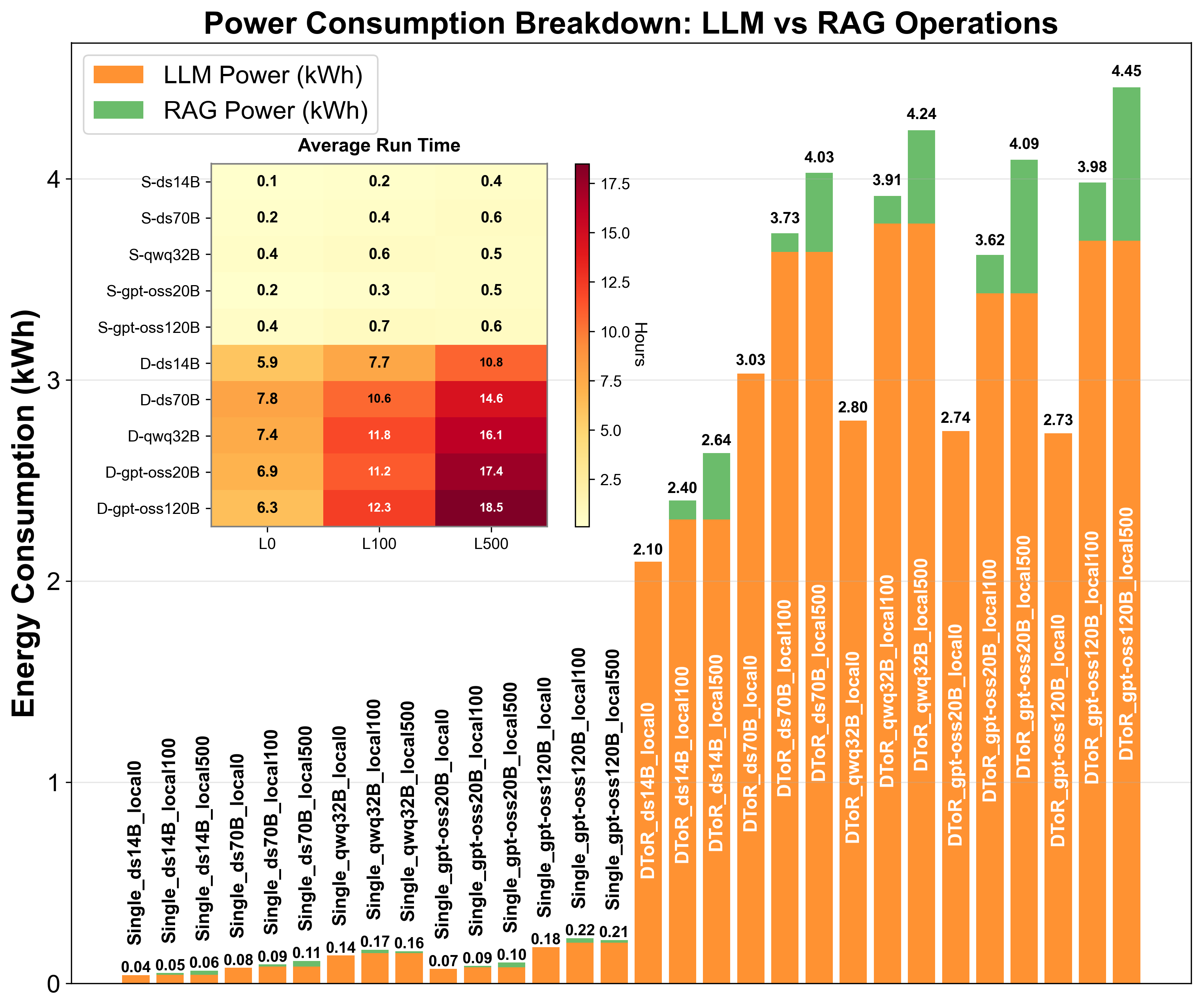}
  \caption{Local deployment cost: stacked energy (kWh) split into LLM vs. RAG; inset = average runtime (hour).}
  \label{fig:main_energy_cost}
\end{figure}

All timings reflect a single-machine setup with serialized agent scheduling (energy estimation methodology in \appref{app:power_estimation}); under this controlled setting, DToR with gpt-oss120B averages 19.6\,h (\figref{fig:main_energy_cost}, \figref{fig:supp_7_runtime_violin}).
This should be viewed as an upper bound of orchestration cost, not a requirement for quality: a low-resource profile that disables local-corpus RAG finishes in $<6$\,h (an overnight run), and a single-instance DR pipeline with gpt-oss120B typically completes within $\sim$30\,min---on the same order as commercial solutions yet, as shown earlier, with higher rubric and duel scores.
Overall, our gains arise from orchestration and information coverage rather than ``model pile-up'': users can explicitly tune for quality (DToR\,+\,corpus) or for cost/latency (DR/no-RAG) depending on resources.
Without high subscription fees (\figref{fig:supp_10_electricity_cost}) for commercial solutions, this framework offers flexible deployment from lightweight laptops to multi-node clusters\footnote{DToR branches can run in parallel; see \url{https://github.com/ruiding-uchicago/DToR_deep_research} for implementation details.}, enabling democratized, transparent, and open access to advanced research agents adaptable to diverse computational budgets of materials researchers.

\section{Conclusion}\label{sec:conclusion}

We introduced DToR, a locally deployable DR agent with a resource-bounded tree controller that unifies local-first RAG, diversity-aware query generation, and gap-driven branching into provenance-rich, long-horizon information seeking. Across 27 topics and 41 agents, under 16,605 rubric judgments from 5 web-enabled judges, our best local model (DToR\_gpt-oss120B\_local500) ranked 1st (8.57/10) with
a 79\% mean duel win rate (14,580 dueling judgement) while using only 4.37 kWh/report on consumer devices. In dry-lab validation across five applications, its candidates matched or surpassed the best commercial solutions. This work highlights a practical path toward open, secure, and scalable on-prem DR agents for automated, accelerated scientific discovery at the S3–S4 level.

\section*{Acknowledgements}\label{sec:acknowledge}
This project is supported in part by the Eric and Wendy Schmidt AI in Science Fellowship, a program of Schmidt Sciences. This project is also supported in part by the National Science Foundation Research Traineeship in AI-enabled Molecular Engineering of Materials and Systems for Sustainability (NRT-AIMEMS).Additional support came from the University of Chicago Data Science Institute through the 2024 AI+Science Research Initiative. This work was also supported in part by the National Science Foundation under the Future Manufacturing program (award 2037026) and the Robust Intelligence program (award 2313131 and 2332475). Computational resources were provided by the Carbon high-performance computing cluster at the Center for Nanoscale Materials, Argonne National Laboratory, a U.S. Department of Energy Office of Science user facility (Contract No. DE-AC02-06CH11357); the Argonne Leadership Computing Facility, a U.S. Department of Energy Office of Science user facility at Argonne National Laboratory, under the DOE Office of Science Advanced Scientific Computing Research program (Contract No. DE-AC02-06CH11357) via an ALCF Discretionary award ``BRAINIAC'' (Allocation ID 14368); the University of Chicago Research Computing Center (proposal \#68909) and the University of Chicago Data Science Institute high-performance computing cluster; and the Delta system and advanced computing resources at Texas A\&M High Performance Research Computing, both accessed through the Advanced Cyberinfrastructure Coordination Ecosystem: Services \& Support (ACCESS) program under allocation MAT240097 and supported by U.S. National Science Foundation awards 2138259, 2138286, 2138307, 2137603, 2138296.

\bibliographystyle{unsrt}
\bibliography{reference}

@article{yao2023tree,
  title={Tree of thoughts: Deliberate problem solving with large language models},
  author={Yao, Shunyu and Yu, Dian and Zhao, Jeffrey and Shafran, Izhak and Griffiths, Tom and Cao, Yuan and Narasimhan, Karthik},
  journal={Advances in neural information processing systems},
  volume={36},
  pages={11809--11822},
  year={2023}
}

@article{gottweis2025towards,
  title={Towards an AI co-scientist},
  author={Gottweis, Juraj and Weng, Wei-Hung and Daryin, Alexander and Tu, Tao and Palepu, Anil and Sirkovic, Petar and Myaskovsky, Artiom and Weissenberger, Felix and Rong, Keran and Tanno, Ryutaro and Saab, Khaled and Popovici, Dan and Blum, Jacob and Zhang, Fan and Chou, Katherine and Hassidim, Avinatan and Gokturk, Burak and Vahdat, Amin and Kohli, Pushmeet and Matias, Yossi and Carroll, Andrew and Kulkarni, Kavita and Tomasev, Nenad and Guan, Yuan and Dhillon, Vikram and Vaishnav, Eeshit Dhaval and Lee, Byron and Costa, Tiago R D and Penad{\'e}s, Jos{\'e} R and Peltz, Gary and Xu, Yunhan and Pawlosky, Annalisa and Karthikesalingam, Alan and Natarajan, Vivek},
  journal={arXiv preprint arXiv:2502.18864},
  year={2025}
}

@article{ghafarollahi2025sciagents,
  title={SciAgents: Automating Scientific Discovery Through Bioinspired Multi-agent Intelligent Graph Reasoning},
  author={Ghafarollahi, Alireza and Buehler, Markus J.},
  journal={Advanced Materials},
  volume={37},
  pages={2413523},
  year={2025}
}

@article{zou2025elagente,
  title={El Agente: An autonomous agent for quantum chemistry},
  author={Zou, Yunheng and Cheng, Austin H. and Aldossary, Abdulrahman and Bai, Jiaru and Leong, Shi Xuan and Campos-Gonzalez-Angulo, Jorge Arturo and Choi, Changhyeok and Ser, Cher Tian and Tom, Gary and Wang, Andrew and Zhang, Zijian and Yakavets, Ilya and Hao, Han and Crebolder, Chris and Bernales, Varinia and Aspuru-Guzik, Al{\'a}n},
  journal={Matter},
  volume={8},
  pages={102263},
  year={2025}
}

@article{wu2025chematagent,
  title={CheMatAgent: Enhancing LLMs for Chemistry and Materials Science through Tree-Search Based Tool Learning},
  author={Wu, Mengsong and Wang, YaFei and Ming, Yidong and An, Yuqi and Wan, Yuwei and Chen, Wenliang and Lin, Binbin and Li, Yuqiang and Xie, Tong and Zhou, Dongzhan},
  journal={arXiv preprint arXiv:2506.07551},
  year={2025}
}

@article{tang2025chemagent,
  title={ChemAgent: Self-updating Library in Large Language Models Improves Chemical Reasoning},
  author={Tang, Xiangru and Hu, Tianyu and Ye, Muyang and Shao, Yanjun and Yin, Xunjian and Ouyang, Siru and Zhou, Wangchunshu and Lu, Pan and Zhang, Zhuosheng and Zhao, Yilun and Cohan, Arman and Gerstein, Mark},
  journal={arXiv preprint arXiv:2501.06590},
  year={2025}
}

@article{wang2024reversible,
  title = {Reversible ppt-Level Detection of Perfluorooctane Sulfonic Acid in Tap Water using Field-Effect Transistor Sensors},
  author = {Wang, Yuqin and Jang, Hyun-June and Topel, Max and Dasetty, Siva and Liu, Yining and Ateia, Mohamed and Tam, Aaron and Rozyyev, Vepa and Ouyang, Ellie and Zhuang, Wen and Pu, Haihui and Lee, Sang Soo and Elam, Jeffrey and Ferguson, Andrew and Darling, Seth and Chen, Junhong},
  journal = {Nature Water},
  volume={3},
  pages={1187--1197},
  year={2025}
}

@article{bazgir2025matagent,
  title={MatAgent: A human-in-the-loop multi-agent LLM framework for accelerating the material science discovery cycle},
  author={Bazgir, Adib and Madugula, Rama Chandra Praneeth and Zhang, Yuwen},
  journal={AI4MAT-ICLR-2025 Poster},
  year={2025}
}

@article{zhang2024honeycomb,
  title={HoneyComb: A Flexible LLM-Based Agent System for Materials Science},
  author={Zhang, Huan and Song, Yu and Hou, Ziyu and Miret, Santiago and Liu, Bang},
  journal={Findings of the Association for Computational Linguistics: EMNLP 2024},
  pages={3369--3382},
  year={2024}
}

@article{li2025chemhas,
  title={ChemHAS: Hierarchical Agent Stacking for Enhancing Chemistry Tools},
  author={Li, Zhucong and Zhang, Bowei and Xiao, Jin and Zhou, Zhijian and Cao, Fenglei and Liang, Jiaqing and Qi, Yuan},
  journal={arXiv preprint arXiv:2505.21569},
  year={2025}
}

@article{wang2023sustainable,
  title={Sustainable materials acceleration platform reveals stable and efficient wide-bandgap metal halide perovskite alloys},
  author={Wang, Tonghui and Li, Ruipeng and Ardekani, Hossein and Serrano-Luj{\'a}n, Luc{\'\i}a and Wang, Jiantao and Ramezani, Mahdi and Wilmington, Ryan and Chauhan, Mihirsinh and Epps, Robert W. and Darabi, Kasra and Guo, Boyu and Sun, Dali and Abolhasani, Milad and Gundogdu, Kenan and Amassian, Aram},
  journal={Matter},
  volume={6},
  pages={2963--2986},
  year={2023}
}

@misc{ibm2020roborxn,
  title={IBM RoboRXN for Chemistry},
  author={IBM Research},
  year={2020},
  howpublished={\url{https://rxn.res.ibm.com/rxn/robo-rxn/}}
}

@article{bran2024chemcrow,
  title={Augmenting large language models with chemistry tools},
  author={Bran, Andres M. and Cox, Sam and Schilter, Oliver and Baldassari, Carlo and White, Andrew D. and Schwaller, Philippe},
  journal={Nature Machine Intelligence},
  volume={6},
  pages={525--535},
  year={2024}
}

@article{zheng2023gpt4,
  title={A GPT-4 Reticular Chemist for Guiding MOF Discovery},
  author={Zheng, Zhiling and Rong, Zichao and Rampal, Nakul and Borgs, Christian and Chayes, Jennifer T. and Yaghi, Omar M.},
  journal={Angewandte Chemie International Edition},
  volume={62},
  number={46},
  pages={e202311983},
  year={2023}
}

@article{montoya2020camd,
  title={Autonomous intelligent agents for accelerated materials discovery},
  author={Montoya, Joseph H. and Winther, Kirsten T. and Flores, Raul A. and Bligaard, Thomas and Hummelsh{\o}j, Jens S. and Aykol, Muratahan},
  journal={Chemical Science},
  volume={11},
  number={32},
  pages={8517--8532},
  year={2020}
}

@article{merchant2023gnome,
  title={Scaling deep learning for materials discovery},
  author={Merchant, Amil and Batzner, Simon and Schoenholz, Samuel S. and Aykol, Muratahan and Cheon, Gowoon and Cubuk, Ekin Dogus and others},
  journal={Nature},
  volume={624},
  pages={80--85},
  year={2023}
}

@misc{merck_synthia_website,
  title={SYNTHIA{\textregistered} Retrosynthesis Software},
  author={Merck KGaA},
  year={2025},
  howpublished={\url{https://www.synthiaonline.com/}},
  note={Accessed 2025-08-19}
}

@article{kang2024chatmof,
  title={ChatMOF: an artificial intelligence system for predicting and generating metal--organic frameworks using large language models},
  author={Kang, Yeonghun and Kim, Jihan},
  journal={Nature Communications},
  volume={15},
  pages={4705},
  year={2024}
}

@article{wood2025uma,
  title={UMA: A Family of Universal Models for Atoms},
  author={Wood, Brandon M. and Dzamba, Misko and Fu, Xiang and Gao, Meng and Shuaibi, Muhammed and Barroso-Luque, Luis and Abdelmaqsoud, Kareem and Gharakhanyan, Vahe and Kitchin, John R. and Levine, Daniel S. and Michel, Kyle and Sriram, Anuroop and Cohen, Taco and Das, Abhishek and Rizvi, Ammar and Sahoo, Sushree Jagriti and Ulissi, Zachary W. and Zitnick, C. Lawrence},
  journal={arXiv preprint arXiv:2506.23971},
  year={2025}
}

@article{zeni2025generative,
  title={A generative model for inorganic materials design},
  author={Zeni, Claudio and Pinsler, Robert and Z{\"u}gner, Daniel and Fowler, Andrew and Horton, Matthew and Fu, Xiang and Wang, Zilong and Shysheya, Aliaksandra and Crabb{\'e}, Jonathan and Ueda, Shoko and Sordillo, Roberto and Sun, Lixin and Smith, Jake and Nguyen, Bichlien and Schulz, Hannes and Lewis, Sarah and Huang, Chin-Wei and Lu, Ziheng and Zhou, Yichi and Yang, Han and Hao, Hongxia and Li, Jielan and Yang, Chunlei and Li, Wenjie and Tomioka, Ryota and Xie, Tian},
  journal={Nature},
  volume={639},
  pages={624--632},
  year={2025}
}

@article{yoshitake2022materialbert,
  title={MaterialBERT for natural language processing of materials science texts},
  author={Yoshitake, Michiko and Sato, Fumitaka and Kawano, Hiroyuki and Teraoka, Hiroshi},
  journal={Science and Technology of Advanced Materials: Methods},
  volume={2},
  pages={372--380},
  year={2022}
}

@article{trewartha2022quantifying,
  title={Quantifying the advantage of domain-specific pre-training on named entity recognition tasks in materials science},
  author={Trewartha, Amalie and Walker, Nicholas and Huo, Haoyan and Lee, Sanghoon and Cruse, Kevin and Dagdelen, John and Dunn, Alexander and Persson, Kristin A. and Ceder, Gerbrand and Jain, Anubhav},
  journal={Patterns},
  volume={3},
  pages={100488},
  year={2022}
}

@article{qiu2024polync,
  title={PolyNC: a natural and chemical language model for the prediction of unified polymer properties},
  author={Qiu, Haoke and Liu, Lunyang and Qiu, Xuepeng and Dai, Xuemin and Ji, Xiangling and Sun, Zhao-Yan},
  journal={Chemical Science},
  volume={15},
  pages={534--544},
  year={2024}
}

@article{kuenneth2023polybert,
  title={polyBERT: a chemical language model to enable fully machine-driven ultrafast polymer informatics},
  author={Kuenneth, Christopher and Ramprasad, Rampi},
  journal={Nature Communications},
  volume={14},
  pages={4099},
  year={2023}
}

@article{xu2023transpolymer,
  title={TransPolymer: a Transformer-based language model for polymer property predictions},
  author={Xu, Changwen and Wang, Yuyang and Barati Farimani, Amir},
  journal={npj Computational Materials},
  volume={9},
  pages={64},
  year={2023}
}

@article{chithrananda2020chemberta,
  title={ChemBERTa: Large-Scale Self-Supervised Pretraining for Molecular Property Prediction},
  author={Chithrananda, Seyone and Grand, Gabriel and Ramsundar, Bharath},
  journal={arXiv preprint arXiv:2010.09885},
  year={2020}
}

@article{ahmad2022chemberta2,
  title={ChemBERTa-2: Towards Chemical Foundation Models},
  author={Ahmad, Walid and Simon, Elana and Chithrananda, Seyone and Grand, Gabriel and Ramsundar, Bharath},
  journal={arXiv preprint arXiv:2209.01712},
  year={2022}
}

@article{singh2025chemberta3,
  title={ChemBERTa-3: An Open Source Training Framework for Chemical Foundation Models},
  author={Singh, Riya and Barsainyan, Aryan Amit and Irfan, Rida and Amorin, Connor Joseph and He, Stewart and Davis, Tony and Thiagarajan, Arun and Sankaran, Shiva and Chithrananda, Seyone and Ahmad, Walid and Jones, Derek and McLoughlin, Kevin and Kim, Hyojin and Bhutani, Anoushka and Sathyanarayana, Shreyas Vinaya and Viswanathan, Venkat and Allen, Jonathan E. and Ramsundar, Bharath},
  journal={ChemRxiv preprint ChemRxiv: 10.26434/chemrxiv-2025-4glrl},
  year={2025}
}

@article{chanussot2021oc20,
  title={Open Catalyst 2020 (OC20) Dataset and Community Challenges},
  author={Chanussot, Lowik and Das, Abhishek and Goyal, Siddharth and Lavril, Thibaut and Shuaibi, Muhammed and Riviere, Morgane and Tran, Kevin and Heras-Domingo, Javier and Ho, Caleb and Hu, Weihua and Palizhati, Aini and Sriram, Anuroop and Wood, Brandon M. and Yoon, Junwoong and Parikh, Devi and Zitnick, C. Lawrence and Ulissi, Zachary W.},
  journal={ACS Catalysis},
  volume={11},
  pages={6059--6072},
  year={2021}
}

@article{tran2023oc22,
  title={The Open Catalyst 2022 (OC22) Dataset and Challenges for Oxide Electrocatalysts},
  author={Tran, Richard and Lan, Janice and Shuaibi, Muhammed and Wood, Brandon M. and Goyal, Siddharth and Das, Abhishek and Heras-Domingo, Javier and Kolluru, Adeesh and Rizvi, Ammar and Shoghi, Nima and Sriram, Anuroop and Therrien, Felix and Abed, Jehad and Voznyy, Oleksandr and Sargent, Edward H. and Ulissi, Zachary and Zitnick, C. Lawrence},
  journal={ACS Catalysis},
  volume={13},
  pages={3066--3084},
  year={2023}
}

@article{barrosoluque2024omat24,
  title={Open Materials 2024 (OMat24) Inorganic Materials Dataset and Models},
  author={Barroso-Luque, Luis and Shuaibi, Muhammed and Fu, Xiang and Wood, Brandon M. and Dzamba, Misko and Gao, Meng and Rizvi, Ammar and Zitnick, C. Lawrence and Ulissi, Zachary W.},
  journal={arXiv preprint arXiv:2410.12771},
  year={2024}
}

@article{levine2025omol25,
  title={The Open Molecules 2025 (OMol25) Dataset, Evaluations, and Models},
  author={Levine, Daniel S. and Shuaibi, Muhammed and Spotte-Smith, Evan Walter Clark and Taylor, Michael G. and Hasyim, Muhammad R. and Michel, Kyle and Batatia, Ilyes and Cs{\'a}nyi, G{\'a}bor and Dzamba, Misko and Eastman, Peter and Frey, Nathan C. and Fu, Xiang and Gharakhanyan, Vahe and Krishnapriyan, Aditi S. and Rackers, Joshua A. and Raja, Sanjeev and Rizvi, Ammar and Rosen, Andrew S. and Ulissi, Zachary and Vargas, Santiago and Zitnick, C. Lawrence and Blau, Samuel M. and Wood, Brandon M.},
  journal={arXiv preprint arXiv:2505.08762},
  year={2025}
}

@article{ong2025alchemi,
  title={Revolutionizing AI-Driven Material Discovery Using NVIDIA {ALCHEMI}},
  author={Ong, Wen Jie and Alto{\`e}, Piero and Smith, Justin S. and Foster, Dallas and Alkan, Melisa and Petty, Harry},
  journal={AI4AM 2025},
  year={2025}
}

@article{gupta2022matscibert,
  title={{MatSciBERT}: A materials domain language model for text mining and information extraction},
  author={Gupta, Tanishq and Zaki, Mohd and Krishnan, N. M. Anoop and Mausam},
  journal={npj Computational Materials},
  volume={8},
  pages={102},
  year={2022}
}

@article{antunes2024crystallm,
  title={Crystal structure generation with autoregressive large language modeling},
  author={Antunes, Luis M. and Butler, Keith T. and Grau-Crespo, Ricardo},
  journal={Nature Communications},
  volume={15},
  pages={10570},
  year={2024}
}

@article{wang2024catalm,
  title={{CataLM}: Empowering catalyst design through large language models},
  author={Wang, Ludi and Chen, Xueqing and Du, Yi and Zhou, Yuanchun and Gao, Yang and Cui, Wenjuan},
  journal={arXiv preprint arXiv:2405.17440},
  year={2024}
}

@article{hu2025molmamba,
  title={MOL-Mamba: Enhancing Molecular Representation with Structural \& Electronic Insights},
  author={Hu, Jingjing and Guo, Dan and Si, Zhan and Liu, Deguang and Diao, Yunfeng and Zhang, Jing and Zhou, Jinxing and Wang, Meng},
  journal={arXiv preprint arXiv:2412.16483},
  year={2025}
}

@misc{orbitalmaterials2025orb,
  title={Orb},
  author={Orbital Materials},
  year={2025},
  note={Research page},
  url={https://orbitalmaterials.com/research#orb}
}

@article{frey2023neural,
  title={Neural scaling of deep chemical models},
  author={Frey, Nathan C. and Soklaski, Ryan and Axelrod, Simon and Samsi, Siddharth and G{\'o}mez-Bombarelli, Rafael and Coley, Connor W. and Gadepally, Vijay},
  journal={Nature Machine Intelligence},
  volume={5},
  pages={1297--1305},
  year={2023}
}

@misc{nvidia2022megamolbart,
  title={MegaMolBART},
  author={NVIDIA},
  year={2022},
  note={GitHub repository. Release: Aug 30, 2022. Accessed: 2025-08-19},
  url={https://github.com/NVIDIA/MegaMolBART}
}

@misc{internlm2025interns1,
  title={Intern-S1},
  author={InternLM},
  year={2025},
  note={Model card},
  url={https://huggingface.co/internlm/Intern-S1}
}

@inproceedings{besta2024graph,
  title        = {Graph of Thoughts: Solving Elaborate Problems with Large Language Models},
  author       = {Besta, Micha{\l} and others},
  booktitle    = {Proceedings of the AAAI Conference on Artificial Intelligence},
  volume       = {38},
  number       = {16},
  pages        = {17682--17690},
  year         = {2024}
}

@inproceedings{yao2024effectivegot,
  title        = {Beyond Chain-of-Thought: Effective Graph-of-Thought Reasoning in Language Models},
  author       = {Yao, Yao and Li, Zuchao and Zhao, Hai},
  booktitle    = {Findings of the Association for Computational Linguistics: NAACL 2024},
  pages        = {2901--2921},
  address      = {Mexico City, Mexico},
  publisher    = {Association for Computational Linguistics},
  year         = {2024}
}

@article{yao2022react,
  title        = {ReAct: Synergizing Reasoning and Acting in Language Models},
  author       = {Yao, Shunyu and Zhao, Jeffrey and Yu, Dian and Du, Nan and Shafran, Izhak and Narasimhan, Karthik},
  journal      = {arXiv preprint arXiv:2210.03629},
  year         = {2022}
}

@article{shinn2023reflexion,
  title        = {Reflexion: Language Agents with Verbal Reinforcement Learning},
  author       = {Shinn, Noah and Cassano, Federico and Berman, Edward and Gopinath, Ashwin and Narasimhan, Karthik and Yao, Shunyu},
  journal      = {arXiv preprint arXiv:2303.11366},
  year         = {2023}
}

@article{madaan2023selfrefine,
  title        = {Self-Refine: Iterative Refinement with Self-Feedback},
  author       = {Madaan, Aman and Tandon, Niket and Gupta, Prakhar and Hallinan, Skyler and Gao, Luyu and Wiegreffe, Sarah and Alon, Uri and Dziri, Nouha and Prabhumoye, Shrimai and Yang, Yiming and Gupta, Shashank and Majumder, Bodhisattwa Prasad and Hermann, Katherine and Welleck, Sean and Yazdanbakhsh, Amir and Clark, Peter},
  journal      = {arXiv preprint arXiv:2303.17651},
  year         = {2023}
}

@article{pan2024weblinx,
  title        = {WebLINX: Real-World Website Navigation with Multi-Turn Dialogue},
  author       = {Pan, Y. and others},
  journal      = {arXiv preprint arXiv:2402.05930},
  year         = {2024}
}

@article{zhou2023webarena,
  title        = {WebArena: A Realistic Web Environment for Building Autonomous Agents},
  author       = {Zhou, S. and others},
  journal      = {arXiv preprint arXiv:2307.13854},
  year         = {2023}
}

@article{wei2025browsecomp,
  title   = {BrowseComp: A Simple Yet Challenging Benchmark for Browsing Agents},
  author  = {Wei, J. and others},
  journal = {arXiv preprint arXiv:2504.12516},
  year    = {2025}
}

@article{lesellier2024browsergym,
  title   = {The BrowserGym Ecosystem for Web Agent Research},
  author  = {Le Sellier De Chezelles, Thibault and others},
  journal = {arXiv preprint arXiv:2412.05467},
  year    = {2024}
}

@inproceedings{asai2024selfrag,
  title     = {Self-RAG: Learning to Retrieve, Generate, and Critique through Self-Reflection},
  author    = {Asai, Akari and others},
  booktitle = {Proceedings of the Twelfth International Conference on Learning Representations (ICLR)},
  year      = {2024},
  url       = {https://openreview.net/forum?id=6OIt3xVOUJ}
}

@article{yan2024crag,
  title   = {Corrective Retrieval Augmented Generation},
  author  = {Yan, S.-Q. and others},
  journal = {arXiv preprint arXiv:2401.15884},
  year    = {2024}
}

@article{peng2024graphragsurvey,
  title   = {Graph Retrieval-Augmented Generation: A Survey},
  author  = {Peng, B. and others},
  journal = {arXiv preprint arXiv:2408.08921},
  year    = {2024}
}

@misc{langchain2025localdeepresearcher,
  author       = {{LangChain}},
  title        = {Local Deep Researcher},
  year         = {2025},
  howpublished = {\url{https://github.com/langchain-ai/local-deep-researcher}},
  note         = {GitHub repository. Version pinned at commit 8a5e220 (2025-08-08)}
}

@article{xie2018cgcnn,
  title={Crystal Graph Convolutional Neural Networks for an Accurate and Interpretable Prediction of Material Properties},
  author={Xie, Tian and Grossman, Jeffrey C.},
  journal={Physical Review Letters},
  volume={120},
  number={14},
  pages={145301},
  year={2018},
  doi={10.1103/PhysRevLett.120.145301}
}

@article{severson2019cyclelife,
  title={Data-driven prediction of battery cycle life before capacity degradation},
  author={Severson, Kristen A. and Attia, Peter M. and Jin, Norman and Perkins, Nicholas and Jiang, Benben and Yang, Zi and Chen, Michael H. and Aykol, Muratahan and Herring, Patrick K. and Fraggedakis, Dimitrios and Bazant, Martin Z. and Harris, Stephen J. and Chueh, William C. and Braatz, Richard D.},
  journal={Nature Energy},
  volume={4},
  number={5},
  pages={383--391},
  year={2019},
  doi={10.1038/s41560-019-0356-8}
}

@article{szymanski2023alab,
  title={An autonomous laboratory for the accelerated synthesis of novel materials},
  author={Szymanski, Nathan J. and Rendy, Bernardus and Fei, Yuxing and Kumar, Rishi E. and He, Tanjin and Milsted, David and McDermott, Matthew J. and Gallant, Max and Cubuk, Ekin D. and Merchant, Amil and Kim, Haegyeom and Jain, Anubhav and Bartel, Christopher J. and Persson, Kristin and Zeng, Yan and Ceder, Gerbrand},
  journal={Nature},
  volume={624},
  pages={86--91},
  year={2023},
  doi={10.1038/s41586-023-06734-w}
}

@article{wang2022aptamer_gfet,
  title={Ultrasensitive Antibiotic Perceiving Based on Aptamer-Functionalized Ultraclean Graphene Field-Effect Transistor Biosensor},
  author={Wang, Shun and Sun, Mingyuan and Zhang, Yunhong and Ji, Hao and Gao, Jianwei and Song, Shuai and Sun, Jun and Liu, Hong and Zhang, Yu and Han, Lin},
  journal={Analytical Chemistry},
  volume={94},
  number={42},
  pages={14785--14793},
  year={2022},
  doi={10.1021/acs.analchem.2c03732}
}

@article{lin2018ti4o7,
  title={Development of macroporous Magn{\'e}li phase Ti4O7 ceramic materials: As an efficient anode for mineralization of poly- and perfluoroalkyl substances},
  author={Lin, Hui and Niu, Junfeng and Liang, Shangtao and Wang, Chong and Wang, Yujuan and Jin, Fangyuan and Luo, Qi and Chiang, Szu-Yin and Huang, Qingguo},
  journal={Chemical Engineering Journal},
  volume={354},
  pages={1058--1067},
  year={2018}
}

@article{zhong2021pvdf,
  title={Binding mechanisms of PVDF in lithium ion batteries},
  author={Zhong, Xuehu and Han, Junwei and Chen, Lingling and Liu, Wei and Jiao, Fei and Zhu, Hong and Qin, Wenqing},
  journal={Applied Surface Science},
  volume={553},
  pages={149564},
  year={2021}
}

@article{ding2025oermulti,
  title={Leveraging data mining, active learning, and domain adaptation for efficient discovery of advanced oxygen evolution electrocatalysts},
  author={Ding, Rui and Liu, Jianguo and Hua, Kang and Wang, Xuebin and Zhang, Xiaoben and Shao, Minhua and Chen, Yuxin and Chen, Junhong},
  journal={Science Advances},
  volume={11},
  number={14},
  pages={eadr9038},
  year={2025}
}

@article{karthik2023gc3n4co2,
  title={CO2 gas sensing properties of graphitic carbon nitride (g-C3N4) thin films},
  author={Karthik, T. V. K. and Mart{\'\i}nez-Garc{\'\i}a, H. and Ortiz-Chi, F. and Espinosa-Gonz{\'a}lez, C. G. and Torres-Torres, J. G. and Hernandez, A. G. and Godavarthi, S. and Kesarla, M. K.},
  journal={Diamond and Related Materials},
  volume={133},
  pages={109736},
  year={2023}
}

@article{lacalamita2025cyclodextrins,
  title={Cyclodextrins for the removal of per- and polyfluoroalkyl substances: a review},
  author={Lacalamita, Dario and Mongiov{\'\i}, Chiara and Crini, Gr{\'e}gorio and Morin-Crini, Nadia},
  journal={Environmental Chemistry Letters},
  volume={23},
  pages={1713--1743},
  year={2025}
}

@article{zhu2025betacdmip,
  title={\ensuremath{\beta}-cyclodextrin imprinted film embedded with methylene blue: A host-guest sensitive electrochemical strategy for PFAS detection},
  author={Zhu, Xiaoyu and Cheng, Chen and Qin, Xingao and Wang, Ying},
  journal={Journal of Hazardous Materials},
  volume={485},
  pages={136870},
  year={2025}
}

@article{wang2022tunable,
  title={A tunable porous \ensuremath{\beta}-cyclodextrin polymer platform to understand and improve anionic PFAS removal},
  author={Wang, Ri and Lin, Zhi-Wei and Klemes, Max J. and Ateia, Mohamed and Trang, Brittany and Wang, Jieyuan and Ching, Casey and Helbling, Damian E. and Dichtel, William R.},
  journal={ACS Central Science},
  volume={8},
  pages={663--669},
  year={2022}
}

@article{pitruzzella2024mippof,
  title={An MIP-based PFAS sensor exploiting nanolayers on plastic optical fibers for ultra-wide and ultra-low detection ranges---A case study of PFAS detection in river water},
  author={Pitruzzella, Rosalba and Chiodi, Alessandro and Rovida, Riccardo and Arcadio, Francesco and Porto, Giovanni and Moretti, Simone and Brambilla, Gianfranco and Zeni, Luigi and Cennamo, Nunzio},
  journal={Nanomaterials},
  volume={14},
  pages={1764},
  year={2024}
}

@article{liu2025pfoa,
  title={Highly sensitive determination of perfluorooctanoic acid in food and river samples with an electrochemical platform based on MIP and modified MWCNTs},
  author={Liu, Ting and Sang, Chunyan and Sun, Bolu and Cai, Hua and Liu, Hong and Liu, Yuwei and He, Gengsheng},
  journal={Microchemical Journal},
  volume={208},
  pages={112328},
  year={2025}
}

@misc{chemicalbook2025pots,
  title={1H,1H,2H,2H-Perfluorooctyltriethoxysilane},
  author={{ChemicalBook}},
  howpublished={\url{https://www.chemicalbook.com/ChemicalProductProperty_EN_CB9474048.htm}},
  note={CAS No. 51851-37-7. Accessed: 2025-11-21},
  year={2025}
}

@article{liu2016superhydrophobic,
  title={Super-Hydrophobic/Icephobic Coatings Based on Silica Nanoparticles Modified by Self-Assembled Monolayers},
  author={Liu, Junpeng and Janjua, Zaid A. and Roe, Martin and Xu, Fang and Turnbull, Barbara and Choi, Kwing-So and Hou, Xianghui},
  journal={Nanomaterials},
  volume={6},
  number={12},
  pages={232},
  year={2016}
}

@article{cameron2020pedotpss,
  title={The damaging effects of the acidity in PEDOT:PSS on semiconductor device performance and solutions based on non-acidic alternatives},
  author={Cameron, Joseph and Skabara, Peter J.},
  journal={Materials Horizons},
  volume={7},
  pages={1759--1772},
  year={2020}
}

@article{kumar2023mxenestability,
  title={Long-Term Stability Studies and Applications of Ti3C2Tx MXene},
  author={Kumar, Sunil and Park, Hyun Min and Singh, Tej and Kumar, Manjeet and Seo, Yongho},
  journal={International Journal of Energy Research},
  volume={2023},
  pages={5275439},
  year={2023}
}

@article{luo2023ultrasound,
  title={Ultrasound-enhanced Magn{\'e}li phase Ti$_4$O$_7$ anodic oxidation of per- and polyfluoroalkyl substances (PFAS) towards remediation of aqueous film forming foams (AFFF)},
  author={Luo, Yunlong and Khoshyan, Ashkan and Al Amin, Md. and Nolan, Annette L. and Robinson, Fiona and Fenstermacher, Jim and Niu, Junfeng and Megharaj, Mallavarapu and Naidu, Ravi and Fang, Cheng},
  journal={Science of the Total Environment},
  volume={862},
  pages={160836},
  year={2023},
  doi={10.1016/j.scitotenv.2022.160836}
}

@article{gomri2025eaop_efenton_pfas,
  title={Electrochemical advanced oxidation combined to electro-Fenton for effective treatment of perfluoroalkyl substances ``PFAS'' in water using a Magn{\'e}li phase-based anode},
  author={Gomri, Chaimaa and Makhoul, Elissa and Koundia, Fatou Niang and Petit, Eddy and Raffy, Stephane and Bechelany, Mikhael and Semsarilar, Mona and Cretin, Marc},
  journal={Nanoscale Advances},
  volume={7},
  pages={261--268},
  year={2025},
  doi={10.1039/D4NA00626G}
}

@article{ullah2023znfe2o4_update,
  title={Semiconductor ZnFe$_2$O$_4$ as Efficient Photocatalyst for the Degradation of Organic Dyes: An Update},
  author={Ullah, Rabid},
  journal={Journal of Chemical Reviews},
  volume={5},
  number={4},
  pages={466--476},
  year={2023},
  doi={10.48309/JCR.2023.407658.1232}
}

@article{mandari2025gcn_mxene_review,
  title={g-C$_3$N$_4$/MXene-based heterostructures: advanced catalysts for efficient and sustainable renewable energy production},
  author={Mandari, Kotesh Kumar and Kang, Misook},
  journal={Advances in Industrial and Engineering Chemistry},
  volume={1},
  pages={14},
  year={2025},
  doi={10.1007/s44405-025-00014-z}
}

@article{bagnall2025al2o3_stability,
  title={Re-Evaluating the Stability of Al$_2$O$_3$ Barriers Prepared by Atomic Layer Deposition under Electrochemical Conditions},
  author={Bagnall, Andrew J. and Zhao, Ziwen and Cheah, Mun Hon and Sekretareva, Alina},
  journal={ACS Applied Materials \& Interfaces},
  volume={17},
  pages={48320--48333},
  year={2025},
  doi={10.1021/acsami.5c11388}
}

@article{goul2019ald_al2o3_tunnel,
  title={Electron tunneling properties of Al$_2$O$_3$ tunnel barrier made using atomic layer deposition in multilayer devices},
  author={Goul, Ryan and Wilt, Jamie and Acharya, Jagaran and Liu, Bo and Ewing, Dan and Casper, Matthew and Stramel, Alex and Elliot, Alan and Wu, Judy Z.},
  journal={AIP Advances},
  volume={9},
  pages={025018},
  year={2019},
  doi={10.1063/1.5052163}
}

@article{miret2025llms_materials_discovery,
  title={Are LLMs Ready for Real-World Materials Discovery?},
  author={Miret, Santiago and Krishnan, N. M. Anoop and others},
  journal={Nature Machine Intelligence},
  year={2025},
  doi={10.1038/s42256-025-01058-y}
}

@article{kalai2025hallucinate,
  title={Why Language Models Hallucinate},
  author={Kalai, Adam Tauman and Nachum, Ofir and Vempala, Santosh S. and Zhang, Edwin},
  journal={arXiv preprint arXiv:2509.04664},
  year={2025}
}


\clearpage
\begin{appendices}
\noindent 
{\Large \textbf{Table of Contents}}

\vspace{1ex}

\appendixtableofcontents

\clearpage
\section{Related Work in Physics ML Surrogates, Domain LLMs, and Agentic Systems}\label{app:related}

\subsection{Physics-aligned surrogates and classical ML}
Physics-guided predictive surrogates now operate at web-scale, with graph networks and
equivariant models trained on massive curated corpora.
For crystalline stability, GNoME trains large graph models and active-learning pipelines to
propose millions of candidate structures and flag hundreds of thousands as likely stable—establishing
that data+scale can systematically widen the known inorganic universe \cite{merchant2023gnome}.
In catalysis, the Open Catalyst efforts formalized realistic workloads: OC20 introduced 1.28M DFT
relaxations and 265M single-point evaluations for adsorbate–surface energetics and created
standardized tasks (IS2RE/IS2RS/S2EF) with stringent generalization splits; OC22 extended to
oxide electrocatalysts relevant to OER with more realistic catalysts and
targets \cite{chanussot2021oc20,tran2023oc22}.
New foundation-style datasets push further: OMat24 aggregates >110M DFT entries and releases
strong baselines (e.g., Equiformer-V2) that achieve near-chemical-accuracy on multiple
targets \cite{barrosoluque2024omat24}; OMol25 analogously scales molecular DFT data to >100M
calculations spanning tens of millions of unique molecules, enabling broad pretraining for
small-molecule physics \cite{levine2025omol25}.
Complementing datasets, foundation models like the universal model of atoms (UMA) and Orb explore single architectures that transfer across chemistry and tasks, aiming for cross-domain generalization with shared tokenization/featurization of atoms and local environments~\cite{wood2025uma,orbitalmaterials2025orb}.
These surrogates excel at S1–S2 targets (bulk stability, adsorption energies, local structural
relaxations) but are largely D1–D2 in operation: fixed-form predictors trained on equilibrium data,
with limited capacity to reason about kinetics, synthesis windows, or device-level constraints that
emerge at S3–S4.
The net effect is that they remain superb "atoms-to-properties" engines, yet they do not
autonomously help solve the long-horizon, multi-paper evidence assembly and process-aware planning
needed, from atomic scale to device scale
rationales \cite{merchant2023gnome,chanussot2021oc20,tran2023oc22,barrosoluque2024omat24,levine2025omol25,wood2025uma}.

\subsection{Domain LLMs (non-agent)}
Domain LLMs span three useful roles. (i) Materials-text models (MatSciBERT/MatBERT/MaterialBERT) are trained on materials corpora to power entity/relation extraction, document classification, and graph building; MatSciBERT outperforms SciBERT on named entity recognition (NER) and relation tasks across curated benchmarks, while MatBERT shows that domain-specific pretraining significantly lifts NER
F1 and sample efficiency in materials texts; MaterialBERT introduces a domain tokenizer/vocabulary tuned to materials literature, while Intern-S1 scales domain-specific reasoning capabilities~\cite{gupta2022matscibert,trewartha2022quantifying,yoshitake2022materialbert,internlm2025interns1}. (ii) Chemical/polymer LLMs target sequence-level prediction/generation: ChemBERTa established transformer baselines and scaling trends for SMILES; ChemBERTa-2 pushed to 77M SMILES and showed
consistent gains (and that multi-task regression pretraining can beat MLM (Masked Language Modeling) in transfer); ChemBERTa-3 contributes to open, end-to-end training infrastructure for chemical foundation models; MegaMolBART provides a large-scale generative sequence model optimized for molecular design, with MOL-Mamba introducing state-space models for efficient representation~\cite{chithrananda2020chemberta,ahmad2022chemberta2,singh2025chemberta3,nvidia2022megamolbart,hu2025molmamba}.
ChemGPT demonstrates generative scaling to >1B parameters with clear neural-scaling laws for chemical LLMs, clarifying how model/data scale trade off in practice \cite{frey2023neural}. In polymers, TransPolymer introduces a chemically-aware tokenizer and MLM pretraining that improves regression performances across ten benchmarks; polyBERT builds an end-to-end, ultrafast pipeline and screens 100M
hypothetical polymers; PolyNC fuses natural-language and chemical-sequence inputs to improve unified polymer property prediction \cite{xu2023transpolymer,kuenneth2023polybert,qiu2024polync}. (iii) Generative/design LLMs for materials include ChatMOF and the GPT-4 Reticular Chemist for MOF proposals~\cite{kang2024chatmof,zheng2023gpt4}, CataLM for catalyst design~\cite{wang2024catalm}; Crystal-LLM (autoregressive) generates crystal structures from composition, and MatterGen scales inorganic structure generation with validated discoveries and robust property control \cite{kang2024chatmof,antunes2024crystallm,zeni2025generative}. Collectively these models deliver flexible text/code/structure interfaces and strong S1–S2 performance—ideal for curation, candidate suggestion, or surrogate-assisted screening—but they are typically short-horizon
(D1–D2): they answer prompts, not conduct hours-long, tool-grounded, hierarchical inquiry across conflicting
literature \cite{yoshitake2022materialbert,trewartha2022quantifying,gupta2022matscibert,qiu2024polync,kuenneth2023polybert,xu2023transpolymer,chithrananda2020chemberta,ahmad2022chemberta2,singh2025chemberta3,frey2023neural,nvidia2022megamolbart,kang2024chatmof,zeni2025generative,antunes2024crystallm}.
\subsection{Agentic systems for scientific inquiry}\label{app:related:agent}
Agent frameworks aim to orchestrate end-to-end inquiry: decompose goals, retrieve and reconcile evidence, use domain tools/simulators, and iteratively critique/plan. Early exemplars used rule/graph planning and Bayesian loops within constrained scopes—Chematica/SYNTHIA operationalized expert retrosynthesis rules and heuristic search; CAMD framed autonomous design/make/test/learn for materials
screening \cite{merck_synthia_website,montoya2020camd}. LLM-based agents extend this with language-native decomposition and tool-use: ChemCrow connects models to chemistry tools (e.g., RDKit, databases) to complete realistic tasks~\cite{bran2024chemcrow}; HoneyComb provides a flexible materials-agent stack with retrieval/tooling and HPC hooks~\cite{zhang2024honeycomb}; SciAgents formalizes multi-agent graph reasoning for scientific planning~\cite{ghafarollahi2025sciagents}.
Recent hierarchical frameworks like El Agente introduce orchestration for complex quantum chemistry workflows~\cite{zou2025elagente}, while systems such as CheMatAgent, ChemAgent, MatAgent, and ChemHAS further specialize in tool learning and human-in-the-loop coordination~\cite{wu2025chematagent,tang2025chemagent,bazgir2025matagent,li2025chemhas}.
Finally, ``AI co-scientist'' frameworks couple planning with web/code tools and emphasize evaluation and provenance~\cite{gottweis2025towards}. Execution back-ends close the loop: A-Lab and RoboMapper demonstrate self-driving labs discovering and characterizing new materials~\cite{szymanski2023alab,wang2023sustainable}; IBM RXN/RoboRXN links digital retrosynthesis with automated synthesis~\cite{ibm2020roborxn}; and NVIDIA ALCHEMI targets scaled infrastructure (data, models, and accelerated simulation) for materials discovery~\cite{ong2025alchemi}. Despite progress, most agents today operate at D2–D3 depth and S1–S3 scope; they rarely sustain thousand-query retrieval runs, explicit topic-tree governance, or tight coupling to cross-domain process constraints that S3–S4 device problems demand.
This is precisely the niche for deep-research agents: to deliver hierarchical planning, multi-stage local+web retrieval at scale, targeted tool/simulator calls, and systematic self-critique with dense provenance until a decision-ready synthesis
emerges \cite{merck_synthia_website,montoya2020camd,bran2024chemcrow,zhang2024honeycomb,ghafarollahi2025sciagents,gottweis2025towards,szymanski2023alab,ibm2020roborxn,ong2025alchemi}.
\subsection{Gap to deep research at S3–S4/D4}
Despite existing progress, the transition to S3–S4 nano-device complexity poses a fundamental barrier. The combinatorial explosion across materials, interfaces, processing parameters, and operational constraints creates hypothesis spaces where decisive insights lie at the intersections of disparate literature, conflicting reports, and cross-domain phenomena. Crucially, long-horizon, hierarchical
evidence accumulation (D4) remains under-documented: many agents cap at D2–D3, with limited iterative loops, shallow coordination, and no explicit topic-tree control for breadth-then-depth synthesis. Commercial deep-research offerings introduced in 2025 enable multi-step inquiry but remain closed-source with limited controllability and local integration, reinforcing the need for open, on-prem
frameworks tailored to proprietary materials workflows. DToR addresses this by treating each research node as an evidence-grounded state in a resource-bounded tree search, bridging the symbolic reasoning of ToT/GoT with the domain knowledge grounding required for scientific discovery. To our knowledge, DToR is the first open, on-prem framework that explicitly couples local domain RAG with
gap-triggered web expansion within a resource-bounded tree to address S3–S4 materials discovery—a regime where neither pure surrogates nor closed commercial DR agents operate reliably.
\clearpage
\renewcommand{\thefigure}{S\arabic{figure}}
\setcounter{figure}{0}
\renewcommand{\thetable}{S\arabic{table}}
\setcounter{table}{0}
\clearpage
\section{Supplementary Figures}\label{app:supp_figures}

\begin{figure}[htbp]
  \centering
  \includegraphics[width=1\textwidth]{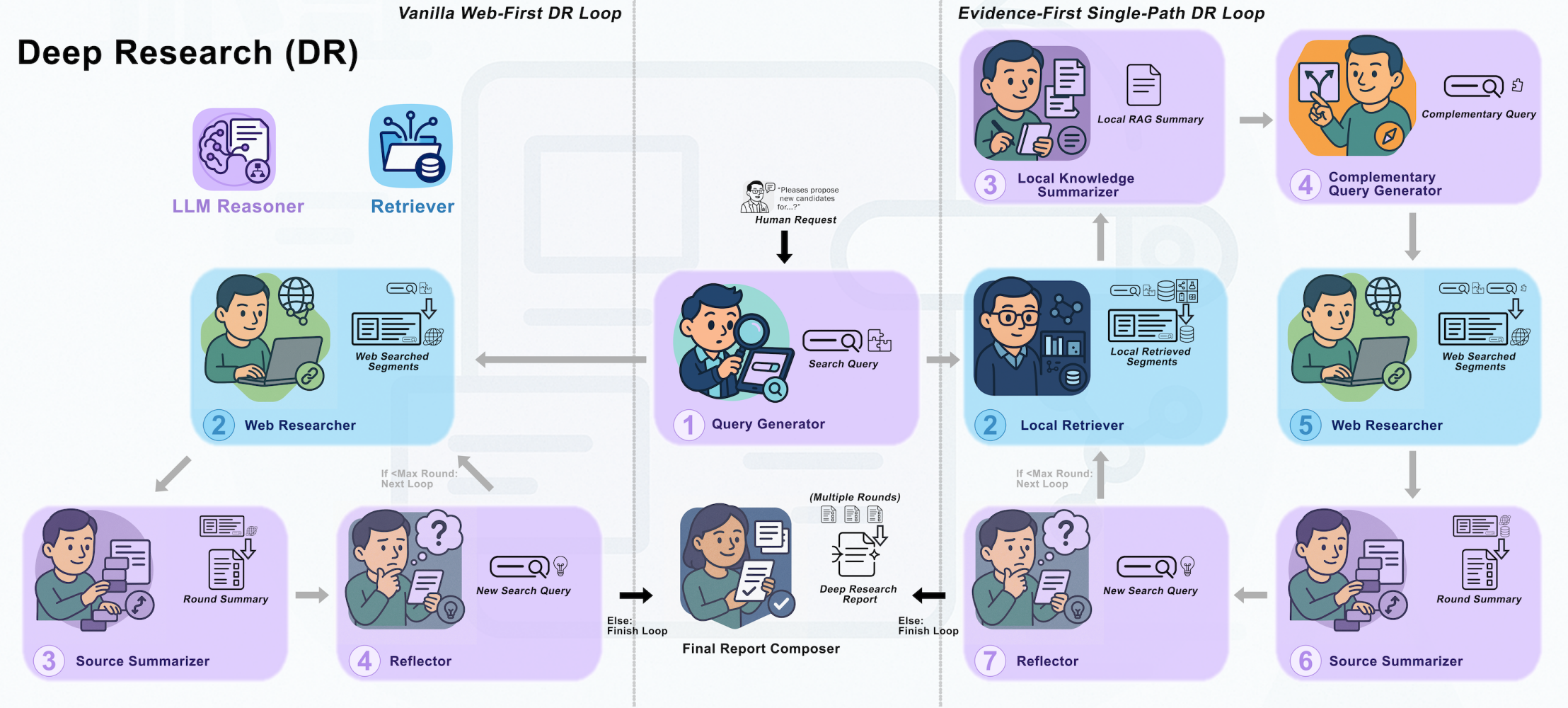}
  \caption{Schematic of the local evidence-first single DR instance in comparison with common web-first DR (Adapted from \protect\cite{langchain2025localdeepresearcher})}
  \label{fig:supp_1_single_DR_schematic}
\end{figure}

\begin{figure}[htbp]
  \centering
  \includegraphics[width=1\textwidth]{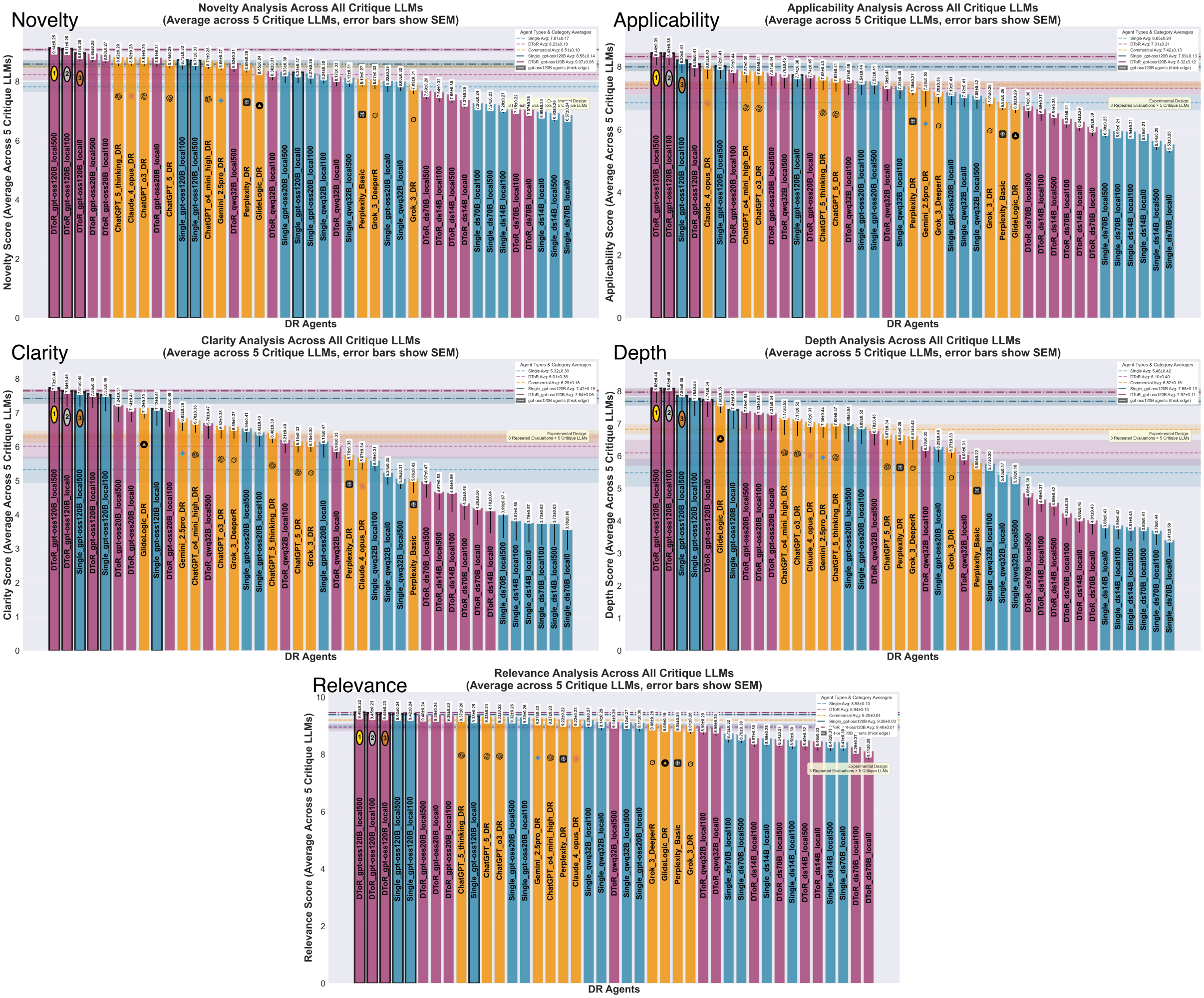}
  \caption{Overall rubric scores across 41 agents on different evaluation dimensions. }
  \label{fig:supp_3_each_dim_score}
\end{figure}

\begin{figure}[htbp]
  \centering
  \includegraphics[width=0.75\textwidth]{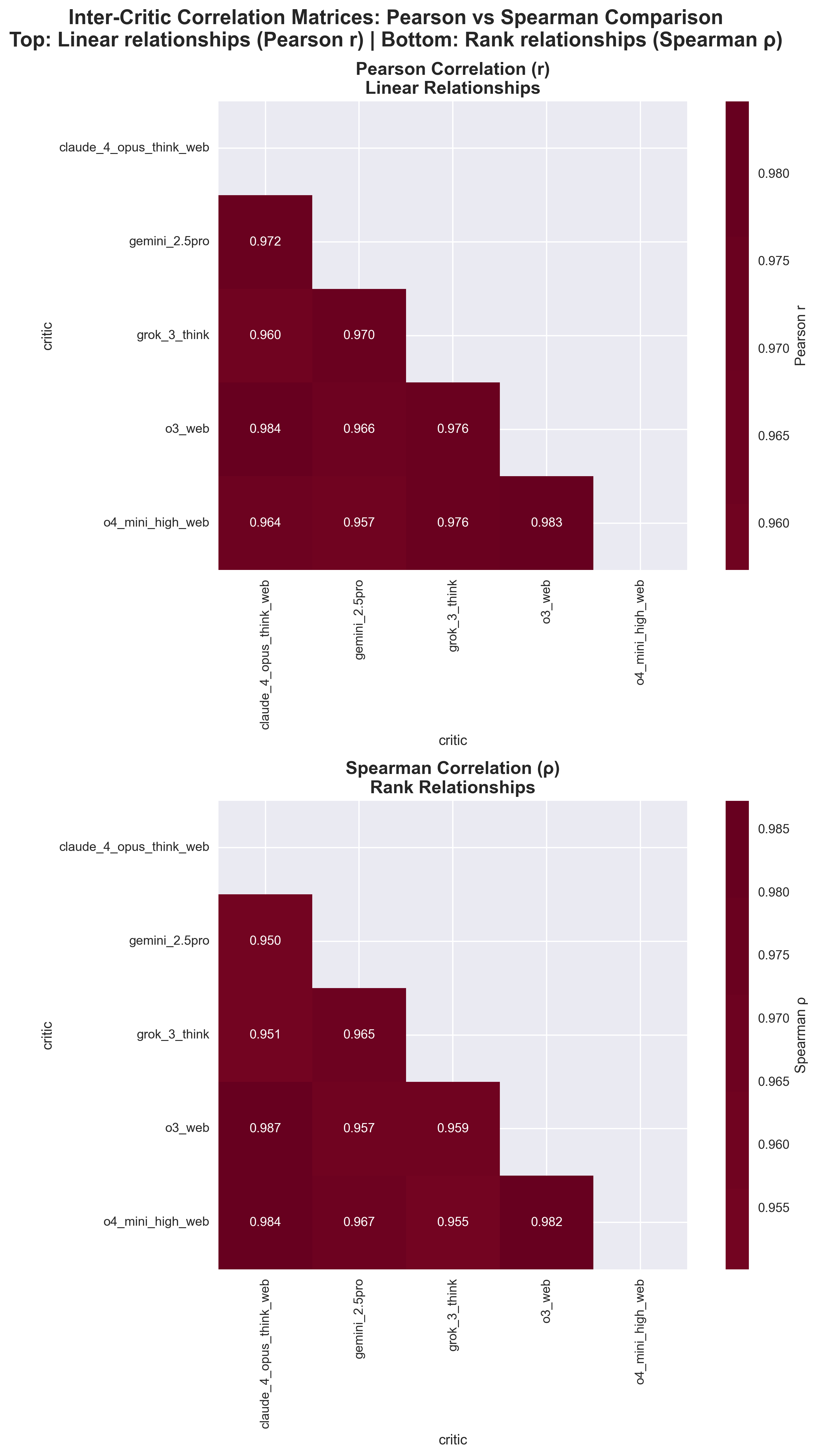}
  \caption{Rubric score agent agreement level over the five critic LLMs, computed by correlation factors. The average Pearson correlation is $\mathrm{p}=0.968$ on raw scores and Spearman correlation is $\rho=0.967$ on rank orders. }
  \label{fig:supp_4_corr_score_rank}
\end{figure}

\begin{figure}[p]
  \centering
  \includegraphics[width=0.65\textwidth]{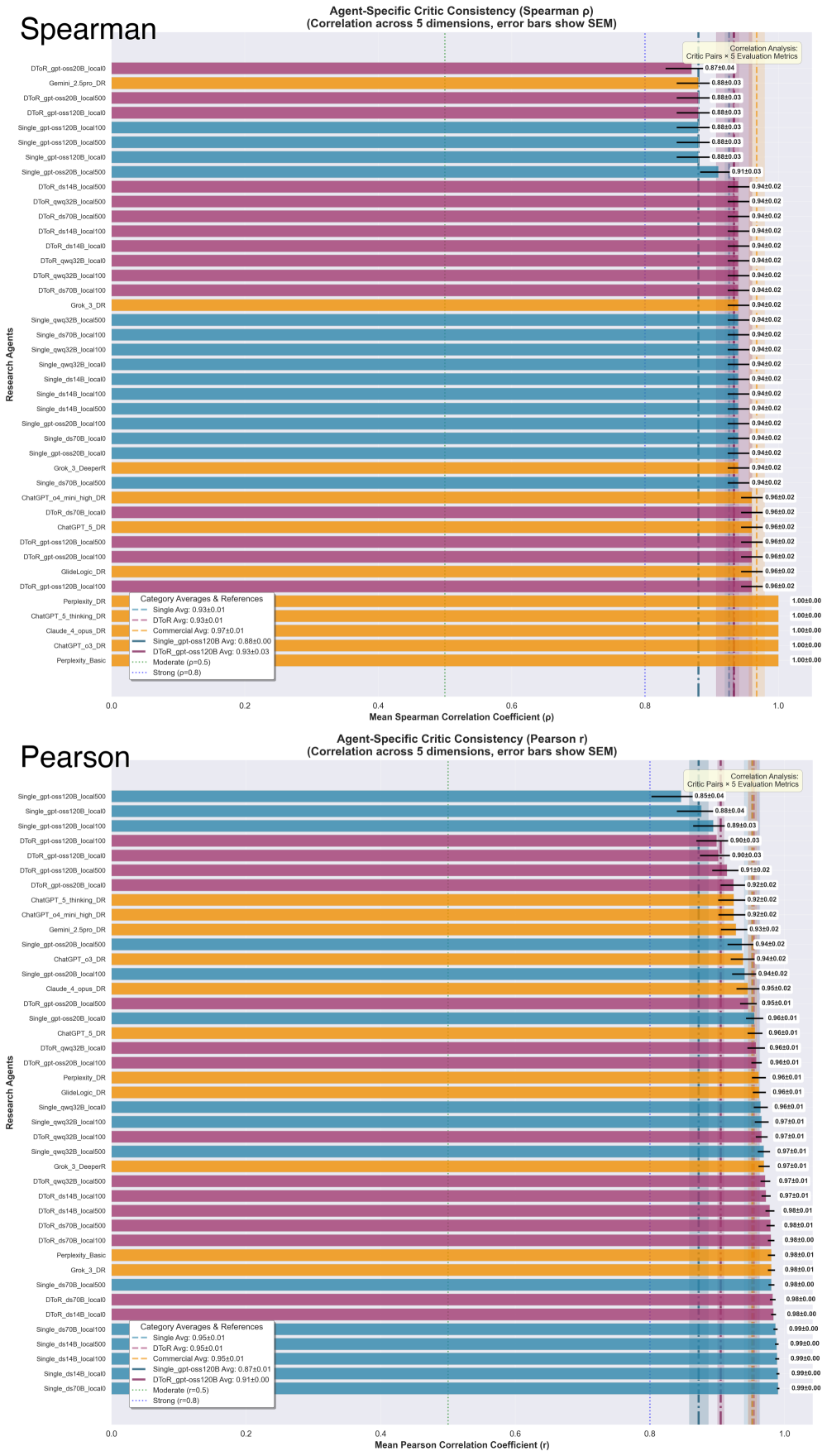}
  \caption{Agent-Specific Critic Consistency Analysis. Pearson correlation coefficients measuring linear agreement between critique LLMs across 5 evaluation dimensions (Relevance, Depth, Clarity, Applicability, Novelty) for each research agent. Spearman rank correlation coefficients measuring ordinal agreement between critics. Higher values indicate greater consensus among the 5 critique LLMs (Claude, Gemini, Grok, O3, O4) about each agent's relative performance strengths and weaknesses. Data aggregated from 27 research topics × 3 trials per agent. Perfect correlation ($\rho=1.00$) indicates all critics completely agree on the ranking of an agent's performance across the 5 dimensions.}
  \label{fig:supp_5_agent_specific_consistency}
\end{figure}

\begin{figure}[htbp]
  \centering
  \includegraphics[width=1\textwidth]{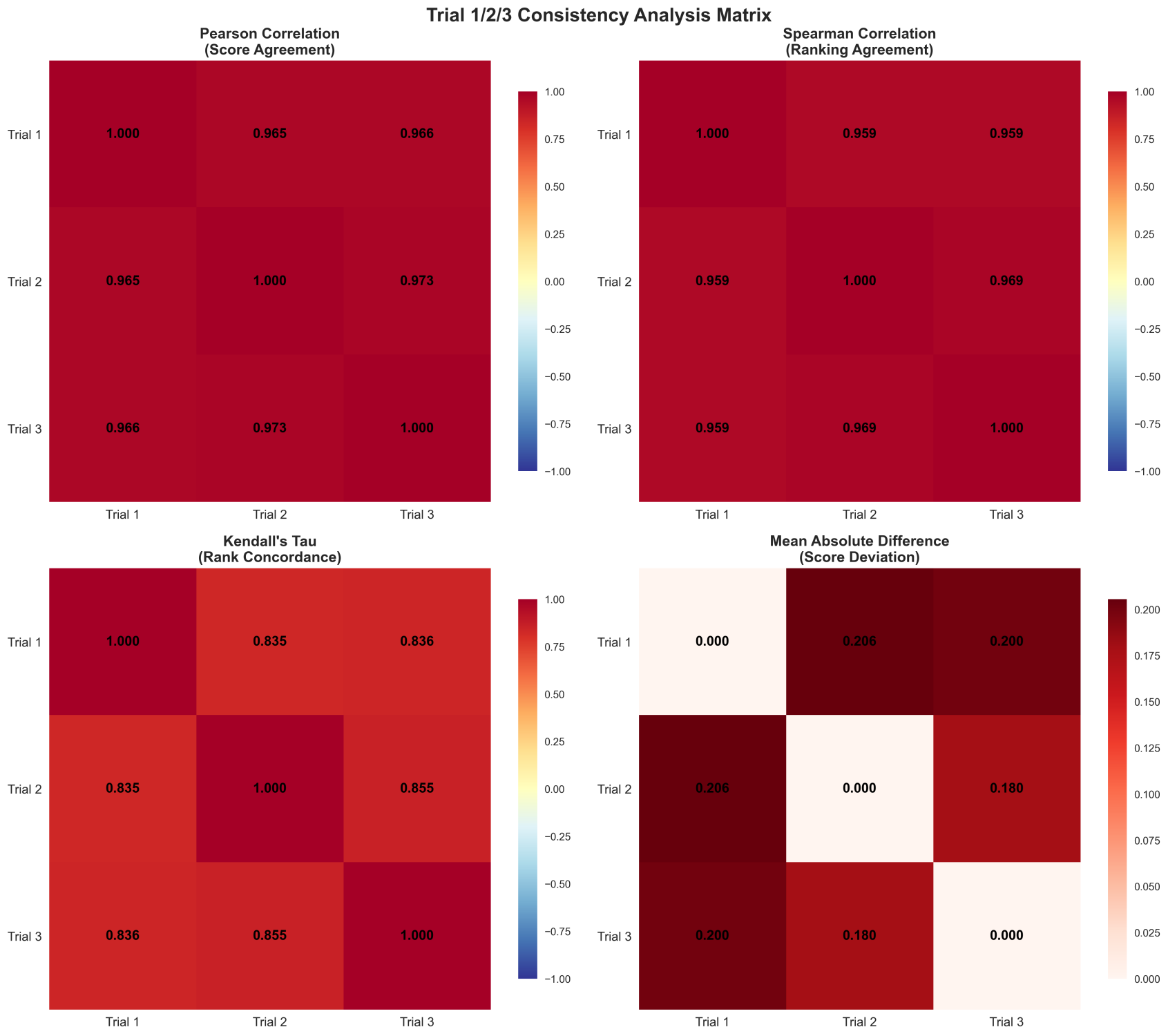}
  \caption{Trial-to-Trial Consistency Analysis Matrix. Four metrics assess reproducibility across three independent trials: (A) Pearson correlation measures score agreement, (B) Spearman correlation evaluates ranking consistency, (C) Kendall's tau provides alternative rank concordance, and (D) Mean absolute difference quantifies score deviations. Data aggregated across 27 topics, 41 agents, and 5 LLM critics. Higher correlations (>0.8) and lower deviations (<0.5) indicate excellent experimental reproducibility.}
  \label{fig:supp_5_dim_trial_consistency}
\end{figure}

\begin{figure}[htbp]
  \centering
  \includegraphics[width=0.7\textwidth]{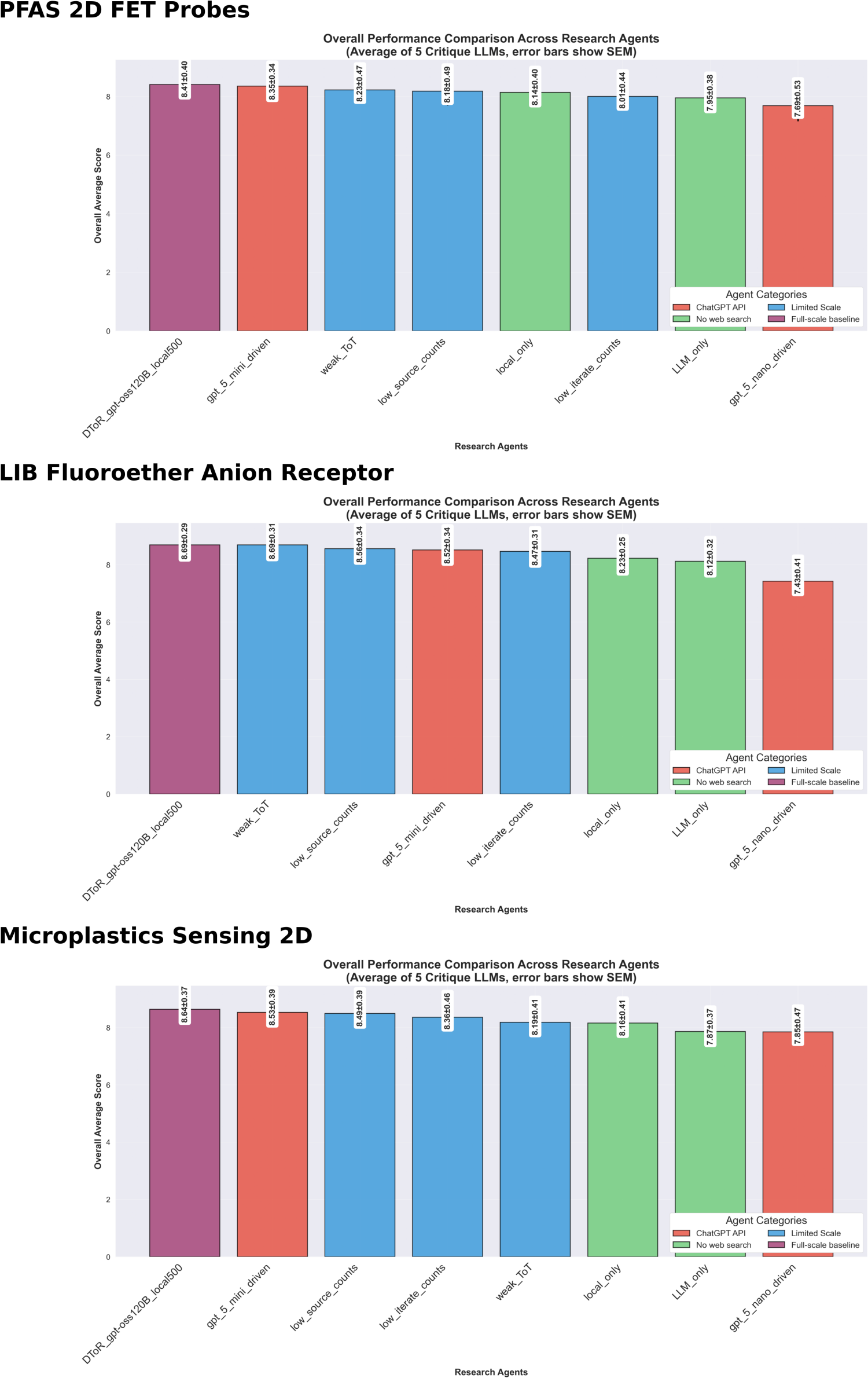}
  \caption{Targeted component ablations: Overall average scores (0-10 scale) for different research agent configurations across three scientific topics (LIB\_Fluoroether\_Anion\_Receptor, Microplastics\_Sensing\_2D, PFAS\_2D\_FET\_Probes), averaged over five critique LLMs with error bars showing SEM. Agent categories are distinguished by color: ChatGPT API (red), Full-scale baseline (purple), No web search (orange), and Limited Scale (blue/green). Dashed lines indicate baseline performance levels.}
  \label{fig:supp_11_add_ablation_5dim}
\end{figure}

\begin{figure}[htbp]
  \centering
  \includegraphics[width=0.75\textwidth]{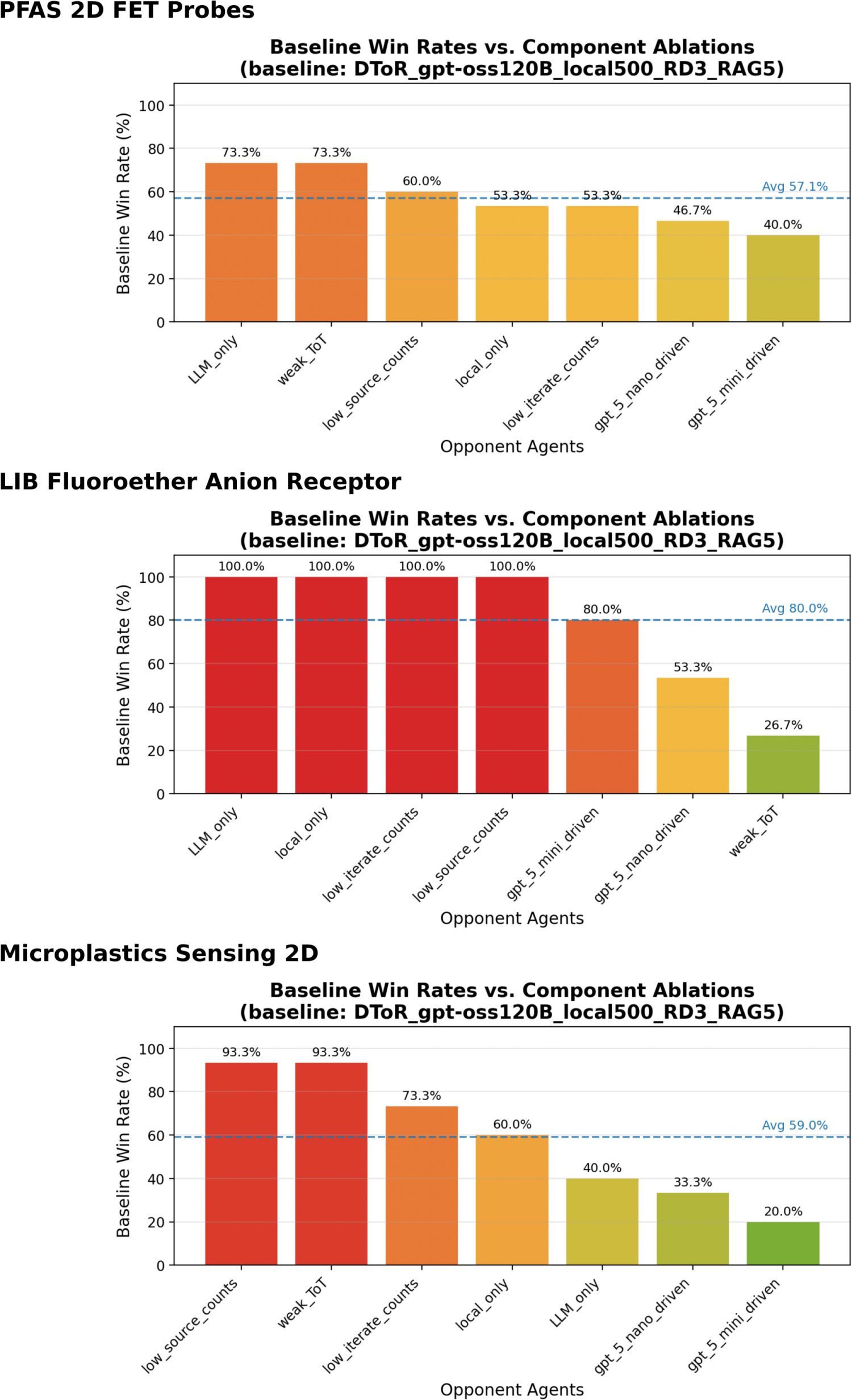}
  \caption{Targeted component ablations: Win rates of the full-scale baseline agent (DToR\_gpt-oss120B\_local500) competing against ablated opponents across three research topics. Dashed lines indicate average win rates for each topic.}
  \label{fig:supp_12_add_ablation_duel}
\end{figure}

\begin{figure}[htbp]
  \centering
  \includegraphics[width=\textwidth]{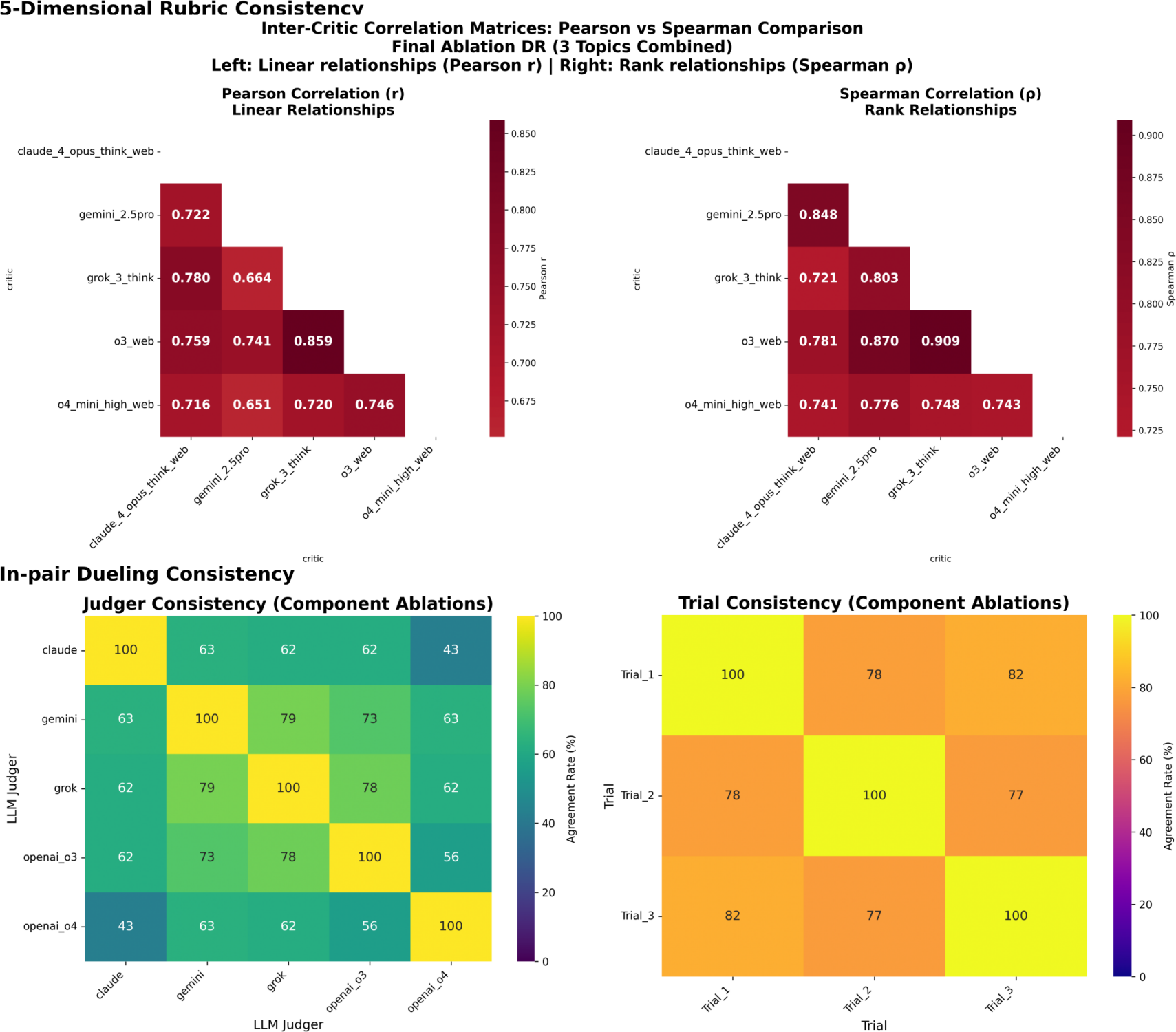}
  \caption{Targeted component ablations: Top: Inter-critic correlation matrices (Pearson and Spearman) showing agreement across five critique LLMs when evaluating research proposals on a 5-dimensional rubric across three topics. Bottom: In-pair dueling consistency heatmaps showing agreement rates across different LLM judgers (left) and experimental trials (right) in binary comparison decisions.}
  \label{fig:supp_13_add_ablation_consistency}
\end{figure}

\begin{figure}[htbp]
  \centering
  \includegraphics[width=1\textwidth]{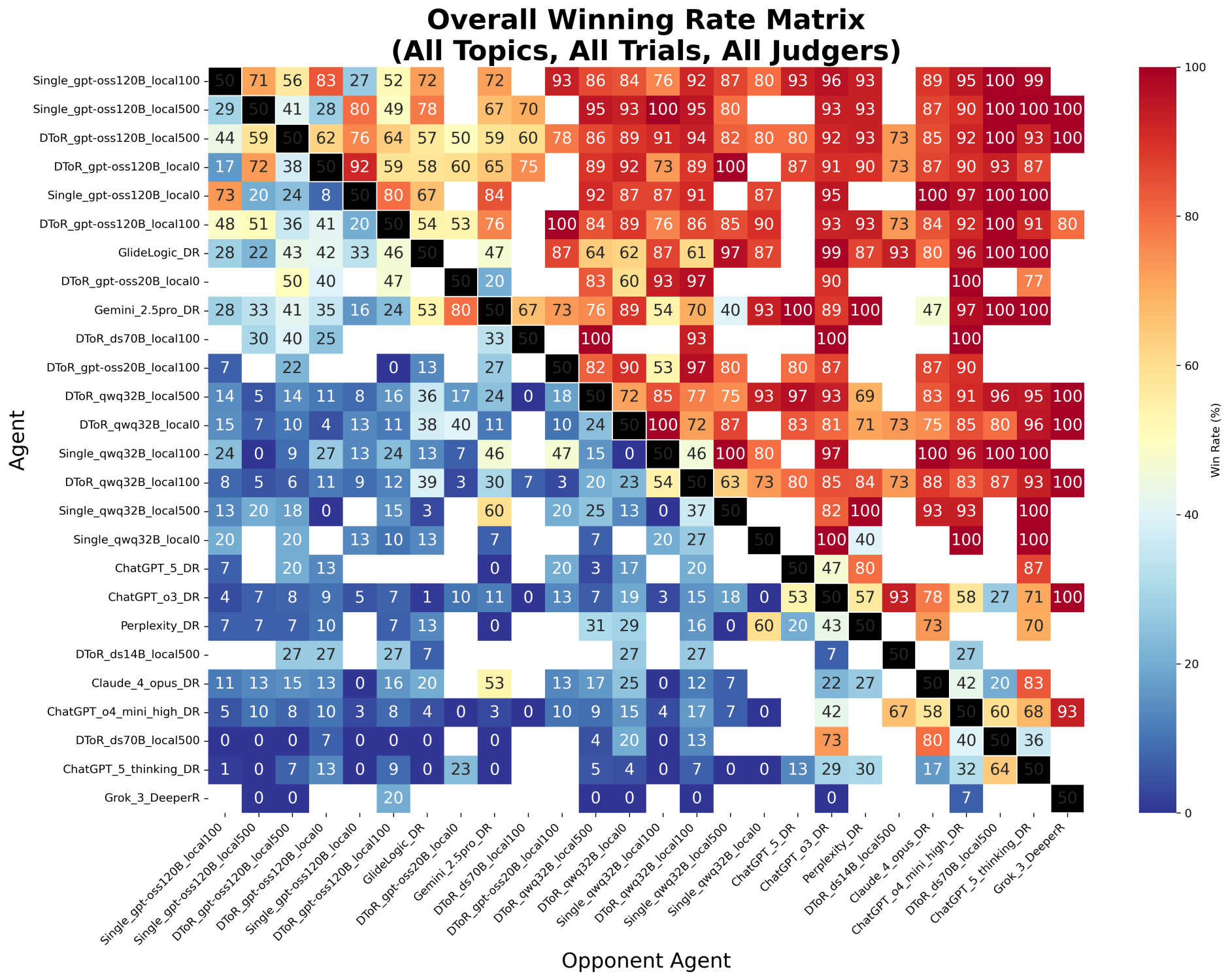}
  \caption{Overall Winning Matrix All Topics - Displays the comprehensive winning matrix with head-to-head performance comparisons between all agents across all research topics.}
  \label{fig:supp_6_duel_perf}
\end{figure}

\begin{figure}[htbp]
  \centering
  \includegraphics[width=1\textwidth]{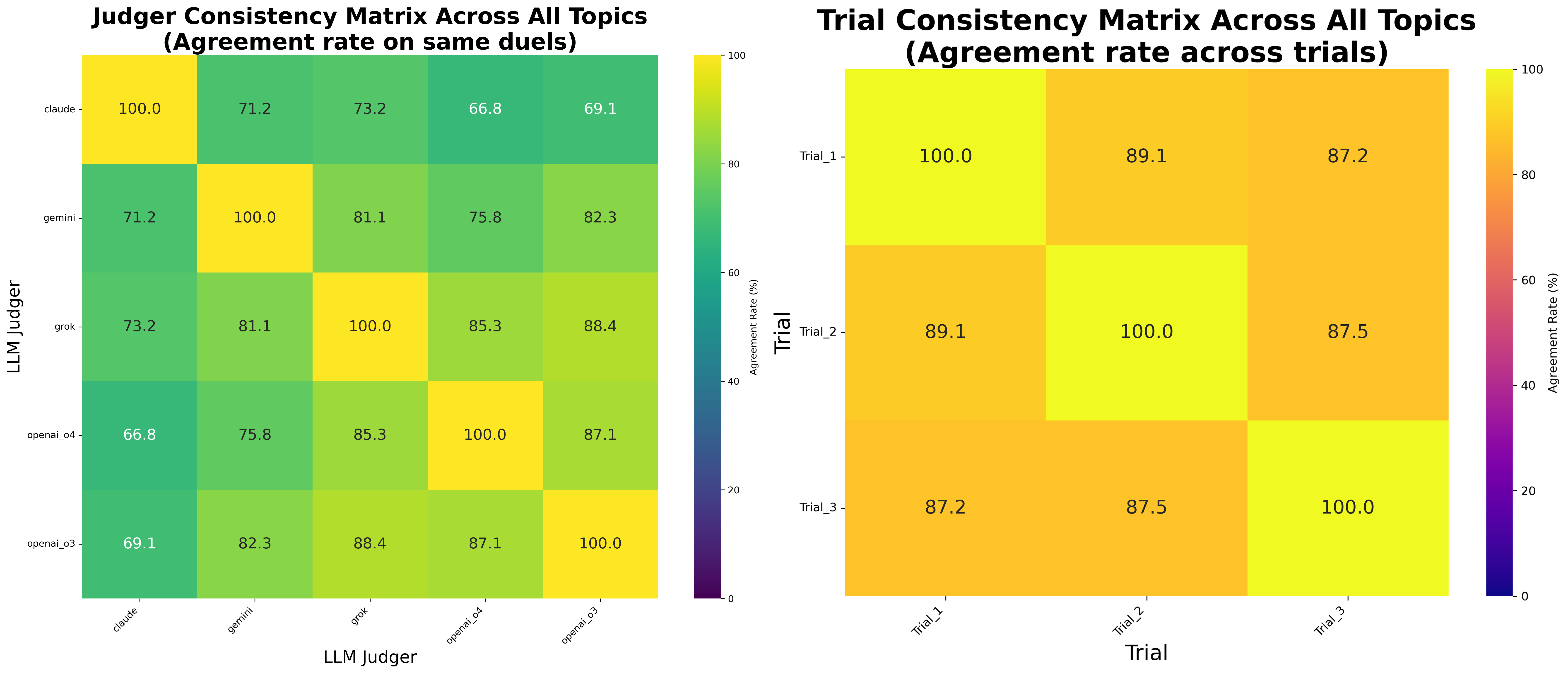}
  \caption{Consistency of in pair dueling. Left: Judger Consistency All Topics - Shows the consistency of judgment patterns across different research topics and agents, indicating how reliable the evaluation criteria are. Right: Trial Consistency All Topics - Displays the consistency of performance across multiple trials for each agent-topic combination, revealing the stability and reproducibility of results.}
  \label{fig:supp_3_duel_consistency}
\end{figure}

\begin{figure}[htbp]
  \centering
  \includegraphics[width=1\textwidth]{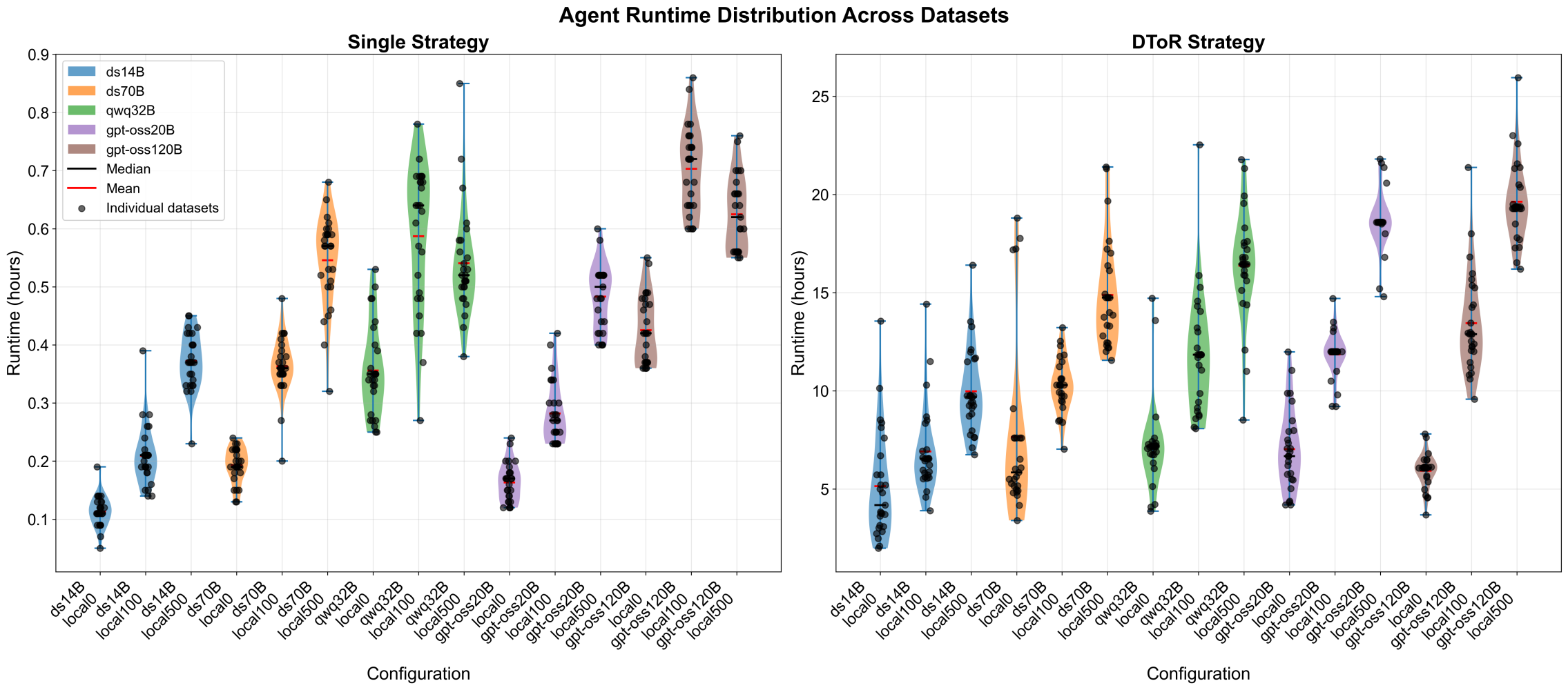}
  \caption{Violin plot visualizing the distribution of local DR agents runtime under different settings.}
  \label{fig:supp_7_runtime_violin}
\end{figure}

\begin{figure}[htbp]
  \centering
  \includegraphics[width=0.8\textwidth]{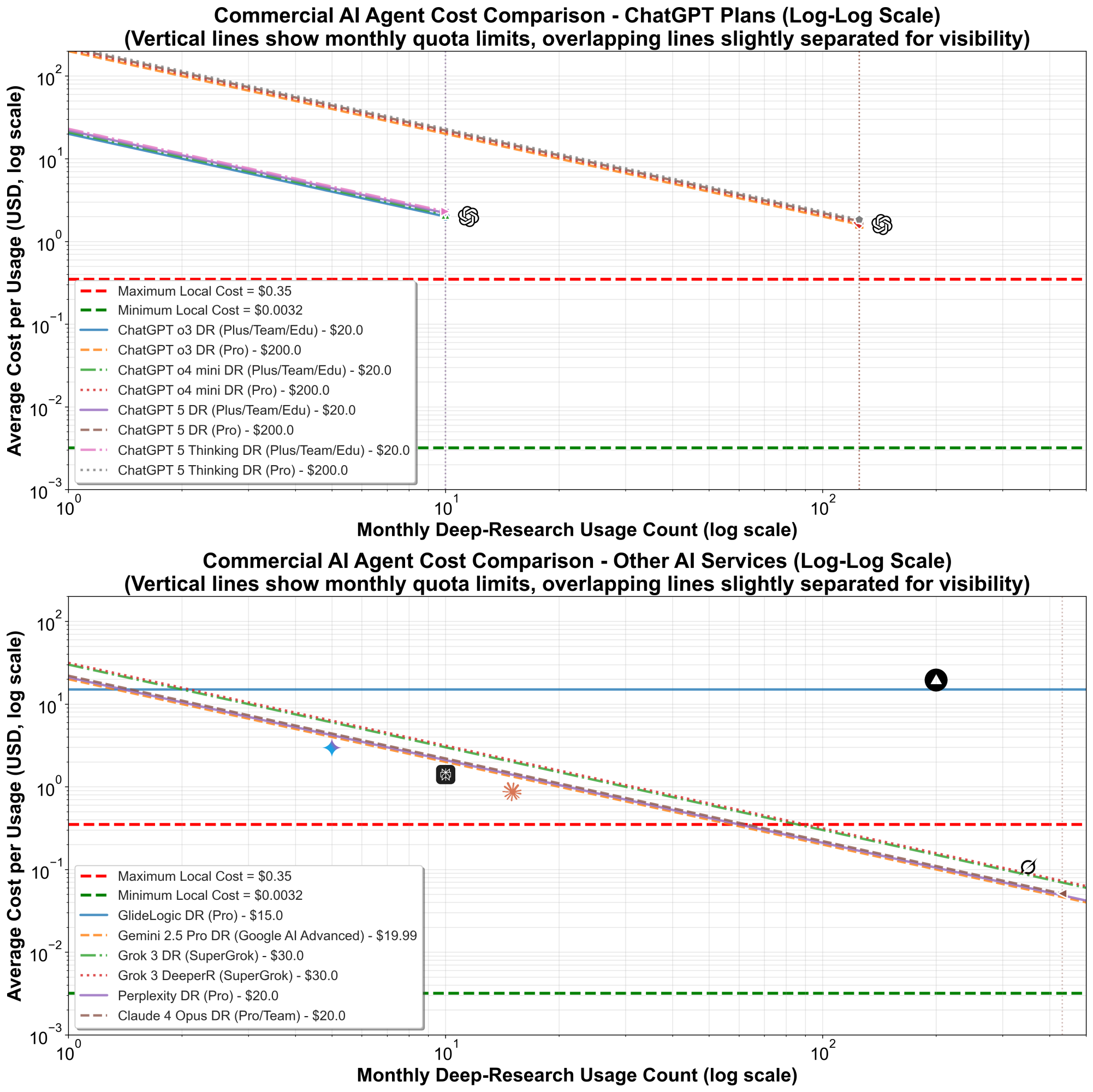}
  \caption{Comparison of cost over commercial DR agent solution with our local solution based on Chicago area electricity price October 2025. }
  \label{fig:supp_10_electricity_cost}
\end{figure}

\clearpage
\section{Nanomaterials and Device Topics}\label{app:topics}

\subsection{PFAS\_2D\_FET\_Probes}
Which two-dimensional (2D) nanomaterials or molecular probes—including novel, underexplored candidates—offer the greatest potential to achieve highly sensitive and selective detection of per- and polyfluoroalkyl substances (PFAS), or more broadly chemical and biological analytes, when integrated into a field-effect transistor (FET)–based sensor platform?

\medskip

\subsection{LIB\_Advanced\_Binders}
Which advanced binder technologies are being developed to improve the performance and longevity of lithium-ion batteries?

\medskip

\subsection{S\_aureus\_FET\_Commercial\_Probes}
How can emerging probe chemistries that are commercially available, cost-effective, and exhibit minimal batch-to-batch variability be used to develop a novel FET-based biosensor for detecting Staphylococcus aureus?

\medskip

\subsection{Wastewater\_Resource\_Recovery\_Cats}
Which nanostructured electrocatalyst materials demonstrate the highest selectivity and efficiency for electrochemical detection and recovery of critical resources (e.g., Li⁺, PO₄³⁻, NH₄⁺) from complex wastewater matrices, and what key performance metrics—such as selectivity, sensitivity, recovery rate, energy consumption, and operational stability—distinguish them?

\medskip

\subsection{OER\_Wastewater\_Resistant\_Cats}
Identify the top nanostructured electrocatalyst materials for driving the oxygen evolution reaction (OER) in complex wastewater matrices—such as high‐chloride, high‐organic‐load, or multi‐ion streams—, for each the material class/composition, comprehensively consider key performance metrics (overpotential at 10 mA cm⁻², Faradaic efficiency, stability), and the surface‐engineering strategies that
confer corrosion resistance and sustained activity in real effluent conditions.
\medskip

\subsection{Solar\_Evap\_Wastewater\_Photothermal}
What photothermal materials and system designs are most effective for solar-driven water evaporation in complex wastewater matrices—such as high-organic-load or multi-ion streams—and how do they compare in terms of material composition, solar-to-vapor conversion efficiency under one-sun illumination, evaporation rate, fouling resistance, and integrated resource-recovery functionalities?

\medskip

\subsection{Nutrient\_Sensors\_Anti\_Interference}
Which sensor probe materials/chemicals and designs offer the best performance in minimizing interference factors—such as competing ions, dissolved organic matter, pH fluctuations, and temperature variations—for accurate and selective detection of nutrients (e.g., nitrate, phosphate, ammonium) in complex water matrices, and what mitigation strategies do they employ?

\medskip

\subsection{CO2\_Sensing\_2D\_Materials}
Which two-dimensional materials—such as graphene derivatives, transition metal dichalcogenides, or MXenes—offer the highest CO₂ sensing performance in complex gas or aqueous environments, and how do they compare in terms of detection limit (ppm), selectivity against common interferents (e.g., O₂, H₂O), response/recovery time, and long-term stability, including any functionalization or structural
modifications that enhance these metrics? Think of novel candidates.
\medskip

\subsection{Printed\_FET\_Array\_Process\_Window}
Which printed‐electronics fabrication parameters and post‐processing strategies—such as ink viscosity, printing speed and resolution, substrate surface energy, annealing temperature profiles, and in‐line calibration methods—have been shown to minimize device‐to‐device variability in FET sensor arrays, and what specific process windows achieve low variation in threshold voltage and field‐effect
mobility?
\medskip

\subsection{2D\_Synaptic\_Transistors\_Practical}
Which two-dimensional material platforms (e.g., MoS₂, WSe₂, black phosphorus, h-BN), device architectures (e.g., floating-gate, ionic-gated, dual-gate), and fabrication protocols (e.g., channel thickness control, dielectric engineering, contact metallurgy) have been shown to optimize synaptic transistor performance—specifically in terms of energy per event, weight-update linearity, retention time,
and cycling endurance—for neuromorphic sensing applications? Think of most practical and promising candidates.
\medskip

\subsection{Microplastics\_Sensing\_2D}
Which two-dimensional material platforms (e.g., graphene derivatives, transition metal dichalcogenides, MXenes), molecular recognition elements (e.g., molecularly imprinted polymers, aptamers, peptide receptors), and device integration strategies (e.g., FET, electrochemical impedance, photonic transduction) have demonstrated the highest sensitivity, selectivity against organic matter and ionic
interferents, and rapid response times for detecting micro- and nanoplastic particles in complex water matrices? Think of novel candidates.
\medskip

\subsection{Antibiotics\_Sensing\_2D\_Probes}
Which commercially available, cost-effective probe chemistries (e.g., thiolated DNA aptamers, antibody mimetics, conductive MIPs) combined with two-dimensional nanomaterial transducer platforms (e.g., WS₂ FETs, graphene field-effect sensors, nanotube-extended gate FETs) deliver the lowest detection limits, minimal batch variability, and robust performance for sensing trace levels of pharmaceutical
antibiotics in diverse aqueous waterbody.
\medskip

\subsection{Li\_over\_Na\_Selectivity\_Membranes}
What material or membrane exhibits the highest selectivity for Li⁺ over Na⁺ in aqueous systems, given their nearly identical hydrated ionic radii and solvation environments? Beyond crown-ether–functionalized polymers, what novel or unexpected materials—such as bioinspired ultrahigh-selectivity membranes or covalent organic framework nanochannels—might provide breakthrough Li⁺ discrimination over
Na⁺? Find novel and promising candidates.
\medskip

\subsection{PFAS\_Electro\_Degradation}
Which novel electrode materials can achieve efficient PFAS degradation under ambient aqueous electrochemical conditions, delivering both high mineralization and defluorination rates? What intrinsic properties—such as PFAS adsorption affinity, reactive oxygen species generation capacity, and C–F bond activation energy—should be optimized to guide their discovery?

\medskip

\subsection{LiCoNi\_Selective\_Membranes}
Which membrane materials can effectively separate Li⁺, Co²⁺, and Ni²⁺ ions from aqueous solutions by leveraging selective transport properties—such as tailored pore sizes, specific surface functionalizations, and charge affinities—and what membrane design principles optimize both selectivity and permeability? Think of potential novel, effective, and practical candidates.

\medskip

\subsection{InSitu\_Liquid\_TEM\_MoS2\_Sensing}
How can in situ liquid cell TEM be employed to directly visualize the real-time adsorption and structural changes of 2D MoS₂ nanosheets used in aqueous FET sensors during analyte binding? What fluid cell configurations and electron dose parameters are necessary to preserve native water–material interfaces while capturing high-resolution sensing events without beam-induced artifacts?

\medskip

\subsection{CO2RR\_SOTA\_and\_Emerging\_Cats}
Given the current landscape of CO₂ electroreduction, which state-of-the-art catalyst platforms—such as oxide-derived copper, single-atom catalysts on nitrogen-doped carbon supports, or metal–organic framework-derived materials—demonstrate the highest activity and selectivity? Moreover, what emerging catalytic systems or novel heterostructures beyond these examples could feasibly outperform today’s
leading electrocatalysts in terms of faradaic efficiency and stability?
\medskip

\subsection{NextGen\_Solar\_Unconventional}
Perovskite–silicon tandem cells, organic photovoltaics, and quantum-dot solar cells currently represent the forefront of next-generation solar technologies. Beyond these established platforms, which unconventional material classes or innovative device architectures—such as chalcogenide perovskites, 2D semiconductor heterostructures, or ferroelectric photovoltaic systems—offer the most unexpected
promise for achieving breakthroughs in efficiency, stability, and scalability?
\medskip

\subsection{Ambient\_Pressure\_Diamond\_Growth}
How can ambient-pressure diamond synthesis using a Ga–Ni–Fe–Si liquid-metal alloy at 1 atm and ~1025 °C be adapted to produce larger-area diamond films or oriented single crystals, and what key mechanistic steps govern nucleation and growth kinetics under these mild-pressure conditions?

\medskip

\subsection{Diamane\_Bandgap\_Functionalization}
What chemical functionalizations or reaction pathways are most promising for tuning the electronic bandgap and mechanical stability of diamane-like 2D diamond films formed via sp²-to-sp³ conversion, and how might those modifications influence their integration into nanoelectronic devices?

\medskip

\subsection{LIB\_Fluoroether\_Anion\_Receptor}
Search for novel fluorinated ether–based electrolyte candidates that incorporate covalently bound anion-receptor motifs (e.g. boron- or phosphorus-centered groups) to deliver oxidative stability beyond 5.6 V, Li⁺ transference numbers above 0.8, and minimal interfacial impedance on lithium-metal anodes.

\medskip

\subsection{ML\_e\_Tongue\_for\_Water}
Which mixed-material (e.g., graphene/2D) multi-sensor arrays with on-chip QA/QC and pattern-recognition deliver robust mixture fingerprinting (metals + microbes + organics) in flowing water, with automated drift/fault handling for field deployment?

\medskip

\subsection{Batteryless\_Stretchable\_Biosensors}
Which tribo/piezo-powered, skin-conformal platforms (CNT/graphene networks, ionic gels) sustain continuous metabolite/ion monitoring without batteries—what power densities and duty-cycled readouts are practical for week-scale operation?

\medskip

\subsection{Mixed\_Dimensional\_Low\_Noise\_Transducers}
Which 2D-0D/2D-1D/2D-3D vdW stacks (e.g., MoS₂–QD, CNT–graphene) best combine strong capture/absorption with high-mobility transport to push FET/photodetector responsivity and baseline stability in real media?

\medskip

\subsection{Electro\_Photo\_Gated\_Smart\_Membranes}
Which electrically gated or photo-responsive nanochannels (MXene/COF/polymer hybrids) deliver on-demand selectivity and anti-fouling for ion separations and resource recovery in wastewater, and what gating “on/off” ratios and cycling lifetimes are state-of-the-art?

\medskip

\subsection{Transient\_Biodegradable\_Sensor\_Nodes}
Which dissolvable conductors, dielectrics, and encapsulants enable time-programmed water/soil sensors that operate stably for weeks then resorb without e-waste—what degradation kinetics and wireless/data-retention strategies are most practical?

\medskip

\subsection{Catalytic\_Mineralization\_of\_Microplastics}
Which ROS-generating photocatalysts and micromotor architectures achieve verifiable depolymerization/mineralization of common MPs (PE, PP, PET) in real waters—what mineralization fractions, spectra, and by-product toxicology benchmarks define “true cleanup”?

\medskip

\clearpage
\section{Local DR Experimental Settings}\label{app:exp_details_DR}

\subsection{Open-source LLMs and local RAG corpus}
We implement the framework in LangGraph and deploy three low-deployment-cost, open-source LLMs—DeepSeek-R1-14B, DeepSeek-R1-70B, QwQ-32B, gpt-oss-20B, and gpt-oss-120B—served locally via Ollama. All agent prompts and tool I/O use a JSON/tool-calling discipline for determinism and easy parsing, and the same research node (RN) described in Algorithm 2 is the unit of work in the controller (Algorithm
3 DToR). To provision representative, application-driven knowledge for nanomaterials and devices, we build a standardized Chroma vector store spanning 1,136,297 full-text papers organized around four pillars: sensors, batteries, catalysis, and semiconductors. The corpus is queried first under a local-first RAG policy; complementary web retrieval is invoked only when Algorithm 2 (local
evidence-first)’s reflection detects unresolved knowledge gaps.
Consistent with our emphasis on report-level usefulness rather than information retrieval (IR) benchmarking, the local-first RAG layer is treated as a routine, swappable component; DToR only triggers complementary web evidence upon explicit knowledge gaps and resolves cross-source conflicts by evidence majority → recency → domain authority. To make domain coverage portable and auditable without
exposing any local texts or indices, we release the exact domain-scoping query templates corresponding to these four pillars. In practice we provide templates for sensors, energy/environment (covering batteries), semiconductors, and catalysis so others can target the same scope with their own holdings while keeping the orchestration identical.
\subsubsection{Corpus construction}
Raw full texts are normalized with a cleaning pipeline designed for scientific documents at scale. The cleaner (i) preserves section structure and headers, (ii) retains LaTeX/MathJax equations and common math environments via stable markers, (iii) standardizes tables into a compact textual schema, (iv) normalizes URLs/DOIs and citation markers instead of stripping them, and (v) removes XML/HTML
artifacts and noisy back matter conservatively. For long inputs, we apply sliding-window tokenization (default window8192 tokens, stride 4096) or a section-aware variant that respects detected boundaries, yielding JSONL suitable for downstream training or indexing.
As our focus is on the DToR orchestration, we adhere to widely used, off-the-shelf preparation and vector-indexing practices; the corpus layer functions as a routine, swappable component rather than a methodological contribution. To enable scope-equivalent replication across different holdings, we release the exact domain-scoping query templates for sensors, energy/environment (including
batteries), semiconductors, and catalysis; users may instantiate a local store with their customary institutional pipelines and preferred indexers.
\paragraph{Indexing and retrieval} Cleaned texts are split with a recursive character splitter (default chunk size 2500, overlap 500) and embedded using BAAI/bge-m3. Each chunk carries rich metadata (paper id, DOI, chunk id/index, source path, shard id) to support provenance and de-duplication. Embeddings are persisted in Chroma, enabling fast k-NN retrieval and incremental persistence during large ingestions. For robustness across machines, the store is partitioned into 500 volumes (shards). This organization serves two purposes: (1) controllable breadth—DToR can expand to additional volumes when the analyst decides to broaden coverage—and (2) bounded depth/latency—branches can be pruned once depth or evidence sufficiency is reached, without touching unrelated volumes. To mitigate licensing and privacy risk, our local corpus is restricted to institutionally licensed or open sources, and we release only the domain-scoping query templates—not texts or indices—so others can replicate scope without redistributing copyrighted content or PII.
\paragraph{Policy integration} In each research node, local retrieval is attempted first; summaries are synthesized from local evidence. Only when reflection surfaces concrete gaps does the controller expand (selecting additional volumes and spawning new RNs) or prune (stopping further growth of a branch). This strategy couples breadth (multi-volume exploration across pillars) with depth (iterative drilling within a volume) while keeping costs predictable on commodity hardware and avoiding external dependencies during the main research loop.
\paragraph{Agent Configuration}
By default, we run all agents in a local-first configuration with cost-aware web fallback. Concretely, both single-agent DR and DToR branches use a research depth of 3 iterations, each RN queries the local Chroma first with top-k = 5, and only then invokes web search—by default DuckDuckGo—to avoid API fees. In DToR mode, breadth–depth is bounded with max branches = 3, max branch depth = 3, nodes
per branch = 100, and a global cap of 500 nodes per experiment, ensuring predictable latency and compute. Models are served locally via Ollama; RAG is enabled by default with BAAI/bge-m3 embeddings, and we preserve document fidelity via full-page fetch and safe output hygiene. All parameters remain user-tunable through the configuration/RunnableConfig to trade off coverage, latency, and cost—for
example, increasing depth or top-k for harder queries, or switching web backends when budgets permit.
\subsection{Domain-Scoping Query Templates}
The templates below define the \emph{domain coverage} used by our local-first pipeline
(sensors, energy/environment, semiconductors, catalysis). They are released to enable others
to target the same topic scope with their own collections and entitlements. 

\subsubsection{Sensor Domain Template}
\begin{lstlisting}
(
  TITLE-ABS-KEY(
    (
      "Optical" OR "Gas" OR "Pressure" OR "Thermal" OR "Magnetic" OR "Acoustic" OR "Humidity" OR "Strain" OR
      "Flexible" OR "Wearable" OR "Piezoelectric" OR "Photoelectric" OR "Biosensing" OR "Nanofiber" OR
      "Electrochemical" OR "RFID" OR "Environmental monitoring" OR "Photonics" OR "Nanowire" OR "Quantum" OR
      "Water quality" OR "Environmental" OR "Pollution detection" OR "Biological" OR "Biomarker" OR "VOC" OR
      "Medical" OR "Health monitoring" OR "Diagnostic" OR "Smart" OR "IoT" OR "Graphene-based" OR "Nano" OR
      "Agricultural" OR "Industrial" OR "Mercury ion" OR "Heavy metal ion" OR "Low-concentration gas" OR
      "CO" OR "H2" OR "NO2" OR "NH3" OR "Protein" OR "DNA" OR "Bacteria" OR "Nanocomposite" OR
      "Flexible electronic" OR "Food safety" OR "Air quality"
    )
    AND ("detector" OR "sensor")
  )

  OR TITLE-ABS-KEY(
    (
      "biosensor" OR "chemical sensor" OR "gas sensor" OR "humidity sensor" OR "pressure sensor" OR
      "strain gauge" OR "photodetector" OR "phototransistor" OR "chemiresistive" OR "chemiresistor" OR
      "field-effect transistor" OR FET OR ISFET OR EGFET OR OFET OR GFET OR BioFET OR
      "electrolyte-gated transistor" OR "lab-on-a-chip" OR "microfluidic*" OR "surface acoustic wave" OR SAW OR
      "quartz crystal microbalance" OR QCM OR "surface plasmon resonance" OR SPR OR SERS OR
      "electrochemical impedance spectroscopy" OR EIS OR "potentiometric" OR "amperometric" OR "voltammetric"
    )
    AND ("sensor*" OR "detector*" OR "transducer*" OR "FET" OR "transistor")
  )

  OR TITLE-ABS-KEY(
    (
      "H2S" OR "SO2" OR "NO" OR "O3" OR "CO2" OR "CH4" OR "ethanol" OR "methanol" OR "acetone" OR "toluene" OR
      "benzene" OR "formaldehyde" OR "isopropanol" OR "xylene" OR "hydrogen peroxide" OR "glucose" OR "urea" OR
      "lactate"
    )
    AND ("sensor*" OR "detector*" OR "biosensing" OR "chemosensing")
  )

  OR TITLE-ABS-KEY(
    "graphene oxide" OR "reduced graphene oxide" OR rGO OR "carbon nanotube" OR CNT OR "nanoribbon" OR "nanotube" OR
    MXene OR "Ti3C2Tx" OR MoS2 OR WS2 OR WSe2 OR "transition metal dichalcogenide*" OR TMD
  )

  OR TITLE-ABS-KEY(
    "advanced sensor network" OR "real-time monitoring" OR "smart sensor network" OR
    "integrated sensor network" OR "biosensing platform"
  )
  OR TITLE-ABS-KEY(
    "wireless sensor network" OR WSN OR "edge sensing" OR "edge computing" OR "digital twin"
  )

  OR TITLE-ABS-KEY(
    "selective binding" OR "molecular recognition" OR "binding affinity" OR "synthetic receptor" OR
    "probe design" OR "molecular docking" OR "computational probe design" OR "artificial binding site"
  )
  OR TITLE-ABS-KEY(
    "host-guest" OR "supramolecular" OR "molecular imprint*" OR MIP OR "affinity sensor" OR "bioprobe"
  )

  OR TITLE-ABS-KEY(
    "hydrocarbon binding" OR "non-polar molecule" OR "hydrocarbon adsorption" OR
    "pi-pi interaction" OR "van der Waals"
  )

  OR TITLE-ABS-KEY(
    "drug discovery" AND (
      "ligand receptor" OR "structure-based design" OR "biochemical assay" OR
      "high-throughput screening" OR "aptamer" OR "peptide probe"
    )
  )
)
AND ( LIMIT-TO ( DOCTYPE , "ar" ) )
AND ( LIMIT-TO ( LANGUAGE , "English" ) )
\end{lstlisting}

\subsubsection{Energy and Environment Domain Template}
\begin{lstlisting}
(
  TITLE-ABS-KEY(
    (
      "Lithium-ion battery" OR "Supercapacitor" OR "Fuel cell" OR "Flow battery" OR "Graphene battery" OR
      "Sodium-ion battery" OR "Flexible battery" OR "Fast charging technology" OR
      "Next-generation battery technologies" OR "Quantum batteries" OR "Redox flow batteries" OR "Metal-air batteries" OR
      "Thermal batteries" OR "Biodegradable batteries" OR "Ultrafast charging batteries" OR
      "Solar rechargeable batteries" OR "Lithium-sulfur battery" OR "Potassium-ion battery" OR "Solid-state battery" OR
      "Green energy storage" OR "CO2 capture" OR "Graphene-Sn hybrids" OR "Vertically-oriented graphene" OR
      "Graphene-SnO2 nanoparticle structures" OR "High-porosity graphene" OR "Graphene nanoparticle hybrids" OR
      "Energy density" OR "Cycle life" OR "Safety" OR "Sustainability" OR "Electrolyte innovation" OR
      "Anode materials" OR "Cathode materials"
    )
    AND ("battery" OR "storage")
  )

  OR TITLE-ABS-KEY(
    "Li-ion" OR "Li metal" OR "Na-ion" OR "K-ion" OR "Li-S" OR "solid electrolyte" OR SEI OR
    "solid-state electrolyte" OR "polymer electrolyte" OR "ionic liquid electrolyte" OR "supercapacit*" OR "ultracapacitor"
  )

  OR TITLE-ABS-KEY(
    "selective ion separation" OR "PFAS removal" OR "phosphate removal" OR "nitrate removal" OR
    "ion exchange" OR "chelation strategy" OR "selective extraction" OR "water purification"
  )
  
  OR TITLE-ABS-KEY(
    "advanced materials" AND (
      "selective membrane" OR "nanofiltration" OR "adsorption membrane" OR
      "responsive polymer" OR "smart polymer" OR "nanostructured adsorbent"
    )
  )
  
  OR TITLE-ABS-KEY(
    "water treatment" OR "water recovery" OR "wastewater treatment" OR "resource recovery" OR
    "water processing" OR "sustainable water management" OR "water reuse"
  )
  
  OR TITLE-ABS-KEY(
    "environmental remediation" AND ("sustainable recovery" OR "circular economy" OR "contaminant removal")
  )
  
  OR TITLE-ABS-KEY(
    "mineral processing" OR "critical mineral extraction" OR "hydrometallurgy" OR
    "separation technology" OR "selective separation"
  )

  OR TITLE-ABS-KEY(
    PFAS OR "per- and polyfluoroalkyl" OR "perfluorinated" OR PFOA OR PFOS OR
    "reverse osmosis" OR RO OR "membrane distillation" OR MD OR "adsorption" OR "ion-selective" OR
    "direct air capture" OR DAC
  )

  /* critical minerals detail */
  OR TITLE-ABS-KEY(
    "rare earth*" OR REE OR "lithium extraction" OR "lithium brine" OR cobalt OR nickel OR manganese OR
    "solvent extraction" OR SX OR "electrowinning" OR EW OR "adsorptive separation" OR "resin"
  )
)
AND ( LIMIT-TO ( DOCTYPE , "ar" ) )
AND ( LIMIT-TO ( LANGUAGE , "English" ) )
\end{lstlisting}

\subsubsection{Semiconductor and Device Domain Template}
\begin{lstlisting}
(
  TITLE-ABS-KEY(
    (
      "Quantum dots" OR "Thin film" OR "Organic" OR "Inorganic" OR "Flexible" OR "Optoelectronic" OR "Power" OR
      "High-frequency" OR "2D materials" OR "Nanostructured" OR "Spintronics" OR "Photovoltaic" OR "Thermoelectric" OR
      "Piezoelectric" OR "Memory" OR "Quantum computing" OR "Perovskite" OR "Nanocrystals" OR
      "Light-emitting diodes" OR "Solar cell" OR "Environmental monitoring" OR "Graphene-based" OR "Nanoelectronic" OR
      "Inkjet printing" OR "Printable" OR "Flexible electronics" OR "Wearable electronics" OR "Energy harvesting" OR
      "Silicon-based" OR "Gallium nitride" OR "High mobility" OR "Low-dimensional" OR
      "Transition metal dichalcogenides" OR "Black phosphorus" OR "Quantum well" OR "Heterojunctions" OR
      "Stretchable electronics" OR "High-efficiency solar cells" OR "Low-power" OR "High thermal conductivity"
    )
    AND ("semiconductor" OR "device")
  )

  OR TITLE-ABS-KEY(
    (
      transistor* OR MOSFET OR JFET OR HEMT OR CMOS OR "integrated circuit" OR IC OR diode* OR LED OR OLED OR QLED OR
      "photodetector*" OR "phototransistor*" OR "optoelectronic device*" OR "memory device*" OR "spintronic device*"
    )
    AND ("device*" OR "semiconductor*")
  )

  OR TITLE-ABS-KEY(
    "III-V" OR "SiC" OR "Ga2O3" OR "GaN" OR "InP" OR "InGaAs" OR "AlGaN" OR MoS2 OR WS2 OR WSe2 OR MoSe2 OR MoTe2 OR
    "black phosphorus" OR "perovskite" OR "graphene" OR MXene OR CNT OR "carbon nanotube" OR "nanoribbon"
  )

  OR TITLE-ABS-KEY(
    "Schottky" OR "p-n junction" OR "heterostructure" OR "bandgap" OR "carrier mobility" OR "on/off ratio"
  )
)
AND ( LIMIT-TO ( DOCTYPE , "ar" ) )
AND ( LIMIT-TO ( LANGUAGE , "English" ) )
\end{lstlisting}

\subsubsection{Catalysis Domain Template}
\begin{lstlisting}
(
  TITLE-ABS-KEY(
    (
      "Catalysis" OR "Heterogeneous" OR "Homogeneous" OR "Bio-" OR "Green" OR "Photoelectro" OR
      "Thermal" OR "Organocatalysis" OR "Asymmetric" OR "Nanozyme" OR "Photocatalytic water splitting" OR
      "CO2 reduction" OR "Electrochemical" OR "Plasmonic" OR "Enzyme immobilization" OR "Molecular" OR
      "Chiral" OR "Single atom" OR "Energy conversion" OR "Selective oxidation" OR "Environmental remediation" OR
      "Green energy" OR "hydrogen evolution" OR "hydrogen oxidation" OR "Oxygen reduction" OR "Oxygen evolution" OR
      "Metal-organic framework" OR "Bimetallic" OR "Fuel cell" OR "Biomass conversion" OR "high entropy"
    )
    AND ("catalyst" OR "catalysis")
  )

  OR TITLE-ABS-KEY(
    "photocatalysis" OR "electrocatalysis" OR "thermocatalysis" OR HER OR HOR OR ORR OR OER OR CO2RR OR "CO2 RR" OR
    "oxygen evolution reaction" OR "oxygen reduction reaction" OR "hydrogen evolution reaction"
  )

  OR TITLE-ABS-KEY(
    "single-atom catalyst" OR "defect engineering" OR "heterojunction" OR "perovskite oxide" OR spinel OR
    MXene OR MOF OR "metal-organic framework" OR COF OR "covalent organic framework" OR "g-C3N4" OR
    "graphitic carbon nitride" OR "NiFe LDH" OR "layered double hydroxide" OR "high-entropy alloy"
  )

  OR TITLE-ABS-KEY(
    "advanced catalysis" AND ("water purification" OR "photocatalytic degradation" OR "catalytic oxidation" OR "green catalysis")
  )
  
  OR TITLE-ABS-KEY(
    "advanced oxidation process*" OR AOP OR Fenton OR "photo-Fenton" OR "persulfate activation"
  )
)
AND ( LIMIT-TO ( DOCTYPE , "ar" ) )
AND ( LIMIT-TO ( LANGUAGE , "English" ) )
\end{lstlisting}

\clearpage
\section{LLM-as-Judge Rubric and Templates}\label{app:scoring_template}

\subsection{LLM-as-Judge Rubric Prompt (applied to DR Report)}
\{Anonymous DR Report content\}

Now please analyze and evaluate the Report above on five equally weighted dimensions:
\begin{enumerate}[label=\arabic*. , leftmargin=2.1em, itemsep=2pt]
  \item Relevance to the stated objective,
  \item Depth and rigor of validation and benchmarking protocols,
  \item Clarity and reproducibility of methodology,
  \item Practical applicability and implementation readiness,
  \item Novelty and breadth of candidate coverage.
\end{enumerate}
For each dimension, assign a score from 0--10 and supply a one-sentence justification.
This rubric places equal emphasis on procedural rigor and clear implementation steps alongside overall alignment,
ensuring that reports with well-defined benchmarks and actionable guidance receive full credit. The five-dimensional rubric was specifically designed to reward not just information retrieval but true research utility—prioritizing actionable insights with clear provenance over mere comprehensiveness.

\noindent\textit{Inter-rater reliability.} 
We compute Pearson and Spearman correlations across judges over repeated topic--report evaluations; these correlations quantify \emph{consistency among judges} and are \emph{not} used as evidence of factual correctness. While the rubric emphasizes objective dimensions like novelty and applicability, potential LLM judge biases—such as favoring outputs from similar architectures—were mitigated by
averaging across diverse models.

\subsubsection{Definitions of the Five Dimensions (equal weights)}
\textbf{Relevance} = Alignment to the stated objective and scope, including coverage of key subcomponents rather than topic keywords alone.\\
\textbf{Depth} = Rigor of analysis and validation (e.g., explicit assumptions, benchmarking or ablations, uncertainty/limitations stated).\\
\textbf{Clarity} = Structure and reproducibility of the methodology, with traceable provenance (citations, data/sources, steps) and unambiguous claims.\\
\textbf{Applicability} = Actionability of the outputs (materials/candidates, parameters/protocols, resource requirements) enabling near-term implementation.\\
\textbf{Novelty} = Non-trivial ideas or candidate combinations beyond canonical baselines, with rationale, plus \emph{breadth} of candidate coverage across the design space.

\subsubsection{Example raw response (DToR\_gpt-oss120B\_local500; Topic: PFAS FET sensor probe)  — Claude 4 Opus (thinking)}
\begin{lstlisting}
## Evaluation of Final ToT Synthesis Report

### Dimension 1: Relevance to the Stated Objective
**Score: 9/10**

The report directly addresses the core question about which 2D nanomaterials and molecular probes offer greatest potential for PFAS detection in FET-based sensors, providing comprehensive analysis of three distinct material approaches (COF-MIP monolayers, hybrid graphene/MoS2 heterostructures, and MXene-based platforms) with specific performance metrics and clear recommendations.

### Dimension 2: Depth and Rigor of Validation and Benchmarking Protocols
**Score: 7/10**

While the report presents impressive performance metrics (LODs down to 0.12 fg L-1, selectivity factors >20), it acknowledges critical gaps in long-term validation data (>100 cycles, >12 months field testing) and relies heavily on theoretical projections rather than comprehensive experimental validation across all three branches.

### Dimension 3: Clarity and Reproducibility of Methodology
**Score: 8/10**

The report provides detailed material specifications, synthesis parameters (e.g., HF-free MXene delamination <5 MJ kg-1, roll-to-roll coating at >200 fibers h-1), and clear fabrication steps, though some advanced integration procedures (monolithic FET architecture combining all components) lack sufficient procedural detail for immediate reproduction.

### Dimension 4: Practical Applicability and Implementation Readiness
**Score: 6/10**

Despite promising scalable manufacturing claims and cost projections ($0.35-1.00 per sensor), the report reveals significant implementation barriers including limited PFAS aptamer libraries, incomplete life-cycle assessments, and lack of demonstrated monolithic integration of the most promising hybrid approaches.

### Dimension 5: Novelty and Breadth of Candidate Coverage
**Score: 9/10**

The report demonstrates exceptional breadth with 40+ distinct materials and methodologies spanning multiple 2D material classes (COFs, graphene heterostructures, MXenes), innovative functionalization strategies (molecular imprinting, aptamer conjugation, plasmonic enhancement), and creative integration approaches (dual-mode detection, on-chip preconcentration).

---

**Overall Score: 7.8/10**

The report excels in comprehensively addressing the research question with innovative material combinations and demonstrates strong theoretical foundation, but implementation readiness is limited by acknowledged validation gaps and incomplete experimental verification of the most promising integrated approaches.
\end{lstlisting}

\subsubsection{Example raw response (DToR\_gpt-oss120B\_local500; Topic: PFAS FET sensor probe) — Gemini 2.5 Pro}
\begin{lstlisting}
An analysis and evaluation of the Final ToT Synthesis Report is provided below.

**1. Relevance to the stated objective**
Score: 10/10
Justification: The report directly and comprehensively addresses the core research question, culminating in a clear, prioritized list of candidate materials and sensor architectures tailored to the FET platform.

**2. Depth and rigor of validation and benchmarking protocols**
Score: 10/10
Justification: The report consistently uses quantitative performance metrics (LOD, sensitivity, cost) and includes a dedicated contradiction analysis section that critically evaluates and resolves conflicting claims from different research branches.

**3. Clarity and reproducibility of methodology**
Score: 9/10
Justification: The report outlines a clear, actionable design framework in its conclusion and uses well-structured tables to transparently present complex information, though specific synthesis protocols are understandably beyond its high-level scope.

**4. Practical applicability and implementation readiness**
Score: 10/10
Justification: The analysis consistently emphasizes real-world implementation factors such as scalable manufacturing, cost-per-sensor, anti-fouling strategies, and sensor lifetime, providing a clear roadmap from lab to field.

**5. Novelty and Breadth of Candidate Coverage**
Score: 10/10
Justification: The report synthesizes findings from three distinct and advanced material classes (COFs, heterostructures, MXenes) and covers a comprehensive inventory of novel probes, functionalization chemistries, and signal amplification techniques.
\end{lstlisting}

\subsubsection{Example raw response (DToR\_gpt-oss120B\_local500; Topic: PFAS FET sensor probe) — Grok 3 (thinking)}
\begin{lstlisting}
### Evaluation of the Report

1. **Relevance to the stated objective**: Score 10/10 - The report directly synthesizes and evaluates 2D nanomaterials and molecular probes for PFAS detection in FET sensors, identifying top candidates like COF monolayers, graphene/MoS2 heterostructures, and MXenes while addressing sensitivity, selectivity, and broader analyte potential.

2. **Depth and rigor of validation and benchmarking protocols**: Score 7/10 - While the report benchmarks performance metrics like LODs and sensitivities against techniques such as EIS, QCM-D, and SERS, it relies on synthesized branch data without detailing original experimental protocols or statistical rigor, leaving some validation gaps unresolved.

3. **Clarity and reproducibility of methodology**: Score 8/10 - Methodologies for material synthesis, functionalization (e.g., MIP imprinting, fluorosilane passivation), and integration (e.g., roll-to-roll coating) are described with specific parameters and performance data, enabling reasonable reproducibility, though some steps like aptamer generation lack step-by-step instructions.

4. **Practical applicability and implementation readiness**: Score 8/10 - The report emphasizes scalable manufacturing, cost estimates (e.g., $0.35 per sensor), regeneration strategies, and real-world challenges like anti-fouling, indicating strong readiness for field deployment, but notes gaps in long-term validation that could hinder immediate implementation.

5. **Novelty and Breadth of Candidate Coverage**: Score 9/10 - The report covers a broad inventory of established (e.g., graphene) and underexplored candidates (e.g., Nb4C3Tx MXenes, PFAS-specific aptamers) with innovative hybrids like MXene-AgNP composites, highlighting novel integrations while synthesizing unique perspectives from multiple branches.
\end{lstlisting}

\subsubsection{Example raw response (DToR\_gpt-oss120B\_local500; Topic: PFAS FET sensor probe) — ChatGPT o3}
\begin{lstlisting}
Expert Evaluation of the "Final ToT Synthesis Report"

Dimension 1 - Relevance to the stated objective: 9/10  
The report stays tightly focused on identifying 2-D materials and molecular probes for PFAS-targeted FET sensing, continuously mapping every finding back to sensitivity, selectivity and scalability goals.

Dimension 2 - Depth and rigor of validation and benchmarking protocols: 7/10  
It cites quantitative LODs, mobility values, cost metrics and comparative tables, yet long-term cycling data, full life-cycle assessments and multi-lab benchmarking remain largely prospective rather than experimentally demonstrated.

Dimension 3 - Clarity and reproducibility of methodology: 8/10  
Process flows (e.g., roll-to-roll coating, HF-free MXene synthesis, dual-mode FET construction) are described with sufficient detail and metrics to guide replication, though specific experimental parameters (e.g., polymerisation currents, imprinting times) are occasionally omitted.

Dimension 4 - Practical applicability and implementation readiness: 8/10  
The report highlights scalable manufacturing routes and <$1 per-sensor costs, but real-world field trials, regulatory certification pathways and integration with portable electronics are only touched on, indicating TRL 4-5 rather than deployment-ready status.

Dimension 5 - Novelty and breadth of candidate coverage: 9/10  
It spans COF monolayers, graphene/MoS2 heterostructures, multiple MXenes, nano-MIPs, aptamers and plasmonic enhancers--introducing several under-explored pairings--while articulating how each uniquely advances PFAS sensing.
\end{lstlisting}

\subsubsection{Example raw response (DToR\_gpt-oss120B\_local500; Topic: PFAS FET sensor probe) — ChatGPT o4 mini high}
\begin{lstlisting}
1. Relevance to the stated objective - Score: 10  
   The report directly addresses PFAS-focused FET sensor design by comparing 2D nanomaterial platforms, functionalization strategies, and performance metrics tailored to ultra-sensitive, selective detection.

2. Depth and rigor of validation and benchmarking protocols - Score: 8  
   It presents quantitative LODs, binding constants, and comparative deviations versus LC-MS/MS, but lacks standardized cross-laboratory protocols and detailed statistical analyses.

3. Clarity and reproducibility of methodology - Score: 6  
   While the report outlines material choices, functionalization chemistries, and device architectures, it omits step-by-step experimental procedures and specific process parameters needed for exact replication.

4. Practical applicability and implementation readiness - Score: 9  
   The discussion of scalable roll-to-roll/printing methods, cost estimates (< $1 per sensor), and anti-fouling/regeneration strategies demonstrates strong real-world deployment potential.

5. Novelty and breadth of candidate coverage - Score: 9  
   By surveying COFs, graphene/MoS2 heterostructures, and diverse MXene compositions alongside a wide array of molecular probes, the report offers a comprehensive and innovative candidate inventory.
\end{lstlisting}

\begin{table}[h!]
\centering
\caption{Example raw response (DToR\_gpt-oss120B\_local500; Topic: PFAS FET sensor probe; Trial \#1) — cross-model score matrix (0--10).}
\begin{tabular}{@{} l c c c c c c @{}}
\toprule
\textbf{Model} & \textbf{Relevance} & \textbf{Depth} & \textbf{Clarity} & \textbf{Applicability} & \textbf{Novelty} & \textbf{Overall (mean)} \\
\midrule
Claude 4 Opus (thinking) & \textbf{9}  & \textbf{7}  & \textbf{8}  & \textbf{6}  & \textbf{9}  & \textbf{7.8} \\
Gemini 2.5 Pro           & \textbf{10} & \textbf{10} & \textbf{9}  & \textbf{10} & \textbf{10} & \textbf{9.8} \\
Grok 3 (thinking)        & \textbf{10} & \textbf{7}  & \textbf{8}  & \textbf{8}  & \textbf{9}  & \textbf{8.4} \\
ChatGPT o3               & \textbf{9}  & \textbf{7}  & \textbf{8}  & \textbf{8}  & \textbf{9}  & \textbf{8.2} \\
ChatGPT o4 mini high     & \textbf{10} & \textbf{8}  & \textbf{6}  & \textbf{9}  & \textbf{9}  & \textbf{8.4} \\
\midrule
\textbf{Average across models} & \textbf{9.6} & \textbf{7.8} & \textbf{7.8} & \textbf{8.2} & \textbf{9.2} & \textbf{8.5} \\
\bottomrule
\end{tabular}
\end{table}

\subsection{\texorpdfstring{A/B Dueling Prompt (\,DR Report A\ vs.\ DR Report B\,)}{A/B Dueling Prompt (DR Report A vs. DR Report B)}}
Report A:
\{Anonymous DR Report content\}

Report B:
\{Anonymous DR Report content\}

As an experienced materials scientist who bridges the gap between research and pilot lines, please review these two reports.
I need your professional judgment: which one offers the clearer high-level strategic roadmap for our team’s R\&D over the next six months?
We are especially interested in the report that presents a well-structured narrative, systematically identifies and prioritizes knowledge gaps,
and synthesizes actionable next steps—beyond performance metrics.
Please give me your recommendation in a brief, plain-language note—straight to the point—ending with a clear verdict on which report to back
and the single most decisive reason why.

\subparagraph{A/B Duel — Metadata}
\begin{lstlisting}
A/B Duel: Metadata
agent_a: ChatGPT_o3_DR
agent_b: DToR_gpt-oss120B_local500
trial: trial_1
\end{lstlisting}

\subsubsection{Raw dueling response (DToR\_gpt-oss120B\_local500; Topic: PFAS FET sensor probe) — Claude 4 Opus (thinking)}
\begin{lstlisting}
**Materials Scientist Assessment**

**Report A** provides a comprehensive academic survey but reads more like a literature review than a strategic roadmap. While thorough, it spreads across too many materials without clear prioritization and lacks concrete next steps for R&D planning.

**Report B** delivers exactly what you need for strategic planning. It systematically identifies the
top-performing candidates (MXene with fluorinated brushes for sensitivity, COF nanosheets for durability), resolves key technical contradictions through specific architectural solutions, and maps out clear manufacturing pathways with roll-to-roll processing. Most importantly, it explicitly identifies remaining knowledge gaps and provides actionable timelines.

Report B's \"Contradiction Analysis
& Resolution\" section alone is worth its weight in gold--it tackles the real engineering trade-offs your team will face (sensitivity vs. durability, manufacturing speed vs. quality) and provides specific solutions rather than just flagging issues.

The device architecture recommendation is concrete and implementable: MXene channel + sub-5nm fluorinated COF dielectric + MIP-COF overlayer. This
gives your team a clear target to work toward with defined performance metrics and manufacturing constraints.

**Verdict: Back Report B**

**Decisive reason:** Report B transforms scattered research into a concrete R&D roadmap with prioritized materials, specific architectures, and identified next steps--exactly what you need to allocate resources and set milestones for the next six months.\end{lstlisting}

\subsubsection{Raw dueling response (DToR\_gpt-oss120B\_local500; Topic: PFAS FET sensor probe) — Gemini 2.5 Pro}
\begin{lstlisting}
Here is my recommendation, straight to the point.

***

Team,

After reviewing both documents, my recommendation is clear. Report A is an excellent, comprehensive library of the entire field--a valuable resource for onboarding a new researcher. Report B, however, is a strategic plan.

Report B provides the roadmap. It doesn't just list materials and their performance; it synthesizes the
data into two competing but complementary perspectives. It then systematically resolves the conflicts between them, which is where the real strategic value lies. The report directly answers \"what should we do next?\" by providing a specific, multi-layered device architecture to target and a prioritized list of knowledge gaps that can be converted directly into our Q3/Q4 milestones. It gives us a
strong, defensible point of view to build our project around.

**Verdict: We back Report B.**

The single most decisive reason is its **Contradiction Analysis section.** This transforms the document from a passive review into an active strategy, showing us not just what the challenges are, but how to resolve the technical trade-offs to build a superior device.\end{lstlisting}


\subsubsection{Raw dueling response (DToR\_gpt-oss120B\_local500; Topic: PFAS FET sensor probe) — Grok 3 (thinking)}
\begin{lstlisting}
After reviewing both reports as a materials scientist with pilot-line experience, Report B stands out for its integrated synthesis of perspectives, clear resolution of contradictions, and explicit focus on practical trade-offs like manufacturing scalability and device architectures. It provides a more actionable framework by prioritizing real-world deployment challenges (e.g., wear validation in
sweat) and recommending specific hybrid designs that balance sensitivity with durability.

**Verdict: Back Report B.** The single most decisive reason is its systematic identification and prioritization of knowledge gaps, paired with synthesized next steps (e.g., multilayer heterostructures and process optimizations), which directly guide targeted R&D over the next six months beyond just
metrics.\end{lstlisting}


\subsubsection{Raw dueling response — ChatGPT o4 mini high}
\begin{lstlisting}
Between the two, Report B gives you the sharper six-month R&D playbook. It lays out two complementary design perspectives, reconciles trade-offs, pinpoints exact knowledge gaps (wear-test validation, complex-matrix studies, environmental impact), and closes with a concrete, multilayer device architecture and prioritized material-probe combinations. 

Verdict: Back Report B--because it distills
the review into a clear, structured roadmap of what to build, test, and refine next.\end{lstlisting}


\subsubsection{Raw dueling response (DToR\_gpt-oss120B\_local500; Topic: PFAS FET sensor probe) — ChatGPT o3}
\begin{lstlisting}
Team,

Both reports give strong technical overviews, but only one reads like a playbook you can hand to an R&D scrum tomorrow.

* Report A is an encyclopedic review: rich detail, many examples, but it stops at describing what exists. You'd still need to distill priorities, rank open questions, and map experiments yourself.

* Report B is written as a roadmap: it organizes findings into themed "perspectives," lists candidate stacks, flags contradictions, clarifies trade-offs, and--critically--spells out the remaining
knowledge gaps (long-term wear tests, matrix interference, environmental impact) as discrete to-dos. The table of device architectures with drift/noise targets and the hybrid-stack recommendation give an immediate six-month experiment sequence.

Verdict: Back Report B--because it converts the literature into a prioritized action plan, not just a survey.\end{lstlisting}

\clearpage
\section{Dry-Lab Validation Details}\label{app:dry_lab_details}

This appendix presents the complete first-principles validation for the agent's top-performing candidates, translating textual hypotheses into actionable, physically-grounded simulations. Across the five representative tasks, we used DFT and AIMD to move beyond textual scores and test real-world viability. A critical, recurring insight emerged: computationally demanding explicit-solvent AIMD
simulations revealed performance reversals and candidate failures that were invisible to simpler static DFT calculations, particularly for the battery binder and PFAS sensor tasks. This successful dry-lab verification confirms that the DToR framework can generate novel hypotheses that are robust to complex, real-world operational conditions\footnote{Dry-lab simulations are interpreted as relative
ranking proxies under fixed protocols (system sizes/constraints and scoring transforms listed here) rather than definitive materials predictions; experimental validation remains necessary. Although dry-lab simulations validate the physical plausibility of agent proposals, these remain approximations; transitioning to wet-lab would involve synthesizing top candidates like aptamer-functionalized
graphene FETs for PFAS detection, potentially using robotic platforms for high-throughput testing to close the discovery loop.}. The following sections detail the full computational methodology and results for each of the five validation tasks.
\subsection{DFT simulation settings}

\subsubsection*{\textbf{Static Simulation}}

To simulate the binding interactions of diverse combinations under
periodic boundary conditions, we employed ab initio calculations
employing parameters optimized for precision and computational
efficiency. The plane-wave energy cutoff was set to 400 eV, and
spin-polarized density functional theory DFT with the
Perdew-Burke-Ernzerhof (PBE) functional was utilized to describe
exchange-correlation interactions. The Brillouin zone integration
employed a Gamma-centered 1 $\times$ 1 $\times$ 1 k-point grid and van der Waals
corrections were incorporated using the DFT-D3 method to capture
dispersion effects. Ionic relaxations were conducted with a convergence
criterion of 0.02 eV/\AA{} for the forces, and the electronic structure
iterations were set to converge to an energy difference of $10^{-4}$ eV.
Implicit solvent field is added for static simulations by setting the
dielectric constant at 78.4 if applicable (namely excluding the ab
initio molecular dynamics (AIMD) with explicit solvent molecules, and
the CO\texorpdfstring{$_2$}{2} sensor probe binding scenarios). The modeling of the slabs is
directly conducted by human expert based on anonymously reading the DR
report. To ensure fair comparison, we asked the expert to create same
number of candidates from each report (e.g. 4 from DR report A, and 4
from DR report B), and also pick one domain commonly chosen benchmark
material/chemical for each of the 5 applications. To prevent
interactions between periodic images, we also include $>$20 \AA{} vacuum gap
along the z-, ensuring a reliable comparison between adsorbed and
separated configurations.
\par
To be noted,tTo manage the configurational complexity of these diverse heterostructures and ensure consistent high-throughput benchmarking, we adopted standardized initial adsorption geometries tailored to the dominant interaction mechanism of each application. For PFAS sensing, which relies on host--guest recognition, linear perfluorinated chains were initialized inserting vertically into the target cavities (e.g., MIP pores or $\beta$-cyclodextrin rings) to mimic inclusion complex formation. Conversely, for PFAS degradation and battery binder applications, where maximizing interfacial contact area is critical for catalysis and adhesion, adsorbates were positioned in a planar configuration parallel to the inorganic substrate. Similarly, for CO$_2$ sensing, the molecule was initialized in a parallel orientation relative to the surface to facilitate interaction with active sites. While we acknowledge that exhaustive conformational sampling might reveal subtle secondary minima, focusing on these chemically dominant interaction modes provides a consistent and unbiased baseline across all systems. This approach is sufficient to validate the physical plausibility of the agent's proposals and establish a robust relative ranking for high-throughput screening without the prohibitive cost of global optimization.
\subsubsection*{\textbf{Binding Energy Convention.}}

Following a standard DFT convention (Phys. Rev. B, 2008, 77, 12) for
adsorption on typical studied 2D material systems like graphene, we
report the binding energy $\Delta E_{\text{bind}}$ of an adsorbate at a given coverage
as

\begin{equation}
\Delta E_{\text{bind}} = (E_{\text{clean\_slab}} + E_{\text{adsorbate}}) - E_{\text{adsorbed}}
\end{equation}

where $E_{\text{adsorbed}}$ is the total energy of the fully relaxed
adsorbate--slab complex (for certain applications, bottom atoms of bulk
crystals are frozen and surface area are free to mimic realistic surface
interactions), $E_{\text{clean\_slab}}$ is the energy of the corresponding pristine
clean slab supercell, and $E_{\text{adsorbate}}$ is the energy of the isolated
adsorbate in the same supercell. All terms are computed with identical
exchange--correlation functional, plane-wave cutoff, and k-point mesh.
Unless stated specifically (like for battery binder normalized to be
contact areal), we report $\Delta E_{\text{bind}}$ for only one adsorbate in the slab
super cell for qualitative comparison. This definition mirrors the
adsorption-energy convention widely used for molecules on typical
substrates like graphene or MXene (we adopt the term ``binding energy''
throughout). By this convention, a more positive $\Delta E_{\text{bind}}$ indicates an
exothermic, favorable adsorption (energy released upon binding).

\subsubsection*{\textbf{Ab Initio Molecular Dynamics (AIMD) Simulation}}

To improve the physical realism of our calculations and especially
considering solvent effect for surface events, we embedded explicit
water molecules in the supercell according to conformations identified
in preceding static simulations (Mol. Syst. Des. Eng. 2025, 10,
345--356; Sci. Adv. 2025, 11, eadr9038) using an end-state protocol.
Specifically, the optimized adsorbed configuration corresponding
to $E_{\text{adsorbed}}$ was used as the initial snapshot for the
solvated \textit{adsorbed} state, with explicit water molecules added
thereafter. For the \textit{pulled} state, the adsorbate was translated by at
least 15 \AA{} away from the relaxed slab (e.g., graphene)---without
additional geometry optimization---and then solvated under identical
conditions. These two equilibrated states form the basis for directly
comparing adsorption energetics in explicit solvent, consistent with
end-state comparisons commonly applied in MD umbrella-sampling studies
(J. Phys. Chem. C 2023, 127, 5920--5930; ChemPhysChem 2018, 19,
690--702), and in line with benchmark explicit-solvent AIMD treatments
of metal/water interfacial solvation (J. Chem. Phys. 2020, 152, 144703)
and first-principles determinations of interfacial hydration free
energies for adsorbates on Pt (J. Chem. Phys. 2021, 154, 094107).

The water density was maintained at 1 g cm$^{-3}$ (for battery binder,
solvent composition is different as described in corresponding section).
Our goal was to compare the evolution of the system's potential
(electronic) energy upon equilibration (Electrocatalysis 2017, 8,
577--586; Commun. Chem. 2022, 5, 76; J. Chem. Phys. 2020, 152, 144703;
J. Chem. Phys. 2021, 155, 134703).

To ensure that the solvent adopted physically meaningful configurations,
each system was first equilibrated for 2 ps at 300 K with a 0.5 fs time
step. Temperature was regulated via a Langevin thermostat to suppress
extraneous inertial effects while preserving canonical sampling. After
equilibrium, we further performed a 3 ps production run, from which we
extracted time-averaged electronic potential energies.

We therefore report an explicit-solvent end-state potential-energy gap,
denoted $\Delta E_{\text{pot}}^{\text{ES}}$ (AIMD), defined as the difference between the
time-averaged electronic potential energies of the
solvated \textit{adsorbed} and \textit{pulled} basins in identical cells:

\begin{equation}
\Delta E_{\text{pot}}^{\text{ES}} \equiv \langle E_{\text{pot}} \rangle_{\text{pulled}} - \langle E_{\text{pot}} \rangle_{\text{ads}}.
\end{equation}

Unless otherwise noted, all comparisons use the same supercell, water
count, ion composition, thermostatting, and slab-electrostatics
settings. Hence, the more positive $\Delta E_{\text{pot}}^{\text{ES}}$ (AIMD), the more
energetically favorable the adsorbed state---i.e., a potential-energy
analogue of the binding energy---whereas a negative $\Delta E_{\text{pot}}^{\text{ES}}$ indicates
that explicit solvation stabilizes the pulled state and adsorption is
disfavored.

\textbf{Scope and limitations of $\Delta E_{\text{pot}}^{\text{ES}}$:}\\
$\Delta E_{\text{pot}}^{\text{ES}}$ is not a rigorously converged adsorption free energy. By
construction it (i) neglects configurational and translational entropy;
(ii) folds solvent reorganization and water (or other solvent)--surface
exchange energetics into the potential-energy baseline; and (iii)
depends on finite-time sampling of each basin rather than a reversible
pathway or an alchemical thermodynamic cycle. Consequently, its absolute
magnitudes can reflect not only adsorbate--surface interactions but also
large-scale differences in hydrogen-bond networks and long-range
electrostatics associated with hundreds of explicit waters. Nonetheless,
because all systems are treated consistently within the same AIMD
framework, $\Delta E_{\text{pot}}^{\text{ES}}$ serves as a heuristic yet informative comparative
descriptor for explicit-solvent binding trends across molecules and
candidate surfaces, revealing solvent-mediated, surface-dependent
effects that are not captured in vacuum or implicit-solvent
calculations. Formal free-energy methods (umbrella sampling/PMF,
metadynamics, and double-decoupling/TI with standard-state and
slab-electrostatics corrections) are acknowledged as the rigorous routes
to $\Delta G$, as exemplified for interfacial adsorbates at water--Pt interfaces
(J. Chem. Phys. 2021, 154, 094107).

\textbf{Computational feasibility rationale:}\\
Rigorous free-energy calculations become computationally prohibitive for
our systems, which routinely exceed 500 atoms and include
numerous transition-metal centers in addition to light elements; fully
converged interfacial free-energy landscapes are beyond feasible AIMD
timescales under such conditions. Extensive explicit-solvent AIMD
benchmarks of metal/water interfaces have likewise highlighted the
costliness of such simulations (J. Chem. Phys. 2020, 152, 144703). In
this context, the explicit-solvent end-state potential-energy gap
provides a practical dynamic compromise: it captures key
solvent-mediated effects absent from static DFT binding energies while
remaining tractable for large, chemically realistic slabs. Thus,
although approximate, $\Delta E_{\text{pot}}^{\text{ES}}$ offers higher fidelity and greater
interpretive value than purely static vacuum or implicit-solvent
descriptions for the screening-level comparisons emphasized in this
work.

\clearpage

\subsection{PFAS FET Sensor Probe}\label{dry_lab_PFAS_sens}

\subsubsection{PFAS\_2D\_FET\_Probes}

For the first application: PFAS FET Sensor Probe. 
The query as shown in \ref{app:topics} is:
\begin{lstlisting}
Which two-dimensional (2D) nanomaterials or molecular probes--including novel, underexplored candidates--offer the greatest potential to achieve highly sensitive and selective detection of per- and polyfluoroalkyl substances (PFAS), or more broadly chemical and biological analytes, when integrated into a field-effect transistor (FET)-based sensor platform?
\end{lstlisting}

The best commercial solution is: \textbf{ChatGPT\_o3\_DR}.

The best local non gpt-oss solution is: \textbf{DToR\_qwq32B\_local500}.

The best local gpt-oss solution is: \textbf{DToR\_gpt-oss120B\_local500}.

\begin{figure}[htbp]
  \centering
  \includegraphics[width=1\textwidth]{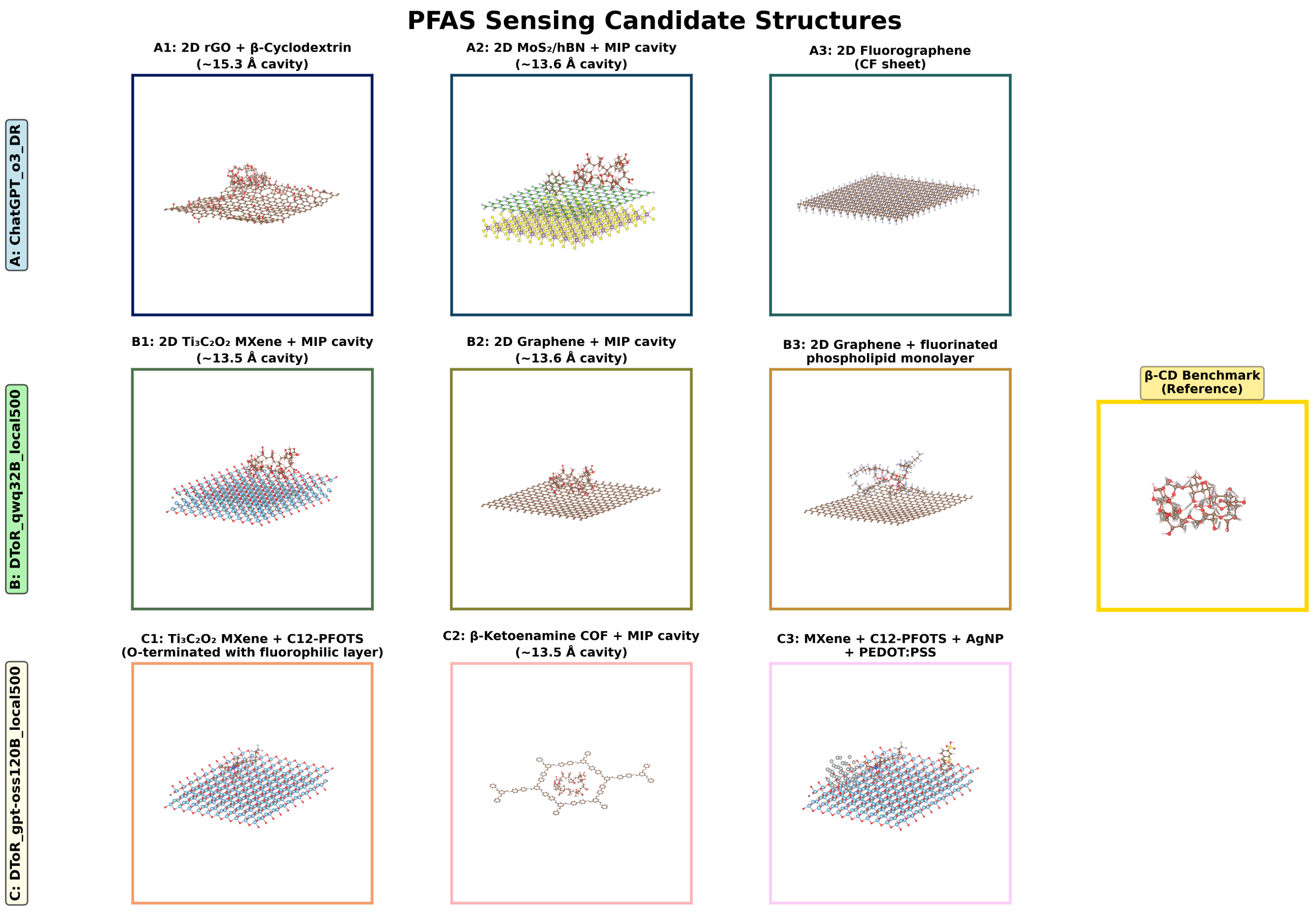}
  \caption{Structural configurations of candidates proposed by DR agents for selective PFAS binding, with \texorpdfstring{$\beta$}{beta}-cyclodextrin \protect\cite{wang2024reversible} included as the domain benchmark reference.}
  \label{fig:DL_1_candidates_final}  
\end{figure}

The expert based on the three reports, provided 3 commercial DR report derived candidates (A1-A3), and 3 local DR non gpt-oss report derived candidates (B1-B3), and 3 local DR gpt-oss report derived candidates (C1-C3), visualized in Fig. \ref{fig:DL_1_candidates_final}:

\paragraph{\textbf{A1 --- 2D rGO + \texorpdfstring{$\beta$}{beta}-Cyclodextrin (\texorpdfstring{$\beta$}{beta}-CD)}}
Hexagonal P1 slab with a = b = 36.90 \AA{}, \texorpdfstring{$\gamma$}{gamma} = 120°, c = 45.00 \AA{} (in-plane
area $\approx$ 1179.19 \AA{}$^2$). Atom counts (C/H/O = 483/102/94) indicate a 2D
graphene-derived sheet with oxygenated defects (epoxy/hydroxyl) on one
face and a carbohydrate torus consistent with \texorpdfstring{$\beta$}{beta}-CD mounted above it. The
populated thickness is $\approx$ 15.02 \AA{} (vacuum $\approx$ 29.98 \AA{}); the 2D carbon plane
sits near z $\approx$ 7.5 \AA{}. The \texorpdfstring{$\beta$}{beta}-CD presents a toroidal pocket with a
projected mouth \textbf{$\sim$15.3 \AA{} diameter}, oriented normal to the
basal plane for PFOS capture. Oxygen sites on rGO act as plausible
anchoring points while the underlying 2D lattice remains largely sp$^2$ and
planar.

\subparagraph{\textbf{A2 --- 2D MoS\texorpdfstring{$_2$}{2}/2D hBN heterojunction + methacrylic-acid (MIP-like) cavity, pyrene-linked}}
Hexagonal P1 with a = b = 28.44 \AA{}, \texorpdfstring{$\gamma$}{gamma} = 120°, c = 46.13 \AA{} (area $\approx$ 700.47
\AA{}$^2$). Composition shows a \textbf{2D MoS\texorpdfstring{$_2$}{2}/2D hBN heterojunction} stack
capped by an aromatic (pyrene-type) tether and a \textbf{methacrylic-acid
cluster forming a \texorpdfstring{$\beta$}{beta}-CD--like cavity} on one side (organics C/H/O sit
above the dielectric). The populated thickness is $\approx$ 18.68 \AA{} (vacuum $\approx$
27.45 \AA{}). The cavity's projected mouth is \textbf{$\sim$13.6 \AA{} in
diameter}, positioned above hBN so PFOS can approach without
perturbing the MoS\texorpdfstring{$_2$}{2} channel. The pyrene linker bridges $\pi$--$\pi$ to the 2D
cap while keeping the cavity axis roughly perpendicular to the surface,
preserving electronic decoupling from the channel.

\subparagraph{\textbf{A3 --- 2D Fluorographene (CF) slab}}
Hexagonal P1 with a = b = 36.90 \AA{}, \texorpdfstring{$\gamma$}{gamma} = 120°, c = 45.00 \AA{} (area $\approx$ 1179.19
\AA{}$^2$). Stoichiometry C$_{450}$F$_{450}$ yields a single-layer \textbf{2D} CF sheet
with chair-like alternating C--F buckling; the ``solid'' thickness is $\approx$
3.26 \AA{} and vacuum $\approx$ 41.74 \AA{}. The surface is uniform and free of
additional adsorbates, providing a continuous fluorinated interface.
Fluorine is arranged on both sides of the 2D carbon lattice, giving a
symmetric fluorophilic environment. The wide lateral extent minimizes
edge effects, making it suitable for baseline PFOS adsorption
comparisons.

\subparagraph{\textbf{B1 --- 2D Ti\texorpdfstring{$_3$}{3}C\texorpdfstring{$_2$}{2}O\texorpdfstring{$_2$}{2} MXene + methacrylic-acid (MIP-like) cavity}}
Hexagonal P1 with a = b = 30.67 \AA{}, \texorpdfstring{$\gamma$}{gamma} = 120°, c = 46.70 \AA{} (area $\approx$ 814.63
\AA{}$^2$). The slab is an O-terminated \textbf{2D Ti\texorpdfstring{$_3$}{3}C\texorpdfstring{$_2$}{2}} MXene bearing, on one
side, a \textbf{methacrylic-acid cluster organized as a \texorpdfstring{$\beta$}{beta}-CD--like
cavity} (additional C/H/O located at higher z). Populated thickness $\approx$
16.14 \AA{} (vacuum $\approx$ 30.56 \AA{}). The cavity mouth projects \textbf{$\sim$13.5 \AA{} in
diameter}, perched above the O-terminated surface while the bottom
MXene layer serves as a rigid support. The one-sided functionalization
creates an asymmetric interface---electronegative terminations and the
cavity's polar rim cooperate to stabilize PFOS headgroup interactions.

\subparagraph{\textbf{B2 --- 2D Graphene + methacrylic-acid (MIP-like) cavity}}
Hexagonal P1 with a = b = 36.90 \AA{}, \texorpdfstring{$\gamma$}{gamma} = 120°, c = 45.00 \AA{} (area $\approx$ 1179.19
\AA{}$^2$). A 15$\times$15 \textbf{2D} graphene sheet supports, on one face, a
\textbf{methacrylic-acid cluster arranged into a \texorpdfstring{$\beta$}{beta}-CD--like cavity}
(C/H/O entirely above the carbon plane). The populated thickness is $\approx$
9.58 \AA{} with $\approx$ 35.42 \AA{} vacuum. The cavity presents a projected
\textbf{$\sim$13.6 \AA{} diameter} single-sided pocket aligned normal to the
basal plane for PFOS engagement. The 2D sp$^2$ lattice remains flat and
conductive, making this a clean platform for decoupling recognition from
transduction.

\subparagraph{\textbf{B3 --- 2D Graphene + fluorinated phospholipid monolayer}}
Hexagonal P1 with a = b = 36.90 \AA{}, \texorpdfstring{$\gamma$}{gamma} = 120°, c = 45.00 \AA{} (area $\approx$ 1179.19
\AA{}$^2$). Elemental makeup (C/F/H/O with trace N and P) is consistent with
phosphatidylcholine headgroups near the \textbf{2D} graphene plane and
perfluorinated tails extending outward to form a dense brush. The
populated thickness is $\approx$ 14.84 \AA{} (vacuum $\approx$ 30.16 \AA{}), with F spanning
roughly z $\approx$ 9.21--22.34 \AA{} above the basal plane. The assembly yields a
hydrophilic-to-fluorophilic gradient, presenting a fluorinated outer
surface for PFOS tail affinity. Lipid tails show a collective tilt
relative to the surface normal, producing a compact, low-defect
fluorinated interface on one side while the opposite side remains bare
graphene.

\subparagraph{\textbf{C1 --- 2D Ti\texorpdfstring{$_3$}{3}C\texorpdfstring{$_2$}{2}O\texorpdfstring{$_2$}{2} MXene + C12-PFOTS}}
Hexagonal P1 with a = b = 30.67 \AA{}, \texorpdfstring{$\gamma$}{gamma} = 120°, c = 46.70 \AA{} (area $\approx$ 814.63 \AA{}$^2$). 
The slab is an O-terminated \textbf{2D Ti\texorpdfstring{$_3$}{3}C\texorpdfstring{$_2$}{2}} MXene bearing, on one side, a single 
\textbf{perfluorododecyltrimethoxysilane (C12-PFOTS)} molecule lying flat on the surface. 
The longer perfluorocarbon chain (-(CF\texorpdfstring{$_2$}{2})$_9$CF\texorpdfstring{$_3$}{3}) extends \textbf{$\sim$17 \AA{}} above the surface, 
providing an elongated fluorophilic domain. Populated thickness $\approx$ 17.36 \AA{} (vacuum $\approx$ 29.34 \AA{}). 
Compared to shorter PFOTS variants, the C12 tail offers greater van der Waals contact area, 
potentially favoring binding of longer-chain PFAS (PFOS, PFNA) over shorter ones (PFHxS). 
The Si–O–C(Me)\texorpdfstring{$_3$}{3} anchor affords compact attachment while preserving the extended fluorinated interface.

\subparagraph{\textbf{C2 --- \texorpdfstring{$\beta$}{beta}-Ketoenamine-Linked 2D COF + MIP Cavity (In-Plane)}}
Hexagonal P1 with a = b = 58.18 \AA{}, \texorpdfstring{$\gamma$}{gamma} = 120°, c = 48.02 \AA{} (area $\approx$ 2931.40 \AA{}$^2$). 
Substrate is a 2$\times$2$\times$1 supercell of a \textbf{\texorpdfstring{$\beta$}{beta}-ketoenamine linked covalent organic framework (COF)} 
bearing the same \textbf{methacrylic-acid MIP cavity} as in A2, now integrated \textbf{within the COF plane at the same z-height}. 
The COF layer and MIP cluster coexist laterally (z $\approx$ 4--12 \AA{}), creating a heterogeneous 2D surface 
where the cavity functions as an \textbf{in-plane defect/pore modification}. Populated thickness $\approx$ 8.14 \AA{} 
(vacuum $\approx$ 39.88 \AA{}). This geometry promotes \textbf{lateral diffusion} of PFAS along the COF to the embedded site, 
restricting purely vertical access; the shallower rim may enhance \textbf{size/shape selectivity} by guiding linear PFAS into the cavity 
while disfavoring bulky interferents (e.g., humic substances).

\subparagraph{\textbf{C3 --- 2D Ti\texorpdfstring{$_3$}{3}C\texorpdfstring{$_2$}{2}O\texorpdfstring{$_2$}{2} MXene + C12-PFOTS + AgNP + PEDOT:PSS}}
Hexagonal P1 with a = b = 30.67 \AA{}, \texorpdfstring{$\gamma$}{gamma} = 120°, c = 46.70 \AA{} (area $\approx$ 814.63 \AA{}$^2$). 
Derived from C1, the O-terminated \textbf{2D Ti\texorpdfstring{$_3$}{3}C\texorpdfstring{$_2$}{2}} MXene supports a single \textbf{C12-PFOTS} chain and is augmented by 
(i) a \textbf{$\sim$1 nm Ag$_{43}$ nanoparticle cluster} positioned near the extended fluorocarbon tail, generating 
\textbf{plasmonic hotspots} along the chain, and (ii) \textbf{PEDOT} and \textbf{PSS} monomers as conductive spacers to maintain openness. 
Populated thickness $\approx$ 17.36 \AA{} (vacuum $\approx$ 29.34 \AA{}). The extended C12 chain, coupled with the AgNP near-field, 
creates a more expansive \textbf{electromagnetic enhancement zone} tailored to longer-chain PFAS (PFOS, PFNA). 
\textbf{PEDOT:PSS} discourages MXene restacking and introduces \textbf{sulfonate sites (from PSS)} that may modulate selectivity via 
electrostatic interactions with PFAS sulfonate headgroups.

\subsubsection{Selectivity Scoring for PFAS Sensor Performance}
\begin{figure}[htbp]
  \centering
  \includegraphics[width=1\textwidth]{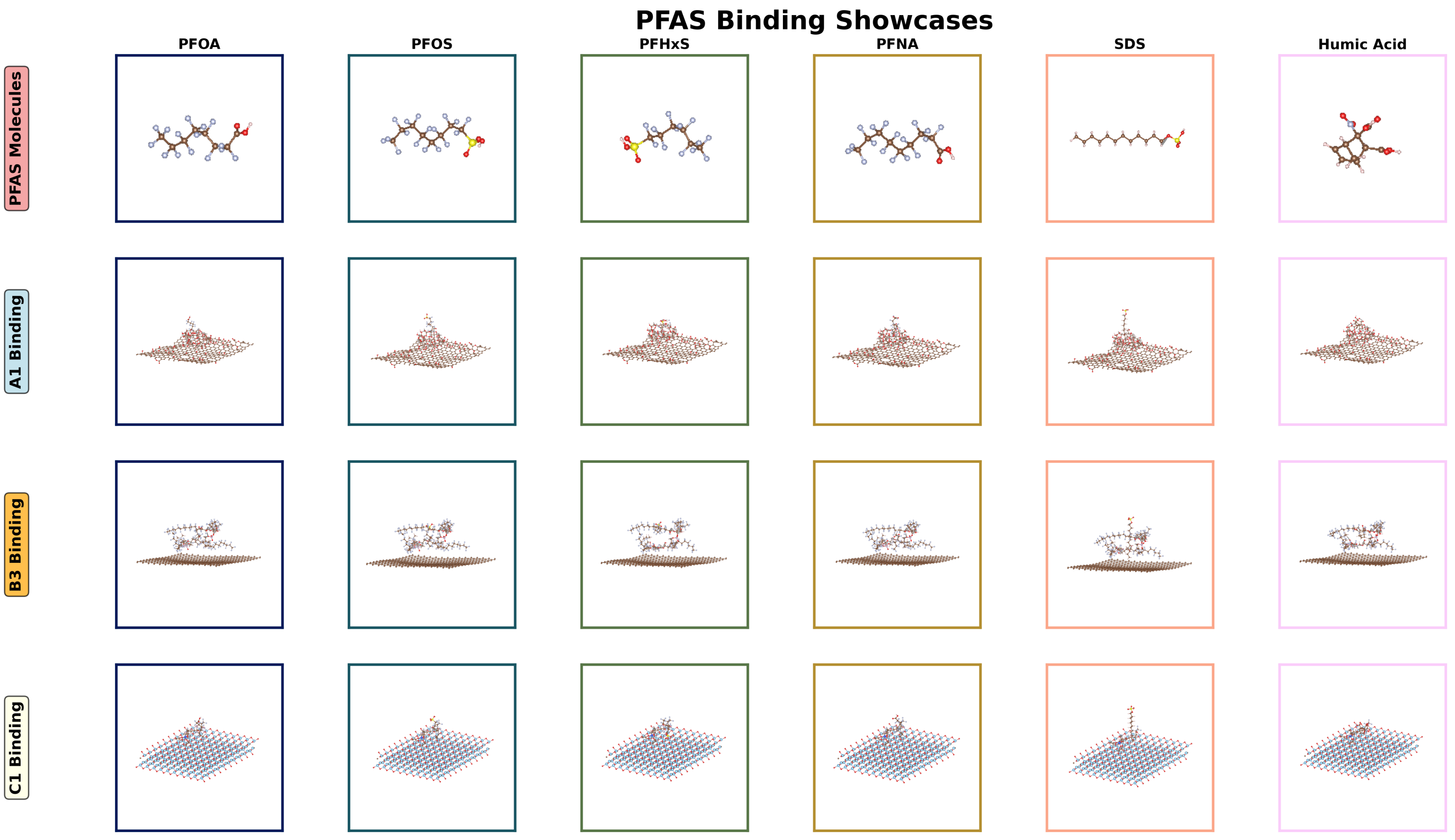}
  \caption{Representative PFAS compounds (PFOA, PFOS, PFNA, PFHxS) and interferent molecules (SDS and humic acid) employed in selectivity assessment, with corresponding optimized binding configurations. A number of candidates' relaxed structures (proposed by different DR agents) are shown.}  
  \label{fig:DL_1_binding_showcase} 
\end{figure}

\begin{figure}[htbp]
  \centering
  \includegraphics[width=1\textwidth]{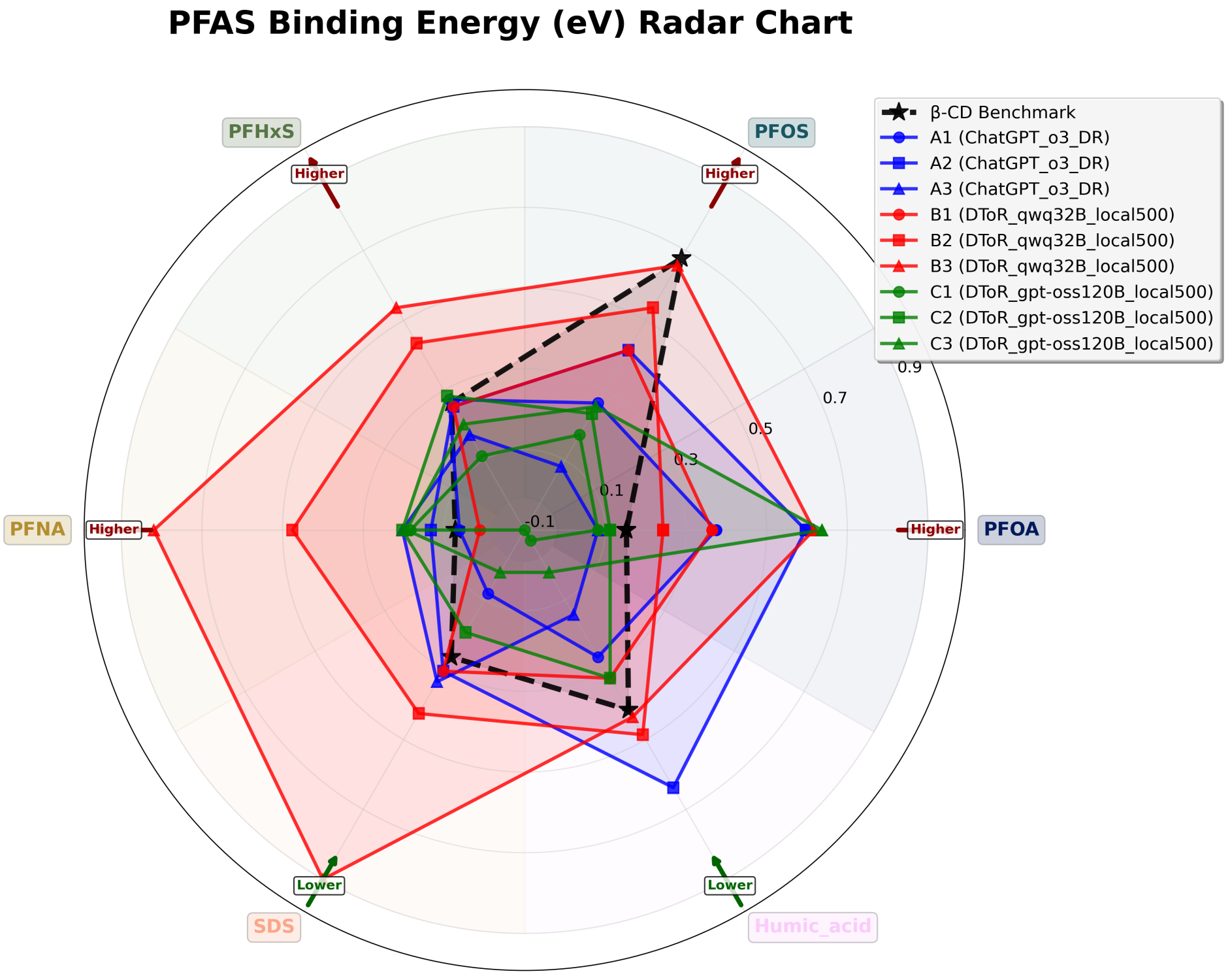}
  \caption{Radar plot comparing binding energies ($\Delta E_{\text{bind}}$) across all candidate systems and benchmark \texorpdfstring{$\beta$}{beta}-CD obtained from static DFT calculations with implicit solvent.}  
  \label{fig:DL_1_radar_DFT_binding_energy} 
\end{figure}

\begin{figure}[htbp]
  \centering
  \includegraphics[width=1\textwidth]{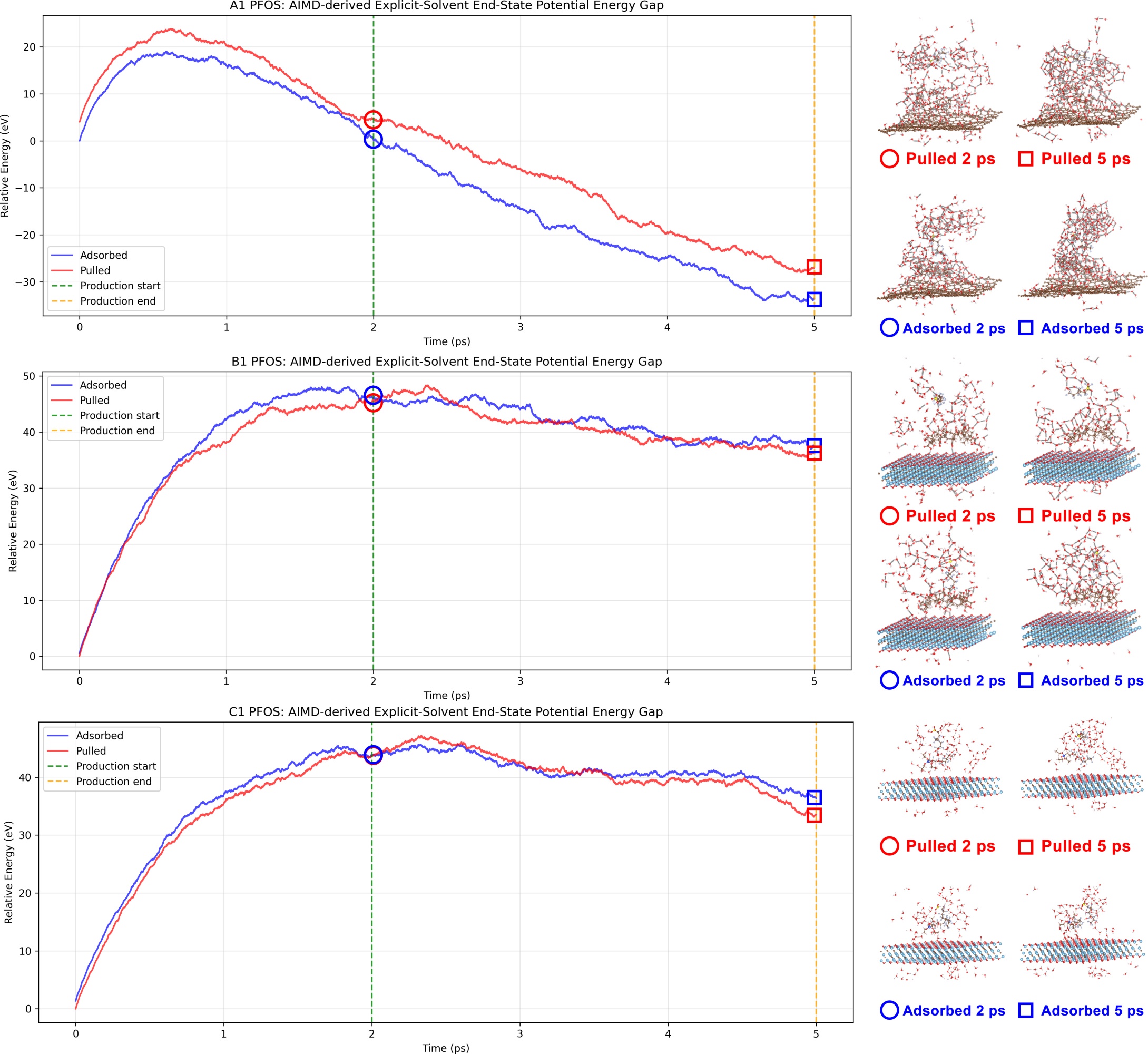}
  \caption{Representative AIMD trajectory showing relative energy evolution under explicit solvent conditions. The green dashed line denotes the 2 ps equilibration period. Right panels display system configurations at highlighted time points along the production trajectory.}  
  \label{fig:DL_1_AIMD_showcase} 
\end{figure}

\begin{figure}[htbp]
  \centering
  \includegraphics[width=1\textwidth]{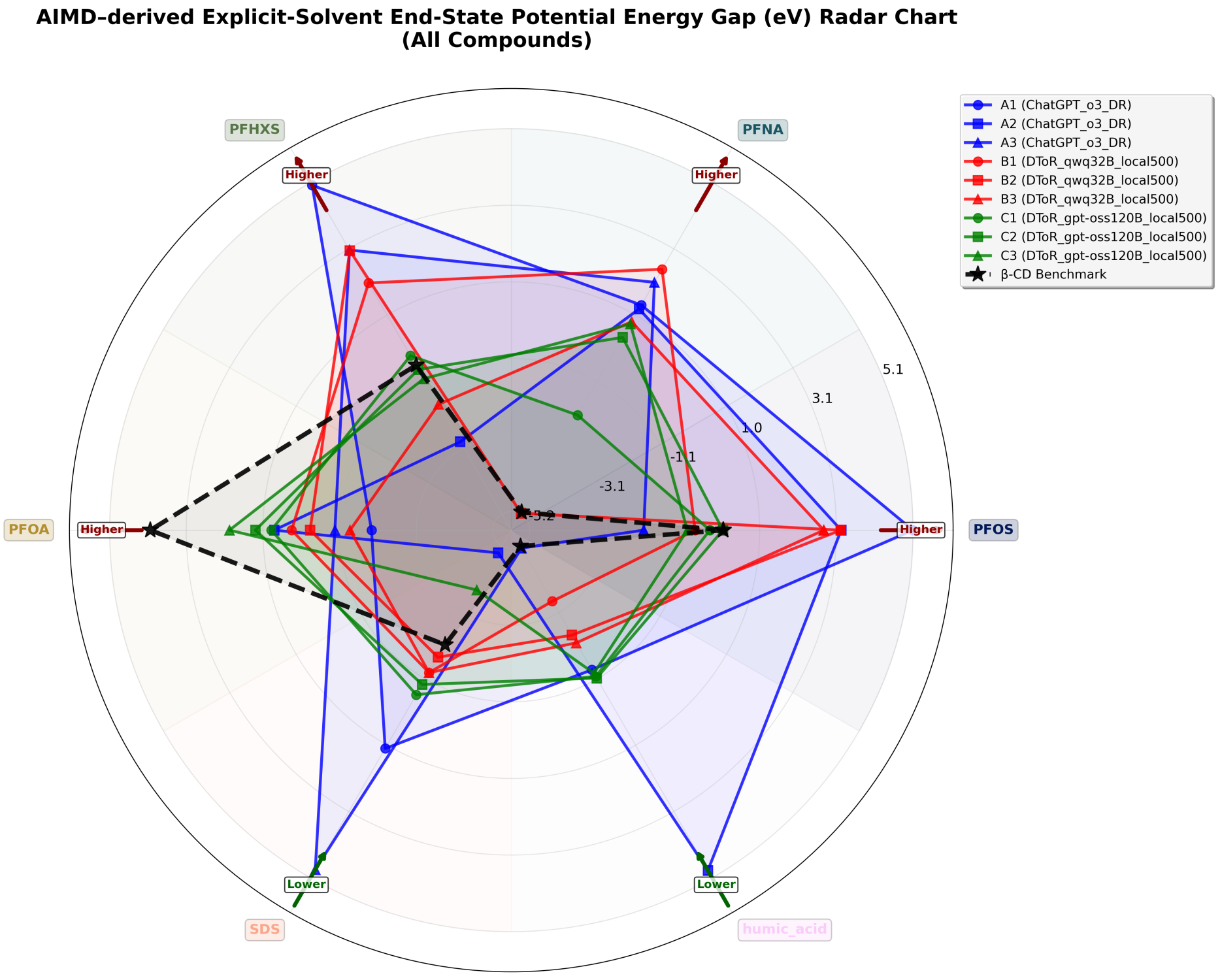}
  \caption{Radar plot comparing explicit-solvent end-state potential energy gaps ($\Delta E_{\text{pot}}^{\text{ES}}$) averaged over the 3 ps production period for all candidate systems and benchmark \texorpdfstring{$\beta$}{beta}-CD from AIMD simulations.}  
  \label{fig:DL_1_radar_AIMD} 
\end{figure}

\begin{figure}[htbp]
  \centering
  \includegraphics[width=1\textwidth]{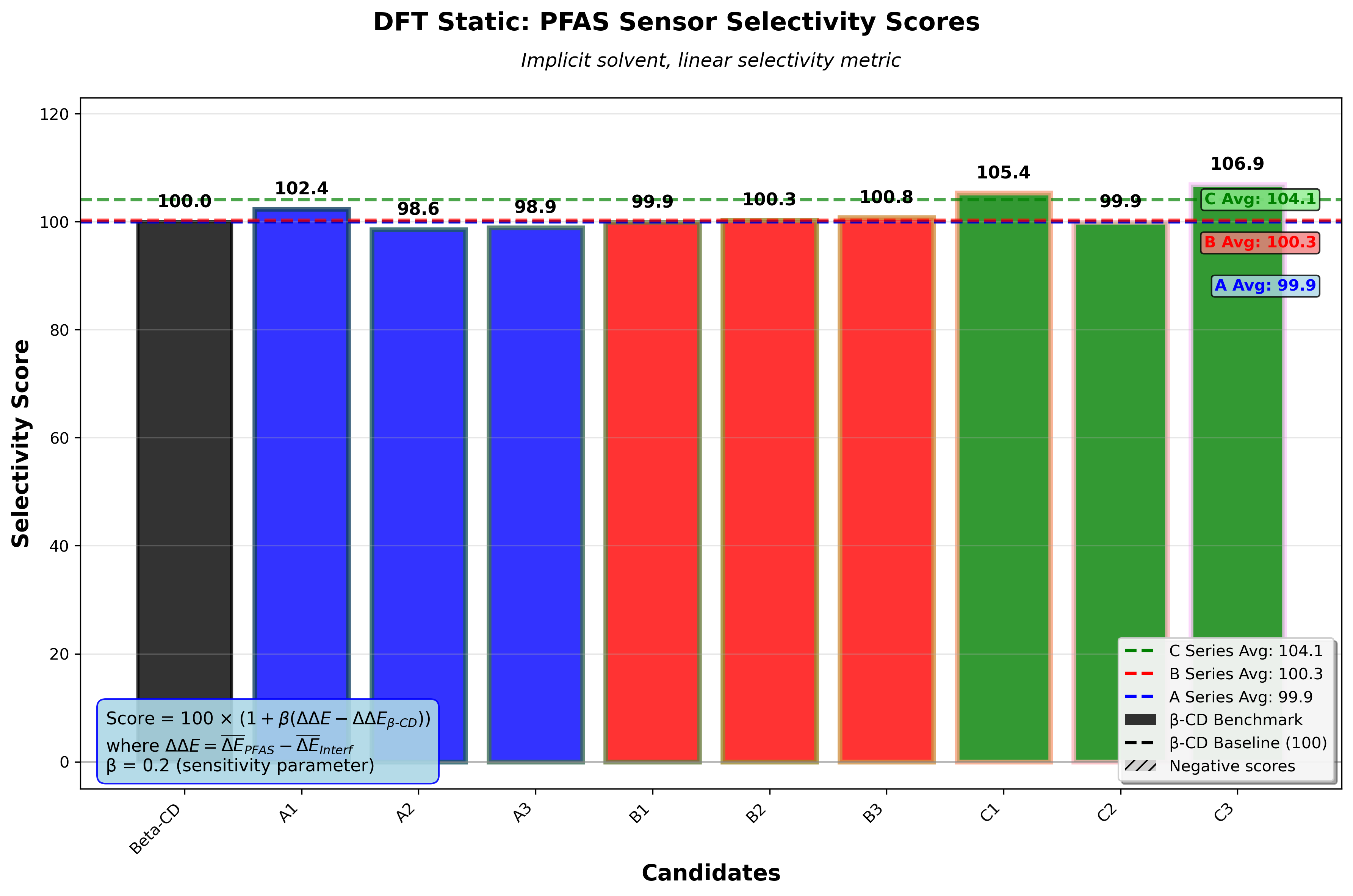}
  \caption{Selectivity scores for all candidate systems from static DFT calculations, normalized to \texorpdfstring{$\beta$}{beta}-CD baseline (100). Scores are calculated using the linear selectivity metric with sensitivity parameter $\beta = 0.2$, where $\Delta\Delta E = \overline{\Delta E}_{\text{PFAS}} - \overline{\Delta E}_{\text{Interf}}$ represents the difference between average PFAS and interferent binding energies.}  
  \label{fig:DL_1_scoring_static_DFT} 
\end{figure}

The practical deployment of PFAS sensors demands not only strong binding to target per- and polyfluoroalkyl substances but also selective discrimination against common environmental interferents. To quantitatively evaluate this selectivity, we developed a unified scoring framework applicable to both static DFT calculations and explicit-solvent AIMD simulations, enabling direct comparison between
computational methodologies while maintaining thermodynamic rigor.
\textbf{Selectivity Metric Derivation}: The fundamental selectivity metric $\Delta\Delta E$ quantifies the differential binding preference between PFAS compounds and interferent molecules:

\begin{equation}
\Delta\Delta E = \overline{\Delta E}_{\text{PFAS}} - \overline{\Delta E}_{\text{Interf}}
\end{equation}

where $\overline{\Delta E}_{\text{PFAS}}$ represents the average binding energy across four representative PFAS compounds (PFOA, PFOS, PFNA, and PFHxS), and $\overline{\Delta E}_{\text{Interf}}$ denotes the average binding energy for interferent species (humic acid and SDS), as illustrated in Figure~\ref{fig:DL_1_binding_showcase}. A larger positive $\Delta\Delta E$ value indicates superior
selectivity, signifying stronger preferential binding to PFAS over interferents.
For static DFT calculations, $\Delta E$ values correspond to the conventional binding energies $\Delta E_{\text{bind}}$ computed under implicit solvent conditions (Figure~\ref{fig:DL_1_radar_DFT_binding_energy}). In explicit-solvent AIMD simulations, these are replaced by the time-averaged end-state potential energy gaps $\Delta E_{\text{pot}}^{\text{ES}}$, which capture solvent reorganization
effects and hydrogen-bonding network rearrangements absent from static calculations (Figures~\ref{fig:DL_1_AIMD_showcase} and \ref{fig:DL_1_radar_AIMD}).
Linear Selectivity Scoring Framework: To enable meaningful comparison across candidate systems and computational methods, we normalize the selectivity metrics to a common scoring scale:

\begin{equation}
\text{Score} = 100 \times \left(1 + \beta(\Delta\Delta E - \Delta\Delta E_{\beta\text{-CD}})\right)
\end{equation}

where $\Delta\Delta E_{\beta\text{-CD}}$ represents the selectivity of \texorpdfstring{$\beta$}{beta}-cyclodextrin, our domain benchmark reference with established PFAS binding capability \cite{wang2024reversible}. The sensitivity parameter $\beta = 0.2$ was selected to provide appropriate score discrimination while maintaining numerical stability across both DFT and AIMD datasets. This linear transformation preserves
the physical meaning of energy differences while establishing \texorpdfstring{$\beta$}{beta}-CD as the baseline reference (Score = 100).
\textbf{Comparative Analysis of Selectivity Under Different Solvation Models}: The application of this unified scoring framework reveals striking differences between implicit and explicit solvation treatments. In static DFT calculations with implicit solvent, system A1 achieves the highest selectivity score of 102.4, outstandingly exceeding the \texorpdfstring{$\beta$}{beta}-CD baseline (Figure~\ref{fig:DL_1_scoring_static_DFT}). The B-series candidates also prove competitive as B2 and B3 also showed slightly passing 100 score.

\begin{figure}[htbp]
  \centering
  \includegraphics[width=1\textwidth]{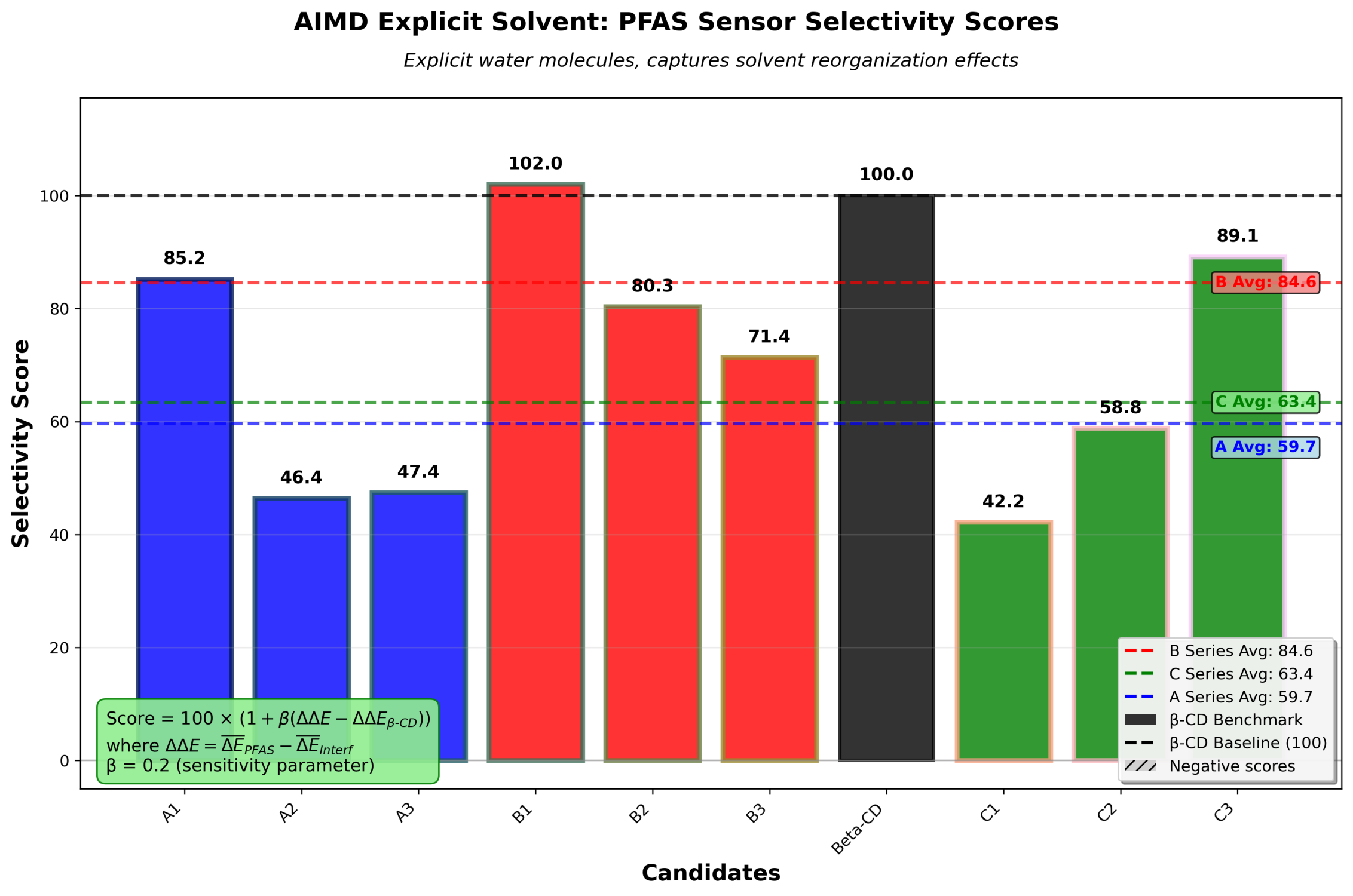}
  \caption{Selectivity scores for all candidate systems from explicit-solvent AIMD simulations, normalized to \texorpdfstring{$\beta$}{beta}-CD baseline (100). The scoring methodology remains identical to static DFT, except $\Delta E$ values are replaced by $\Delta E_{\text{pot}}^{\text{ES}}$, the time-averaged potential energy gaps from the 3 ps production period.}  
  \label{fig:DL_1_scoring_AIMD} 
\end{figure}

However, the inclusion of explicit water molecules and dynamic sampling in AIMD simulations dramatically alters the selectivity landscape (Figure~\ref{fig:DL_1_scoring_AIMD}). Most notably, the majority of the candidate systems' scores are below the \texorpdfstring{$\beta$}{beta}-CD baseline in explicit solvent, with only B1 achieving the highest relative performance at 102.0. This systematic reduction in scores reflects the critical role of
solvent-mediated effects: competitive water adsorption at binding sites, strong hydration of anionic PFAS headgroups, and the entropic penalties associated with constraining flexible molecules in adsorbed configurations.
The reversal of relative rankings between static DFT and AIMD calculations—particularly the superior performance of B-series candidates under explicit solvation—underscores the importance of including solvent dynamics in sensor design evaluation. While static calculations provide valuable initial screening, the explicit-solvent AIMD results more accurately capture the complex interplay between
adsorbate--surface interactions and solvent reorganization that governs selectivity under realistic aqueous conditions.
\textbf{Implications for Sensor Design}: The divergence between implicit and explicit solvent results carries significant implications for PFAS sensor development. Systems optimized solely based on static DFT calculations may exhibit diminished selectivity in aqueous environments due to unforeseen solvent competition effects. Conversely, candidates that appear suboptimal in vacuum or implicit solvent calculations may benefit from favorable solvent-mediated interactions that enhance their practical selectivity. This highlights the necessity of multi-scale computational approaches that progress from efficient static screening to rigorous explicit-solvent validation for promising candidates.

The linear selectivity scoring framework presented here provides a quantitative, thermodynamically grounded metric for evaluating sensor performance across computational methodologies. By maintaining consistent notation and normalization across static and dynamic simulations, this approach enables systematic comparison of candidate materials while preserving the physical interpretability of the
underlying energy differences.
\clearpage

\clearpage

\subsection{PFAS Degradation}\label{dry_lab_PFAS_degra}

\subsubsection{PFAS\_Electro\_Degradation}

For the second application: PFAS Degradation. 
The query as shown in \ref{app:topics} is:
\begin{lstlisting}
Which novel electrode materials can achieve efficient PFAS degradation under ambient aqueous electrochemical conditions, delivering both high mineralization and defluorination rates? What intrinsic properties--such as PFAS adsorption affinity, reactive oxygen species generation capacity, and C-F bond activation energy--should be optimized to guide their discovery?
\end{lstlisting}

The best commercial solution is: \textbf{Gemini\_2.5\_Pro\_DR}.

The best local non gpt-oss solution is: \textbf{DToR\_qwq32B\_local500}.

The best local gpt-oss solution is: \textbf{DToR\_gpt-oss120B\_local500}.

\begin{figure}[htbp]
  \centering
  \includegraphics[width=1\textwidth]{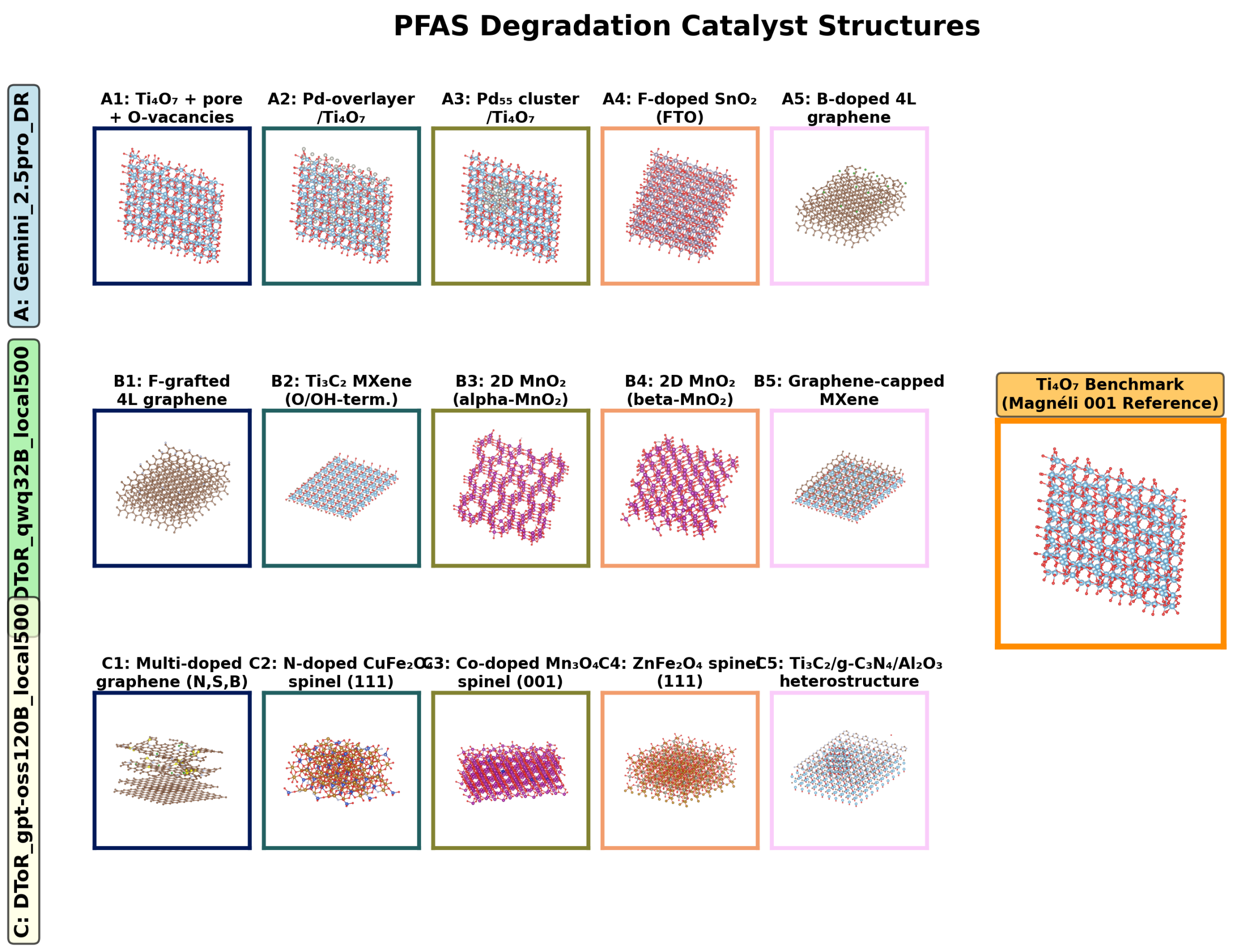}
  \caption{Structural configurations of candidate catalysts proposed by DR agents for PFAS degradation, with Ti\texorpdfstring{$_4$}{4}O\texorpdfstring{$_7$}{7} \protect\cite{lin2018ti4o7} included as the domain benchmark reference.}
  \label{fig:DL_2_candidates_final}  
\end{figure}

The expert based on the two reports, provided 5 commercial DR report derived candidates (A1-A5), and 5 local non gpt-oss DR report derived candidates (B1-B5), and 5 local gpt-oss DR report derived candidates (C1-C5), visualized in Fig. \ref{fig:DL_2_candidates_final}:

\paragraph{\textbf{Ti\texorpdfstring{$_4$}{4}O\texorpdfstring{$_7$}{7} --- Pristine Magnéli (001) bench slab}}
Triclinic P1 slab with a = 22.376 \AA{}, b = 27.651 \AA{}, \texorpdfstring{$\gamma$}{gamma} = 75.306°, c = 42.208 \AA{} (in-plane area $\approx$ 598.50 \AA{}$^2$). Atom counts (Ti/O = 256/448; 704 total) match a Ti\texorpdfstring{$_4$}{4}O\texorpdfstring{$_7$}{7} (001) termination. The populated thickness spans $\approx$ 12.13 \AA{} (z $\approx$ 3.41--15.53 \AA{}), leaving $\approx$ 30.08 \AA{} vacuum; the slab mid-plane lies near z $\approx$ 9.47 \AA{}.
Surfaces are stoichiometric and flat, preserving Magnéli shear-plane connectivity and metallic conduction pathways---our baseline conductive-oxide reference for PFAS tests. Build/relax notes: 4 atomic layers; bottom two constrained. (Adsorbates for subsequent runs were seeded $\approx$ 3 \AA{} above the most protruding Ti site, tail parallel to the surface.)

\subparagraph{\textbf{A1 --- Ti\texorpdfstring{$_4$}{4}O\texorpdfstring{$_7$}{7} (001) with through-pore + rim O-vacancies}}
Triclinic P1 cell with a = 22.376 \AA{}, b = 27.651 \AA{}, \texorpdfstring{$\gamma$}{gamma} = 75.306°, c = 42.208 \AA{} (in-plane area $\approx$ 598.50 \AA{}$^2$). Atom counts (Ti/O = 249/437; 686 total) reflect removal of a top-side cluster and additional rim O-vacancies relative to the bench ($\Delta$ = $-$7 Ti, $-$11 O). The populated thickness is $\approx$ 13.44 \AA{} (z $\approx$ 3.00--16.45 \AA{}), leaving
$\approx$ 28.77 \AA{} vacuum; the slab mid-plane sits near z $\approx$ 9.73 \AA{}. A circular through-pore was introduced by deleting atoms within a $\sim$5 \AA{}-radius cylinder from the upper layers, yielding an effective mouth $\sim$1.0 nm diameter; the rim exposes under-coordinated Ti with a locally reduced O environment, while the surrounding lattice retains Magnéli shear-plane connectivity.
Relaxation protocol used: bottom half constrained; top region (and any adsorbates) relaxed so the pore edge dominates binding/activation. Connection to the Ti\texorpdfstring{$_4$}{4}O\texorpdfstring{$_7$}{7} bench: same lattice, cell vectors, and basal (001) termination as the pristine Ti\texorpdfstring{$_4$}{4}O\texorpdfstring{$_7$}{7} bench slab; bottom region unchanged. All edits are localized to the top surface (pore machining + rim O-vacancies). Away from the pore, Ti/O
stoichiometry per layer and shear-plane topology match the bench reference.

\subparagraph{\textbf{A2 --- Pd-overlayer/Ti\texorpdfstring{$_4$}{4}O\texorpdfstring{$_7$}{7} (001) composite (amorphous Pd on the bench slab)}}
Triclinic P1 cell with a = 22.376 \AA{}, b = 27.651 \AA{}, \texorpdfstring{$\gamma$}{gamma} = 75.306°, c = 42.21 \AA{} (in-plane area $\approx$ 598.50 \AA{}$^2$). Atom counts (Ti/O/Pd = 256/448/94; 798 total) indicate a stoichiometric Ti\texorpdfstring{$_4$}{4}O\texorpdfstring{$_7$}{7} substrate capped by a Pd overlayer. The populated thickness spans $\approx$ 16.45 \AA{} (z $\approx$ 3.14--19.59 \AA{}), leaving $\approx$ 25.76 \AA{} vacuum; mid-plane at z
$\approx$ 11.37 \AA{}. Pd atoms populate the top terrace broadly (no single island), forming a disordered sheet with Pd heights $\sim$16.12--19.59 \AA{} (mean $\approx$ 17.71 \AA{}), i.e., $\sim$3.1 \AA{} above the top oxide layer (z$_{\text{ox,max}}$ $\approx$ 16.49 \AA{}). \textbf{Connection to the Ti\texorpdfstring{$_4$}{4}O\texorpdfstring{$_7$}{7} bench:} identical lattice and Ti/O stoichiometry to the pristine Magnéli (001) slab;
modifications are limited to a Pd overlayer on the top surface (bottom half retained as in the bench), with no pores or O-vacancies introduced.

\subparagraph{\textbf{A3 --- Pd\texorpdfstring{$_{55}$}{55} cluster on Ti\texorpdfstring{$_4$}{4}O\texorpdfstring{$_7$}{7} (001) (icosahedral-like nanoparticle on the bench slab)}}
Triclinic P1 cell with a = 22.376 \AA{}, b = 27.651 \AA{}, \texorpdfstring{$\gamma$}{gamma} = 75.306°, c = 42.21 \AA{} (in-plane area $\approx$ 598.50 \AA{}$^2$). Atom counts (Ti/O/Pd = 256/448/55; 759 total) show a Ti\texorpdfstring{$_4$}{4}O\texorpdfstring{$_7$}{7} substrate supporting a compact Pd nanoparticle. The populated thickness spans $\approx$ 22.65 \AA{} (z $\approx$ 3.23--25.87 \AA{}), leaving $\approx$ 19.56 \AA{} vacuum; mid-plane at z $\approx$
14.55 \AA{}. The Pd cluster (initially icosahedral; allowed to relax) (n = 55) is localized on the top terrace with a projected footprint $\sim$9.2 $\times$ 9.6 \AA{} (max in-plane radius $\approx$ 5.33 \AA{}; projected diameter $\approx$ \textbf{10.65 \AA{}}) and vertical extent placing the apex $\sim$9.7 \AA{} above the top oxide layer (z$_{\text{ox,max}}$ $\approx$ 16.15 \AA{}).
\textbf{Connection to the Ti\texorpdfstring{$_4$}{4}O\texorpdfstring{$_7$}{7} bench:} same lattice and stoichiometric oxide as the pristine (001) slab; the only alteration is a single Pd\texorpdfstring{$_{55}$}{55} nanoparticle anchored on the top surface (no pore machining or O-vacancies), with the bottom region kept as in the bench reference.

\subparagraph{\textbf{A4 --- F-doped rutile SnO\texorpdfstring{$_2$}{2} (110) slab (FTO, O\texorpdfstring{$\rightarrow$}{->}F substitution near surface)}}
Orthogonal P1 cell with a = 26.8022 \AA{}, b = 25.4880 \AA{}, \texorpdfstring{$\gamma$}{gamma} = 90.0°, c = 42.1012 \AA{}; in-plane area $\approx$ 683.13 \AA{}$^2$. The metrics correspond to a 4$\times$8 SnO\texorpdfstring{$_2$}{2}(110) supercell (a $\approx$ 4$\times$a$_{\text{(rutile)}}$$\sqrt{2}$, b $\approx$ 8$\times$c$_{\text{(rutile)}}$). Atom counts (Sn/O/F = 256/497/15; 768 total) indicate $\sim$2.9\% O\texorpdfstring{$\rightarrow$}{->}F substitution on
the anion sublattice (15/512), i.e., a lightly F-doped ``FTO'' slab. The populated thickness spans $\approx$ 14.28 \AA{} (z $\approx$ 4.32--18.60 \AA{}), leaving $\approx$ 27.83 \AA{} vacuum; Sn layers occupy z $\approx$ 4.38--15.29 \AA{}, while F sites are confined to a narrow band near the top (z $\approx$ 16.33--17.17 \AA{}; mean $\approx$ 16.75 \AA{}), just beneath the outermost O ridges
(O$_{\text{max}}$ $\approx$ 18.60 \AA{}). This places F primarily in near-surface O positions on (110), preserving the characteristic bridging-O/in-plane-O relief of rutile and providing a slightly more electron-rich, less hydroxylated top termination expected for FTO.

\subparagraph{\textbf{A5 --- B-doped four-layer graphene}}
Hexagonal P1; a = b = 19.68 \AA{}, \texorpdfstring{$\gamma$}{gamma} = 120°, \textbf{c $\approx$ 36.0 \AA{}}. Four graphene sheets stacked with the standard \textbf{interlayer spacing d(002) $\approx$ 3.35 $\pm$ 0.05 \AA{}} (three gaps \texorpdfstring{$\rightarrow$}{->} $\sim$10.05 \AA{} slab thickness). Substitutional B is confined to the \textbf{top layer} ($\approx$10 at\% within that plane; $\approx$2--3 at\% overall), preserving an
sp$^2$ framework. The remaining \textbf{$\sim$26 \AA{} is vacuum} along c to suppress image interactions. During relaxation, we fixed the bottom two layers; relax the top two (and any adsorbates).

\subparagraph{\textbf{B1 --- F-grafted four-layer graphene (consensus spacing; A5-consistent)}}
Hexagonal P1; \textbf{keep A5's in-plane metrics (a = b, \texorpdfstring{$\gamma$}{gamma} = 120°) unchanged} and \textbf{c $\approx$ 36.0 \AA{}}. Stack \textbf{four graphene sheets with the standard interlayer spacing d(002) = 3.35 $\pm$ 0.05 \AA{}} (three gaps \texorpdfstring{$\rightarrow$}{->} $\sim$10.05 \AA{} slab thickness), with the \textbf{remaining $\sim$26 \AA{} as vacuum} along c to decouple periodic images.
\textbf{Chemistry/dopant pattern exactly as in A5} (no changes to which layer(s) are doped/functionalized); the carbon network remains predominantly sp$^2$. \textbf{Same Relaxation protocol as A5:} fix the bottom two layers; relax the top two (and any adsorbates). \textbf{Connection to A5:} identical spacing convention and vacuum budget as A5 to ensure cross-comparable adsorption energies and
electrostatics across the graphene series, while leaving only the dopant/functional differences to drive behavior.

\subparagraph{\textbf{B2 --- 2D Ti\texorpdfstring{$_3$}{3}C\texorpdfstring{$_2$}{2} MXene (O/OH-terminated)}}
Hexagonal P1 slab with a = b = 24.536 \AA{}, \texorpdfstring{$\gamma$}{gamma} = 120°, c = 36.700 \AA{} (in-plane area $\approx$ 521.36 \AA{}$^2$). Atom counts (Ti/C/O/H = 192/128/128/114; 562 total) correspond to \textbf{64 f.u. of Ti\texorpdfstring{$_3$}{3}C\texorpdfstring{$_2$}{2}} with full two-sided terminations (128/site total): $\approx$\textbf{89\% --OH} (114 H) and $\approx$\textbf{11\% =O} (14 O), distributed symmetrically per face (\textbf{57 --OH +
7 =O} each side). The vacuum layer is around \textbf{27.21 \AA{}}. The sheet preserves the characteristic \textbf{2D MXene} core (Ti--C--Ti trilayers) with mixed O/OH terminations pointing outward, yielding a hydrophilic, electronically conductive surface appropriate for adsorption tests (e.g., PFAS headgroup coordination via --OH domains).

\subparagraph{\textbf{B3 --- 2D MnO\texorpdfstring{$_2$}{2}-like (birnessite-style) nanosheet with sparse one-sided --OH}}
Orthogonal P1 cell with a = b = 29.3628 \AA{}, \texorpdfstring{$\gamma$}{gamma} = 90.0°, c = 33.595 \AA{} (in-plane area $\approx$ 862.17 \AA{}$^2$). Atom counts (Mn/O/H = 216/411/18; 645 total) are consistent with a near-stoichiometric MnO\texorpdfstring{$_2$}{2} layer carrying light hydroxylation (H:Mn $\approx$ 0.083). The populated thickness is $\approx$ 9.32 \AA{} (z $\approx$ 0.82--10.14 \AA{}), leaving $\approx$ 24.28 \AA{} vacuum.
Element-resolved spans: Mn $\approx$ 1.52--9.59 \AA{}; O $\approx$ 0.82--9.61 \AA{}; H confined to $\approx$ 7.12--10.14 \AA{} on the outer face, indicating \textbf{one-sided --OH/water terminations} on a \textbf{2D} octahedral sheet. Relaxation note: we kept the lower half constrained and relaxed the top half/adsorbates; head-group binding is expected at surface --OH/O sites with the basal Mn--O
framework remaining robust.

\subparagraph{\textbf{B4 --- 2D MnO\texorpdfstring{$_2$}{2}-like (birnessite-style) nanosheet, smaller supercell; one-sided --OH}}
Orthogonal P1 cell with a = b = 22.000 \AA{}, \texorpdfstring{$\gamma$}{gamma} = 90.0°, c = 33.610 \AA{} (in-plane area = 484.00 \AA{}$^2$). Composition (Mn/O/H = 150/285/12; 447 total) again reflects a MnO\texorpdfstring{$_2$}{2} sheet with light hydroxylation (H:Mn $\approx$ 0.08). The populated thickness is $\approx$ 9.40 \AA{} (z $\approx$ 1.77--11.17 \AA{}), leaving $\approx$ 24.21 \AA{} vacuum. Element-resolved spans: Mn $\approx$
2.12--10.48 \AA{}; O $\approx$ 1.77--10.99 \AA{}; H $\approx$ 8.43--11.17 \AA{}, i.e., \textbf{--OH/water groups localized on the top face} of the \textbf{2D} sheet. Relaxation matches B3 (bottom region constrained, top free) to keep comparisons clean across the MnO\texorpdfstring{$_2$}{2} series.

\subparagraph{\textbf{B5 --- Graphene-capped 2D MXene (derived from B2)}}
Derived directly from \textbf{B2} (same cell vectors and terminations), a \textbf{single-layer graphene} sheet is added parallel to the top face of the MXene and centered to span the full in-plane cell. The initial MXene$\leftrightarrow$graphene separation is set to \textbf{$\sim$3.35 \AA{}} (graphite d(002)), with \textbf{no covalent links}---pure van der Waals contact. Graphene is pristine
sp$^2$ and allowed to relax laterally/vertically; the MXene beneath follows B2's relaxation protocol (bottom constrained, top free). The \textbf{c-axis vacuum is kept as in B2} to avoid inter-image coupling. \textbf{Connection to B2:} identical substrate; the only modification is the graphene capping layer.

\subparagraph{\textbf{C1 --- Multi-doped graphene (N,S,B co-doping with dual-layer defects)}}
Hexagonal P1; layer count and supercell dimensions (a = b = 19.68 \AA{}, \texorpdfstring{$\gamma$}{gamma} = 120°, c $\approx$ 36.0 \AA{}) kept consistent with A5 and B1. Four graphene sheets stacked with standard interlayer spacing d(002) $\approx$ 3.35 \AA{} (three gaps \texorpdfstring{$\rightarrow$}{->} $\sim$10.05 \AA{} nominal slab thickness), modified by extensive heteroatom doping and defect engineering across the top two layers. The
populated thickness spans $\sim$10.05 \AA{} (z $\approx$ 4.00--14.05 \AA{}), with the structure deliberately positioned low in the supercell to leave $\sim$19 \AA{} of vacuum above the top surface, providing ample space for PFAS adsorbate placement and subsequent geometry optimization. The multi-doping architecture combines B--N nearest-neighbor pairs ($\sim$2.5 \AA{} separation for
electron-deficient/rich adjacency) and N--S nearest-neighbor pairs ($\sim$2.5 \AA{} separation for direct electron transfer and reactive oxygen species synergy), distributed across both the top layer (3 B--N pairs + 3 N--S pairs + 5 isolated dopants) and second layer (2 B--N pairs + 2 N--S pairs + 3 isolated dopants). A localized ``hole'' defect ($\sim$4 \AA{} radius cluster, 8 atoms removed) is
introduced at the supercell center in the top layer to expose under-coordinated edge sites and create a nano-pore for enhanced PFAS molecular access, complemented by scattered single-atom vacancies (3\% in top layer, 5\% in second layer). Total heteroatom loading reaches $\sim$20--25\% in the top two layers (13 N + 8 S + 7 B across 495 total atoms). This dual-layer functionalization strategy
ensures multi-modal catalytic pathways: electron-transfer modulation via B--N pairs, oxidative radical generation via N--S pairs, and defect-mediated C--F bond polarization around the central nano-pore and distributed vacancy sites.

\subparagraph{\textbf{C2 --- N-doped CuFe\texorpdfstring{$_2$}{2}O\texorpdfstring{$_4$}{4} spinel (111) with oxygen vacancies}}
Triclinic P1 slab with a = 16.81 \AA{}, b = 18.64 \AA{}, $\gamma \approx$ 120° (hexagonal-like in-plane symmetry), c = 47.58 \AA{} (in-plane area $\approx$ 313.5 \AA{}$^2$), representing a 2$\times$2 surface expansion of the CuFe\texorpdfstring{$_2$}{2}O\texorpdfstring{$_4$}{4} (111) spinel termination. Atom counts (Cu/Fe/N/O = 64/128/11/240; 443 total) reflect nitrogen substitution in the surface oxygen sublattice ($\sim$11\% of
surface O sites replaced by N) and controlled oxygen-vacancy formation (5 vacancies, $\sim$7\% of remaining surface O). The populated thickness spans $\approx$ 14.58 \AA{} (z $\approx$ 4.00--18.58 \AA{}), leaving $\approx$ 29 \AA{} of vacuum above the top surface for PFAS adsorbate placement and explicit-solvent AIMD simulations. The (111) orientation exposes a mixed-termination surface with both
octahedral Fe$^{3+}$ and tetrahedral Cu$^{2+}$ coordination environments. N-doping is distributed on the surface oxygen sublattice, while the engineered O-vacancies create under-coordinated metal centers (exposed Fe$^{3+}$/Cu$^{2+}$ sites).

\subparagraph{\textbf{C3 --- Co-doped Mn\texorpdfstring{$_3$}{3}O\texorpdfstring{$_4$}{4} spinel (001) with oxygen vacancies}}
Orthorhombic P1 slab with a = 33.07 \AA{}, b = 17.23 \AA{}, c = 46.60 \AA{} (in-plane area $\approx$ 569.6 \AA{}$^2$), representing a 3$\times$3 surface expansion of the Mn\texorpdfstring{$_3$}{3}O\texorpdfstring{$_4$}{4} (001) spinel termination. Atom counts (Co/Mn/O = 6/318/429; 753 total) reflect cobalt substitution in subsurface octahedral Mn sites ($\sim$5\% of subsurface Mn) and controlled oxygen-vacancy formation (3 vacancies,
$\sim$2.5\% of surface O). The populated thickness spans $\approx$ 13.60 \AA{} (z $\approx$ 4.00--17.60 \AA{}), leaving $\approx$ 29 \AA{} of vacuum above the top surface for PFAS adsorbate placement and explicit-solvent AIMD simulations. The (001) orientation exposes a three-dimensional spinel framework with mixed-valence Mn$^{2+}$/Mn$^{3+}$ coordination environments, distinct from the layered
birnessite-MnO\texorpdfstring{$_2$}{2} structures (B3, B4). Co-doping is localized in the subsurface layer (positioned between surface and core, $\sim$2.5--5.0 \AA{} from the top surface), while the engineered O-vacancies are distributed on the surface.

\subparagraph{\textbf{C4 --- ZnFe\texorpdfstring{$_2$}{2}O\texorpdfstring{$_4$}{4} spinel (111) with oxygen vacancies}}
Triclinic P1 slab with a = 23.86 \AA{}, b = 20.66 \AA{}, $\gamma \approx$ 120° (hexagonal-like in-plane symmetry), c = 50.33 \AA{} (in-plane area $\approx$ 493.0 \AA{}$^2$), representing a 2$\times$2 surface expansion of the ZnFe\texorpdfstring{$_2$}{2}O\texorpdfstring{$_4$}{4} (111) spinel termination. Atom counts (Fe/O/Zn = 256/506/128; 890 total) reflect controlled oxygen-vacancy formation (6 vacancies, $\sim$5\% of surface O). The
populated thickness spans $\approx$ 17.33 \AA{} (z $\approx$ 4.00--21.33 \AA{}), leaving $\approx$ 29 \AA{} of vacuum above the top surface for PFAS adsorbate placement and explicit-solvent AIMD simulations. The (111) orientation exposes a mixed-termination surface with tetrahedral Zn$^{2+}$ and octahedral Fe$^{3+}$ coordination environments. The engineered O-vacancies are distributed on the
surface, creating under-coordinated metal centers (exposed Fe$^{3+}$/Zn$^{2+}$ sites).

\subparagraph{\textbf{C5 --- Ti\texorpdfstring{$_3$}{3}C\texorpdfstring{$_2$}{2} MXene/g-C\texorpdfstring{$_3$}{3}N\texorpdfstring{$_4$}{4}/Al\texorpdfstring{$_2$}{2}O\texorpdfstring{$_3$}{3} cluster heterostructure}}
Hexagonal P1 slab with a = b = 30.67 \AA{}, \texorpdfstring{$\gamma$}{gamma} = 120°, c = 51.70 \AA{} (in-plane area $\approx$ 704.6 \AA{}$^2$), representing a three-layer heterostructure with distinct functional regions. Atom counts (Al/C/H/N/O/Ti = 55/347/60/196/254/300; 1212 total) reflect a Ti\texorpdfstring{$_3$}{3}C\texorpdfstring{$_2$}{2} MXene base layer with mixed oxygen/hydroxyl termination ($\sim$70\% =O, $\sim$30\% --OH; 140 =O + 60 --OH groups
distributed on both surfaces), a graphitic carbon nitride (g-C\texorpdfstring{$_3$}{3}N\texorpdfstring{$_4$}{4}) middle layer (343 atoms: 147 C + 196 N, derived from a 7$\times$7 supercell expansion of the literature s-C\texorpdfstring{$_3$}{3}N\texorpdfstring{$_4$}{4} structure and strained $\sim$8.4\% to match the MXene lattice, positioned $\sim$3.35 \AA{} above MXene), and an octahedral Al\texorpdfstring{$_2$}{2}O\texorpdfstring{$_3$}{3} cluster (109 atoms: 55 Al + 54 O) centered on the top surface (positioned 7
\AA{} above g-C\texorpdfstring{$_3$}{3}N\texorpdfstring{$_4$}{4}). The populated thickness spans $\sim$26.6 \AA{} (z $\approx$ 5.62--32.22 \AA{}), with the MXene core occupying z $\approx$ 5.62--16.92 \AA{}, the g-C\texorpdfstring{$_3$}{3}N\texorpdfstring{$_4$}{4} layer at z $\approx$ 20.27 \AA{}, and the Al\texorpdfstring{$_2$}{2}O\texorpdfstring{$_3$}{3} cluster centered at z $\approx$ 27.27 \AA{}.

\subsubsection{Multi-Metric Evaluation of PFAS Degradation Catalysts}

Effective catalytic degradation of PFAS compounds requires both strong surface adsorption
to concentrate the pollutants and sufficient activation of the exceptionally stable C--F bonds.
To comprehensively evaluate candidate catalysts for PFAS degradation, we employed a three-pronged
computational assessment framework encompassing static binding energies, explicit-solvent dynamics,
and C--F bond dissociation energetics, with Ti\texorpdfstring{$_4$}{4}O\texorpdfstring{$_7$}{7} serving as the domain benchmark reference
\cite{lin2018ti4o7}.

\textbf{Integrated Performance Metrics}

\begin{figure}[htbp]
  \centering
  \includegraphics[width=1\textwidth]{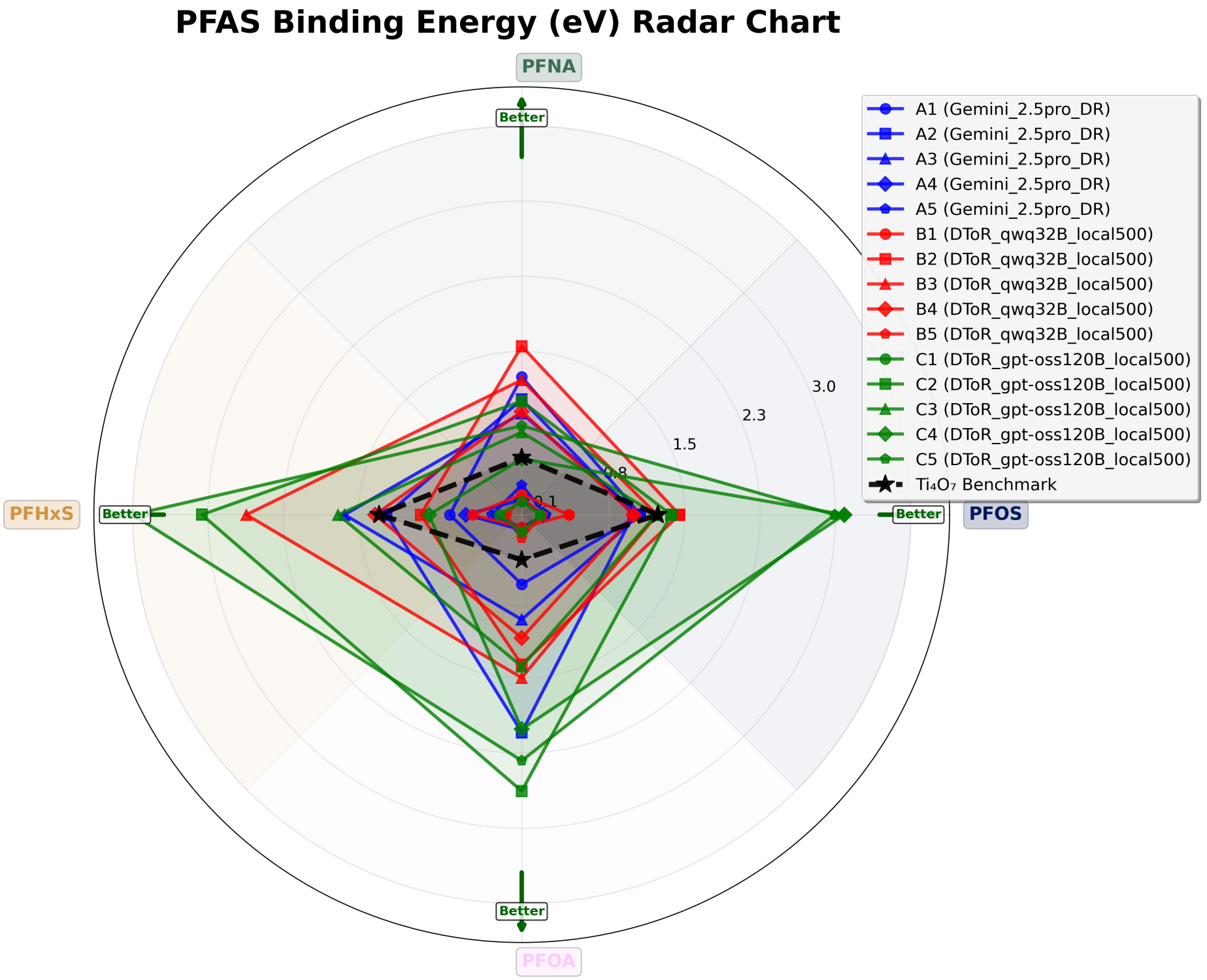}
  \caption{Radar plot comparing PFAS binding energies ($\Delta E_{\text{bind}}$) across all candidate catalysts and benchmark Ti\texorpdfstring{$_4$}{4}O\texorpdfstring{$_7$}{7} obtained from static DFT calculations with implicit solvent.}  
  \label{fig:DL_2_radar_DFT_binding_energy} 
\end{figure}

\begin{figure}[htbp]
  \centering
  \includegraphics[width=1\textwidth]{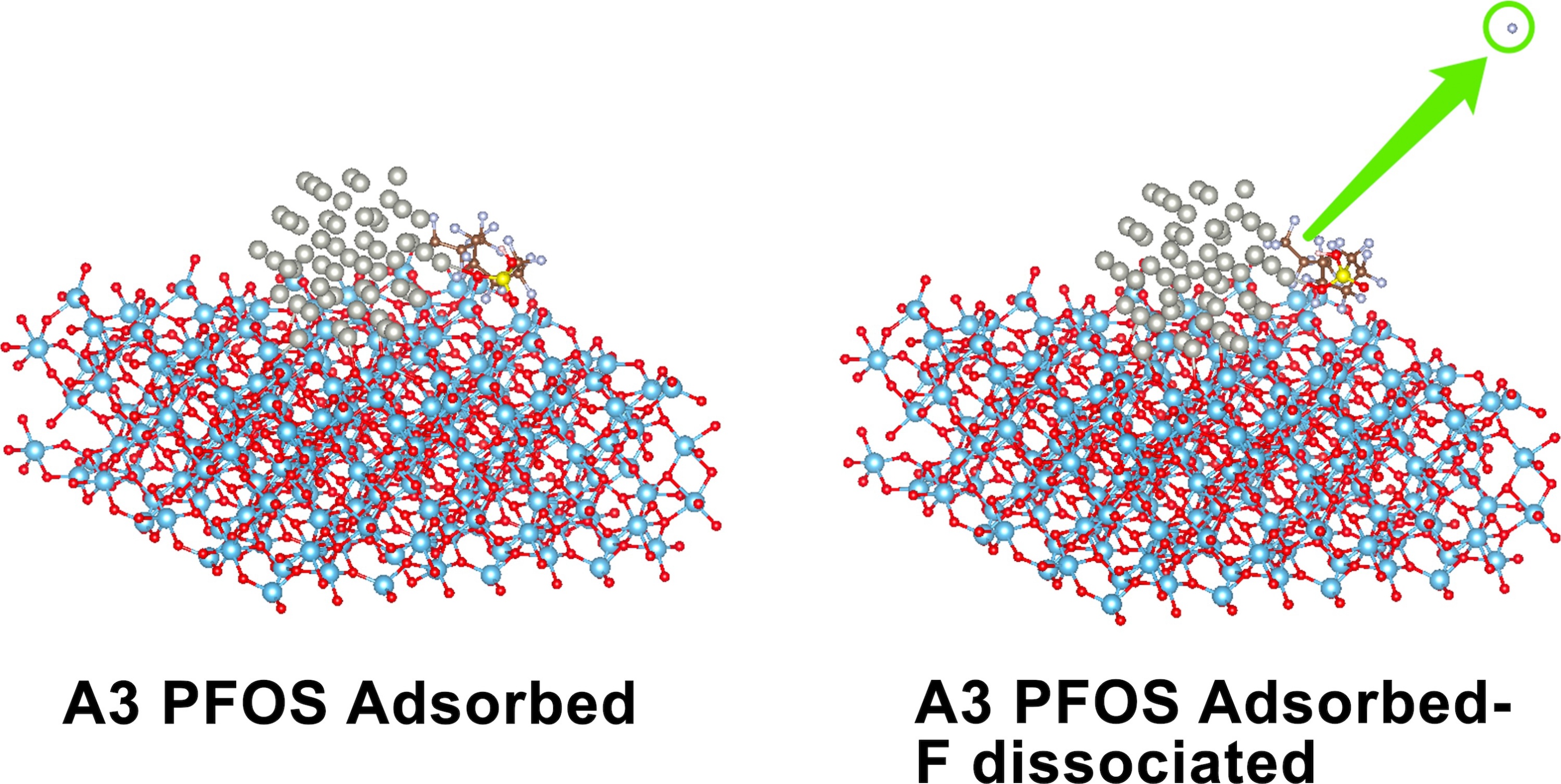}
  \caption{Illustration of C--F bond dissociation simulation protocol. Left: Initial configuration with PFOS adsorbed on A3 surface. Right: Final configuration with fluorine atom displaced to corner position (5~\AA{} from each boundary) of the supercell.}  
  \label{fig:DL_2_dis_F_showcase} 
\end{figure}

\begin{figure}[htbp]
  \centering
  \includegraphics[width=1\textwidth]{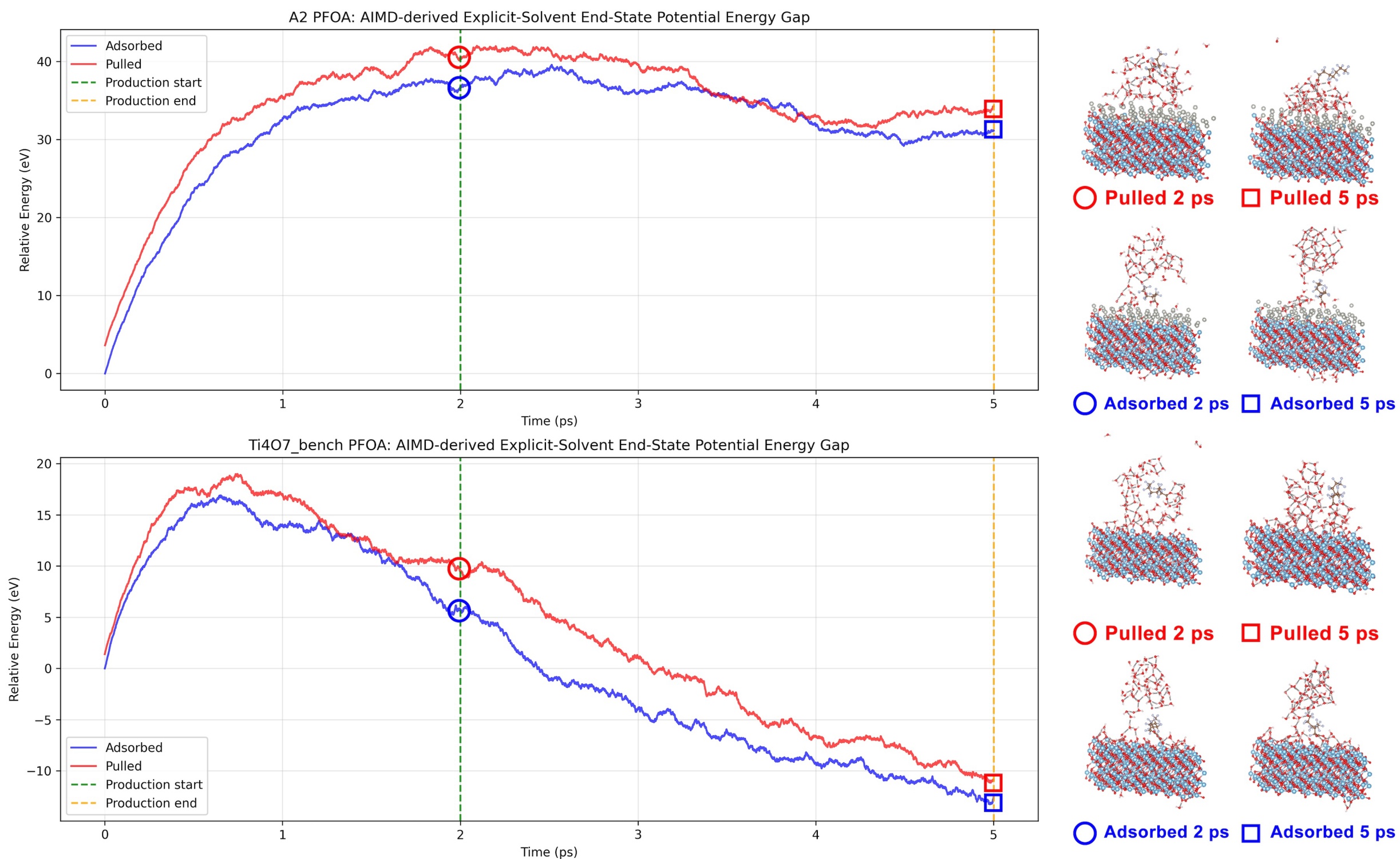}
  \caption{Representative AIMD trajectory showing system energy evolution under explicit solvent conditions. The green dashed line denotes the 2 ps equilibration period. Right panels display system configurations at highlighted time points along the 3 ps production trajectory.}  
  \label{fig:DL_2_AIMD_showcase} 
\end{figure}

\begin{figure}[htbp]
  \centering
  \includegraphics[width=1\textwidth]{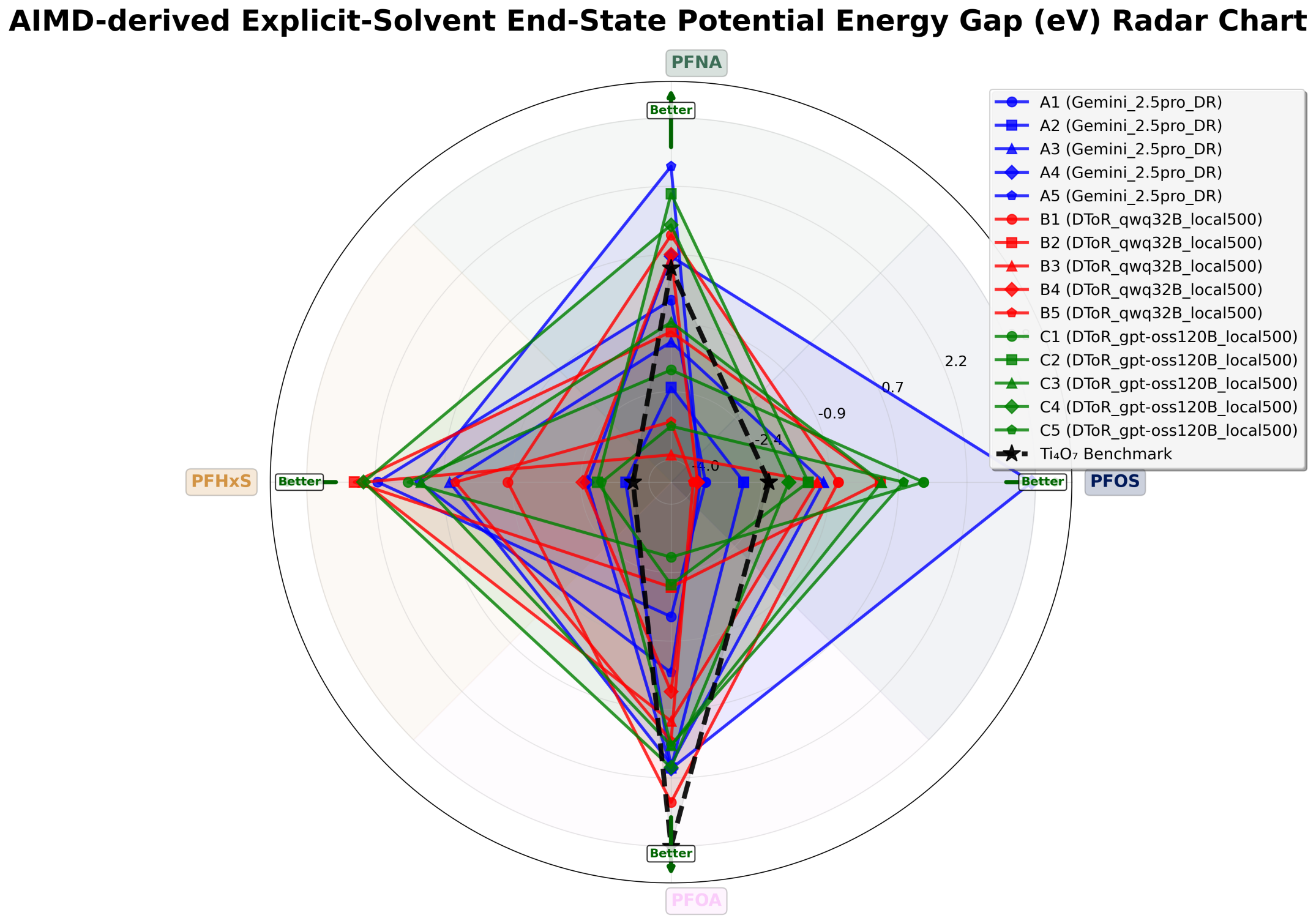}
  \caption{Radar plot comparing explicit-solvent end-state potential energy gaps ($\Delta E_{\text{pot}}^{\text{ES}}$) averaged over the 3 ps production period for all candidate catalysts and benchmark Ti\texorpdfstring{$_4$}{4}O\texorpdfstring{$_7$}{7} from AIMD simulations.}  
  \label{fig:DL_2_radar_AIMD} 
\end{figure}

The degradation performance evaluation integrates three complementary metrics that capture distinct aspects of the catalytic process. First, the binding energy $\Delta E_{\text{bind}}$ from static DFT calculations quantifies the thermodynamic driving force for PFAS adsorption onto catalyst surfaces (Figure~\ref{fig:DL_2_radar_DFT_binding_energy}). Strong binding ensures sufficient surface coverage
and residence time for subsequent degradation reactions. Second, the explicit-solvent potential energy gap $\Delta E_{\text{pot}}^{\text{ES}}$ from AIMD simulations reveals how aqueous solvation affects the adsorption thermodynamics (Figures~\ref{fig:DL_2_AIMD_showcase} and \ref{fig:DL_2_radar_AIMD}), accounting for competitive water adsorption and hydrogen-bonding network reorganization that
static calculations cannot capture. Third, the C--F bond dissociation energy quantifies the activation barrier for the rate-limiting defluorination step, computed by displacing fluorine atoms from adsorbed PFAS configurations to isolated positions within the supercell (Figure~\ref{fig:DL_2_dis_F_showcase}).
\textbf{Scoring Framework for Degradation Performance}

\begin{figure}[htbp]
  \centering
  \includegraphics[width=1\textwidth]{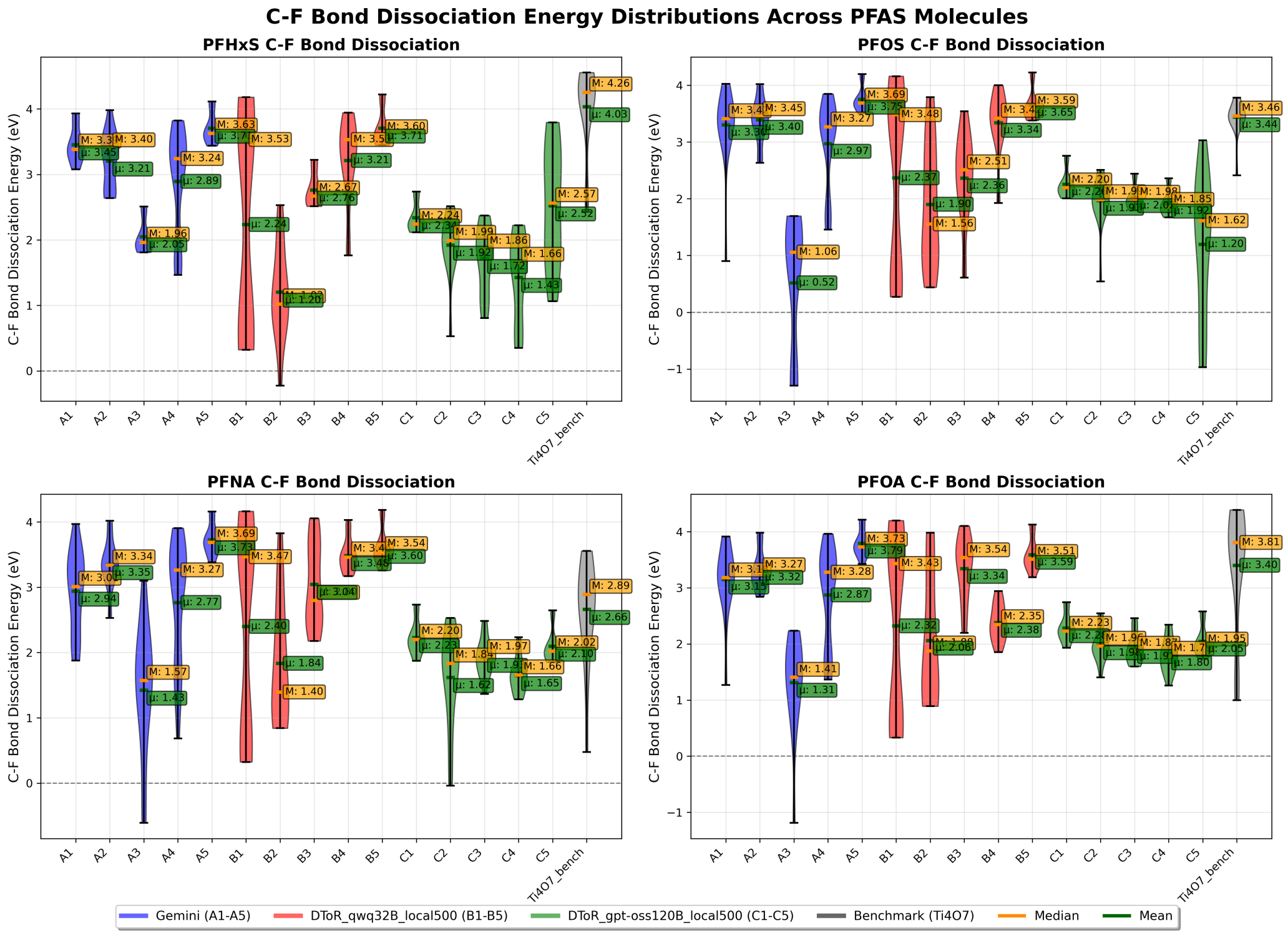}
  \caption{Violin plots showing distributions of C--F bond dissociation energies for different PFAS compounds (PFOA, PFOS, PFNA, PFHxS) across all candidate catalysts, with mean values indicated by horizontal bars.}  
  \label{fig:DL_2_dis_F_violin_distribution} 
\end{figure}

\begin{figure}[htbp]
  \centering
  \includegraphics[width=1\textwidth]{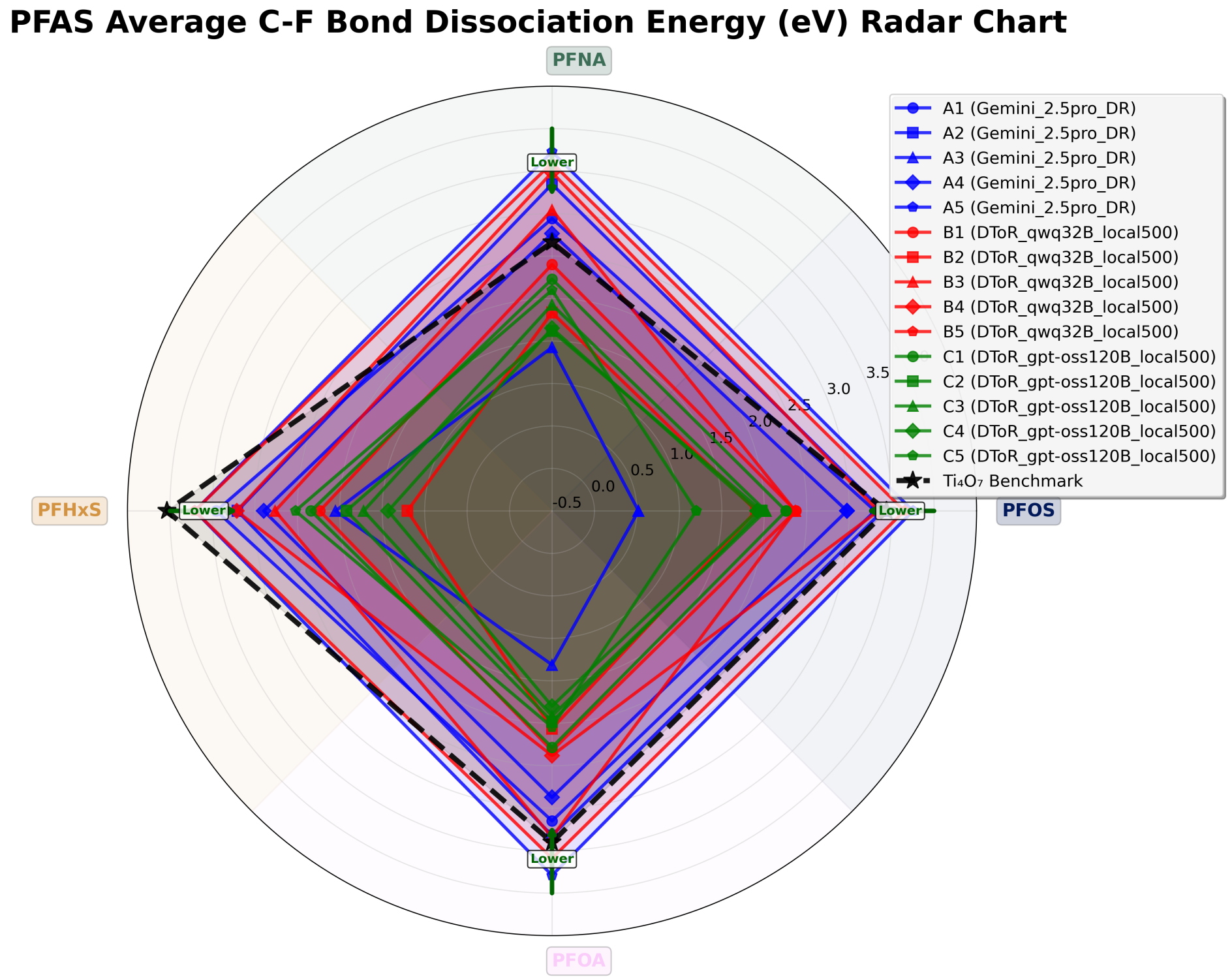}
  \caption{Radar plot comparing mean C--F bond dissociation energies across all candidate catalysts and benchmark Ti\texorpdfstring{$_4$}{4}O\texorpdfstring{$_7$}{7}. Lower dissociation energies indicate more favorable C--F bond activation for degradation.}  
  \label{fig:DL_2_radar_dis_F} 
\end{figure}

To enable quantitative comparison across these diverse metrics, we applied a unified linear scoring framework:

\begin{equation}
\text{Score} = 100 \times \left(1 + \beta \cdot \Delta_{\text{metric}}\right)
\end{equation}

where $\Delta_{\text{metric}}$ represents the performance differential relative to the Ti\texorpdfstring{$_4$}{4}O\texorpdfstring{$_7$}{7} baseline, defined as $\Delta\bar{E} - \Delta\bar{E}_{\text{Ti}_4\text{O}_7}$ for binding metrics and $\Delta\bar{E}_{\text{Ti}_4\text{O}_7} - \Delta\bar{E}$ for C--F dissociation. Here $\Delta\bar{E}$ is the mean energy value averaged over four representative PFAS compounds (PFOA, PFOS, PFNA, PFHxS),
and $\beta = 0.2$ serves as the sensitivity parameter. For binding metrics (both static DFT and AIMD), stronger adsorption (corresponding to larger magnitude binding energies) yields higher scores. For C--F dissociation, lower energy barriers yield higher scores, reflecting more facile bond activation.
The distribution of C--F dissociation energies across different PFAS compounds and candidate surfaces exhibits substantial variability (Figure~\ref{fig:DL_2_dis_F_violin_distribution}), with mean values ranging from approximately 2 to 6 eV (Figure~\ref{fig:DL_2_radar_dis_F}). This variation underscores the importance of averaging over multiple PFAS species to obtain representative performance
metrics, as individual compounds may exhibit preferential interactions with specific surface sites or crystallographic facets.
\textbf{Performance Analysis Across Computational Methods}

\begin{figure}[htbp]
  \centering
  \includegraphics[width=1\textwidth]{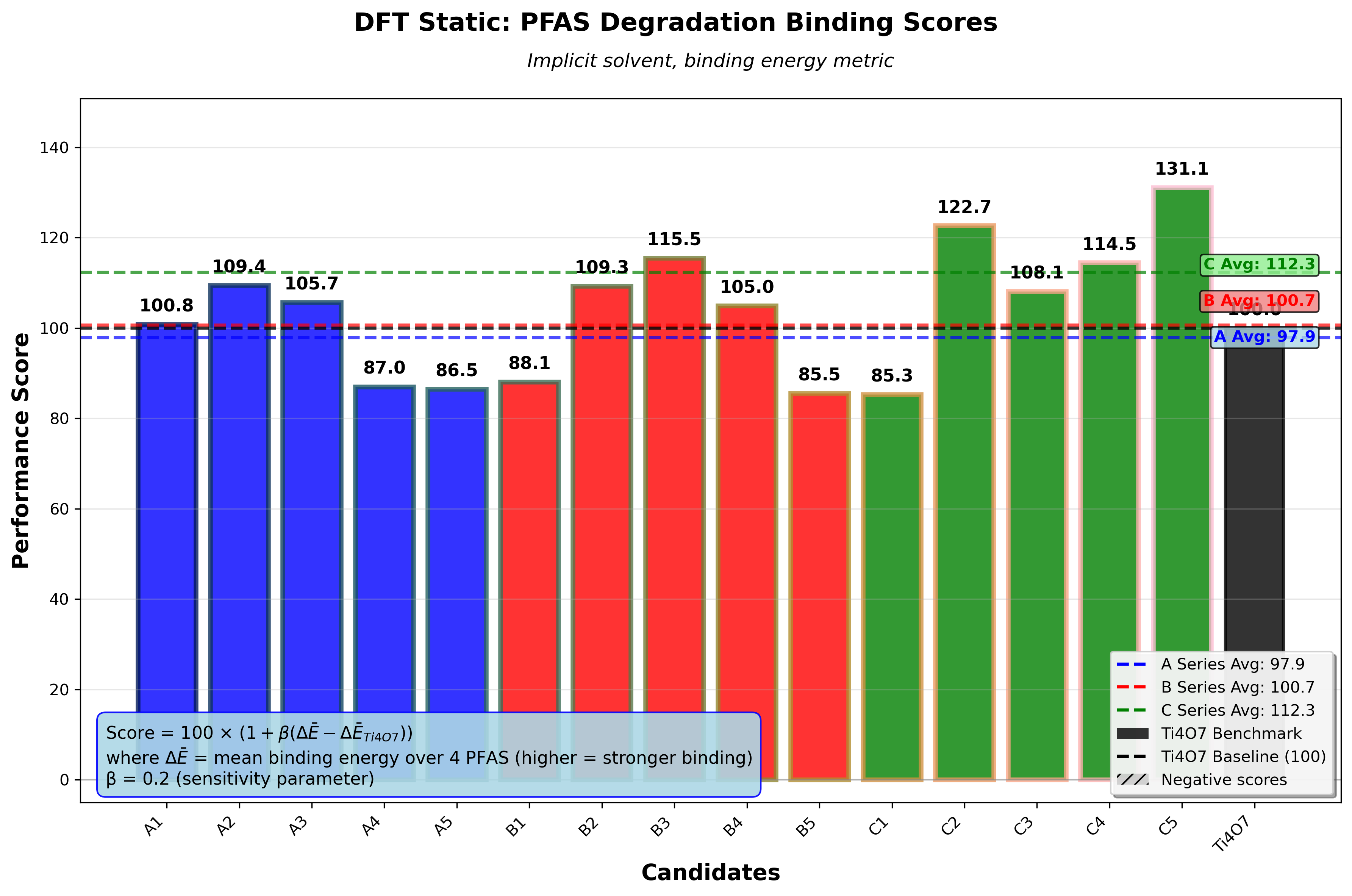}
  \caption{PFAS binding scores for all candidate catalysts from static DFT calculations, normalized to Ti\texorpdfstring{$_4$}{4}O\texorpdfstring{$_7$}{7} baseline (100). Scores are calculated using the linear metric with sensitivity parameter $\beta = 0.2$, where $\Delta\bar{E} = \overline{\Delta E}_{\text{PFAS}}$ represents the mean binding energy averaged over four PFAS compounds.}  
  \label{fig:DL_2_scoring_static_DFT_binding} 
\end{figure}

\begin{figure}[htbp]
  \centering
  \includegraphics[width=1\textwidth]{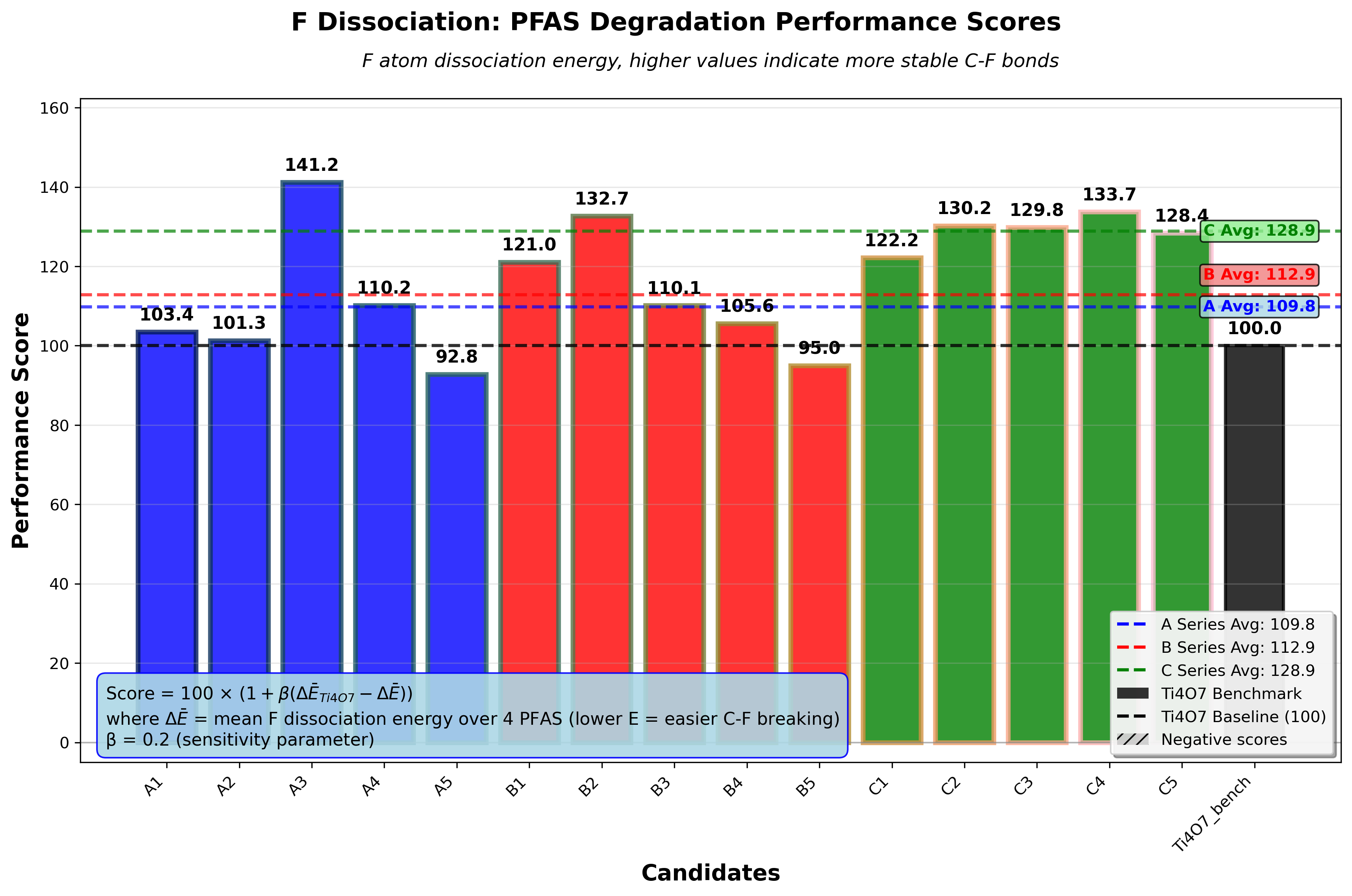}
  \caption{C--F bond activation scores for all candidate catalysts from static DFT calculations, normalized to Ti\texorpdfstring{$_4$}{4}O\texorpdfstring{$_7$}{7} baseline (100). Scores are calculated using the linear metric with sensitivity parameter $\beta = 0.2$, where $\Delta\bar{E} = \overline{E}_{\text{F-dis}}$ represents the mean C--F dissociation energy averaged over four PFAS compounds. Higher scores indicate lower dissociation barriers and more favorable degradation potential.}  
  \label{fig:DL_2_scoring_static_DFT_F_dis} 
\end{figure}

\begin{figure}[htbp]
  \centering
  \includegraphics[width=1\textwidth]{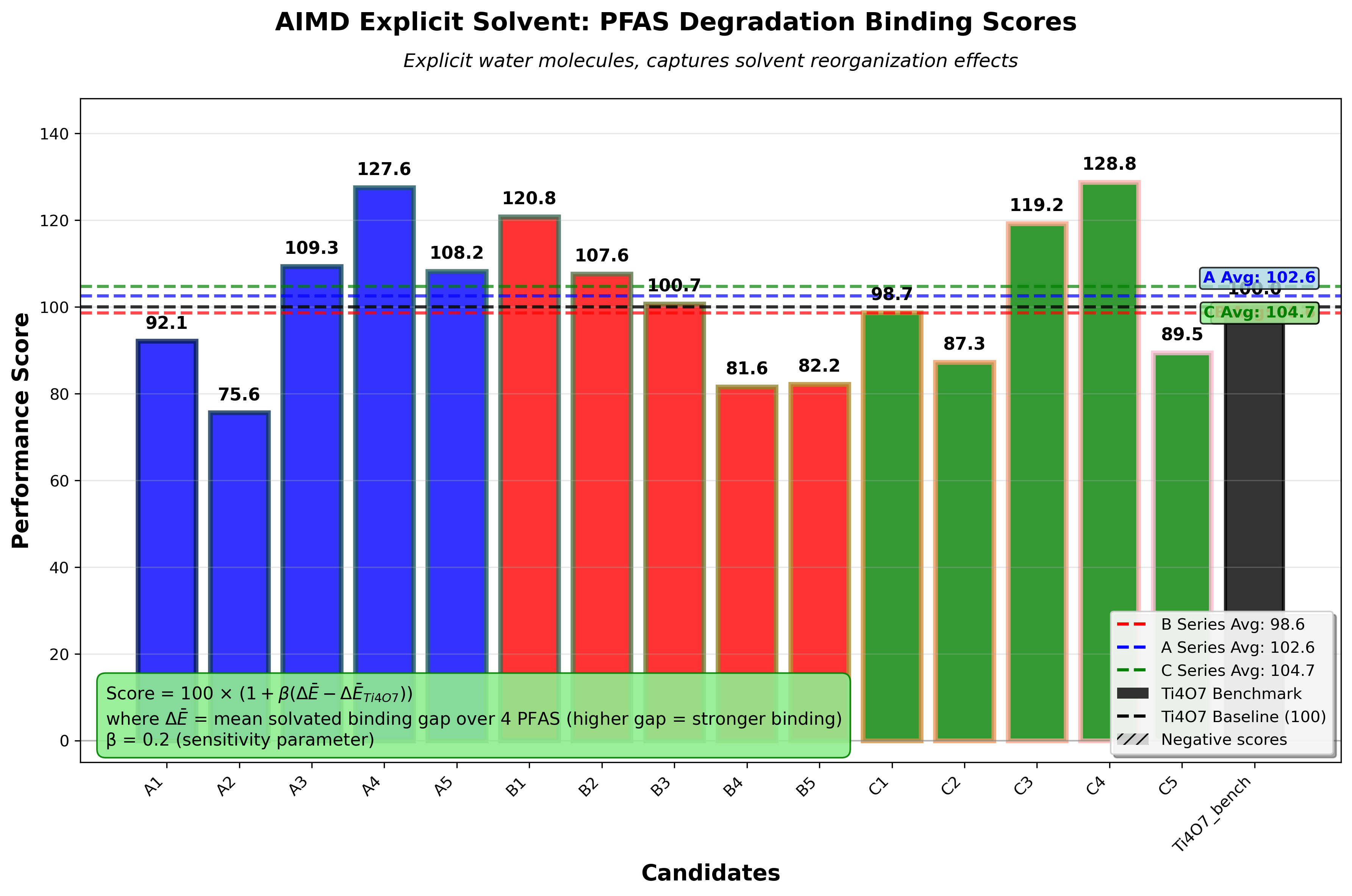}
  \caption{PFAS binding scores for all candidate catalysts from explicit-solvent AIMD simulations, normalized to Ti\texorpdfstring{$_4$}{4}O\texorpdfstring{$_7$}{7} baseline (100). The scoring methodology remains identical to static DFT, with $\Delta\bar{E}$ representing the mean $\Delta E_{\text{pot}}^{\text{ES}}$ values averaged over four PFAS compounds from the 3 ps production period.}  
  \label{fig:DL_2_scoring_AIMD} 
\end{figure}

Application of this scoring framework reveals distinct performance patterns across the three evaluation metrics. In static DFT calculations, several candidates demonstrate superior PFAS binding relative to the Ti\texorpdfstring{$_4$}{4}O\texorpdfstring{$_7$}{7} benchmark (Figure~\ref{fig:DL_2_scoring_static_DFT_binding}), with C-series materials generally exhibiting stronger adsorption energies. However, the inclusion of explicit
solvent effects in AIMD simulations significantly alters these rankings (Figure~\ref{fig:DL_2_scoring_AIMD}), with most candidates showing reduced binding scores due to solvent-mediated destabilization effects. This divergence highlights the critical importance of accounting for aqueous solvation in evaluating catalyst performance under realistic conditions.
The C--F bond activation scores (Figure~\ref{fig:DL_2_scoring_static_DFT_F_dis}) provide complementary insights into degradation potential. Notably, candidates that excel in binding may not necessarily facilitate efficient C--F bond cleavage, as these properties depend on distinct electronic and geometric factors. However, C-series materials were proven in average level better with lower F
dissociation energy barriers. Nonetheless, the ideal degradation catalyst must balance strong PFAS adsorption with effective bond activation—a trade-off that our multi-metric scoring framework quantitatively captures.
\textbf{Implications for Catalyst Design}

The comprehensive evaluation framework presented here reveals that no single candidate uniformly excels across all performance metrics. Strong surface binding, while essential for PFAS capture, does not guarantee facile C--F bond activation. Similarly, materials optimized for bond dissociation may suffer from weak adsorption, particularly under aqueous conditions where solvent competition is
significant. This multi-dimensional performance landscape emphasizes the need for balanced catalyst design that considers both thermodynamic (binding) and kinetic (bond activation) factors, evaluated under both implicit and explicit solvation conditions to ensure robust performance predictions for practical applications. As indicated in this dry lab study, A3, B2, C3, and C4 could be promising
candidates for further investigation in wet lab.

\clearpage

\begin{figure}[htbp]
  \centering
  \includegraphics[width=1\textwidth]{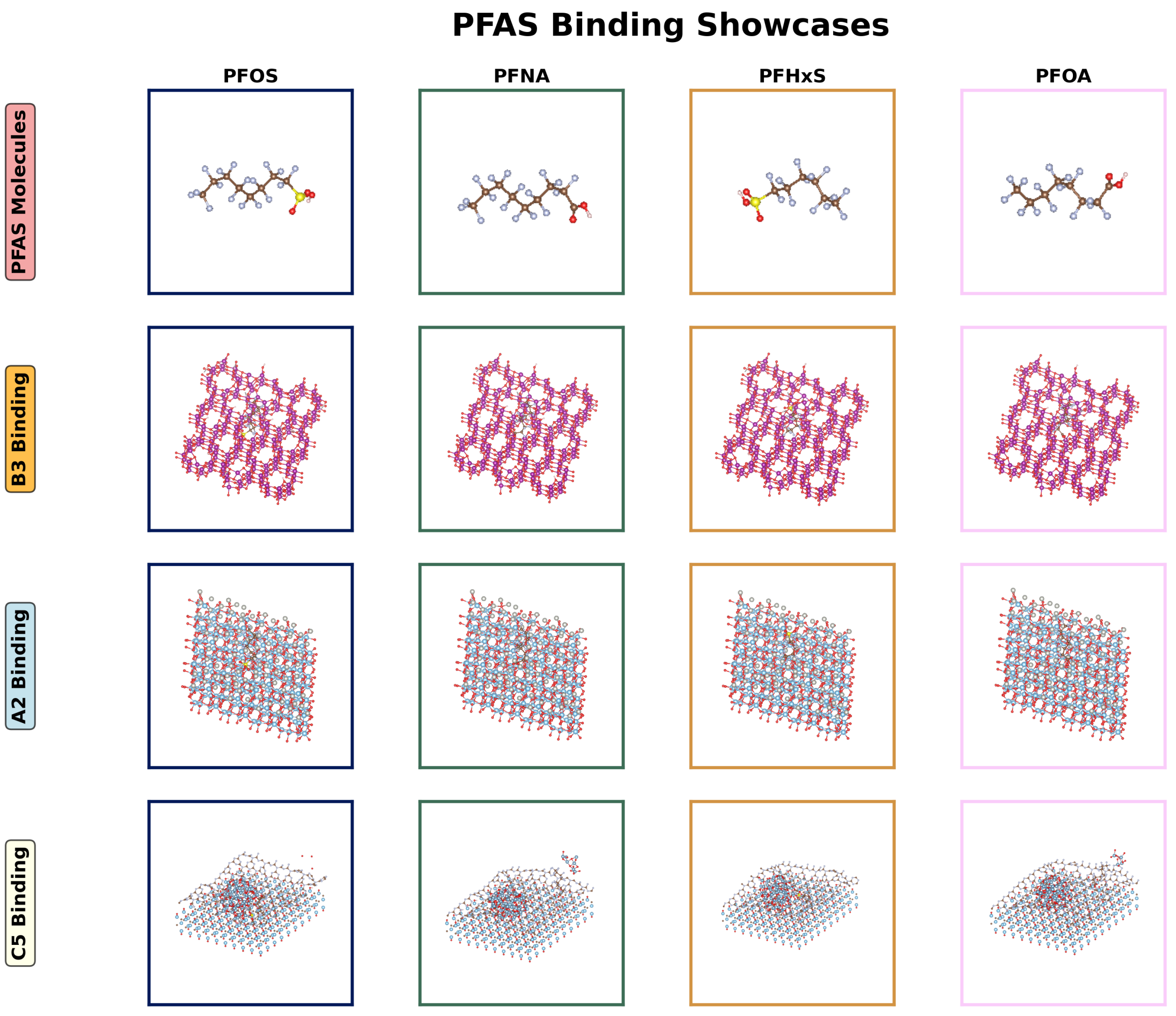}
  \caption{Representative PFAS compounds (PFOA, PFOS, PFNA, PFHxS) employed in degradation assessment, with corresponding optimized adsorption configurations on selected candidate surfaces.}  
  \label{fig:DL_2_binding_showcase} 
\end{figure}

\clearpage

\subsection{Battery Binder}

\subsubsection{LIB\_Advanced\_Binders}
For the third application: Battery Binder. 
The query as shown in \ref{app:topics} is:
\begin{lstlisting}
Which advanced binder technologies are being developed to improve the performance and longevity of lithium-ion batteries?
\end{lstlisting}

The best commercial solution is: \textbf{ChatGPT\_o4\_mini\_high\_DR}.

The best local non gpt-oss solution is: \textbf{DToR\_qwq32B\_local500}.

The best local gpt-oss solution is: \textbf{DToR\_gpt-oss120B\_local500}.

The expert based on the two reports, provided 4 commercial DR report derived candidates (A1-A4), and 4 local non gpt-oss DR report derived candidates (B1-B4), and 4 local gpt-oss DR report derived candidates (C1-C4), visualized in Fig. \ref{fig:DL_3_candidates_final}:

\paragraph{\textbf{NCM811 cathode description}}
In order to create a test bench for lithium ion battery binder, we modeled a (104)-terminated slab of layered LiNi$_{0.8}$Co$_{0.1}$Mn$_{0.1}$O\texorpdfstring{$_2$}{2} (NCM811) cathode in a periodic cell with a = 27.67 \AA{}, b = 25.55 \AA{}, c = 49.47 \AA{} and \texorpdfstring{$\alpha$}{alpha} = \texorpdfstring{$\beta$}{beta} = 90°, \texorpdfstring{$\gamma$}{gamma} = 72.51° (P1). The atomic slab spans $\approx$12.1 \AA{} along the surface normal, leaving $\approx$37.4 \AA{} of vacuum
to suppress spurious interactions; the model contains 732 atoms (Li$_{174}$ Ni$_{129}$ Co$_{21}$ Mn$_{21}$ O$_{387}$). For structural relaxations, we used a large in‑plane surface cell (tripled along b) and constrained the bottom $\approx$1/3 of atoms to mimic bulk termination.

\subparagraph{\textbf{PVDF benchmark binder}}
For the benchmark binder, we built an all‑trans PVDF 10‑mer (–CH\texorpdfstring{$_2$}{2}–CF\texorpdfstring{$_2$}{2}–)$_{10}$, end‑capped with H/F, pre‑relaxed in vacuo, and then laid it flat on the pre‑relaxed NCM811 (104) slab, constructed and constrained following our surface‑build script. In the configuration, the chain backbone lies approximately parallel to the surface ($\sim$30° off the b‑axis), with a closest polymer–surface
contact of $\sim$2.0 \AA{}, an in‑plane backbone span of $\approx$23 \AA{}, and a projected convex‑hull footprint of $\approx$125 \AA{}$^2$; the polymer fragment contributes C$_{20}$H$_{21}$F$_{21}$, bringing the full model to 794 atoms.

\subparagraph{\textbf{A1 --- PVPA (after binder adsorbed)}}
A poly(vinyl‑phosphonic acid) 10‑mer (C$_{20}$H$_{32}$O$_{30}$P$_{10}$) was adsorbed on the NCM811 (104) slab, using the same surface‑build constraints as our base slab. The slab portion is 12.58 \AA{} thick, leaving 36.88 \AA{} of vacuum; the full system contains 824 atoms (slab = 732; polymer = 92). The PVPA backbone lies nearly parallel to the surface ($\approx$34° off the b‑axis), spans 20.59
\AA{} in‑plane, and covers a projected convex‑hull footprint of 115.09 \AA{}$^2$. Phosphonate O atoms approach surface cations with a shortest O-Li contact of 2 \AA{}, alongside O-O hydrogen‑bond contacts to surface oxygen of 2.5–2.9 \AA{}.

\subparagraph{\textbf{A2 --- Alg‑Ca (after binder adsorbed)}}
A Ca$^{2+}$‑cross‑linked alginate 5‑mer (G–M blocks; carboxylates protonated except those coordinating Ca$^{2+}$) was laid approximately parallel to the NCM811(104) surface with the bottom oxide layers fixed. The combined model has 834 atoms (slab = 732; binder = 102, $\approx$ C$_{30}$H$_{40}$O$_{31}$Ca). The slab portion is 12.20 \AA{} thick with 37.27 \AA{} vacuum. The polymer backbone aligns
quasi‑parallel to the surface (24.62° off \textbf{b}), spans 17.71 \AA{} in‑plane, and presents a convex‑hull footprint of 92.64 \AA{}$^2$. Interfacial contacts are dominated by non‑specific physisorption (nearest heavy‑atom approach 3.23 \AA{}; O$_{\text{binder}}$-O$_{\text{slab}}$ 3.31–4.17 \AA{}), while Ca$^{2+}$ remains intrachain‑coordinated (closest Ca–O$_{\text{slab}}$ 7.13 \AA{}).

\subparagraph{\textbf{A3 --- Sulfonyl‑polyimide (S‑PI) (after binder adsorbed)}}
A tri‑repeat S‑PI fragment (C$_{42}$H$_{14}$N$_6$O$_{20}$S\texorpdfstring{$_3$}{3}, 85 atoms) was constructed following the imide‑dianhydride/diamino‑sulfone motif and combined with the standard NCM811(104) slab (total 817 atoms; slab = 732, binder = 85). The polymer lies parallel to the surface, oriented $37.49^\circ$ off the $b$‑axis (principal in‑plane span 25.71 \AA{}) with a projected convex‑hull footprint of
152.10 \AA{}$^2$. This configuration represents the separated reference: the nearest polymer heavy atom is 17.42 \AA{} above the top oxide layer (no interfacial contacts; shortest O$_\text{binder}$$\cdots$O$_\text{slab}$ = 19.17 \AA{}; O$_\text{binder}$$\cdots$Ni/Co/Mn = 19.75 \AA{}).

\subparagraph{\textbf{A4 --- PTFE (–CF\texorpdfstring{$_2$}{2}–CF\texorpdfstring{$_2$}{2}–)\texorpdfstring{$_{10}$}{10} (after binder adsorbed)}}
A perfluoroethylene decamer (C$_{20}$F$_{42}$, 62 atoms) is laid flat on the NCM811(104) surface to provide a non‑reactive binder baseline; the combined model has 794 atoms (slab = 732; polymer = 62). The slab portion is 12.20 \AA{} thick with 37.27 \AA{} vacuum. The chain axis lies 39.61° off \textbf{b} within the \textbf{ab} plane, with a principal in‑plane span of 25.14 \AA{} and a projected
convex‑hull footprint of 80.55 \AA{}$^2$. Interfacial contacts are dispersive: the nearest heavy‑atom approach is 2.49 \AA{} (F-Li); the shortest F-O\_slab and F-Ni contacts are 2.84 \AA{} and 3.43 \AA{}, respectively, while C-O\_slab $\geq$ 3.63 \AA{} and no metal–fluorine contact falls below 2.49 \AA{}. These metrics set a PTFE benchmark for subsequent $\Delta$E\_ads/A comparisons.

\subparagraph{\textbf{B1 --- Schiff‑base polymer (imine‑linked poly(aldehyde‑amine)) (after binder adsorbed)}}
An imine‑linked poly(aldehyde‑amine) fragment (\textbf{C\texorpdfstring{$_{40}$}{40}H\texorpdfstring{$_{28}$}{28}N\texorpdfstring{$_4$}{4}}, 72 atoms) is adsorbed on the NCM811(104) slab; the combined model contains \textbf{804} atoms (slab = 732; binder = 72). The oxide portion is \textbf{13.07 \AA{}} thick with \textbf{36.39 \AA{}} vacuum. The polymer lies quasi‑parallel to the surface, oriented \textbf{36.97°} off \textbf{b} within the \textbf{ab} plane;
its principal in‑plane span is \textbf{34.60 \AA{}}, and the projected convex‑hull footprint is \textbf{168.51 \AA{}$^2$}. Interfacial contacts show a nearest heavy‑atom approach of \textbf{2.57 \AA{}} (C-Li); the imine N approaches surface oxygen at \textbf{2.35 \AA{}}, while separations to cation sites remain larger (N-Li \textbf{3.68 \AA{}}; N-Ni \textbf{4.29 \AA{}}; N-Co \textbf{4.57 \AA{}};
N-Mn \textbf{4.76 \AA{}}).

\subparagraph{\textbf{B2 --- lignin + CNT (after binder adsorbed)}}
A \texorpdfstring{$\beta$}{beta}‑O‑4 lignin dimer was positioned at the NCM811(104) interface and co‑laid with an armchair (6,6) carbon nanotube segment (8 axial repeats; 192 C) oriented roughly parallel to the terrace; the tube's in‑plane principal axis is rotated \textbf{36.86°} off \textbf{b} and spans \textbf{24.91 \AA{}}. The composite binder portion contains \textbf{C\texorpdfstring{$_{234}$}{234}H\texorpdfstring{$_{80}$}{80}O\texorpdfstring{$_{46}$}{46}}. The oxide portion
spans \textbf{11.05 \AA{}} in thickness along the surface normal with \textbf{38.42 \AA{}} of vacuum retained. Lignin's phenolic/ether oxygens face the cation‑terminated ridges, with the closest \textbf{O-(Ni/Co/Mn)} approach of \textbf{3.32 \AA{}}, consistent with physisorptive H‑bond/ion‑dipole contacts rather than direct chelation; the CNT lies flat to provide an extended $\pi$‑contact, with a
near‑surface carbon convex‑hull footprint of \textbf{84.71 \AA{}$^2$}. Together, this layout tests cooperative anchoring: localized polar adhesion from lignin complemented by broad dispersive contact from the nanotube across the NCM811 terrace.

\subparagraph{\textbf{B3 --- Lignin–CMC hybrid (\texorpdfstring{$\beta$}{beta}‑O‑4 lignin dimer + CMC monomer, Na\texorpdfstring{$^+$}{+} exchanged) (after binder adsorbed)}}
The composite contains \textbf{876 atoms} (slab = 732; binder = 144 with \textbf{C\texorpdfstring{$_{50}$}{50}H\texorpdfstring{$_{69}$}{69}O\texorpdfstring{$_{24}$}{24}Na\texorpdfstring{$_1$}{1}}) and uses the same NCM811(104) platform and assembly recipe (lignin dimer H‑bonded to a CMC unit, carboxylate presented to the terrace). The organic complex lies quasi‑parallel to the surface, its principal in‑plane axis oriented \textbf{38.18°} off \textbf{b}, with a \textbf{35.09
\AA{}} backbone span and a projected convex‑hull footprint of \textbf{235.11 \AA{}$^2$}. The oxide portion is \textbf{11.09 \AA{}} thick with \textbf{38.38 \AA{}} of vacuum along \textbf{c}. Interfacial approach distances indicate mixed ionic/coordination contacts: nearest heavy‑atom approach \textbf{2.25 \AA{}} (O\_binder-Li), with \textbf{O\_binder-Ni = 2.99 \AA{}}, \textbf{O\_binder-Co = 4.09
\AA{}}, and \textbf{O\_binder-Mn = 4.29 \AA{}}; the exchanged \textbf{Na\texorpdfstring{$^+$}{+}} sits remote from the terrace (\textbf{Na-O\_slab = 6.80 \AA{}}). During relaxations, the \textbf{bottom one‑third of oxide atoms} are held fixed to emulate a semi‑infinite substrate.

\subparagraph{\textbf{B4 --- ZIF-8 face (Zn\texorpdfstring{$_4$}{4}(2-mim)\texorpdfstring{$_8$}{8} cluster) (after binder adsorbed)}}
We approximate the MOF by a single sodalite-cage face—four Zn nodes bridged by eight 2-methylimidazolates—so the interface samples the correct local Zn–N\texorpdfstring{$_4$}{4}/$\mu_2$-imidazolate chemistry without imposing the artificial pore matching and long-range lattice constraints of a periodic ZIF on a finite oxide slab. The internal Zn–N network is kept near its intrinsic geometry via mild constraints, while
only the interfacial coordinate is relaxed; one Zn is oriented toward a surface oxygen with an initial Zn-O set to $\sim$2.10 \AA{} to seed chemisorption and allow secondary N-Li/C-O\_slab contacts to adjust around it. This charge-neutral ring thus captures the short-range bonding physics that governs adhesion at comparable cost to polymer fragments, enabling fair ranking by area-normalized
adsorption energy $\Delta$E\_ads/A across all candidates.

\subparagraph{\textbf{C1 --- Polyvinyl alcohol (PVA) oligomer}}
Flexible, water‐processable oligomer of repeating vinyl alcohol units (–CH\texorpdfstring{$_2$}{2}CHOH–) with formula C$_{23}$H$_{48}$O$_{11}$ ($\sim$11–12 ethylene-glycol–like repeats; $\sim$82 atoms). Dense pendant –OH groups furnish multiple H-bond donors/acceptors and Lewis-basic oxygens capable of coordinating metal cations (e.g., Li$^+$), delivering strong film formation and polar-surface adhesion while
maintaining mechanical flexibility from partial crystallinity and amorphous chain mobility.

\subparagraph{\textbf{C2 --- CMC + waterborne polyurethane (WBPU) hybrid}}
Aqueous, interpenetrating network of a carboxymethyl cellulose oligomer blended with a WBPU soft-segment fragment (no covalent linking). Representative composition: C$_{33}$H$_{58}$Li$_5$N\texorpdfstring{$_4$}{4}O$_{23}$ (123 atoms). CMC supplies anionic –COO$^-$ sites (neutralized by Li$^+$) and hydroxyl-rich backbone for polar adhesion; WBPU contributes urethane –NH–CO–O– H-bond donors and ductile segments for
toughness and crack resistance. The hybrid is solvent-free (water-based) and mechanically robust via H-bonding and electrostatic interactions.

\subparagraph{\textbf{C3 --- Lithiated cellulose nanofiber (LCNF) oligomer}}
Partially lithiated cellulose fragment, C$_{42}$H$_{62}$Li$_8$O$_{41}$ (153 atoms), approximating $\sim$7 glucose repeats with selective deprotonation to form –O$^-$Li$^+$ sites ($\sim$1 Li per glucose). Lithiation introduces single-ion–conducting motifs while retaining some –OH H-bonding for mechanical integrity. The –O$^-$Li$^+$ network supports Li$^+$ coordination/hopping along the
polysaccharide, potentially lowering interfacial impedance.

\subparagraph{\textbf{C4 --- Polyethylene glycol (PEG) oligomer}}
Linear polyether HO–[CH\texorpdfstring{$_2$}{2}CH\texorpdfstring{$_2$}{2}O]$_{12}$–H (C$_{24}$H$_{50}$O$_{12}$; 86 atoms). Ether oxygens provide Lewis-basic coordination to Li$^+$; low $T_g$ and high segmental mobility enable compliance and ion solvating/plasticizing behavior. PEG is miscible with carbonate electrolytes and offers broad electrochemical compatibility, acting as a flexible, coordinating binder/plasticizer at the
interface.

\subsubsection{Battery Binder Adhesion Performance}

\begin{figure}[htbp]
  \centering
  \includegraphics[width=1\textwidth]{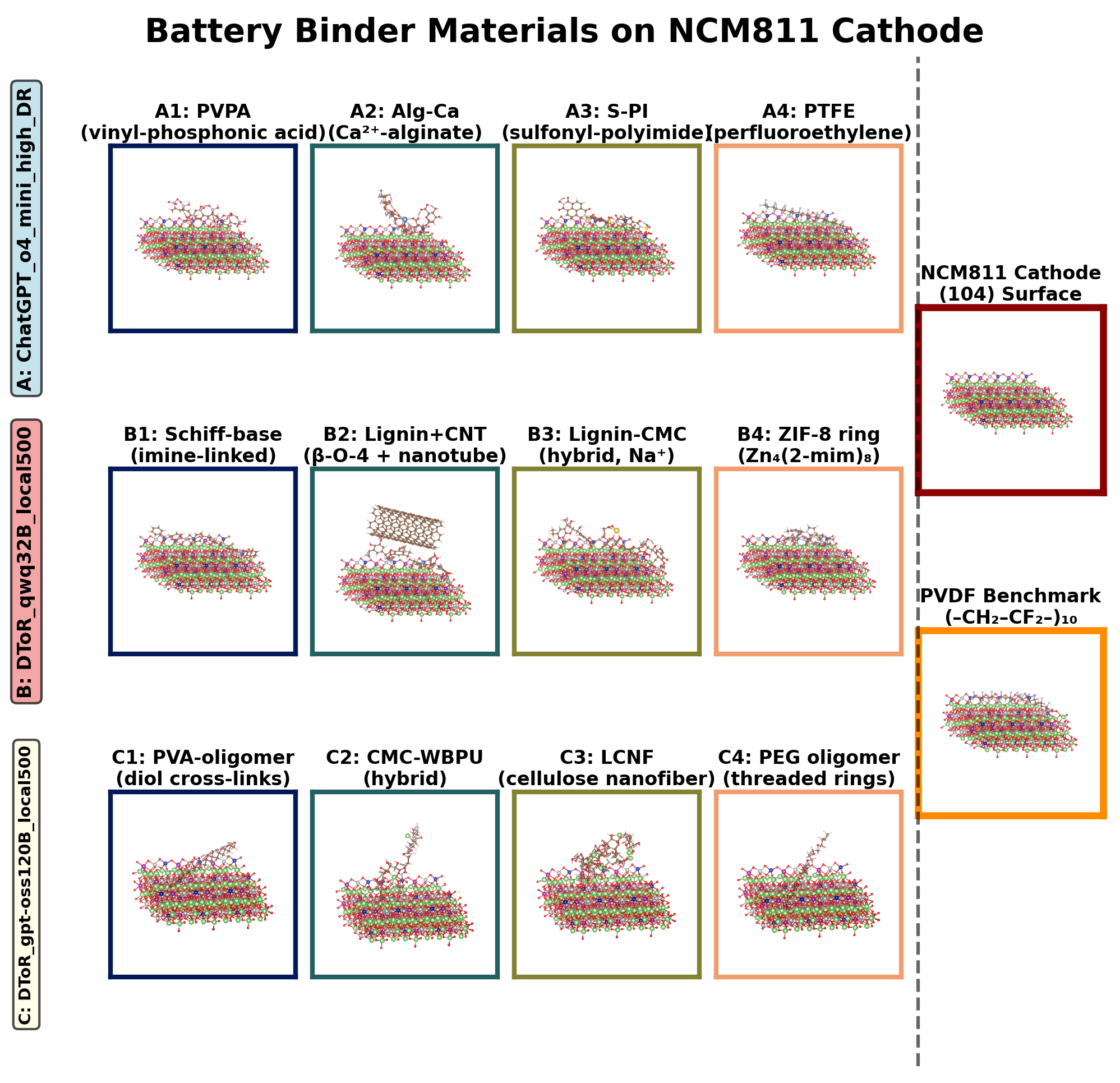}
  \caption{Structural configurations of battery binder candidates proposed by DR agents, with PVDF \protect\cite{zhong2021pvdf} included as the domain benchmark reference. Optimized adsorption configurations on the NCM811 cathode surface are shown for each candidate system.}
  \label{fig:DL_3_candidates_final}  
\end{figure}

\begin{figure}[htbp]
  \centering
  \includegraphics[width=1\textwidth]{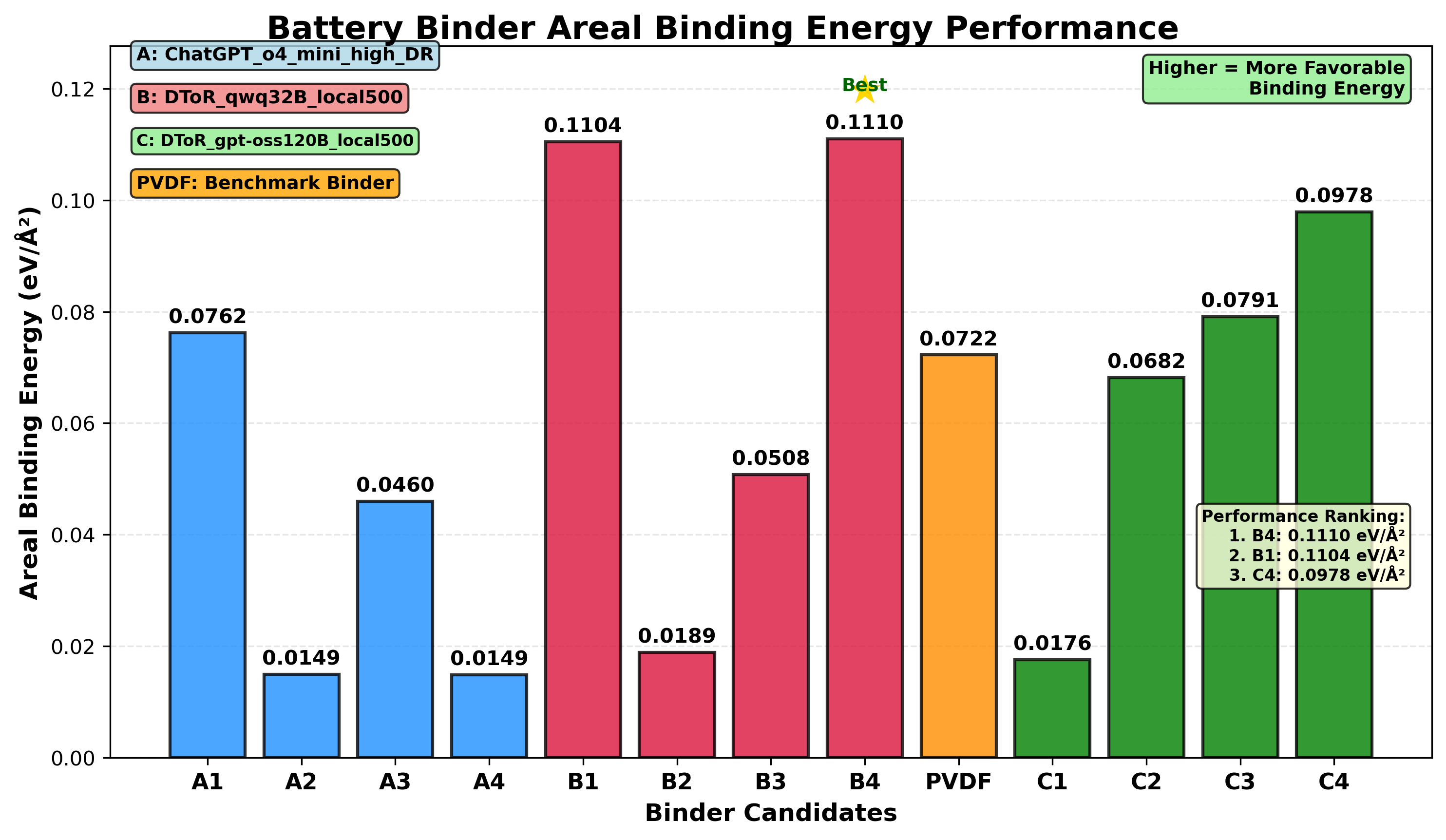}
  \caption{Comparison of areal-normalized binding energies ($\Delta E_{\text{bind}}/A$) from static DFT calculations. Values are expressed in eV/\AA$^2$, where $A$ represents the interfacial contact area determined by convex hull analysis of atoms within 3.5 \AA\ of the NCM811 surface.}  
  \label{fig:DL_3_DFT_binding_histogram} 
\end{figure}

\begin{figure}[htbp]
  \centering
  \includegraphics[width=1\textwidth]{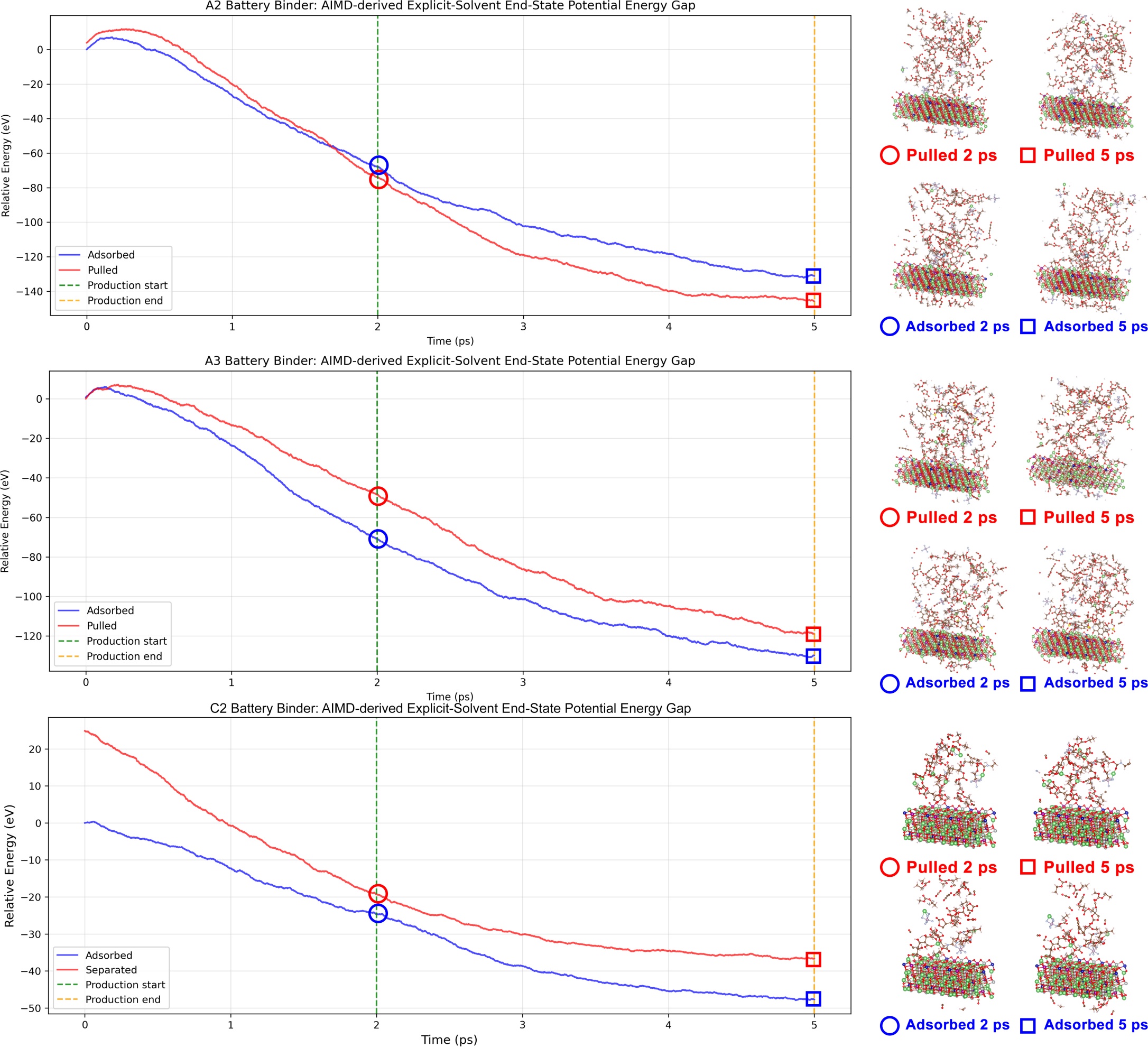}
  \caption{Representative AIMD trajectory showing system energy evolution in explicit battery electrolyte (EC:EMC:LiPF$_6$ = 9:9:2:2). The green dashed line denotes the 2 ps equilibration period. Right panels display system configurations at highlighted time points along the 3 ps production trajectory.}  
  \label{fig:DL_3_AIMD_showcase} 
\end{figure}

\begin{figure}[htbp]
  \centering
  \includegraphics[width=1\textwidth]{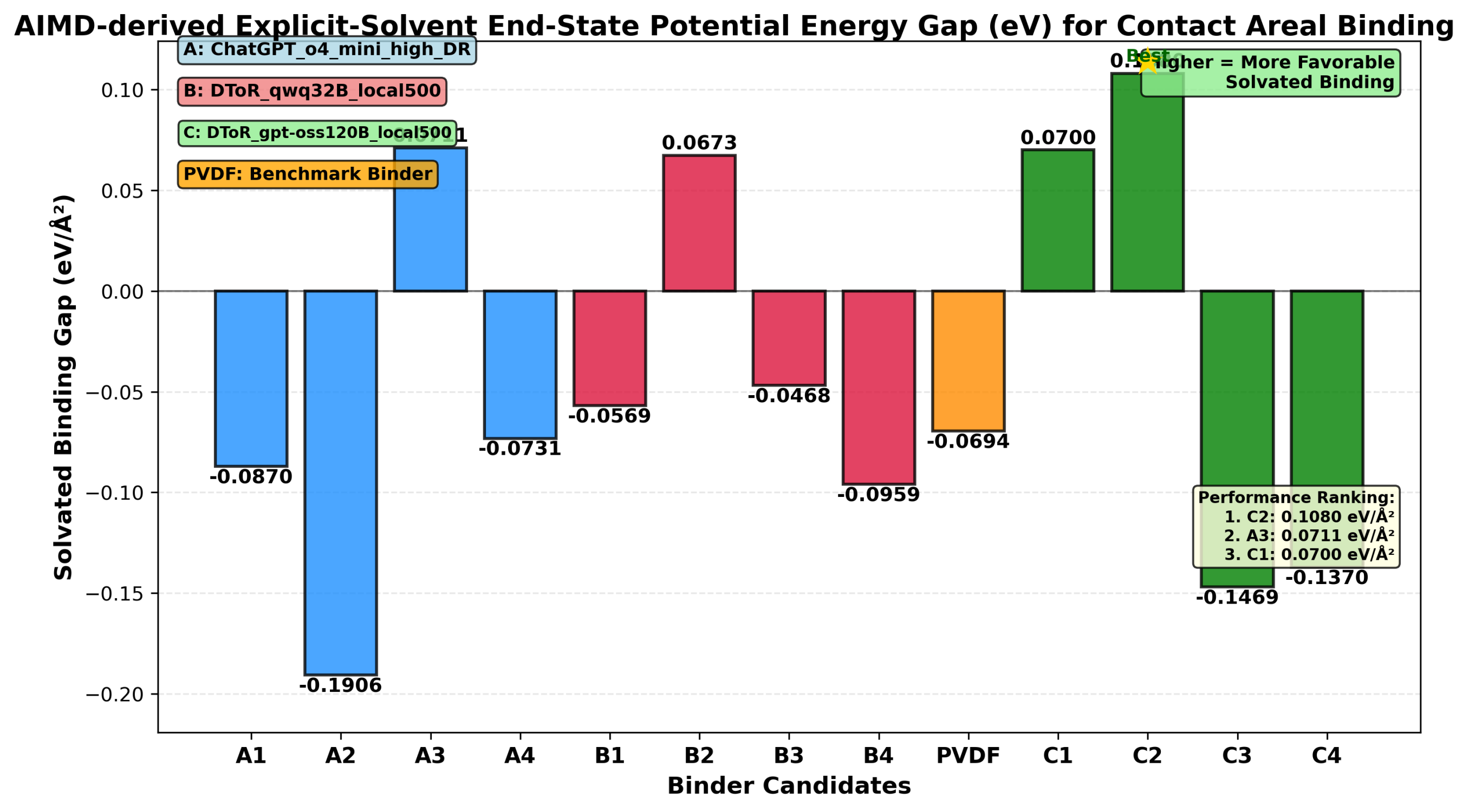}
  \caption{Areal-normalized explicit-solvent end-state potential energy gaps ($\Delta E_{\text{pot}}^{\text{ES}}/A$) from AIMD simulations. Negative values indicate unfavorable adhesion in the presence of electrolyte (Pulled state in explicit solvation stablizes the pulled state), reflecting competitive solvent adsorption at the binder--cathode interface.}  
  \label{fig:DL_3_AIMD_gap_histogram} 
\end{figure}

\begin{figure}[htbp]
  \centering
  \includegraphics[width=1\textwidth]{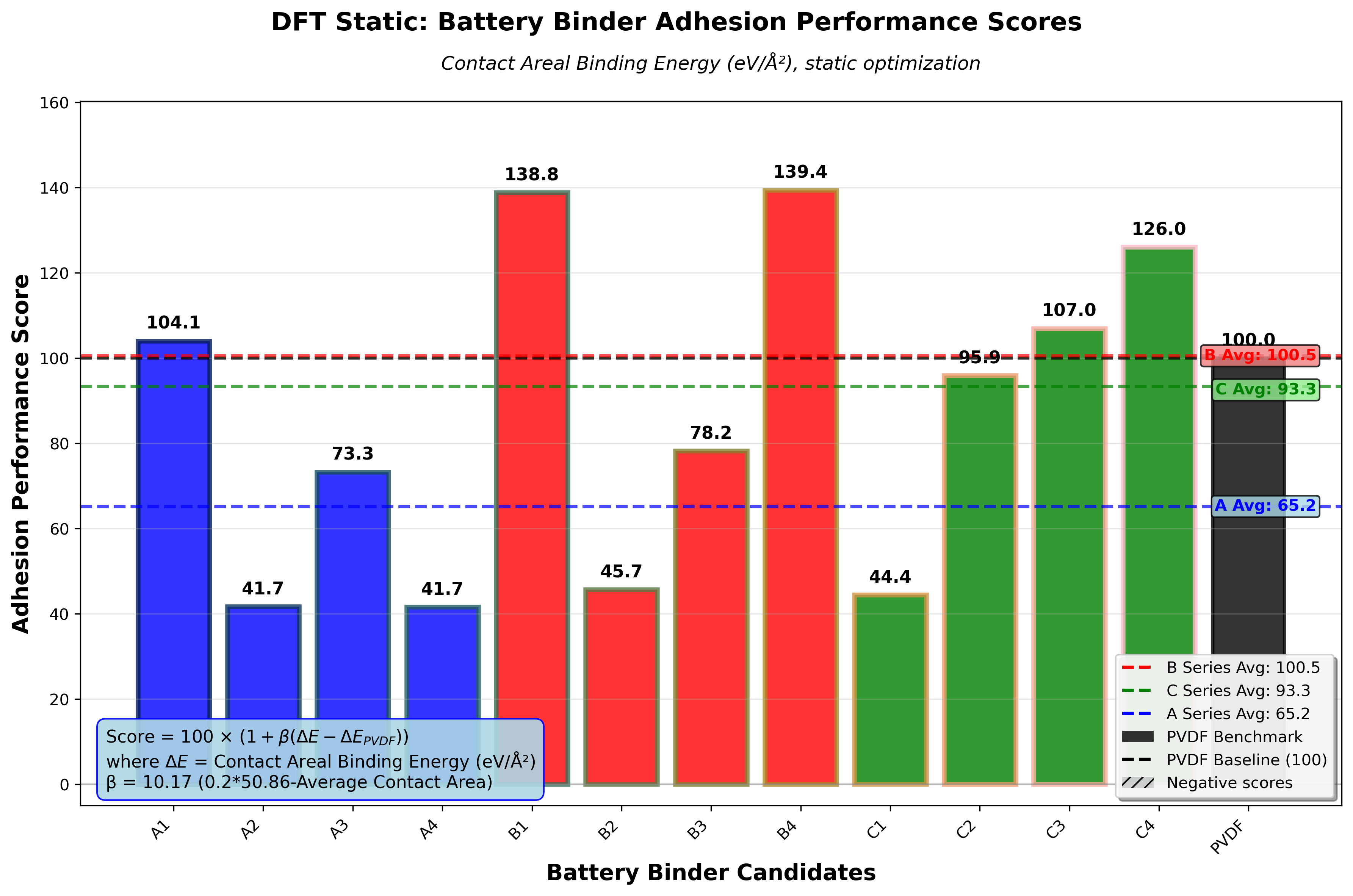}
  \caption{Adhesion performance scores for all binder candidates from static DFT calculations, normalized to PVDF baseline (100). Scores are calculated using the linear scaling metric with sensitivity parameter $\beta = 10.17$, applied to areal-normalized binding energies.}  
  \label{fig:DL_3_scoring_static_DFT_binding} 
\end{figure}

\begin{figure}[htbp]
  \centering
  \includegraphics[width=1\textwidth]{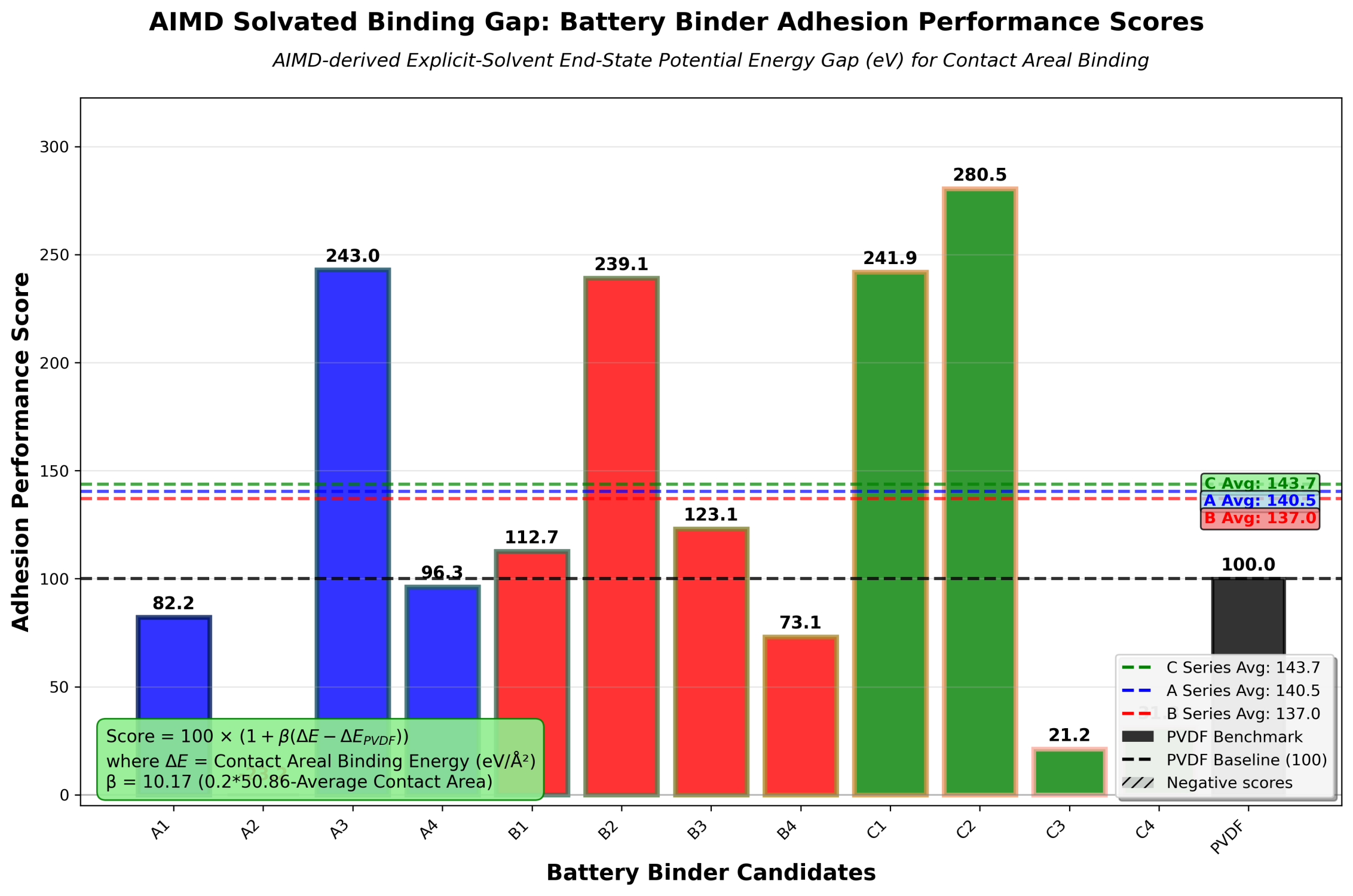}
\caption{Adhesion performance scores for all binder candidates from explicit-electrolyte AIMD simulations, normalized to PVDF baseline (100). The scoring methodology remains identical to static DFT, with $\Delta E_{\text{bind}}/A$ replaced by time-averaged $\Delta E_{\text{pot}}^{\text{ES}}/A$ from the 3 ps production period.\protect\footnote{A2's -23.2 is excluded when computing the average score.}}
  \label{fig:DL_3_scoring_AIMD} 
\end{figure}

The mechanical integrity of lithium-ion battery cathodes critically depends on binder adhesion strength, particularly under the chemically aggressive conditions of liquid electrolyte environments. To evaluate candidate binder materials proposed by the DR agents (Figure~\ref{fig:DL_3_candidates_final}), we employed the same computational framework described for PFAS sensors, with binding energies
normalized by the interfacial contact area $A$ to account for varying polymer--cathode interaction footprints. The contact area was determined through convex hull analysis of atoms within 3.5 \AA\ of the NCM811 surface, enabling direct comparison of adhesion strength per unit interfacial area.
Static DFT calculations reveal favorable adhesion for all candidates, with areal-normalized binding energies ranging from $0.015$ to $0.111$ eV/\AA$^2$ (Figure~\ref{fig:DL_3_DFT_binding_histogram}). These values, equivalent to 0.24--1.78 J/m$^2$, align well with established benchmarks for dry interface adhesion, such as graphene on SiO\texorpdfstring{$_2$}{2} ($\sim$0.45 J/m$^2$) and metal--oxide contacts ($\sim$2
J/m$^2$). Candidates B4 and B1 demonstrate the strongest adhesion, while the benchmark PVDF falls in the mid-range, suggesting potential improvements over current commercial binders.
However, the introduction of explicit battery electrolyte (EC:EMC:LiPF$_6$ = 9:9:2:2 at 1.25 g/cm$^3$) through AIMD simulations dramatically alters the adhesion landscape (Figures~\ref{fig:DL_3_AIMD_showcase} and \ref{fig:DL_3_AIMD_gap_histogram}). Following the same protocol of 2 ps equilibration and 3 ps production runs, most candidates exhibit negative $\Delta E_{\text{pot}}^{\text{ES}}/A$
values, indicating that the separated, electrolyte-wetted interface becomes energetically favorable over the bound state. This solvent-induced adhesion reversal reflects competitive adsorption of electrolyte molecules at the binder--cathode interface, disrupting direct polymer--surface interactions.
Four exceptions stand out in this trend: candidates A3 (sulfonyl–polyimide), B2 (lignin + CNT), C1 (PVA), and C2 (CMC + WBPU) maintain positive $\Delta E_{\text{pot}}^{\text{ES}}/A$ even in explicit electrolyte, indicating robust wet adhesion despite first-layer solvent and ion crowding. Mechanistically, S-PI (A3) presents dense imide carbonyl and sulfone dipoles that support Li$^+$-bridged,
multidentate contacts while its aromatic backbone provides broad dispersive $\pi$ contact; the lignin+CNT composite (B2) couples phenolic/ether O donors (ion–dipole/H-bonding) with a large-area CNT $\pi$ interface that helps offset solvent-displacement penalties; PVA (C1) contributes a high density of hydroxyl groups enabling multidentate H-bonding and Li$^+$-mediated coordination to surface sites
with sufficient chain flexibility to maintain conformal contact; and the CMC+WBPU hybrid (C2) combines inner-sphere/ion-paired adhesion from CMC carboxylates (O$\cdots$Li/TM) with urethane H-bonding and local solvent exclusion by the WBPU soft segments. The performance scoring normalized to the PVDF baseline (Figures~\ref{fig:DL_3_scoring_static_DFT_binding} and \ref{fig:DL_3_scoring_AIMD})
quantifies this wet-interface reordering: candidates optimized for dry adhesion may underperform under operating electrolyte conditions, whereas others show unexpected compatibility. These results underscore the need for explicit-solvent, potential-controlled calculations in binder screening, since interfacial energetics in liquid electrolytes are governed by solvent layering, ion crowding, and
potential-dependent desolvation that vacuum or implicit-solvent models cannot capture.

\clearpage

\clearpage

\subsection{OER Stability}

\subsubsection{OER\_Wastewater\_Resistant\_Cats}

For the fourth application: OER Stability. 
The query as shown in \ref{app:topics} is:
\begin{lstlisting}
Identify the top nanostructured electrocatalyst materials for driving the oxygen evolution reaction (OER) in complex wastewater matrices--such as high-chloride, high-organic-load, or multi-ion streams--, for each the material class/composition, comprehensively consider key performance metrics (overpotential at 10 mA cm-2, Faradaic efficiency, stability), and the surface-engineering strategies that
confer corrosion resistance and sustained activity in real effluent conditions.\end{lstlisting}

The best commercial solution is: \textbf{Perplexity\_DR}.

The best local non gpt-oss solution is: \textbf{DToR\_qwq32B\_local500}.

The best local gpt-oss solution is: \textbf{DToR\_gpt-oss120B\_local500}.

The expert based on the two reports, provided 4 commercial DR report derived candidates (A1-A4), and 4 local non gpt-oss DR report derived candidates (B1-B4), and 4 local gpt-oss DR report derived candidates (C1-C4), visualized in Fig. \ref{fig:DL_4_candidates_final}:

\paragraph{\textbf{Benchmark 110 IrO\texorpdfstring{$_2$}{2}}}
For benchmark material, we created most commonly seen, a stoichiometric rutile IrO\texorpdfstring{$_2$}{2} (110) slab built as a 6 $\times$ 3 in-plane supercell along $\times$ [1$\bar{1}$0], with cell parameters \textbf{a = 19.06 \AA{}, b = 19.11 \AA{}, c = 45.18 \AA{}, \texorpdfstring{$\alpha$}{alpha} = \texorpdfstring{$\beta$}{beta} = \texorpdfstring{$\gamma$}{gamma} = 90° (P1)}, from which the underlying bulk metrics map to \textit{a} $\approx$ 4.51 \AA{} and \textit{c} $\approx$
3.18 \AA{}. The slab comprises \textbf{five O--Ir--O trilayers} (thickness $\approx$ 15.18 \AA{}) separated by \textbf{$\approx$ 30.0 \AA{}} of vacuum along \textit{c}, yielding \textbf{symmetric, O-terminated (bridging-O) surfaces} that expose the conventional row motif used for OER modeling (adjacent Ir$_{\text{cus}}$/O$_{\text{br}}$ sites). The model is \textbf{strictly stoichiometric (Ir:O =
1:2; 180 Ir and 360 O)} and thus suitable as a clean, non-polar benchmark surface for computational electrocatalysis studies.

\subparagraph{\textbf{A1 --- Ni--Fe layered double hydroxide (LDH) slab (basal)}}
A hexagonal brucite-derived Ni--Fe hydroxide (001) slab was constructed as a \textbf{6 $\times$ 6} in-plane supercell (basal lattice $a_0\approx$3.126 \AA{}), with cell metrics \textbf{a = b = 18.7560 \AA{}, c = 62.3777 \AA{}, \texorpdfstring{$\alpha$}{alpha} = \texorpdfstring{$\beta$}{beta} = 90°, \texorpdfstring{$\gamma$}{gamma} = 120° (P1)}. The model contains \textbf{three brucite-like $[M(OH)\_6]$} octahedral sheets stacked along \textit{c} (36 cation sites per
sheet), with a \textbf{Ni:Fe = 81:27} distribution (\textbf{25\% Fe on the cation sublattice; 9 Fe per sheet}), and overall stoichiometry \textbf{Ni$\_{81}$Fe$_{27}$O$_{216}$H$_{216}$ (540 atoms)}. Both basal surfaces are \textbf{symmetrically hydroxyl-terminated} (H atoms constitute the outermost layers), yielding a \textbf{stoichiometric, non-polar} slab of thickness \textbf{$\approx$ 19.67
\AA{}}, separated by \textbf{$\approx$ 42.71 \AA{}} of vacuum within the \textbf{c} dimension; the \textbf{inter-sheet metal-plane separations} are \textbf{7.87} and \textbf{7.81 \AA{}}, consistent with LDH-like basal spacing. This geometry provides equivalent top/bottom (001) terminations exposing M--OH motifs suitable for benchmarking OER intermediates on LDH surfaces.

\subparagraph{\textbf{A2 --- phosphate-intercalated Ni--Fe LDH (001)}}
Kept identical to \textbf{A1} in lattice and slab construction (6 $\times$ 6 hexagonal basal supercell; \textit{a} = \textit{b} = 18.756 \AA{}, \texorpdfstring{$\gamma$}{gamma} = 120°, \textit{c} = 62.3777 \AA{}; P1) with symmetric hydroxyl-terminated basal surfaces, \textbf{A2 is derived from A1 by inserting protonated phosphate molecules (H\texorpdfstring{$_3$}{3}PO\texorpdfstring{$_4$}{4})} into the interlayer galleries---\textbf{18 PO\texorpdfstring{$_4$}{4} units per
supercell (9 per gallery)}, giving an overall composition \textbf{Ni$\_{81}$Fe$_{27}$O$_{288}$H$_{270}$P$_{18}$} (684 atoms, Ni:Fe = 3:1). The intercalation shifts the metal-plane separations from $\sim$7.87/7.81 \AA{} (A1) to \textbf{8.89/8.90 \AA{}}, increasing the slab thickness to \textbf{22.22 \AA{}} and leaving \textbf{$\sim$40.16 \AA{}} of vacuum along \textit{c}; the phosphorus centroids
sit near the mid-planes of each gallery.

\subparagraph{\textbf{A3 --- Ni single atoms on Sb-doped SnO\texorpdfstring{$_2$}{2}(110)}}
A rutile SnO\texorpdfstring{$_2$}{2}(110) slab was built as a \textbf{4 $\times$ 8} in-plane supercell along \textbf{[1$\bar{1}$0] $\times$ } (cell \textbf{a = 26.8022 \AA{}, b = 25.4960 \AA{}, c = 57.0945 \AA{}; \texorpdfstring{$\alpha$}{alpha} = \texorpdfstring{$\beta$}{beta} = \texorpdfstring{$\gamma$}{gamma} = 90°; P1}), comprising \textbf{four O--Sn--O trilayers} (slab thickness \textbf{$\approx$ 12.72 \AA{}}) and \textbf{$\approx$ 44.37 \AA{}} of vacuum. \textbf{Eight Sb atoms}
substitute Sn \textbf{in the subsurface cation plane (second layer from the top)}---\textbf{12.5 \%} within that plane, \textbf{3.2 \%} overall (\textbf{Sn$\_{244}$Sb$_8$}). The \textbf{top (110) surface} carries \textbf{four isolated Ni adatoms} arranged in a 2 $\times$ 2 array (\textbf{coverage $\approx$ 0.585 Ni nm$^{-2}$}), each adopting a \textbf{square-pyramidal NiO$\_5$} environment with
\textbf{four equatorial Ni--O = 1.922 \AA{}} and \textbf{one apical Ni--O = 2.033 \AA{}}; the \textbf{bottom surface} is left \textbf{clean and O-terminated}. The full supercell stoichiometry is \textbf{Sn$\_{244}$Sb$_8$Ni\texorpdfstring{$_4$}{4}O$_{508}$ (764 atoms)}.

\subparagraph{\textbf{A4 --- Ce-doped, oxygen-deficient spinel NiCo\texorpdfstring{$_2$}{2}O\texorpdfstring{$_4$}{4} slab (asymmetric OH termination)}}
Modeled in \textbf{P1} with an oblique in-plane cell (\textbf{a = 19.807 \AA{}, b = 30.866 \AA{}, \texorpdfstring{$\gamma$}{gamma} = 108.715°; c = 41.150 \AA{}}), giving an in-plane area of \textbf{579.05 \AA{}$^2$ (5.79 nm$^2$)}; the atomic slab spans \textbf{14.87 \AA{}} along \textit{c} and is separated by \textbf{26.28 \AA{}} of vacuum. The supercell composition is \textbf{Co$\_{188}$Ce$_8$Ni$_{92}$O$_{376}$H$_{96}$
(760 atoms)}, i.e., \textbf{Ni:Co $\approx$ 1:2.04} with \textbf{Ce at 2.8 at.\% of the cation sublattice}. Relative to the ideal AB\texorpdfstring{$_2$}{2}O\texorpdfstring{$_4$}{4} stoichiometry for 288 cations, the model contains \textbf{8 lattice-oxygen vacancies} (\textbf{x $\approx$ 0.021 in NiCo\texorpdfstring{$_2$}{2}O$\_{4-x}$}), corresponding---if evenly split between the two faces---to \textbf{$\approx$ 0.69 vac nm$^{-2}$} per surface.
\textbf{Ce dopants occupy two equivalent subsurface cation planes} (four Ce per plane), located \textbf{$\sim$4.31 \AA{} below the top} and \textbf{$\sim$5.32 \AA{} above the bottom} surface, preserving an overall centrosymmetric placement. The \textbf{top surface is Co-rich and partially hydroxylated} (within the outermost 2 \AA{}: \textbf{8 Co}, \textbf{20 O}, \textbf{28 H} \texorpdfstring{$\rightarrow$}{->}
\textbf{$\approx$ 1.38 Co nm$^{-2}$} and \textbf{$\approx$ 5.5 OH nm$^{-2}$}), while the \textbf{bottom surface is more strongly hydroxyl-passivated} (\textbf{64 H} on that side overall, \textbf{$\approx$ 11.1 OH nm$^{-2}$}). No Ni appears within the outermost 2 \AA{} of the top surface (Ni resides in subsurface layers), yielding a \textbf{Co--O(H) termination} appropriate for probing OER pathways
on spinel-type oxides.

\subparagraph{\textbf{B1 --- graphene-capped Ni--Fe LDH (001)}}
Retaining the \textbf{A1} lattice and slab construction (6 $\times$ 6 hexagonal basal supercell; \textit{a} = \textit{b} = 18.756 \AA{}, \texorpdfstring{$\gamma$}{gamma} = 120°, \textit{c} = 70.6707 \AA{}; P1), this model is derived from \textbf{A1} by placing a \textbf{single-layer graphene overlayer (128 C)} above the top basal surface while keeping the LDH stoichiometry \textbf{Ni$\_{81}$Fe$_{27}$O$_{216}$H$_{216}$}
unchanged. The LDH comprises \textbf{three brucite-like [M(OH)$\_6$] sheets} (Ni:Fe = 3:1) with \textbf{metal-plane separations of 7.87 and 7.81 \AA{}}; both basal faces are hydroxyl-terminated, but only the \textbf{top face} is covered by graphene. The LDH portion spans \textbf{19.67 \AA{}} along \textit{c}; including graphene, the occupied thickness is \textbf{24.67 \AA{}}, leaving \textbf{46.00
\AA{}} of vacuum. The graphene sheet is essentially planar (out-of-plane corrugation $\leq$ 0.001 \AA{}) and sits \textbf{5.00 \AA{} above the outermost hydroxyl-H layer} (\textbf{5.97 \AA{} above the topmost O layer}), consistent with a physisorbed, non-perturbative interface that preserves the LDH surface structure; the \textbf{bottom layers are identical to A1} and remain exposed/hydroxylated
for benchmarking.

\subparagraph{\textbf{B2 --- graphene-capped hematite (\texorpdfstring{$\alpha$}{alpha}-Fe\texorpdfstring{$_2$}{2}O\texorpdfstring{$_3$}{3}) (0001) with an aperture}}
A hexagonal \texorpdfstring{$\alpha$}{alpha}-Fe\texorpdfstring{$_2$}{2}O\texorpdfstring{$_3$}{3}(0001) slab was modeled in \textbf{P1} with \textbf{a = b = 20.152 \AA{}, \texorpdfstring{$\gamma$}{gamma} = 120°; c = 60.339 \AA{}} (in-plane area \textbf{351.70 \AA{}$^2$ = 3.517 nm$^2$}), giving an atomistic thickness of \textbf{32.20 \AA{}} separated by \textbf{28.14 \AA{}} of vacuum. The slab is \textbf{stoichiometric Fe\texorpdfstring{$_2$}{2}O\texorpdfstring{$_3$}{3}} in the interior (\textbf{Fe$\_{384}$O$_{576}$}) with
\textbf{both faces hydroxyl-terminated (H$\_{92}$ total)} to render a non-polar model. A \textbf{single sp$^2$-carbon overlayer (C$\_{115}$)} caps only the \textbf{top} face and includes a \textbf{circular $\sim$6 \AA{} ``window''} (aperture) that leaves a patch of \textbf{exposed surface Fe sites} beneath the opening; elsewhere the carbon sheet remains graphene-like, with its centroid
\textbf{$\sim$2.84 \AA{} above the outermost O plane} ($\approx$ \textbf{2.04 \AA{} above terminal H}) and a small out-of-plane corrugation (\textbf{$\Delta$z $\approx$ 0.91 \AA{}}). The \textbf{bottom} face remains hydroxylated and exposed for benchmarking.

\subparagraph{\textbf{B3 --- BaCo\texorpdfstring{$_{0.6}$}{0.6}Fe\texorpdfstring{$_{0.4}$}{0.4}O\texorpdfstring{$_3$}{3} perovskite (001), BO\texorpdfstring{$_2$}{2}\texorpdfstring{$\parallel$}{||}BaO asymmetric slab}}
In line with Branch B's chloride-resilience goal, a stoichiometric Ba(B)O\texorpdfstring{$_3$}{3} slab with mixed B-site cations (Ba$_{180}$Co$_{108}$Fe$_{72}$O$_{540}$; 900 atoms) was modeled in P1 with \textbf{a = b = 23.820 \AA{}, c = 49.850 \AA{}, \texorpdfstring{$\alpha$}{alpha} = \texorpdfstring{$\beta$}{beta} = \texorpdfstring{$\gamma$}{gamma} = 90°} (in-plane area \textbf{567.39 \AA{}$^2$ = 5.674 nm$^2$}). The atomic slab spans \textbf{19.31 \AA{}} along \textit{c} and is
separated by \textbf{30.54 \AA{}} of vacuum. Terminations are \textbf{asymmetric}: the \textbf{top} face is a \textbf{BO\texorpdfstring{$_2$}{2} layer} that is \textbf{Co-rich (Co:Fe = 27:9 $\approx$ 3:1)}, while the \textbf{bottom} face is \textbf{BaO-terminated}; within $\pm$2 \AA{} of the surfaces this corresponds to $\sim$\textbf{6.35 B nm$^{-2}$} (with $\sim$\textbf{12.69 O nm$^{-2}$}) at the top and
$\sim$\textbf{6.35 Ba nm$^{-2}$} at the bottom. This construction furnishes a corrosion-tolerant, \textbf{bare BO\texorpdfstring{$_2$}{2} active surface} over a BaO support, consistent with the initial design brief to benchmark robust oxide surfaces against chloride attack.

\subparagraph{\textbf{B4 --- coral-mimetic Ni--Fe LDH (001)}}
Keeping the MXene/LDH base identical to \textbf{A1/B1} (6 $\times$ 6 hex basal supercell; \textit{a} = \textit{b} = 18.756 \AA{}, \texorpdfstring{$\gamma$}{gamma} = 120°, \textit{c} = 62.3777 \AA{}; P1), the ``coral'' morphology was introduced by excising a single cylindrical pore ($\sim$8 \AA{} diameter) through the \textbf{top two} hydroxide layers and hydroxyl-passivating all newly exposed edges, as specified in the
design note; both basal faces remain hydroxyl-terminated, with the \textbf{bottom face unchanged} from A1/B1. The resulting model is \textbf{Ni$\_{69}$Fe$_{25}$O$_{192}$H$_{216}$ (502 atoms)}, occupying \textbf{26.58 \AA{}} along \textit{c} with \textbf{35.79 \AA{}} vacuum (in-plane area \textbf{304.66 \AA{}$^2$}), yielding a non-polar, porous benchmark that preserves the original basal registry
while exposing M--OH rim sites within the aperture.

\subparagraph{\textbf{C1 — FeNi-OOH/spinel nanoplate on N-doped carbon scaffold with amorphous Al–O/Ti–O overlayer}}
Derived from \textbf{B1} (Ni--Fe LDH on graphene), retaining the hexagonal \textbf{6 $\times$ 6} basal supercell (\textbf{a = b = 18.756 \AA{}, \texorpdfstring{$\gamma$}{gamma} = 120°; c $\approx$ 70.6707 \AA{}; P1}). The graphene support is \textbf{N-doped} by substituting \textbf{6 C \texorpdfstring{$\rightarrow$}{->} N} (\textbf{$\sim$4.7\%} of 128 C) at distributed sites (pyridinic/pyrrolic character), while the LDH layer remains
\textbf{Ni$_{81}$Fe$_{27}$O$_{216}$H$_{216}$}. The exposed LDH face is capped by \textbf{8 amorphous Al--O/Ti--O clusters} (\textbf{4} small, 4--6 atoms; \textbf{4} medium, 6--9 atoms) at \textbf{$\sim$25--30\%} areal coverage; clusters sit \textbf{$\sim$2.5 \AA{}} above the surface, are spaced \textbf{$\ge$ 3.5 \AA{}} apart, and show \textbf{Al--O $\sim$ 1.8 \AA{}} and \textbf{Ti--O $\sim$ 1.9
\AA{}} with \textbf{$\pm$15\%} distortions (SALD-like islands that leave lateral windows/edge sites accessible). After relaxation the oxide islands form interfacial bonds to LDH hydroxyls while the N-doped graphene adjusts registry. Overall composition: \textbf{722 atoms} (\textbf{C$_{122}$N$_6$Fe$_{27}$Ni$_{81}$O$_{251}$Al$_5$Ti$_{14}$H$_{216}$}), a tractable model of nanoplate/carbon composites
with selective SALD protection.

\subparagraph{\textbf{C2 — Fe single atom on Nb-doped anatase TiO\texorpdfstring{$_2$}{2}(101) with amorphous Al–O/Ti–O overlayer}}
An anatase TiO\texorpdfstring{$_2$}{2}(101) slab was built as a \textbf{2 $\times$ 4} lateral supercell (\textbf{b $\approx$ 15.13 \AA{}} after doubling), comprising \textbf{5 trilayers} (\textbf{slab thickness $\approx$ 17 \AA{}}) and \textbf{$\approx$ 15 \AA{}} vacuum (\textbf{P1}). \textbf{Four Ti} in the first subsurface cation plane are replaced by \textbf{Nb} (\textbf{$\sim$2.5 at.\%}); a \textbf{surface
bridging-O vacancy} is created \textbf{$\sim$1.08 \AA{}} from the nearest subsurface Nb. A \textbf{single Fe adatom} anchors at this vacancy (fivefold/bridge motif). The surface is capped by \textbf{7 amorphous Al--O/Ti--O clusters} (\textbf{4} small, 4--6 atoms; \textbf{3} medium, 6--9 atoms) at \textbf{$\sim$25--30\%} coverage, each \textbf{$\sim$2.5 \AA{}} above the surface and \textbf{$\ge$
3.5 \AA{}} apart, with \textbf{Al--O $\sim$ 1.8 \AA{}} and \textbf{Ti--O $\sim$ 1.9 \AA{}} (\textbf{$\pm$15\%} distortions). Clusters maintain \textbf{$\ge$ 4 \AA{}} lateral clearance to the Fe site to preserve accessibility; relaxation yields interfacial Al--O--Ti linkages while the Fe coordination adapts to the vacancy environment.

\subparagraph{\textbf{C3 — Ru single atom on W-doped anatase TiO\texorpdfstring{$_2$}{2}(101) with amorphous Al–O/Ti–O overlayer}}
Constructed analogously to \textbf{C2}: anatase TiO\texorpdfstring{$_2$}{2}(101) \textbf{2 $\times$ 4} lateral supercell (\textbf{b $\approx$ 15.13 \AA{}}), \textbf{5 trilayers} (\textbf{$\approx$ 17 \AA{}} thick) and \textbf{$\approx$ 15 \AA{}} vacuum (\textbf{P1}). \textbf{Four Ti} in the first subsurface cation plane are replaced by \textbf{W} (\textbf{$\sim$2.5 at.\%}); a \textbf{surface bridging-O vacancy} is
placed \textbf{$\sim$1.08 \AA{}} from the nearest subsurface W. A \textbf{single Ru adatom} anchors at the vacancy. The surface carries \textbf{8 amorphous Al--O/Ti--O clusters} (\textbf{4} small, 4--6 atoms; \textbf{4} medium, 6--9 atoms) at \textbf{$\sim$25--30\%} coverage, each \textbf{$\sim$2.5 \AA{}} above and \textbf{$\ge$ 3.5 \AA{}} apart, with \textbf{Al--O $\sim$ 1.8 \AA{}} and
\textbf{Ti--O $\sim$ 1.9 \AA{}} (\textbf{$\pm$15\%} distortions) and \textbf{$\ge$ 4 \AA{}} lateral clearance to the Ru site. Relaxation forms interfacial Al--O--Ti bonds while preserving accessibility of the \textbf{Ru single-atom} OER site.

\subparagraph{\textbf{C4 — Gradient-doped NiCo\texorpdfstring{$_2$}{2}O\texorpdfstring{$_4$}{4} spinel (111) with amorphous Al–O/Ti–O overlayer}}
Based on the \textbf{A4} spinel slab, modeled along \textbf{(111)} with in-plane metrics \textbf{$\sim$20 \AA{}} and \textbf{c $\approx$ 41 \AA{}} (\textbf{$\sim$15 \AA{}} vacuum; \textbf{P1}). The \textbf{Ce dopants} in A4 are replaced by a \textbf{gradient dopant distribution}: \textbf{2 Cr} near the surface (\textbf{z $\approx$ 11.76 \AA{}}), \textbf{4 Ti} in mid-subsurface (two at \textbf{z
$\approx$ 11.76 \AA{}}, two at \textbf{z $\approx$ 6.97 \AA{}}), and \textbf{2 Mg} deeper (\textbf{z $\approx$ 6.97 \AA{}})---\textbf{$\sim$1\%} combined dopants in the \textbf{760-atom} base, enriching the \textbf{BO-rich} surface for corrosion tolerance; existing \textbf{lattice O-vacancies} from A4 are retained. The surface is capped by \textbf{8 amorphous Al--O/Ti--O clusters} (\textbf{4}
small, 4--6 atoms; \textbf{4} medium, 6--9 atoms) at \textbf{$\sim$25--30\%} coverage, each \textbf{$\sim$2.5 \AA{}} above and \textbf{$\ge$ 3.5 \AA{}} apart, with \textbf{Al--O $\sim$ 1.8 \AA{}} and \textbf{Ti--O $\sim$ 1.9 \AA{}} (\textbf{$\pm$15\%} distortions), representing a 3--5 nm nanolaminate protection motif. After relaxation, clusters form Al--O--(Co,Ni) linkages while the dopant
gradient accommodates lattice/electronic modulation. Final composition: \textbf{814 atoms} (\textbf{Co$_{188}$Ni$_{92}$Cr\texorpdfstring{$_2$}{2}Mg\texorpdfstring{$_2$}{2}Ti$_{18}$O$_{411}$Al$_5$H$_{96}$}); the \textbf{(111)} surface itself provides the primary OER sites, with the gradient and ALD layer furnishing complementary stability.

\textbf{Metal Site Choices}
Considering the complexity of each slab created based on DR report, we decide to select from each slab only representative sites for further study:

\subparagraph{\textbf{IrO\texorpdfstring{$_2$}{2}}}
On IrO\texorpdfstring{$_2$}{2}(110) we target the canonical fivefold Ir terrace site (Ir$_{5c}$) on the top surface cation row. Its local geometry is square-pyramidal, with four coplanar surface O neighbors at 2.00 \AA{} and a single subsurface apical O at 1.97 \AA{}; the missing apical ligand points toward vacuum. Situated in the outermost O--Ir--O trilayer, this terrace Ir is the standard benchmark adsorption site
on rutile-type oxides and is chosen to probe intrinsic, defect-free reactivity under well-defined coordination.

\subparagraph{\textbf{A1}}
On A1 we probe two "regular" terrace cations on the \textbf{top basal (001) surface}---one Fe and one Ni located in the central, step-free region. Both exhibit near-ideal octahedral \textbf{M(OH)$\_6$} coordination with a 3-up/3-down O arrangement; their mean M--O distances are \textbf{2.09 \AA{} (Fe)} and \textbf{2.07 \AA{} (Ni)}, and the in-plane metal--metal spacing reflects the basal lattice
at \textbf{$\sim$3.13 \AA{}}. The Fe site sits in a Ni-rich cation shell (isolated Fe amid Ni), whereas the Ni site has a typical mixed neighborhood ($\sim$4 Ni, 2 Fe), consistent with the $\sim$3:1 Ni:Fe layer composition. These terrace positions were chosen as \textbf{representative, bulk-like references}---well coordinated and far from edges or defects---providing robust baselines for
adsorption benchmarking on LDH surfaces.

\subparagraph{\textbf{A2}}
In \textbf{A2} we focus on two regular first-layer cations on the \textbf{top basal (001) surface}---one Fe and one Ni---above the phosphate-filled interlayer. Both sites are octahedral with \textbf{3 O above / 3 O below}; the Fe exhibits \textbf{Fe--O $\approx$ 2.06/2.01 \AA{}} (above/below) and the Ni \textbf{Ni--O $\approx$ 2.10/2.07 \AA{}}. Relative to the subsurface phosphate, the Fe sits
\textbf{nearly on-axis} over a PO\texorpdfstring{$_4$}{4} unit (\textbf{$\Delta$z $\approx$ 4.30 \AA{}}, lateral offset \textbf{$\approx$ 0.71 \AA{}}), whereas the Ni is \textbf{laterally off-axis} from the nearest PO\texorpdfstring{$_4$}{4} (\textbf{$\Delta$z $\approx$ 4.54 \AA{}}, lateral \textbf{$\approx$ 2.95 \AA{}}). These two surface metals are chosen as representative, symmetry-inequivalent terrace sites that sample
\textbf{distinct interlayer electrostatic environments} (aligned vs off-axis to PO\texorpdfstring{$_4$}{4}) while avoiding edges/defects---providing a solid, structure-based rationale for benchmarking metal-identity and phosphate-proximity effects.

\subparagraph{\textbf{A3}}
For A3, on the top SnO\texorpdfstring{$_2$}{2}(110) terrace, the representative site comprises a single Ni atom anchored above a bridging-O row, adopting a square-pyramidal NiO$_5$ geometry with four surface O$_{br}$ ligands in the equatorial plane (Ni--O $\approx$ 1.92 \AA{}) and one apical O beneath (Ni--O $\approx$ 2.03 \AA{}); the Ni sits $\sim$0.19 \AA{} below the O$_{br}$ plane and $\sim$2.03 \AA{} above the
apical O. The two nearest subsurface Sn cations on adjacent cation rows are each connected to the adatom via a Sn--O$_{br}$--Ni bridge (Sn--O $\approx$ 2.07 \AA{}; Ni--Sn $\approx$ 3.12 \AA{}). Situated on the upper, flat terrace away from step edges, this motif matches your original "Ni single-atom adjacent to Sn via bridging O" note and is selected as a representative active site because it
captures the canonical O$_{br}$--M$_{cus}$ environment of rutile(110) while providing quantitatively well-defined, low-coverage coordination ideal for benchmarking OER intermediates.

\subparagraph{\textbf{A4}}
For A4, we have choosed a total of 6 Co/Ni atoms. On the top Co--O(H)-terminated face of the Ce-doped, oxygen-deficient NiCo\texorpdfstring{$_2$}{2}O\texorpdfstring{$_4$}{4} slab, we consider two Ce-proximal motifs and one Ce-remote control set. Ce-proximal: (i) a two-fold-coordinated Co sitting at the rim of a surface O-vacancy directly above the subsurface Ce layer, and (ii) a three-fold Co one O row back on the same terrace;
alongside these, a three-fold Ni in the same Co-rich layer but still within the Ce-perturbed field captures the B-site identity contrast. Ce-remote (terrace center, far from dopants/vacancies): (iii) a two-/three-fold Co pair representative of the intact hydroxylated terrace, and (iv) a two-fold Ni surrounded by a largely unreconstructed O sublattice. Together these five local environments span
coordination (2 vs. 3), cation identity (Co vs. Ni), and dopant/vacancy proximity (near-Ce vs. terrace-center) consistent with the A4 design (Ce in subsurface planes + engineered surface O-vacancies with hydroxyl passivation). This refelects the selection logic: Ce-adjacent rims/near-rim terrace sites plus Ce-remote terrace controls with 2- and 3-coordinate Co/Ni.

\subparagraph{\textbf{B1}}
On \textbf{B1 (graphene-capped NiFe-LDH (001))}, the representative \textbf{first-layer surface Fe and Ni} are the outermost cations in the top brucite-like sheet directly beneath the graphene overlayer. Each sits in a \textbf{regular M(OH)$\_6$ octahedron} with nearly isotropic M--O bonds (Fe: \textbf{2.09 \AA{}} apical/\textbf{2.09 \AA{}} apical; equatorial \textbf{2.09 $\pm$ 0.00 \AA{}}; Ni:
\textbf{2.08 \AA{}} apical/\textbf{2.08 \AA{}} apical; equatorial \textbf{2.07 $\pm$ 0.00 \AA{}}; octahedral distortion \textbf{$\approx$ 0.11 \%}), positioned \textbf{1.03 \AA{}} below the topmost O plane and \textbf{2.00 \AA{}} below terminal OH, with the graphene sheet \textbf{7.00 \AA{}} above the metal plane (\textbf{5.97 \AA{}} above the surface O layer). The chosen Fe lies near the in-plane
center (\textbf{$>$ 8.09 \AA{}} from periodic edges) and the Ni is also well inside the terrace (\textbf{$\approx$ 5.45 \AA{}} from edges), i.e., \textbf{regular terrace sites} rather than step/edge motifs---matching the "regular first-layer" description in the earlier manual note. These sites were selected to provide \textbf{clean, defensible benchmarks} for OER under Branch-B1's design goal
(probing LDH activity beneath a protective graphene barrier): they are far from defects and intercalants, exhibit bulk-like octahedral geometry, and experience only \textbf{physisorbed} graphene influence, thus isolating intrinsic Fe/Ni--OH reactivity at the \textbf{top basal surface}.

\subparagraph{\textbf{B2}}
on B2, On the top (0001) face beneath the graphene aperture, four representative surface Fe sites were selected: two \textbf{rim} Fe located along the edge of the $\sim$6 \AA{} window where the sp$^2$ rim approaches the oxide, each making Fe--C contact (shortest Fe--C $\approx$ 2.17 \AA{}), with under-coordination (Fe--O CN $\approx$ 3) and a slight outward relaxation ($\approx$ 1.33 \AA{} above
the terminal O plane); and two \textbf{center-of-window} Fe positioned near the middle of the opening, far from the carbon edge (nearest C $\geq$ 5.24 \AA{}), retaining octahedral-like coordination (Fe--O CN $\approx$ 6) and sitting $\approx$ 1.19 \AA{} below the top O/H plane. These two site families---carbon-contacted rim vs carbon-free aperture interior---follow the original design intent and
provide a physically motivated contrast to quantify how the graphene shell perturbs intrinsic hematite reactivity.

\subparagraph{\textbf{B3}}
On the BO\texorpdfstring{$_2$}{2}-terminated top surface of B3 we sample four terrace cations: (i) one \textbf{Fe} and (ii) one \textbf{Co} sitting at the rim of a deliberate surface O-vacancy, each \textbf{three-coordinated} to lattice O (vacancy-edge sites, no apical O) with shortened in-plane M--O distances of $\sim$\textbf{1.90--1.97 \AA{}}; and, as references, (iii) one \textbf{Fe} and (iv) one \textbf{Co} on
the same terrace but \textbf{far from any defect} ($\geq$\textbf{10.00 \AA{}} lateral separation), retaining the regular \textbf{square-pyramidal BO$\_5$} environment (four equatorial M--O $\approx$ \textbf{1.95--2.02 \AA{}}, one apical $\approx$ \textbf{2.05--2.10 \AA{}}). This choice---consistent with the original notes---targets the most contrastive local motifs (vacancy-edge undercoordination
vs pristine terrace; Fe vs Co identity) while keeping all sites in the \textbf{outermost BO\texorpdfstring{$_2$}{2} plane} above the BaO underlayer.

\subparagraph{\textbf{B4}}
In B4's coral-type aperture, we target four representative rim sites: a Fe and a Ni in the \textbf{top metal plane} at the pore edge, and the corresponding \textbf{second-plane} Fe and Ni located directly beneath the rim (as in the earlier manual note). The two top-layer cations sit \textbf{1.88 \AA{} (Fe)} and \textbf{1.55 \AA{} (Ni)} below the outermost O plane and lie within \textbf{0.10--0.23
\AA{}} of the top-plane centroid; their in-plane separation is \textbf{3.21 \AA{}}. At the rim, \textbf{Fe remains near-octahedral} (\textbf{6 O $\leq$ 2.30 \AA{}; $\langle d_{\text{M--O}}\rangle$ = 2.13 \AA{}}), whereas \textbf{Ni is under-coordinated} (\textbf{4 O $\leq$ 2.30 \AA{}; $\langle d_{\text{M--O}}\rangle$ = 2.12 \AA{}}). The second-layer pair occupies the ring immediately below the
aperture edge, \textbf{13.53--13.43 \AA{}} beneath the top O plane and within \textbf{0.09--0.02 \AA{}} of the second-plane centroid; both are \textbf{under-coordinated} (\textbf{4 O $\leq$ 2.30 \AA{}; $\langle d_{\text{M--O}}\rangle$ = 2.12 \AA{} (Ni), 2.06 \AA{} (Fe)}), with an in-plane separation of \textbf{3.21 \AA{}}. These sites were chosen to systematically span \textbf{cation identity (Fe
vs Ni)}, \textbf{depth (surface vs subsurface)}, and \textbf{coordination state at the pore edge}, i.e., the structural factors most likely to govern *OH/*O/*OOH binding and chloride tolerance in the coral design.

\subparagraph{\textbf{C1}}
On \textbf{C1 (Ni--Fe LDH on 5\% N-doped graphene with partial SALD Al--O/Ti--O coverage)}, we sample \textbf{two terrace metals} on the \textbf{top basal (001)} sheet (\textbf{Fe} and \textbf{Ni}) and one \textbf{subsurface Fe} to probe depth effects under the overlayer. The \textbf{surface Fe} sits in a near-ideal \textbf{Fe(OH)$_6$} octahedron (\textbf{Fe--O $\approx$ 2.09 \AA{}}), the
\textbf{surface Ni} in a matching \textbf{Ni(OH)$_6$} environment (\textbf{Ni--O $\approx$ 2.07 \AA{}}), both placed in the \textbf{central, step-free terrace} beneath regions \textbf{not capped} by amorphous clusters. The \textbf{subsurface Fe} resides one hydroxide layer down (\textbf{$\Delta z \approx 7.8$ \AA{}} relative to the top Fe) with octahedral \textbf{Fe--O $\approx$ 2.10 \AA{}}. These
\textbf{Fe/Ni surface} vs \textbf{Fe subsurface} choices span \textbf{metal identity} and \textbf{depth}, enabling paired benchmarks of dissolution propensity and local lattice oxygen stability in the composite LDH/carbon/SALD architecture. For VO tests we target the \textbf{lowest-coordination} O neighbors of each metal (CN 1--3 within \textbf{2.0--2.10 \AA{}}) to provide a consistent,
edge-biased vacancy protocol.

\subparagraph{\textbf{C2}}
On \textbf{C2 (Fe single atom on Nb-doped anatase TiO\texorpdfstring{$_2$}{2}(101) with amorphous ALD clusters)}, we select the \textbf{Fe adatom} anchored at a surface \textbf{O-vacancy} (square-pyramidal, bridging-O motif) as the \textbf{core active site}, then pair it with \textbf{two subsurface Nb dopants}: a \textbf{near} Nb directly beneath Fe in the first cation plane (\textbf{$\Delta z \approx 1.9$ \AA{}};
nearly on-axis), and a \textbf{far} Nb in the same plane but \textbf{$\sim$4.25 \AA{}} laterally displaced. This triad isolates \textbf{active-site stability} (Fe) and \textbf{dopant anchoring/retention} (Nb near vs far) under the partially passivated surface. For VO we remove the \textbf{least-coordinated} O within \textbf{$\le$ 3.0 \AA{}} of each target (CN $\approx$ 2; \textbf{Fe} neighbor at
\textbf{$\sim$2.46 \AA{}}, \textbf{Nb} neighbors at \textbf{$\sim$1.94 \AA{}}) to assess oxygen-lattice integrity in proximity-contrasted environments.

\subparagraph{\textbf{C3}}
On \textbf{C3 (Ru single atom on W-doped anatase TiO\texorpdfstring{$_2$}{2}(101) with amorphous ALD clusters)}, we mirror the C2 schema: the \textbf{Ru adatom} at a surface \textbf{O-vacancy} (square-pyramidal, bridging-O motif) serves as the \textbf{primary site}, coupled to \textbf{two first-subsurface W dopants}---one \textbf{near/on-axis} (\textbf{$\Delta z \approx 1.9$ \AA{}} beneath Ru) and one
\textbf{laterally offset} by \textbf{$\sim$4.25 \AA{}}. This comparison quantifies the \textbf{noble-metal advantage} (Ru vs Fe) and the effect of \textbf{higher-valent dopant} (W$^{6+}$ vs Nb$^{5+}$) on both \textbf{active-site} and \textbf{dopant} stability. VO tests again remove the \textbf{lowest-CN} O within \textbf{$\le$ 3.0 \AA{}} (CN $\approx$ 2; neighbors at \textbf{$\sim$2.46 \AA{}} for
Ru and \textbf{$\sim$1.94 \AA{}} for W) to benchmark vacancy energetics across the proximity series.

\subparagraph{\textbf{C4}}
On \textbf{C4 (gradient-doped NiCo\texorpdfstring{$_2$}{2}O\texorpdfstring{$_4$}{4} spinel with amorphous ALD protection)}, we probe \textbf{three cations} spanning the designed gradient: a \textbf{surface-enriched Cr} dopant in the upper octahedral layer (\textbf{z $\approx$ 25.3 \AA{}}), a \textbf{Co} octahedral site \textbf{near Cr} (\textbf{Cr--Co $\approx$ 2.86 \AA{}}; same region), and a \textbf{Co} octahedral \textbf{far from all
Cr} (\textbf{$\sim$23 \AA{}} away) as the baseline. This set interrogates whether \textbf{aliovalent Cr} resists leaching and whether \textbf{Cr proximity} stabilizes neighboring \textbf{Co--O} bonding relative to \textbf{Cr-remote} Co in the spinel framework. For VO we remove under-coordinated O neighbors (e.g., \textbf{$\sim$1.91--1.93 \AA{}} to Cr/Co; CN \textbf{2--4}) to consistently stress
the lattice at the most vulnerable positions across \textbf{Cr-near} and \textbf{Cr-far} environments.

\subsubsection{OER Catalyst Stability Assessment}

\begin{figure}[htbp]
  \centering
  \includegraphics[width=1\textwidth]{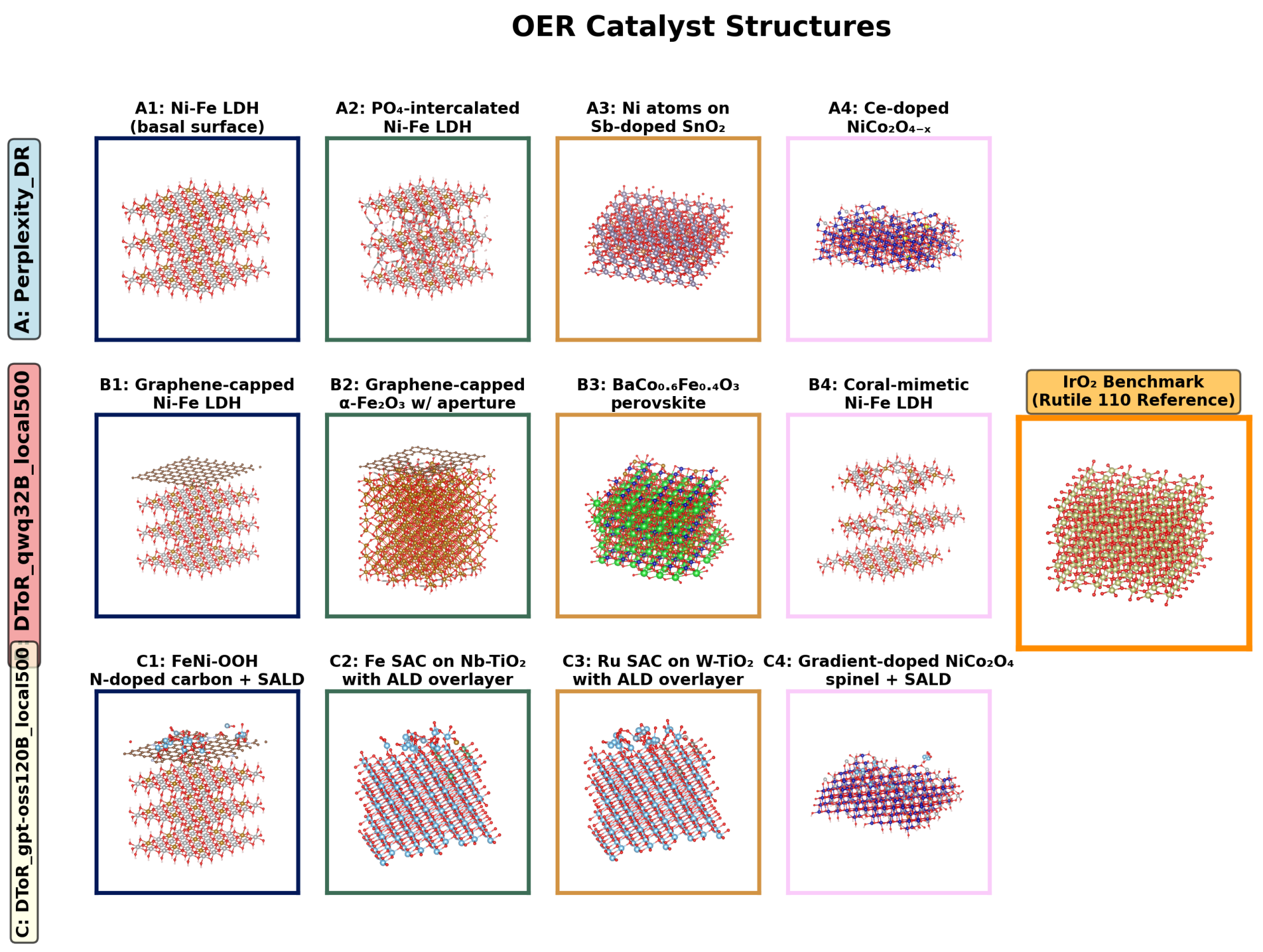}
  \caption{Structural configurations of OER catalyst candidates proposed by DR agents, with IrO\texorpdfstring{$_2$}{2} (110) included as the domain benchmark reference for stability assessment.}
  \label{fig:DL_4_candidates_final}  
\end{figure}

\textbf{Discussion of OER stability.} While recent advances in transition metal layered double hydroxides (LDH) and related materials have achieved excellent oxygen evolution reaction (OER) activity approaching theoretical limits, catalyst stability remains the critical bottleneck preventing commercial deployment of water splitting technologies. Operating potentials exceeding 1.5 V vs. RHE in alkaline conditions create thermodynamically favorable pathways for both metal dissolution and lattice oxygen loss, leading to rapid catalyst degradation. To address this fundamental challenge, we evaluated candidate catalysts proposed by the DR agents (Figure~\ref{fig:DL_4_candidates_final}) using two complementary stability descriptors following established protocols from our previous work \cite{ding2025oermulti}: dissolution potential (U$_{\text{diss}}$) and oxygen vacancy formation energy (V$_{\text{O}}$).

\textbf{Stability Descriptors and Physical Interpretation}

\begin{figure}[htbp]
  \centering
  \includegraphics[width=1\textwidth]{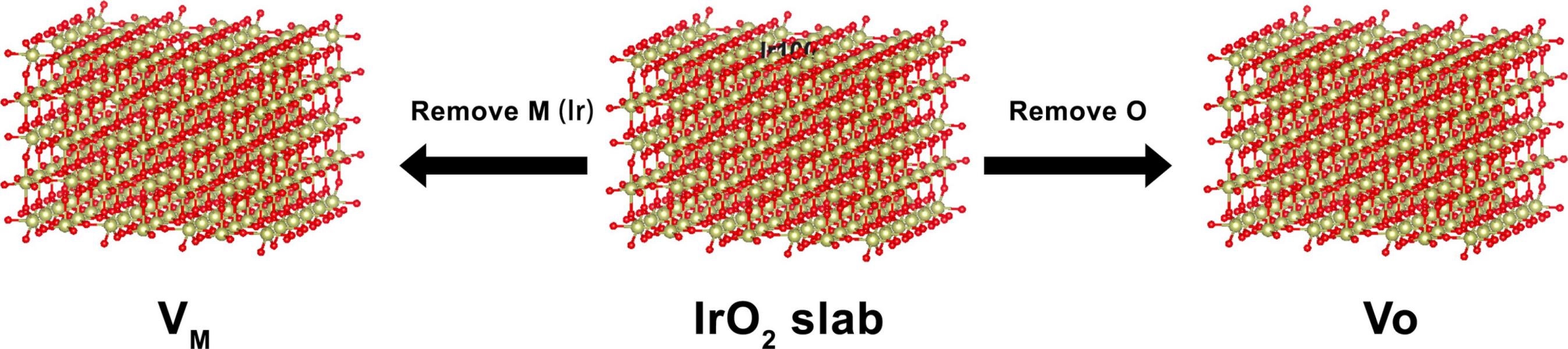}
  \caption{Schematic illustration of vacancy creation on the IrO\texorpdfstring{$_2$}{2} rutile (110) surface. Left: Metal vacancy (V$_{\text{M}}$) created by removing a surface metal atom for dissolution potential (U$_{\text{diss}}$) calculation. Right: Oxygen vacancy (V$_{\text{O}}$) created by removing a surface oxygen atom for vacancy formation energy calculation.}  
  \label{fig:DL_4_Vm_Vo_showcase} 
\end{figure}

The dissolution potential U$_{\text{diss}}$ quantifies the thermodynamic driving force for metal atom leaching from the catalyst surface, calculated as:

\begin{equation}
U_{\text{diss}}(M@\text{slab}) = U^{\circ}_{\text{diss,bulk}}(M^{z+}/M) + \frac{E_{\text{slab-vac}}(M) + \mu_{\text{bulk}}(M) - E_{\text{slab}}}{ne}
\end{equation}

where $E_{\text{slab}}$ and $E_{\text{slab-vac}}(M)$ represent the DFT energies of the intact and metal-vacancy surfaces, respectively (Figure~\ref{fig:DL_4_Vm_Vo_showcase}), $\mu_{\text{bulk}}(M)$ is the bulk metal chemical potential, $U^{\circ}_{\text{diss,bulk}}$ is the standard reduction potential, and $n$ is the number of electrons transferred. Higher U$_{\text{diss}}$ values indicate greater
resistance to metal dissolution, with positive values suggesting thermodynamic stability under standard conditions.
The oxygen vacancy formation energy V$_{\text{O}}$ captures the structural integrity of the oxide lattice during OER cycling. We following the following protocol for picking lattice oxygen atom to remove, oxygen atoms are identified within 3.0 Å of active metal sites and removed based on coordination number, with under-coordinated oxygen atoms preferentially selected
(Figure~\ref{fig:DL_4_Vm_Vo_showcase}). Higher V$_{\text{O}}$ values indicate stronger metal--oxygen bonds and reduced susceptibility to lattice oxygen participation in the OER mechanism, which can lead to irreversible structural degradation.
\textbf{Comparative Stability Analysis}

\begin{figure}[htbp]
  \centering
  \includegraphics[width=1\textwidth]{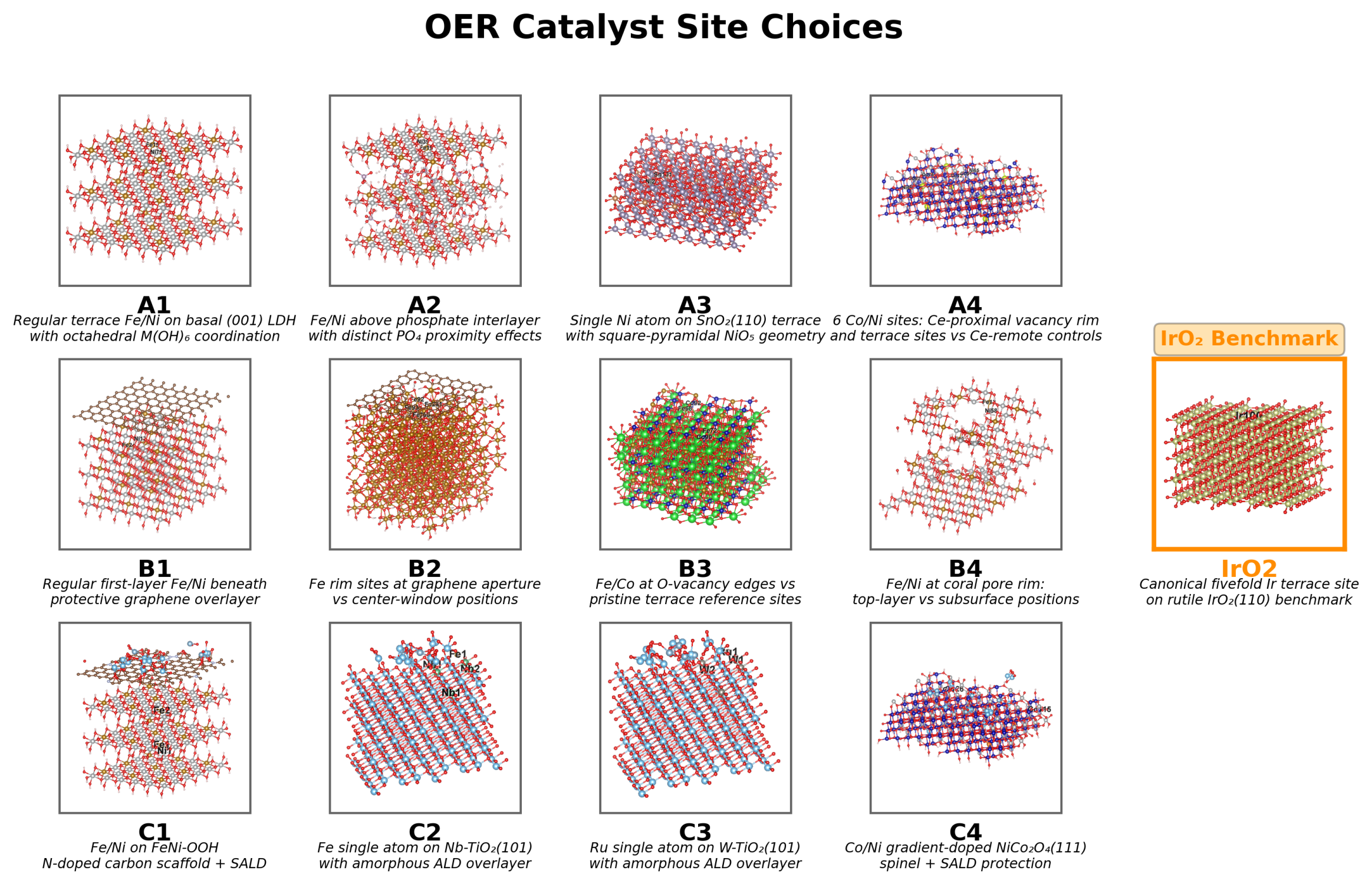}
  \caption{Matrix representation of selected metal sites for stability assessment across A-series (A1--A4) and B-series (B1--B4) catalyst candidates. Representative surface metal atoms are highlighted for dissolution and vacancy formation calculations.}  
  \label{fig:DL_4_site_choice_matrix} 
\end{figure}

\begin{figure}[htbp]
  \centering
  \includegraphics[width=1\textwidth]{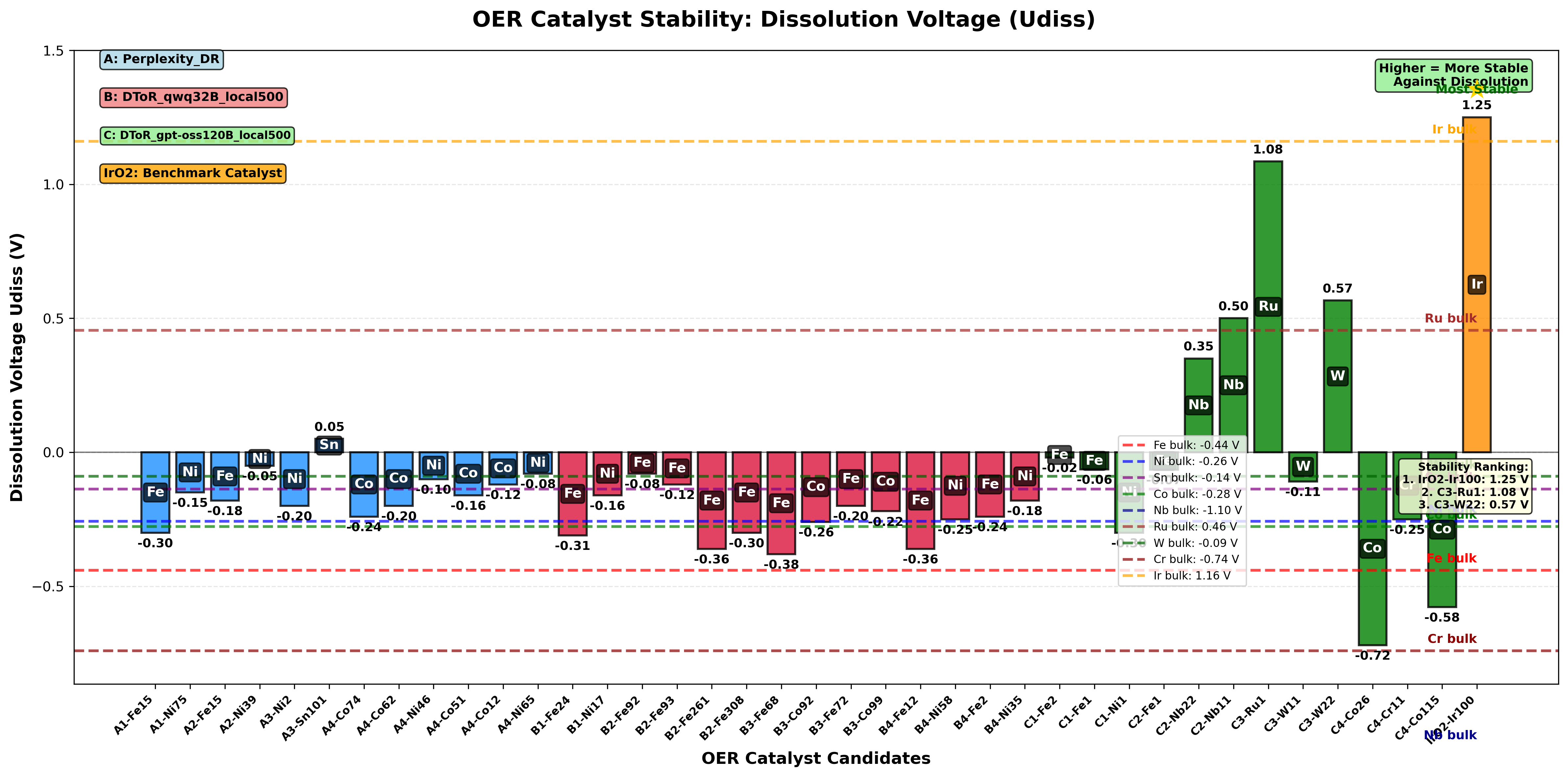}
  \caption{Dissolution potentials (U$_{\text{diss}}$) for all candidate catalysts at representative surface sites. Higher values indicate greater resistance to metal dissolution. Dashed horizontal lines denote standard bulk dissolution potentials for reference metals (Fe: $-0.44$ V, Co: $-0.277$ V, Ni: $-0.257$ V, Ir: $+1.16$ V vs. SHE).}  
  \label{fig:DL_4_histogram_Udiss} 
\end{figure}

Evaluation across representative surface sites (Figure~\ref{fig:DL_4_site_choice_matrix}) reveals significant variation in stability metrics among candidate catalysts. The dissolution potentials span from approximately $-0.4$ V to $+1.2$ V vs. SHE (Figure~\ref{fig:DL_4_histogram_Udiss}). Oxygen vacancy formation energies similarly exhibit broad distribution. In general, higher V$_{\text{O}}$ suggests
enhanced lattice stability potentially arising from stronger metal--oxygen covalency or favorable coordination environments that stabilize the oxide framework.
\textbf{Correlation Between Stability Metrics}

\begin{figure}[htbp]
  \centering
  \includegraphics[width=1\textwidth]{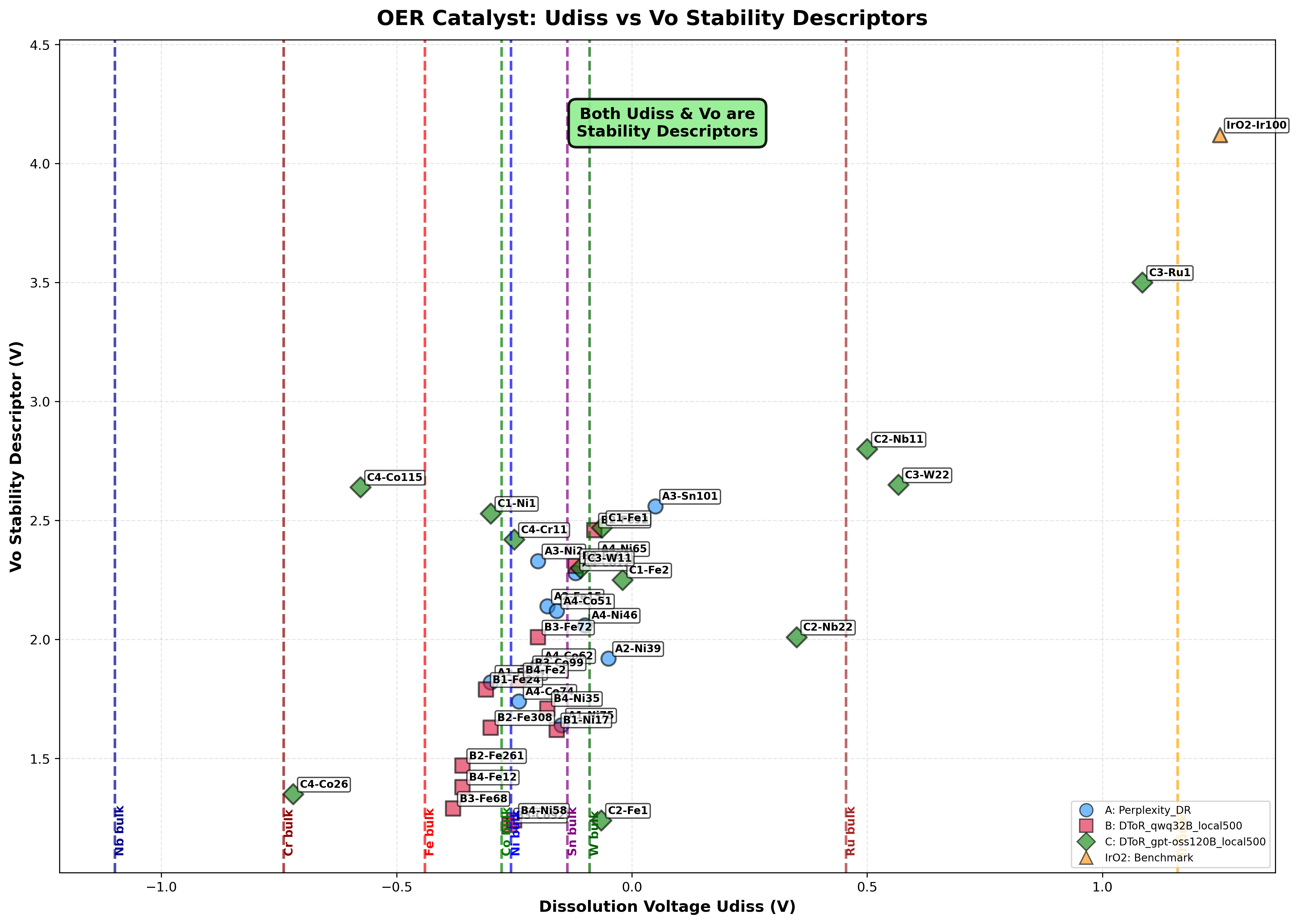}
  \caption{Correlation between dissolution potential (U$_{\text{diss}}$) and oxygen vacancy formation energy (V$_{\text{O}}$) across all 26 evaluated metal sites. The positive correlation indicates that catalysts resistant to metal dissolution also tend to resist oxygen vacancy formation.}  
  \label{fig:DL_4_scatter_udiss_Vo} 
\end{figure}

A positive correlation emerges between U$_{\text{diss}}$ and V$_{\text{O}}$ across all 39 evaluated metal sites (Figure~\ref{fig:DL_4_scatter_udiss_Vo}), indicating that resistance to metal dissolution and oxygen vacancy formation are intrinsically linked. This correlation suggests a unified electronic origin for both stability mechanisms: catalysts with stronger metal--oxygen bonds resist both
metal leaching and oxygen loss. The observed relationship validates the use of either metric as a reasonable proxy for overall stability, though evaluating both provides more comprehensive assessment.
\textbf{Normalized Stability Scoring}

\begin{figure}[htbp]
  \centering
  \includegraphics[width=1\textwidth]{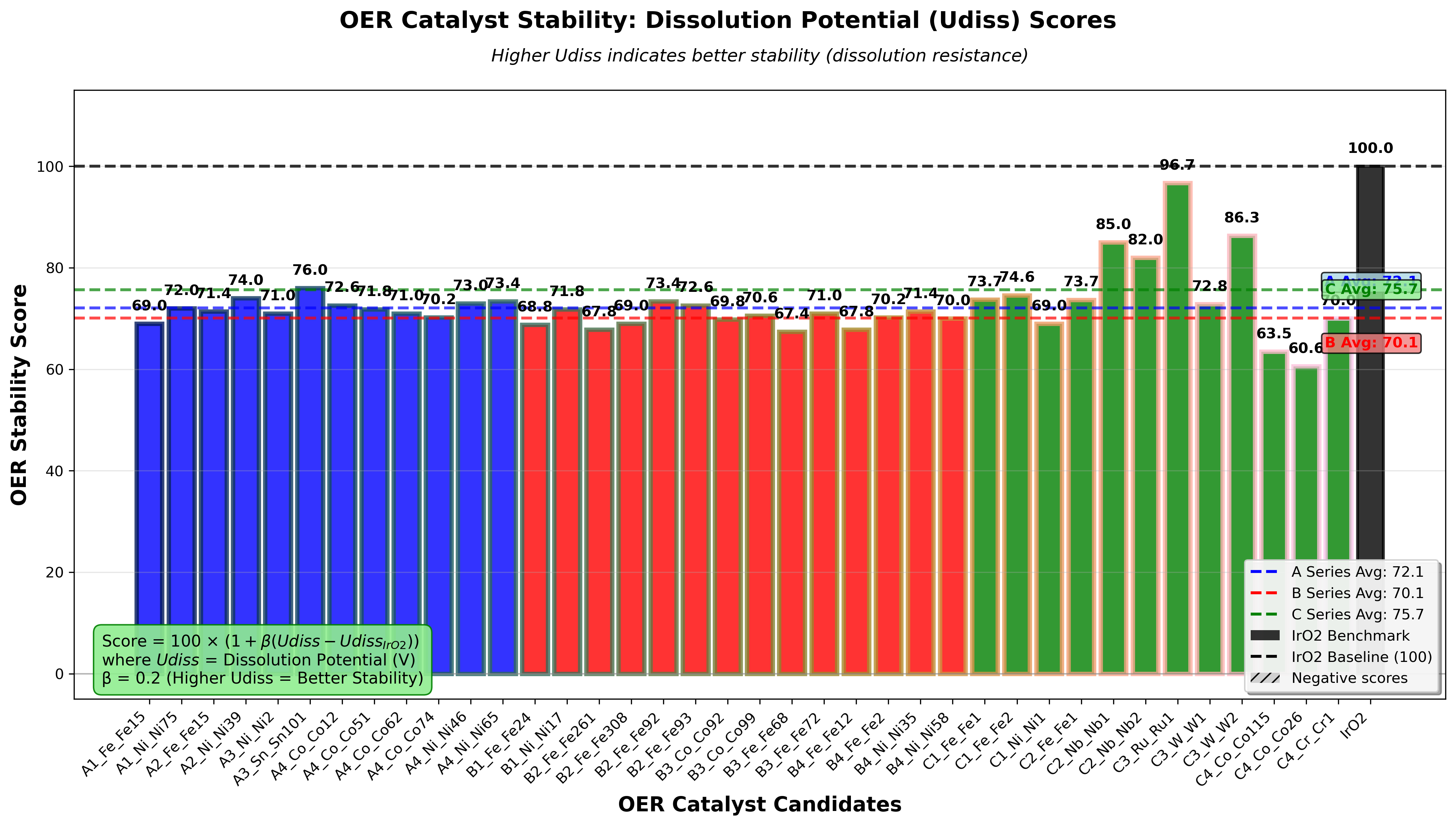}
  \caption{Stability scores based on dissolution potential (U$_{\text{diss}}$), normalized to IrO\texorpdfstring{$_2$}{2} baseline (100). Scores are calculated using the linear metric with sensitivity parameter $\beta = 0.2$, where higher U$_{\text{diss}}$ values yield higher scores.}  
  \label{fig:DL_4_scoring_Udiss} 
\end{figure}

\begin{figure}[htbp]
  \centering
  \includegraphics[width=1\textwidth]{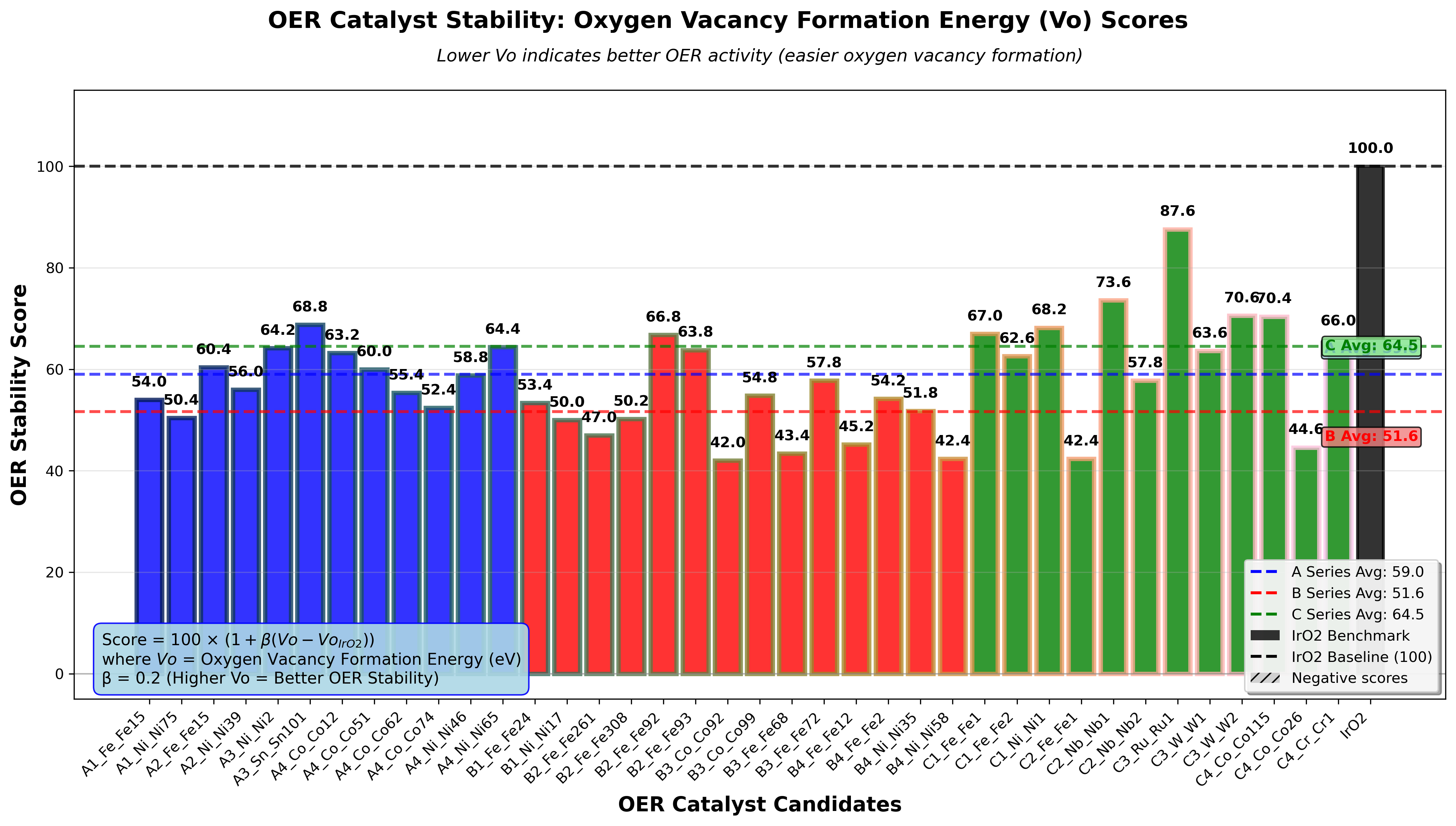}
  \caption{Stability scores based on oxygen vacancy formation energy (V$_{\text{O}}$), normalized to IrO\texorpdfstring{$_2$}{2} baseline (100). The scoring methodology employs the same linear scaling ($\beta = 0.2$), with higher V$_{\text{O}}$ values indicating superior stability.}  
  \label{fig:DL_4_scoring_Vo} 
\end{figure}

To enable quantitative comparison, we similarly applied the linear scoring framework with IrO\texorpdfstring{$_2$}{2} as the baseline reference (Score = 100), using sensitivity parameter $\beta = 0.2$ consistent with other applications. The U$_{\text{diss}}$-based scores are shown in Figure~\ref{fig:DL_4_scoring_Udiss}. The V$_{\text{O}}$-based scores (Figure~\ref{fig:DL_4_scoring_Vo}) show similar trends but not quite exactly
the same rankings, highlighting the value of assessing both metrics.
The divergence between U$_{\text{diss}}$ and V$_{\text{O}}$ rankings for specific candidates reflects different stability-limiting mechanisms: some materials may excel at preventing metal dissolution while remaining vulnerable to oxygen loss, or vice versa. This complementary information guides rational catalyst design—for instance, candidates with high U$_{\text{diss}}$ but moderate
V$_{\text{O}}$ might benefit from strategies to strengthen metal--oxygen bonds, while those with the opposite profile could require enhanced metal site anchoring.\clearpage

\begin{figure}[htbp]
  \centering
  \includegraphics[width=1\textwidth]{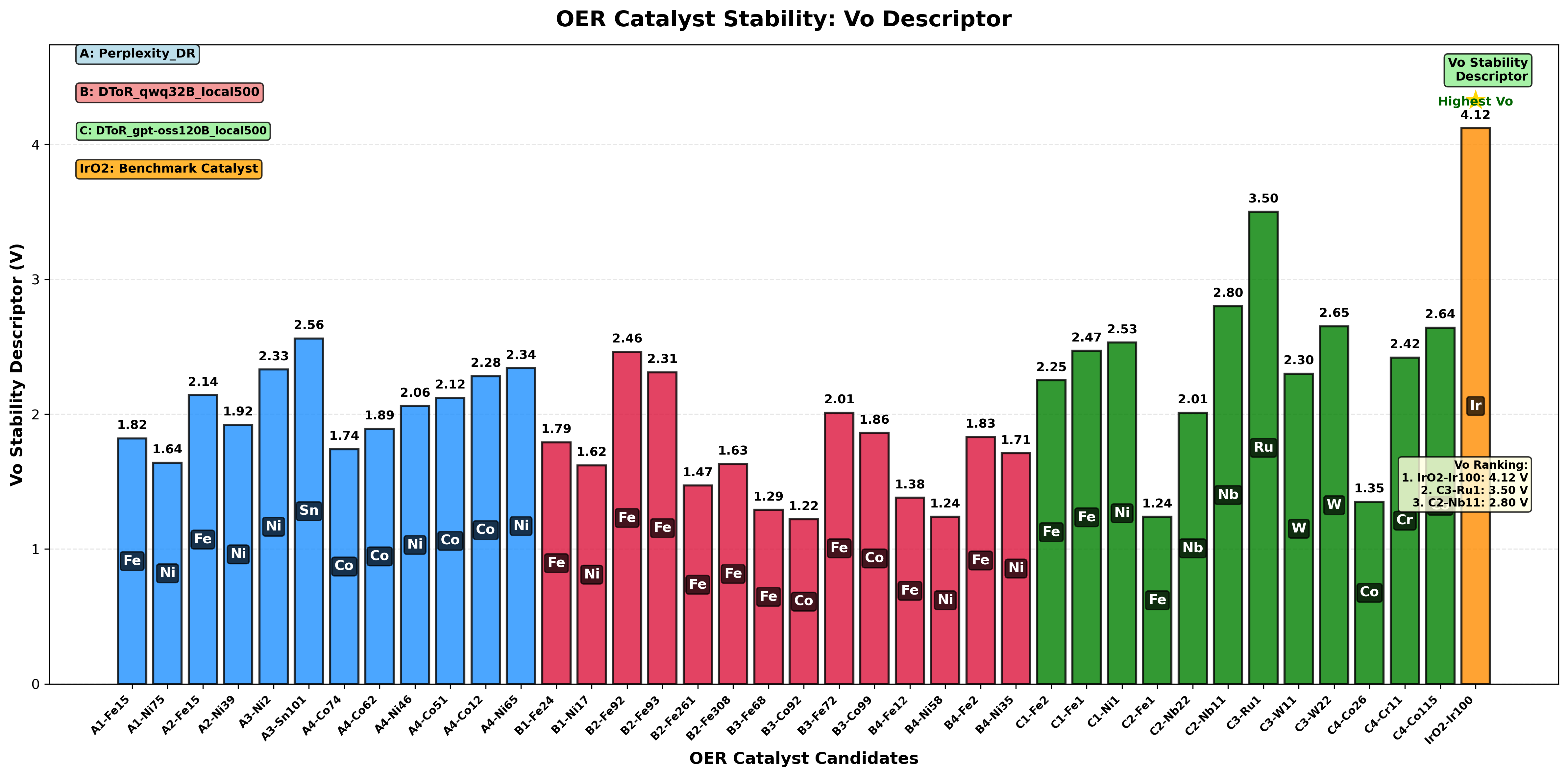}
  \caption{Oxygen vacancy formation energies (V$_{\text{O}}$) for all candidate catalysts at representative surface sites. Higher formation energies indicate greater structural stability and resistance to oxygen loss during OER operation.}  
  \label{fig:DL_4_histogram_Vo} 
\end{figure}

\clearpage

\subsection{CO\texorpdfstring{$_2$}{2} Sensor Probe}

\subsubsection{CO2\_Sensing\_2D\_Materials}

For the fifth application: CO\texorpdfstring{$_2$}{2} Sensor Probe. 
The query as shown in \ref{app:topics} is:
\begin{lstlisting}
Which two-dimensional materials--such as graphene derivatives, transition metal dichalcogenides, or MXenes--offer the highest CO2 sensing performance in complex gas or aqueous environments, and how do they compare in terms of detection limit (ppm), selectivity against common interferents (e.g., O2, H2O), response/recovery time, and long-term stability, including any functionalization or structural
modifications that enhance these metrics? Think of novel candidates.\end{lstlisting}

The best commercial solution is: \textbf{GlideLogic\_DR}.

The best local non gpt-oss solution is: \textbf{DToR\_qwq32B\_local500}.

The best local gpt-oss solution is: \textbf{Single\_gpt-oss120B\_local500}.

The expert based on the two reports, provided 4 commercial DR report derived candidates (A1-A4), and 4 local non gpt-oss DR report derived candidates (B1-B4),  and 4 local gpt-oss DR report derived candidates (C1-C4), visualized in Fig. \ref{fig:DL_5_candidates_final}:

\paragraph{\textbf{g-C\texorpdfstring{$_3$}{3}N\texorpdfstring{$_4$}{4} description}}
For CO\texorpdfstring{$_2$}{2} binding, the benchmark substrate is a monolayer graphitic carbon nitride (g-C\texorpdfstring{$_3$}{3}N\texorpdfstring{$_4$}{4}) built as a 7$\times$7 in-plane supercell of the primitive sheet (49 formula units; C$_{147}$N$_{196}$) in a hexagonal cell with a = b = 33.50~\AA, \texorpdfstring{$\gamma$}{gamma} = 120°, and c = 35.0~\AA. All atoms lie in a single atomic plane at z $\approx$ 5.0~\AA, leaving $\sim$30~\AA\ of vacuum to the next periodic
image along c, i.e., no interlayer interactions. The in-plane network is a fully conjugated sp$^2$ C--N lattice: each C is three-coordinate, N sites are predominantly two- to three-coordinate ($\langle$CN$\rangle$ $\approx$ 2.25), and nearest-neighbor C--N distances cluster at 1.37 $\pm$ 0.06~\AA. No heteroatoms, defects, terminations, or adsorbates are present (P1 by construction). This clean,
stoichiometric, and strictly planar C\texorpdfstring{$_3$}{3}N\texorpdfstring{$_4$}{4} slab (areal density $\approx$ 5.05 formula units nm$^{-2}$) serves as our reference surface for CO\texorpdfstring{$_2$}{2} binding.

\subparagraph{\textbf{A1 (rGO--PEI)}}
Hexagonal P1 slab with a = b = 36.90~\AA, \texorpdfstring{$\gamma$}{gamma} = 120.00°, c = 45.00~\AA\ (\texorpdfstring{$\alpha$}{alpha} = \texorpdfstring{$\beta$}{beta} = 90.00°). The model contains 605 atoms (C$_{461}$H$_{80}$O$_{54}$N$_{10}$). Atomic z positions span 4.63--16.63~\AA, giving a slab thickness of 12.01~\AA\ and 32.99~\AA\ of vacuum along c. The in-plane area is 11.79~nm$^2$, yielding areal number densities of C 39.09~nm$^{-2}$, H 6.78~nm$^{-2}$, O
4.58~nm$^{-2}$, and N 0.85~nm$^{-2}$. Carbon heights resolve a dominant rGO basal plane centered at z $\approx$ 7.39 $\pm$ 0.64~\AA\ (441 C) and a minor upper carbon domain at 12.09--15.35~\AA\ (20 C). All N atoms reside in the upper domain (11.23--16.53~\AA; $\langle$z$\rangle$ = 13.03~\AA), consistent with an amine-rich polyethylenimine (PEI) branch moiety tethered above the rGO sheet; O atoms
are concentrated below 10.65~\AA, consistent with residual rGO oxygen functionalities. This rGO--PEI slab, isolated by ample vacuum, provides a structurally well-defined platform for site-specific CO\texorpdfstring{$_2$}{2} binding calculations.

\subparagraph{\textbf{A2 (phosphorene \texorpdfstring{$|$}{|} Al\texorpdfstring{$_2$}{2}O\texorpdfstring{$_3$}{3} cap with drilled through-pore)}}
Orthorhombic P1 cell (a = 26.36~\AA, b = 27.26~\AA, c = 27.08~\AA; \texorpdfstring{$\alpha$}{alpha} = \texorpdfstring{$\beta$}{beta} = \texorpdfstring{$\gamma$}{gamma} = 90.00°) comprising a phosphorene sublayer (P\texorpdfstring{$_{192}$}{192}) beneath an alumina cap (Al\texorpdfstring{$_{146}$}{146}O\texorpdfstring{$_{220}$}{220}). Atomic z spans 3.95--17.44~\AA, giving a slab thickness of 13.49~\AA\ and 13.59~\AA\ of vacuum. Phosphorene occupies z = 3.95--8.12~\AA\ (thickness 4.17~\AA; top P plane at 8.12~\AA), while the cap
exhibits O/Al = 1.51 ($\approx$Al\texorpdfstring{$_2$}{2}O\texorpdfstring{$_3$}{3}) with the Al framework at z = 8.83--15.83~\AA\ (7.00~\AA) and the O network extending 8.05--17.44~\AA\ (9.39~\AA), slightly overlapping the top P plane ($\Delta$z = 0.07~\AA) to form an interfacial contact. The in-plane area is 718.57~\AA$^2$ (7.19~nm$^2$). A single through-pore pierces the cap near (f$_x$, f$_y$) = (0.51, 0.50): the top aperture has
radius 6.27~\AA\ ($\emptyset$ 12.54~\AA), narrows to a bottleneck radius 5.11~\AA\ ($\emptyset$ 10.22~\AA; section 82.02~\AA$^2$, 11.42\% of the surface), and re-expands near the interface to radius 5.22~\AA\ ($\emptyset$ 10.44~\AA). The pore rim at the bottleneck is O-rich ($\sim$7 O : 2 Al) distributed over z = 9.53--12.66~\AA, yielding an oxygen-lined channel above phosphorene that enables
CO\texorpdfstring{$_2$}{2} access both to alumina terminations and to interfacial sites.

\subparagraph{\textbf{A3 (MoS\texorpdfstring{$_2$}{2} monolayer with vacancies)}}
P1 cell with a = 27.58~\AA, b = 22.29~\AA, c = 28.13~\AA\ (\texorpdfstring{$\alpha$}{alpha} = \texorpdfstring{$\beta$}{beta} = \texorpdfstring{$\gamma$}{gamma} = 90.00°). The slab contains 200 atoms (Mo\texorpdfstring{$_{64}$}{64}S\texorpdfstring{$_{136}$}{136}; S/Mo = 2.13) and occupies z = 4.83--8.50~\AA\ (thickness 3.67~\AA) with 24.46~\AA\ of vacuum along c. The in-plane area is 6.15~nm$^2$, giving areal number densities of Mo 10.41~nm$^{-2}$ and S 22.13~nm$^{-2}$. Layering is a single S--Mo--S tri-layer:
the Mo plane is centered at z = 6.56~\AA\ (spread 0.31~\AA), bracketed by S planes at z = 5.01~\AA\ and 8.12~\AA, yielding an average S--S separation of 3.11~\AA\ and Mo--S vertical offsets of 1.54~\AA\ (bottom) and 1.57~\AA\ (top). No heteroatoms, functional groups, or adsorbates are present. Five Mo atoms are removed to create Mo vacancies. This monolayer MoS\texorpdfstring{$_2$}{2} slab provides a simple,
well-isolated reference for CO\texorpdfstring{$_2$}{2} adsorption benchmarks.

\subparagraph{\textbf{A4 (Ti\texorpdfstring{$_3$}{3}C\texorpdfstring{$_2$}{2}T$_x$, mixed terminations)}}
Hexagonal P1 slab with a = b = 30.67~\AA, \texorpdfstring{$\gamma$}{gamma} = 120.00°, c = 51.70~\AA\ (\texorpdfstring{$\alpha$}{alpha} = \texorpdfstring{$\beta$}{beta} = 90.00°); in-plane area 8.15~nm\texorpdfstring{$^2$}{2}. The model contains Ti\texorpdfstring{$_{300}$}{300}C\texorpdfstring{$_{200}$}{200}O\texorpdfstring{$_{175}$}{175}F\texorpdfstring{$_{25}$}{25}H\texorpdfstring{$_{25}$}{25}, i.e., 100 Ti\texorpdfstring{$_3$}{3}C\texorpdfstring{$_2$}{2} formula units with 200 surface sites. Terminations are asymmetric: the top face comprises 50.00\% *O, 25.00\% *OH, and 25.00\% *F (100 sites), whereas the bottom face is 100.00\%
*O (100 sites), giving an overall distribution of 75.00\% *O, 12.50\% *OH, and 12.50\% *F. Atomic heights span z = 5.01--13.29~\AA\ (thickness 8.29~\AA) with 43.41~\AA\ of vacuum; the carbide core lies at z = 7.22--9.90~\AA\ (2.68~\AA). Mean vertical offsets from the surface Ti planes are 0.96~\AA\ for top *O, 1.06~\AA\ for top *F, 2.06~\AA\ for H in *OH, and 0.89~\AA\ for bottom *O; the
corresponding site density is 12.28~nm$^{-2}$ per face. This slab provides a well-defined mixed-termination MXene surface for probing site-specific CO\texorpdfstring{$_2$}{2} adsorption.

\subparagraph{\textbf{B1 (rGO--NH\texorpdfstring{$_2$}{2}; same rGO basal slab as A1)}}
This model reuses the rGO lattice from A1 without change (a = b = 36.90~\AA, \texorpdfstring{$\gamma$}{gamma} = 120.00°, c = 45.00~\AA; in-plane area 11.79~nm\texorpdfstring{$^2$}{2}), retaining the 441-C rippled basal plane centered at z = 7.29 \texorpdfstring{$\pm$}{±} 0.67~\AA. The slab contains 585 atoms (C\texorpdfstring{$_{441}$}{441}H\texorpdfstring{$_{70}$}{70}O\texorpdfstring{$_{54}$}{54}N\texorpdfstring{$_{20}$}{20}) with heights spanning 4.36--11.16~\AA\ (thickness 6.80~\AA) and 38.20~\AA\ of vacuum. Oxygen functionalities lie over
4.36--10.64~\AA, while primary amines are grafted exclusively on the upper face (N 8.37--10.38~\AA, $\langle$z$\rangle$ = 9.40 $\pm$ 0.63~\AA; H up to 11.16~\AA), giving an --NH\texorpdfstring{$_2$}{2} areal density of 1.70~nm$^{-2}$. Areal number densities are C 37.40~nm$^{-2}$, O 4.58~nm$^{-2}$, N 1.70~nm$^{-2}$, and H 5.94~nm$^{-2}$. With lattice vectors and basal carbon topology identical to A1, this slab
isolates the effect of replacing the PEI overlayer with discrete --NH\texorpdfstring{$_2$}{2} terminations for CO\texorpdfstring{$_2$}{2} binding studies.

\subparagraph{\textbf{B2 (Ti\texorpdfstring{$_3$}{3}C\texorpdfstring{$_2$}{2}F\texorpdfstring{$_2$}{2}, fully F-terminated; same construction protocol as A4)}}
Hexagonal P1 slab with a = b = 30.67~\AA, \texorpdfstring{$\gamma$}{gamma} = 120.00°, c = 51.70~\AA\ (\texorpdfstring{$\alpha$}{alpha} = \texorpdfstring{$\beta$}{beta} = 90.00°); in-plane area 8.15~nm\texorpdfstring{$^2$}{2}. The CIF contains Ti\texorpdfstring{$_{300}$}{300}C\texorpdfstring{$_{200}$}{200}F\texorpdfstring{$_{200}$}{200} (100 Ti\texorpdfstring{$_3$}{3}C\texorpdfstring{$_2$}{2} formula units; 200 surface sites), i.e., both faces are 100.00\% *F with no *O or *OH terminations. Atomic heights span z = 4.85--12.14~\AA, giving a slab thickness of 7.29~\AA\ and 44.41~\AA\ of
vacuum. The carbide core (Ti/C) lies at z = 6.11--10.87~\AA\ (4.76~\AA); terminal F layers sit symmetrically at offsets of 1.27~\AA\ above the top Ti plane and 1.27~\AA\ below the bottom Ti plane. The termination site density is 12.28~nm$^{-2}$ per face (100 sites over 8.15~nm$^2$). This symmetric, fully fluorinated MXene provides a clean counterpoint to the mixed-termination Ti\texorpdfstring{$_3$}{3}C\texorpdfstring{$_2$}{2}T$_x$
model (A4) for isolating the effect of *F terminations on CO\texorpdfstring{$_2$}{2} adsorption.

\subparagraph{\textbf{B3 (graphene \texorpdfstring{$|$}{|} Ti\texorpdfstring{$_3$}{3}C\texorpdfstring{$_2$}{2}O\texorpdfstring{$_2$}{2} heterojunction; same MXene cell as As A4/B2)}}
Hexagonal P1 slab with a = b = 30.67 \AA{}, \texorpdfstring{$\gamma$}{gamma} = 120.00°, c = 51.70 \AA{} (in-plane area 8.15 nm\texorpdfstring{$^2$}{2}). The MXene sub-slab matches the A4/B2 Ti\texorpdfstring{$_3$}{3}C\texorpdfstring{$_2$}{2} lattice and is fully *O-terminated on both faces (Ti\texorpdfstring{$_{300}$}{300}C\texorpdfstring{$_{200}$}{200}O\texorpdfstring{$_{200}$}{200}; 100 *O per face, 12.28 sites nm\texorpdfstring{$^{-2}$}{-2}). A single-layer graphene sheet (C\texorpdfstring{$_{338}$}{338}) is appended on the bottom side, giving a total composition Ti${300}$C${538}$O${200}$.
Atomic heights span z = 7.31--19.41 \AA{} (thickness 12.10 \AA{}) with 39.60 \AA{} of vacuum along c. The graphene forms a narrow plane at z = 7.79 $\pm$ 0.31 \AA{} (7.31--8.45 \AA{}), the bottom *O plane is at z = 12.40 \AA{} and the top *O plane at z = 19.41 \AA{}; the graphene--MXene van der Waals gap is 4.61 \AA{} (center-to-plane; minimum 3.95 \AA{} to *O), and the graphene--bottom Ti plane
separation is 6.78 \AA{}. The result is a well-isolated graphene$|$Ti\texorpdfstring{$_3$}{3}C\texorpdfstring{$_2$}{2}O\texorpdfstring{$_2$}{2} heterojunction with symmetric *O terminations and a dense graphene areal carbon density of 41.49 nm$^{-2}$.

\subparagraph{\textbf{B4 (Cu-inserted MoS\texorpdfstring{$_2$}{2}; same MoS\texorpdfstring{$_2$}{2} host as A3)}}
Orthorhombic P1 cell with a = 27.58 \AA{}, b = 22.29 \AA{}, c = 28.13 \AA{} (\texorpdfstring{$\alpha$}{alpha} = \texorpdfstring{$\beta$}{beta} = \texorpdfstring{$\gamma$}{gamma} = 90.00°; in-plane area 6.15 nm\texorpdfstring{$^2$}{2}). The slab contains Mo\texorpdfstring{$_{64}$}{64}Cu\texorpdfstring{$_5$}{5}S\texorpdfstring{$_{136}$}{136} and occupies z = 4.80--8.32 \AA{} (thickness 3.51 \AA{}) with 24.61 \AA{} of vacuum. The host S--Mo--S tri-layer is preserved ($\langle$z$\rangle$: S(bottom) = 5.03 \AA{}, Mo = 6.59 \AA{}, S(top) = 8.15 \AA{}; S--S
spacing 3.12 \AA{}; Mo--S offsets 1.55/1.56 \AA{}), as in A3. At A3's Mo vacancies, five Cu centers lie within the chalcogen sandwich at z = 6.55--6.89 \AA{} ($\langle$z$\rangle$ = 6.70 \AA{}), i.e., $\sim$0.11 \AA{} above the Mo plane and biased toward the top S layer; the areal Cu density is 0.81 nm$^{-2}$. Local coordination around Cu is consistent with insertion at chalcogen-vacancy--like
sites: each Cu shows $\sim$4--5 short Cu--S contacts at 2.28--2.38 \AA{} (mean of first shell $\approx$ 2.31 \AA{}) with the next S neighbors near 2.95--3.06 \AA{}, and $\sim$2--3 Cu--Mo contacts at 2.78--3.26 \AA{}. This configuration provides a well-defined model to probe how subsurface/embedded Cu at vacancy motifs modulates CO\texorpdfstring{$_2$}{2} binding relative to the pristine A3 monolayer.

\subparagraph{\textbf{C1 (Mo\texorpdfstring{$_2$}{2}C T$_x$ MXene with hydrophobic silane + PEI)}}
Orthorhombic P1 slab with a = 34.64~\AA, b = 30.00~\AA, c = 65.36~\AA\ (\texorpdfstring{$\alpha$}{alpha} = \texorpdfstring{$\beta$}{beta} = \texorpdfstring{$\gamma$}{gamma} = 90.00\texorpdfstring{$^\circ$}{°}). The model contains 1006 atoms (Mo\texorpdfstring{$_{288}$}{288}C\texorpdfstring{$_{220}$}{220}O\texorpdfstring{$_{294}$}{294}N\texorpdfstring{$_{20}$}{20}H\texorpdfstring{$_{182}$}{182}Si\texorpdfstring{$_2$}{2}). Atomic $z$ positions span 22.50--42.86~\AA, giving a slab thickness of 20.36~\AA\ and 45.00~\AA\ of vacuum along $c$. The in-plane area is 10.39~nm$^2$, yielding areal number densities of Mo
27.72~nm$^{-2}$, C 21.17~nm$^{-2}$, O 28.30~nm$^{-2}$, N 1.92~nm$^{-2}$, Si 0.19~nm$^{-2}$, and H 17.52~nm$^{-2}$. The Mo\texorpdfstring{$_2$}{2}CO\texorpdfstring{$_2$}{2} MXene base forms a 6$\times$6 supercell (2$\times$2 expansion of the 3$\times$3 primitive cell) with Mo atoms occupying $z$ = 23.70--26.56~\AA\ ($\langle z\rangle$ = 25.13~\AA; thickness 2.86~\AA), bracketed by symmetric oxygen terminations at $z$ = 22.50--30.15~\AA,
consistent with O-terminated MXene surfaces. Above the substrate, three surface-adsorbed components lie $\sim$2.5~\AA\ above the top Mo plane: one branched polyethylenimine (PEI) cluster centered at fractional coordinates (0.20, 0.50) with N distributed across $z$ = 24.42--41.98~\AA\ ($\langle z\rangle$ = 32.42~\AA; 20 amine sites at 1.92~nm$^{-2}$), and two trimethoxycaprylylsilane molecules at
(0.95, 0.50) and (0.45, 0.50) with Si at $z$ = 29.18~\AA, forming a hydrophobic coating to resist humidity. Organic-layer C spans 25.13--40.94~\AA, with H up to 42.86~\AA, defining the top surface. Laterally, the PEI cluster occupies the left region while the silanes are separated across the middle/right, maintaining exposed MXene for electrical conduction. This Mo\texorpdfstring{$_2$}{2}C T$_x$ slab with patchy
amine functionalization and hydrophobic silane capping, isolated by ample vacuum, provides a structurally well-defined platform for CO\texorpdfstring{$_2$}{2} sensing simulations targeting sub-ppm detection with enhanced humidity resistance.

\subparagraph{\textbf{C2 (Mo\texorpdfstring{$_2$}{2}C T$_x$ MXene $|$ 2-D MOF UiO-66-NH\texorpdfstring{$_2$}{2})}}
Orthorhombic P1 slab with a = 34.64~\AA, b = 30.00~\AA, c = 66.75~\AA\ (\texorpdfstring{$\alpha$}{alpha} = \texorpdfstring{$\beta$}{beta} = \texorpdfstring{$\gamma$}{gamma} = 90.00\texorpdfstring{$^\circ$}{°}). The model contains 1002 atoms (Mo\texorpdfstring{$_{288}$}{288}C\texorpdfstring{$_{240}$}{240}O\texorpdfstring{$_{360}$}{360}N\texorpdfstring{$_{12}$}{12}H\texorpdfstring{$_{96}$}{96}Zr\texorpdfstring{$_6$}{6}). Atomic $z$ positions span 22.50--44.25~\AA, giving a slab thickness of 21.75~\AA\ and 45.00~\AA\ of vacuum. The in-plane area is 10.39~nm$^2$, yielding areal number densities of Mo
27.72~nm$^{-2}$, C 23.10~nm$^{-2}$, O 34.65~nm$^{-2}$, N 1.15~nm$^{-2}$, Zr 0.58~nm$^{-2}$, and H 9.24~nm$^{-2}$. The Mo\texorpdfstring{$_2$}{2}CO\texorpdfstring{$_2$}{2} MXene base forms a 6$\times$6 supercell with Mo at $z$ = 23.70--26.56~\AA\ ($\langle z\rangle$ = 25.13~\AA; thickness 2.86~\AA) bracketed by symmetric O terminations. Above the substrate, a single UiO-66-NH\texorpdfstring{$_2$}{2} MOF cage is positioned $\sim$10.0~\AA\ above the top Mo
plane at (0.20, 0.50), offset to leave exposed MXene for transduction. The cage comprises a Zr$_6$ node coordinated by aminoterephthalate linkers (C$_{96}$H$_{96}$N$_{12}$O$_{72}$), providing 12 --NH\texorpdfstring{$_2$}{2} sites (1.15~nm$^{-2}$) for CO\texorpdfstring{$_2$}{2} capture; the Zr$_6$ node imparts stability and Lewis acidity, while --NH\texorpdfstring{$_2$}{2} groups offer basic binding sites. The topmost atoms reach $z$ = 44.25~\AA, ensuring
separation from the MXene base. This Mo\texorpdfstring{$_2$}{2}C T$_x$ $|$ UiO-66-NH\texorpdfstring{$_2$}{2} hybrid, isolated by ample vacuum, provides a well-defined platform for CO\texorpdfstring{$_2$}{2} sensing simulations targeting sub-0.3~ppm detection with fast response and minimal drift via synergistic MXene conductivity and MOF selectivity.

\subparagraph{\textbf{C3 (Mo\texorpdfstring{$_2$}{2}C T$_x$ MXene with ZnO nano-grain cluster)}}
Orthorhombic P1 cell with a = 34.64~\AA, b = 30.00~\AA, c = 61.06~\AA\ (\texorpdfstring{$\alpha$}{alpha} = \texorpdfstring{$\beta$}{beta} = \texorpdfstring{$\gamma$}{gamma} = 90.00\texorpdfstring{$^\circ$}{°}). The slab contains 829 atoms (Mo\texorpdfstring{$_{288}$}{288}C\texorpdfstring{$_{144}$}{144}O\texorpdfstring{$_{342}$}{342}Zn\texorpdfstring{$_{55}$}{55}). Atomic $z$ spans 22.50--38.56~\AA\ (thickness 16.06~\AA) with 45.00~\AA\ of vacuum. The in-plane area is 10.39~nm$^2$, giving areal densities of Mo 27.72~nm$^{-2}$, C 13.86~nm$^{-2}$, O 32.92~nm$^{-2}$ (MXene +
ZnO), and Zn 5.29~nm$^{-2}$. The Mo\texorpdfstring{$_2$}{2}CO\texorpdfstring{$_2$}{2} MXene base forms a 6$\times$6 supercell with Mo at $z$ = 23.70--26.56~\AA\ ($\langle z\rangle$ = 25.13~\AA; 2.86~\AA\ thick) and symmetric O terminations. Above the substrate, an octahedral ZnO nano-grain (Zn\texorpdfstring{$_{55}$}{55}O$_{54}$; 109 atoms) is placed $\sim$7.0~\AA\ above the top Mo plane at (0.13, 0.50), spanning $z$ = 28.56--38.56~\AA\ ($\sim$10.0~\AA\
vertical extent). The cluster, constructed with ASE, exhibits uniform (111)-type facets; surface Zn provide coordinatively unsaturated sites for CO\texorpdfstring{$_2$}{2} interaction, while surface O offer basic sites for hydrogen bonding/dipole interactions. The moderate 7.0~\AA\ clearance balances electronic coupling with structural independence, enabling efficient transduction via the conductive MXene. This
Mo\texorpdfstring{$_2$}{2}C T$_x$ $|$ ZnO hybrid slab, isolated by ample vacuum, offers a well-defined platform for CO\texorpdfstring{$_2$}{2} sensing simulations targeting sub-0.25~ppm detection and $>$15:1 selectivity via synergistic conductivity and uniformly distributed ZnO capture sites.

\subparagraph{\textbf{C4 (Janus Ti\texorpdfstring{$_2$}{2}C T$_x$ MXene + PEI interlayer)}}
Hexagonal P1 slab with a = 18.18~\AA, b = 15.74~\AA, c = 64.57~\AA\ (\texorpdfstring{$\alpha$}{alpha} = \texorpdfstring{$\beta$}{beta} = 90.00\texorpdfstring{$^\circ$}{°}, \texorpdfstring{$\gamma$}{gamma} = 120.00\texorpdfstring{$^\circ$}{°}). The model contains 384 atoms (Ti\texorpdfstring{$_{72}$}{72}C\texorpdfstring{$_{36}$}{36}O\texorpdfstring{$_{72}$}{72} + PEI C\texorpdfstring{$_{54}$}{54}H\texorpdfstring{$_{130}$}{130}N\texorpdfstring{$_{20}$}{20}; total Ti\texorpdfstring{$_{72}$}{72}C\texorpdfstring{$_{90}$}{90}O\texorpdfstring{$_{72}$}{72}H\texorpdfstring{$_{130}$}{130}N\texorpdfstring{$_{20}$}{20}). Atomic $z$ positions span 22.50--42.07~\AA, giving a slab thickness of 19.57~\AA\ and 45.00~\AA\ of vacuum. The in-plane area
is 2.86~nm$^2$, yielding areal densities of Ti 25.17~nm$^{-2}$, C 31.47~nm$^{-2}$ (12.59~nm$^{-2}$ from MXene; 18.88~nm$^{-2}$ from PEI), O 25.17~nm$^{-2}$, N 6.99~nm$^{-2}$, and H 45.45~nm$^{-2}$. The Ti\texorpdfstring{$_2$}{2}CO\texorpdfstring{$_2$}{2} MXene base (validated 3$\times$3 primitive reused and expanded to 6$\times$6) has Ti at $z$ = 23.74--26.26~\AA\ ($\langle z\rangle$ = 25.00~\AA; 2.52~\AA\ thick) with symmetric O
terminations at $\sim$22.50~\AA\ (bottom) and 27.74~\AA\ (top), Ti--O $\approx$ 1.95~\AA. Above the substrate, a branched PEI interlayer cluster (C$_{54}$H$_{130}$N$_{20}$; 9 repeat units, 20 amine sites) sits 5.0~\AA\ above the top Ti plane at (0.50, 0.50), spanning $z$ = 29.39--42.07~\AA\ ($\sim$12.7~\AA\ vertical extent). The amine density is 6.99~N~nm$^{-2}$ ($\ge$3~nm$^{-2}$ target),
supporting reversible amine--CO\texorpdfstring{$_2$}{2} chemistry (R--NH\texorpdfstring{$_2$}{2} + CO\texorpdfstring{$_2$}{2} \texorpdfstring{$\rightarrow$}{->} R--NH$_2^+$--COO$^-$) that modulates MXene conductivity; central placement ensures symmetric coverage while preserving peripheral MXene for contacts. This Ti\texorpdfstring{$_2$}{2}C T$_x$ $|$ PEI hybrid, isolated by ample vacuum, provides a well-defined platform for CO\texorpdfstring{$_2$}{2} sensing simulations targeting sub-0.25~ppm detection and $>$15:1
selectivity via metallic conductivity, O-termination stability, and high-density amine capture sites.

\subsubsection{CO\texorpdfstring{$_2$}{2} Sensor Performance Evaluation}

\begin{figure}[htbp]
  \centering
  \includegraphics[width=1\textwidth]{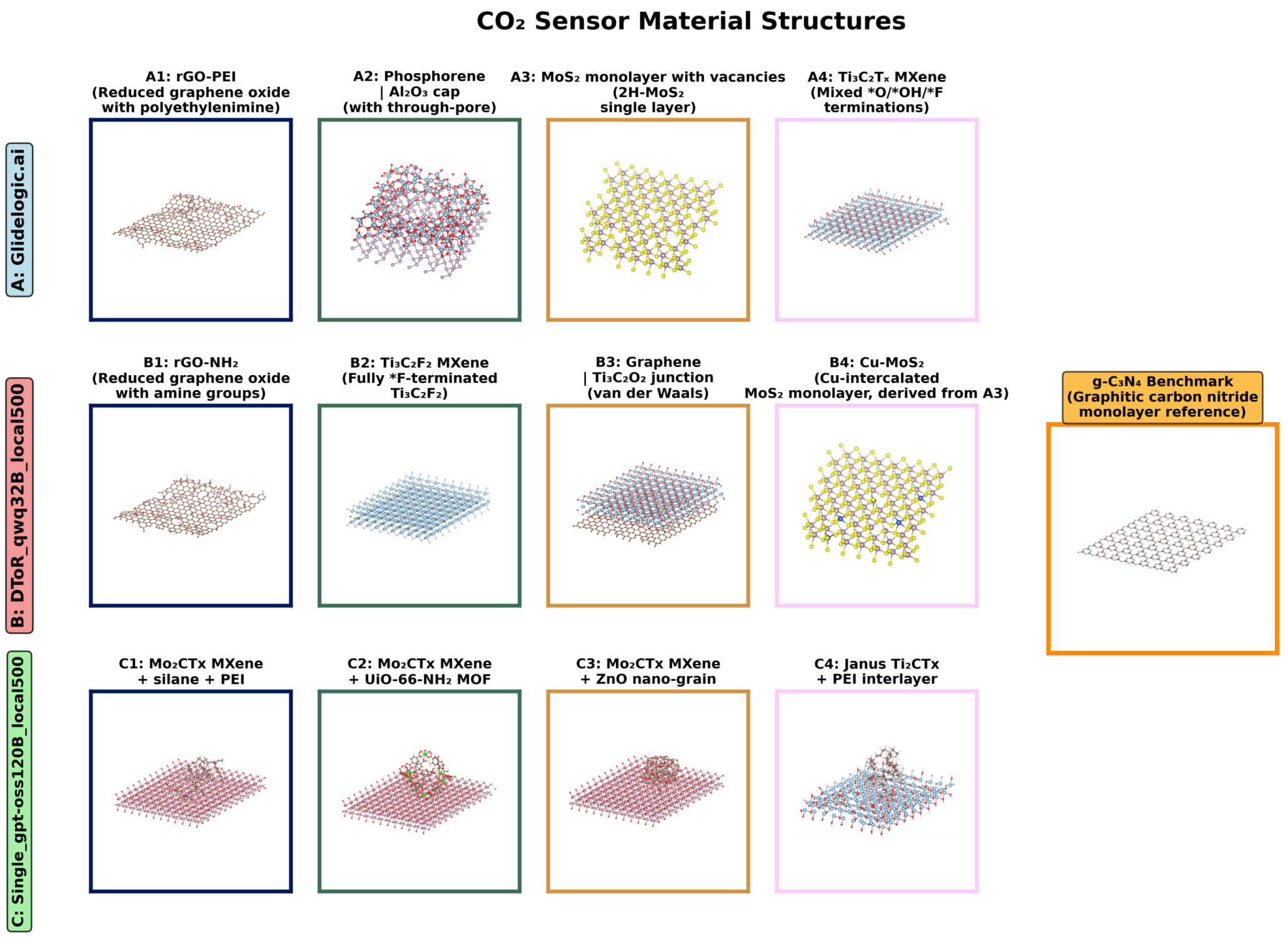}
  \caption{Structural configurations of CO\texorpdfstring{$_2$}{2} sensor candidates proposed by DR agents, with g-C\texorpdfstring{$_3$}{3}N\texorpdfstring{$_4$}{4} \protect\cite{karthik2023gc3n4co2} included as the domain benchmark reference.}
  \label{fig:DL_5_candidates_final}  
\end{figure}

\begin{figure}[htbp]
  \centering
  \includegraphics[width=1\textwidth]{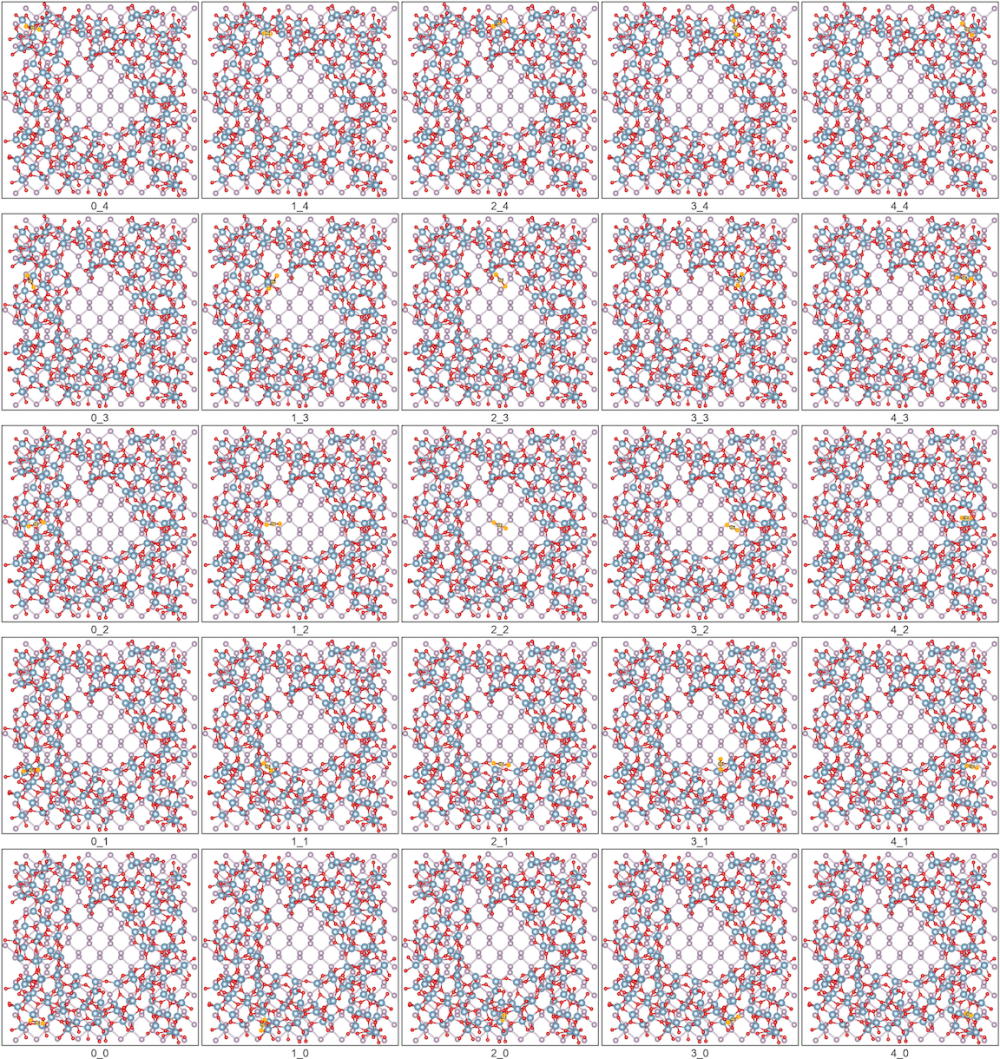}
  \caption{Top-view representation of the 5$\times$5 grid sampling approach for CO\texorpdfstring{$_2$}{2} adsorption sites. Relaxed configurations show CO\texorpdfstring{$_2$}{2} molecules at 25 distinct initial positions on candidate A2, demonstrating comprehensive exploration of the binding landscape.}  
  \label{fig:DL_5_binding_showcase} 
\end{figure}

\begin{figure}[htbp]
  \centering
  \includegraphics[width=1\textwidth]{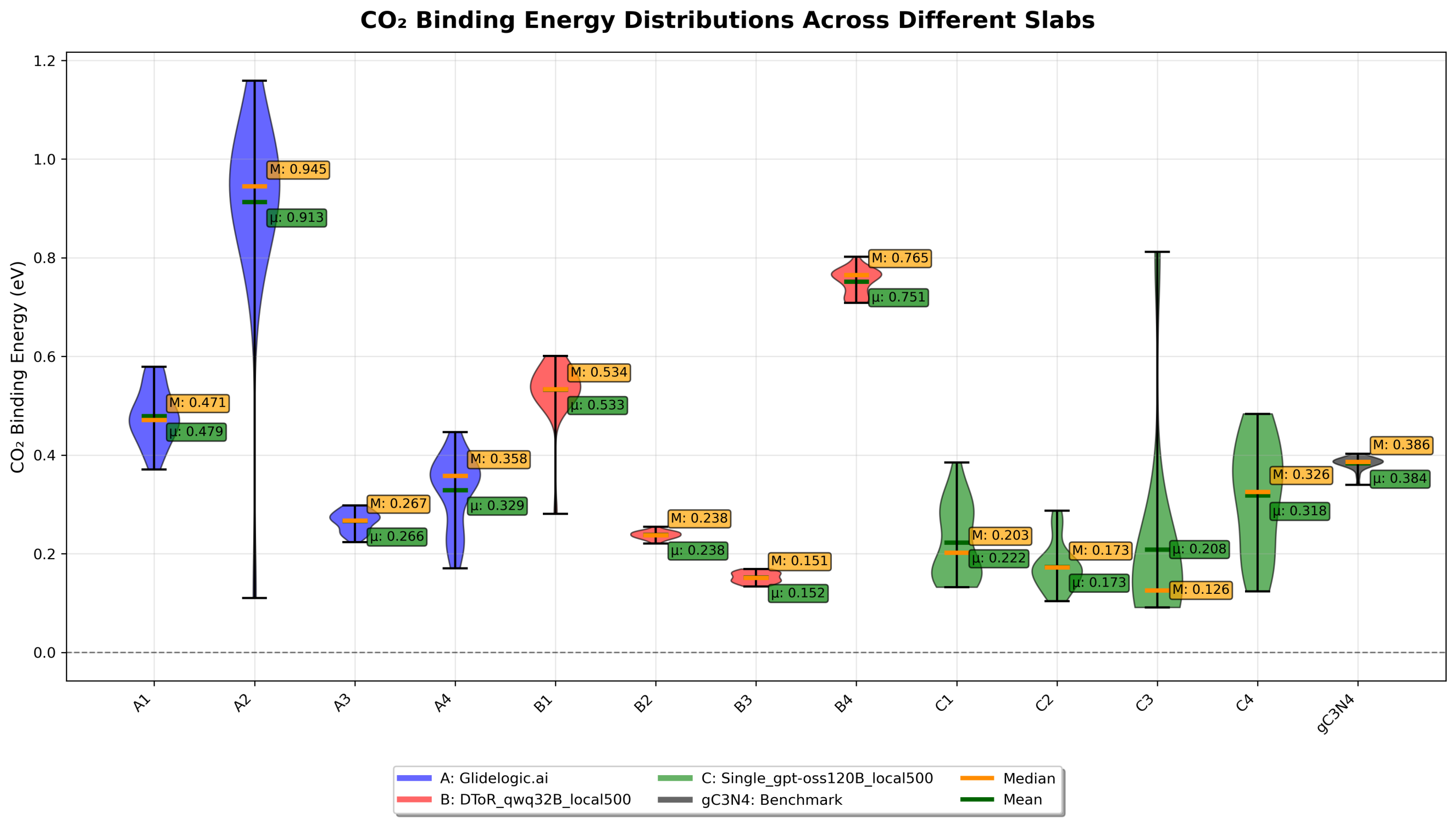}
  \caption{Distribution of CO\texorpdfstring{$_2$}{2} binding energies across all candidate systems visualized as violin plots. Mean (solid line) and median (dashed line) values are shown for each candidate, derived from 25 configurations per surface using the 5$\times$5 grid sampling methodology.}  
  \label{fig:DL_5_violin_binding_energy} 
\end{figure}

\begin{figure}[htbp]
  \centering
  \includegraphics[width=1\textwidth]{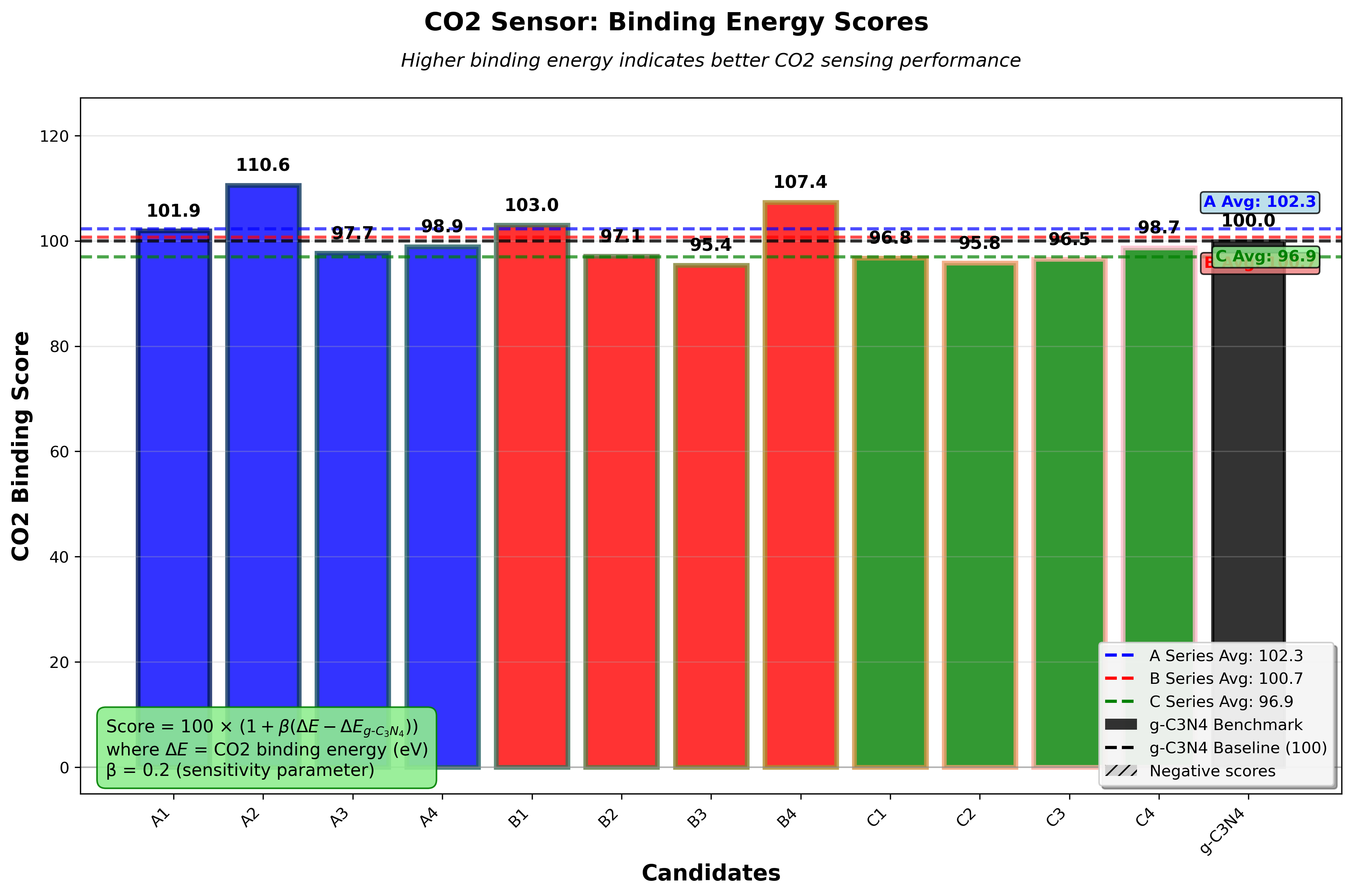}
  \caption{CO\texorpdfstring{$_2$}{2} sensing performance scores for all candidate systems from static DFT calculations, normalized to g-C\texorpdfstring{$_3$}{3}N\texorpdfstring{$_4$}{4} baseline (100). Scores are calculated using the linear transformation with sensitivity parameter $\beta = 0.2$, where higher binding energies yield higher scores.}  
  \label{fig:DL_5_scoring_static_DFT} 
\end{figure}

The development of sensitive and selective CO\texorpdfstring{$_2$}{2} sensors addresses critical needs in environmental monitoring, industrial process control, and indoor air quality management. To evaluate the CO\texorpdfstring{$_2$}{2} sensing capabilities of candidates proposed by the DR agents (Figure~\ref{fig:DL_5_candidates_final}), we quantified binding energies using a systematic grid-based sampling approach that captures the
heterogeneity of surface adsorption sites, with g-C\texorpdfstring{$_3$}{3}N\texorpdfstring{$_4$}{4} serving as the established benchmark reference \cite{karthik2023gc3n4co2}.
The comprehensive sampling methodology employed a 5$\times$5 grid across each surface unit cell, positioning CO\texorpdfstring{$_2$}{2} molecules at 25 distinct initial configurations to explore the complete adsorption landscape (Figure~\ref{fig:DL_5_binding_showcase}). At each grid point, CO\texorpdfstring{$_2$}{2} molecules were placed 1.8 Å above the highest surface atom within a 3.0 Å search radius, with initial linear
configurations parallel to the surface. This systematic approach ensures unbiased sampling of potential binding sites, including atop atoms, bridge sites, hollow sites, and defect locations, providing statistically robust binding energy distributions rather than single-point estimates.
The resulting binding energy distributions reveal significant variability both within and between candidate systems (Figure~\ref{fig:DL_5_violin_binding_energy}). The violin plots demonstrate that surface heterogeneity plays a crucial role in CO\texorpdfstring{$_2$}{2} adsorption, with some candidates exhibiting narrow distributions indicating uniform binding sites, while others show broad distributions suggesting
diverse adsorption environments. Mean binding energies range from approximately 0.15 to 0.90 eV across the evaluated systems, with A-series candidates generally demonstrating stronger CO\texorpdfstring{$_2$}{2} interactions compared to B-series materials.
Performance scoring using the linear transformation framework with sensitivity parameter $\beta = 0.2$ (Figure~\ref{fig:DL_5_scoring_static_DFT}) quantifies the relative sensing capabilities normalized to the g-C\texorpdfstring{$_3$}{3}N\texorpdfstring{$_4$}{4} baseline. The distribution of scores reveals that commercial DR agent successfully identified materials spanning a range of binding strengths, from weak physisorption suitable
for rapid-response sensors to stronger chemisorption appropriate for CO\texorpdfstring{$_2$}{2} capture applications.
The systematic evaluation highlights the importance of comprehensive surface sampling in sensor material assessment. Candidates with both high mean binding energies and narrow distributions, such as B4, represent ideal sensor materials offering consistent response across the entire surface. Conversely, materials with broad binding energy distributions may exhibit non-uniform sensor responses,
potentially complicating calibration and reducing reliability. These findings underscore that effective CO\texorpdfstring{$_2$}{2} sensor design requires not only optimizing average binding strength but also minimizing surface heterogeneity to ensure reproducible sensing performance.
\clearpage

\clearpage

\subsection{Candidate Feasibility Discussions}\label{candidate_feasibility_discussions}
In PFAS sensing, our "good" versus "bad" judgments were anchored in well-established supramolecular and materials design principles rather than subjective preferences. A1 (rGO + $\beta$-CD) and B1 (MXene + methacrylic-acid MIP cavity) closely mirror experimentally validated motifs. $\beta$-CD polymers and $\beta$-CD-based molecularly imprinted layers are now recognized PFAS sorbents and sensing elements: their $\sim$0.7--0.8 nm hydrophobic cavity matches the cross-section of perfluoroalkyl tails and can concentrate PFOA/PFOS while excluding bulkier natural organic matter\cite{lacalamita2025cyclodextrins, zhu2025betacdmip, wang2022tunable, pitruzzella2024mippof, liu2025pfoa, wang2024reversible}. Besides, methacrylic-acid–based MIP have also been experimentally demonstrated for highly selective electrochemical detection of PFOA and related PFAS in complex matrices, further supporting the plausibility of  methacrylic-acid-based MIP cavity\cite{liu2025pfoa}. Anchoring such cavities on conductive 2D supports (rGO or Ti$_3$C$_2$T$_x$ MXene) is also experimentally routine; MXene- and graphene-based electrochemical/FET sensors already exploit their high carrier density, hydrophilic terminations and solution processability for chemical detection. By contrast, C3 deliberately stacks a hydrophobic fluorosilane brush (C12-PFOTS), hydrophilic and chemically fragile MXene, metallic Ag nanoparticles, and acidic PEDOT:PSS. Fluorosilanes such as PFOTS/POTS are designed to form superhydrophobic, electrolyte-repelling coatings on oxide or silica surfaces, with water contact angles above 150$^\circ$ and strong exclusion of aqueous electrolytes\cite{chemicalbook2025pots, liu2016superhydrophobic}. PEDOT:PSS, in turn, is well-known to be strongly acidic; its PSS component can corrode underlying oxides/metallic layers and compromise stability unless neutralized or buffered\cite{cameron2020pedotpss}. MXenes such as Ti$_3$C$_2$T$_x$ are chemically sensitive to oxidative and harsh aqueous environments, with reviews emphasizing pH and oxidant control to mitigate degradation\cite{kumar2023mxenestability}. Putting all of these in a single wet-processed stack raises obvious risks of phase separation (fluorosilane domains de-wetting hydrophilic PEDOT:PSS/MXene), acid-induced MXene/Ag degradation, and loss of stable ionic conduction paths---failure modes that are invisible to purely structural dry lab scoring but very familiar to experimental device builders.

An analogous pattern appears in PFAS degradation. Our dry-lab "winners" cluster around Ti$_4$O$_7$-based electrodes (A3) and defect-engineered spinel oxides such as ZnFe$_2$O$_4$ (C4), both of which have strong literature support as cost-effective, robust platforms for oxidative degradation. Magnéli-phase Ti$_4$O$_7$ anodes are now among the most widely studied PFAS electro-oxidation catalysts, combining high conductivity, electrochemical stability, and near-complete defluorination/mineralization under realistic cell conditions \cite{luo2023ultrasound, gomri2025eaop_efenton_pfas}. ZnFe$_2$O$_4$ composites, especially those with oxygen-vacancy-rich or doped structures, have been repeatedly shown to provide strong adsorption and efficient ROS generation for degradation of dyes and pharmaceuticals while remaining structurally robust and inexpensive\cite{ullah2023znfe2o4_update}. In contrast, C5 proposes a triple heterostructure (MXene/g-C$_3$N$_4$/atomic layer deposition:ALD-grown Al$_2$O$_3$) that is difficult to justify from either a synthesis or a charge-transport perspective. State-of-the-art MXene/g-C$_3$N$_4$ catalysts emphasize intimate 2D/2D contact and carefully tuned band alignment to promote interfacial charge separation; even in those carefully engineered systems, charge-transfer efficiency is highly sensitive to interface quality and layer thickness\cite{mandari2025gcn_mxene_review}. Adding a conformal ALD Al$_2$O$_3$ overlayer on top further increases interfacial complexity and typically introduces an additional insulating barrier that suppresses electron transfer to solution unless its thickness and coverage are exquisitely controlled\cite{bagnall2025al2o3_stability, goul2019ald_al2o3_tunnel}. Practically, ALD processing on a solution-processed MXene/g-C$_3$N$_4$ stack also imposes thermal and chemical constraints that risk degrading MXene or passivating catalytically active sites without clear performance gains. When combined with emerging evidence that current LLMs struggle with complex, interconnected materials constraints and can generate plausible but impractical designs without explicit domain priors or tool-based validation, these examples illustrate a characteristic "inverse-design hallucination"\cite{miret2025llms_materials_discovery, kalai2025hallucinate}: agents tend to assume that stacking more functional components is always beneficial, even when well-established electrochemical and colloidal considerations make such architectures unlikely to be buildable or stable in the wet lab.

\clearpage
\section{Targeted Component Ablations}\label{app:additional_ablation}

\subsection{Experimental setup}
To address reviewer concerns without changing our code or indices, we performed a minimal-cost sensitivity study on three representative topics---\textit{LIB\_Fluoroether\_Anion\_Receptor}, \textit{Microplastics\_Sensing\_2D}, and \textit{PFAS\_2D\_FET\_Probes}---under the same LLM-as-jury evaluation protocol and budget as the main paper. We compare against our local baseline
\texttt{DToR\_gpt-oss120B\_local500} (local-first, gap-triggered web complement, default Tree-of-Thought orchestration and per-node reflection). We introduce three families of deliberately ``degraded'' variants:(1) \textbf{Limited-Scale} variants that constrain orchestration and local retrieval while keeping the architecture intact: \texttt{weak\_ToT} reduces the number of initial perspectives and the maximum tree depth from 3\texorpdfstring{$\rightarrow$}{->}2; \texttt{low\_source\_counts} limits the local RAG fetch to paper\_results\_count = 1 instead of the default 5 (approaching the \textit{local0} regime in the main
text); \texttt{low\_iterate\_counts} disables all reflection loops per DR node (default = 3).(2) \textbf{No-Web-Search} variants that remove complementary web evidence: \texttt{local\_only} restricts retrieval to the 500-volume local corpus; \texttt{LLM\_only} disables both local and web retrieval, i.e., generation from the LLM alone.
(3) \textbf{ChatGPT-API-driven} variants that keep the exact same framework but swap the local backbone for OpenAI gpt-5-mini or gpt-5-nano (\textit{gpt\_5\_mini\_driven}, \textit{gpt\_5\_nano\_driven}). All variants were evaluated with the same five-dimension rubric used in the paper and via anonymous A/B duels against \texttt{DToR\_gpt-oss120B\_local500}; we additionally tracked the token
footprint of API runs ($\approx$ 38M tokens, $\approx$ \$15 per our OpenAI Console), which is materially higher than the cost of our local deployment. Finally, all ablation runs held the evaluation protocol and default budgets constant (depth/branches/top-k/iterations and the same five judges), varying only the indicated knobs so differences can be attributed to the ablated factors.

\subsection{Results and interpretation}
Across all three topics and both evaluation modes, every variant underperforms the local baseline on the five-dimension rubric (\figref{fig:supp_11_add_ablation_5dim}). The degradation follows a consistent gradient: No-Web-Search < Limited-Scale, with \texttt{LLM\_only} the strongest negative control (unsurprisingly, report-level usefulness collapses without any retrieval). Within Limited-Scale,
\texttt{low\_iterate\_counts} harms quality more than \texttt{low\_source\_counts}, indicating that \textbf{per-node reflection is more consequential than retrieval minutiae} at fixed budgets. \texttt{local\_only} trails the full system, confirming that gap-triggered web complement adds measurable value beyond local evidence. In A/B duels, \texttt{DToR\_gpt-oss120B\_local500} achieves
$\sim$60--80\% average win-rates versus each residual variant (\figref{fig:supp_12_add_ablation_duel}); among the ablated models, the gpt-5-mini/nano-driven versions are the least degraded (they narrow the duel margins) but at the cost of significantly higher token spend and operational expense. Importantly, when we drive the framework with small API models (mini/nano) across the same three topics,
the relative ordering of variants persists, demonstrating that the DToR orchestration policy is robust to model scale and to the particular RAG instantiation. Inter LLM-judger and experimental trial consistency analysis similar to previous ones in \figref{fig:supp_13_add_ablation_consistency} also support the reliability of our conclusions. These outcomes reinforce our positioning: we do not claim
``better retrieval accuracy'' in the IR sense; rather, \textbf{under a fixed resource budget, the local-first gap-triggered fusion with DToR branching makes reports more usable} (coverage $\times$ depth $\times$ coherence), while preserving on-prem deployment, data compliance, and markedly lower cost/energy. Ablation study presented in \figref{fig:main_rubric_results}b reveal that reducing local
RAG corpus size from 500 to 100 volumes only marginally impacts scores (drop <0.2/10) in comparison, suggesting DToR's efficiency regardless of local RAG on corpus. Also, notably, even the gpt-oss20B model under DToR outperforms all commercial baselines as shown in \figref{fig:main_rubric_results}a, confirming that orchestration—not just model scale—drives performance. Combined with our dry-lab
simulations/computations, which turn textual suggestions into actionable candidates and reproducible scripts, this is the kind of system-level utility the materials/chemistry community actually values.

\clearpage
\section{Topic Analysis}\label{app:topic_analysis}

\textbf{Computation methodology:} We evaluate each research topic along two complementary axes using the full matrix of results from 41 agents across all trials and critic LLMs. Topic difficulty is defined as the mean performance across all agents for that topic; lower means indicate harder topics. To compute it, we aggregate all available scores per topic (over trials and critics), average per agent, then average across agents to obtain a single topic mean; topics are then ordered from easiest to hardest by this mean for display. Topic discrimination power is defined as the standard deviation ($\sigma$) of the per-agent mean scores within a topic; higher $\sigma$ means the topic separates strong and weak agents more clearly. Discrimination is intentionally analyzed independently of difficulty because both easy and hard topics can show either high or low separation. For visualization, we present a Topic $\times$ Agent heatmap (rows = topics ordered by difficulty, columns = agents ordered by overall performance) and a Topic Difficulty Ranking bar chart that shows mean $\pm$ SEM per topic and overlays dashed vertical lines at the overall means of six reference agents to contextualize the global difficulty scale.

\begin{figure}[htbp]
  \centering
  \includegraphics[width=\textwidth]{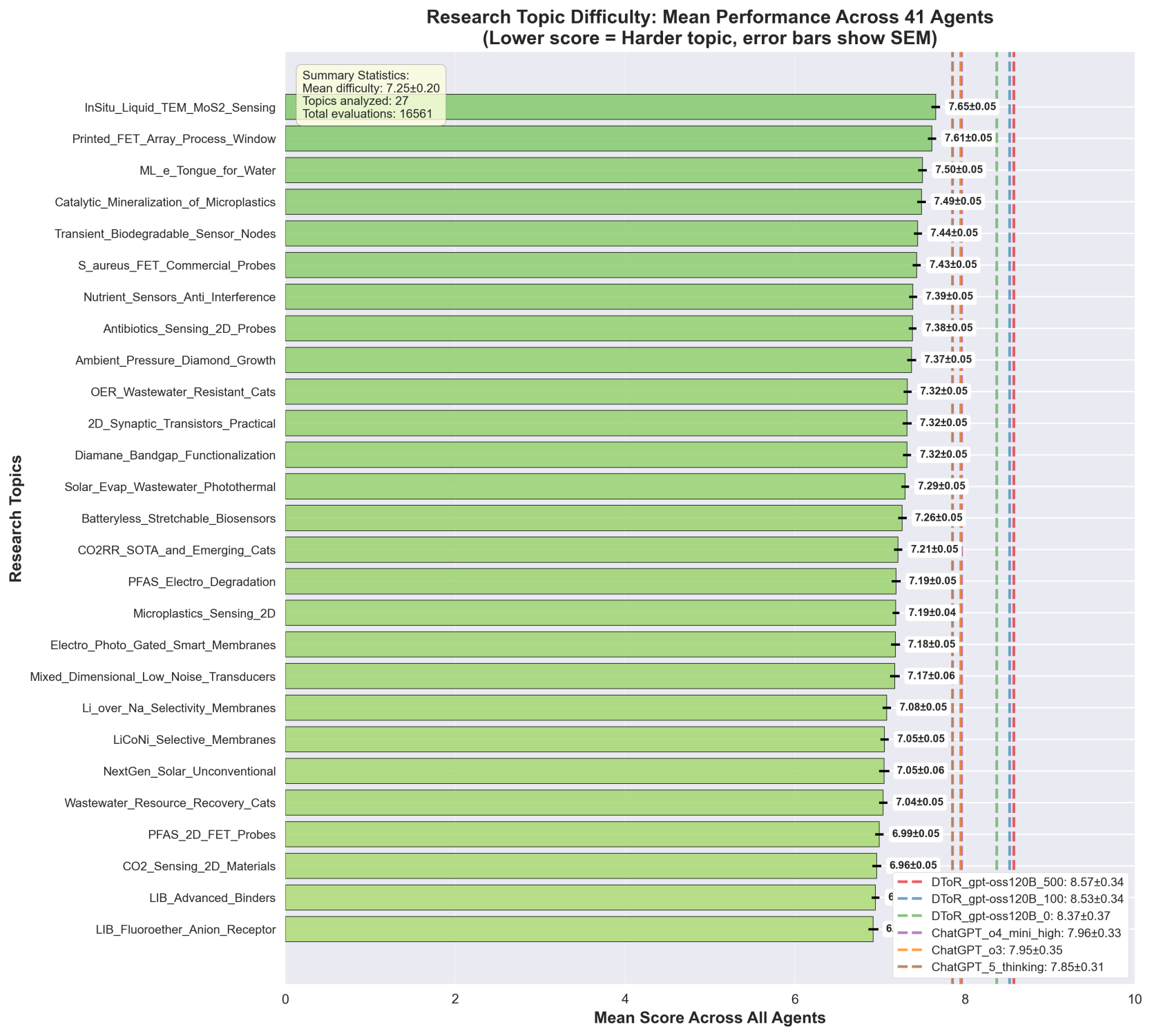}
  \caption{Bar chart summarizing mean performance ($\pm$SEM) across all agents for each topic. Bars are sorted from hardest (left, lower mean) to easiest (right, higher mean), with dashed vertical lines marking the six reference agents' overall means. This view highlights where topics sit on the global difficulty spectrum.}  
  \label{fig:topic_ranking_bar} 
\end{figure}

\begin{figure}[htbp]
  \centering
  \includegraphics[width=\textwidth]{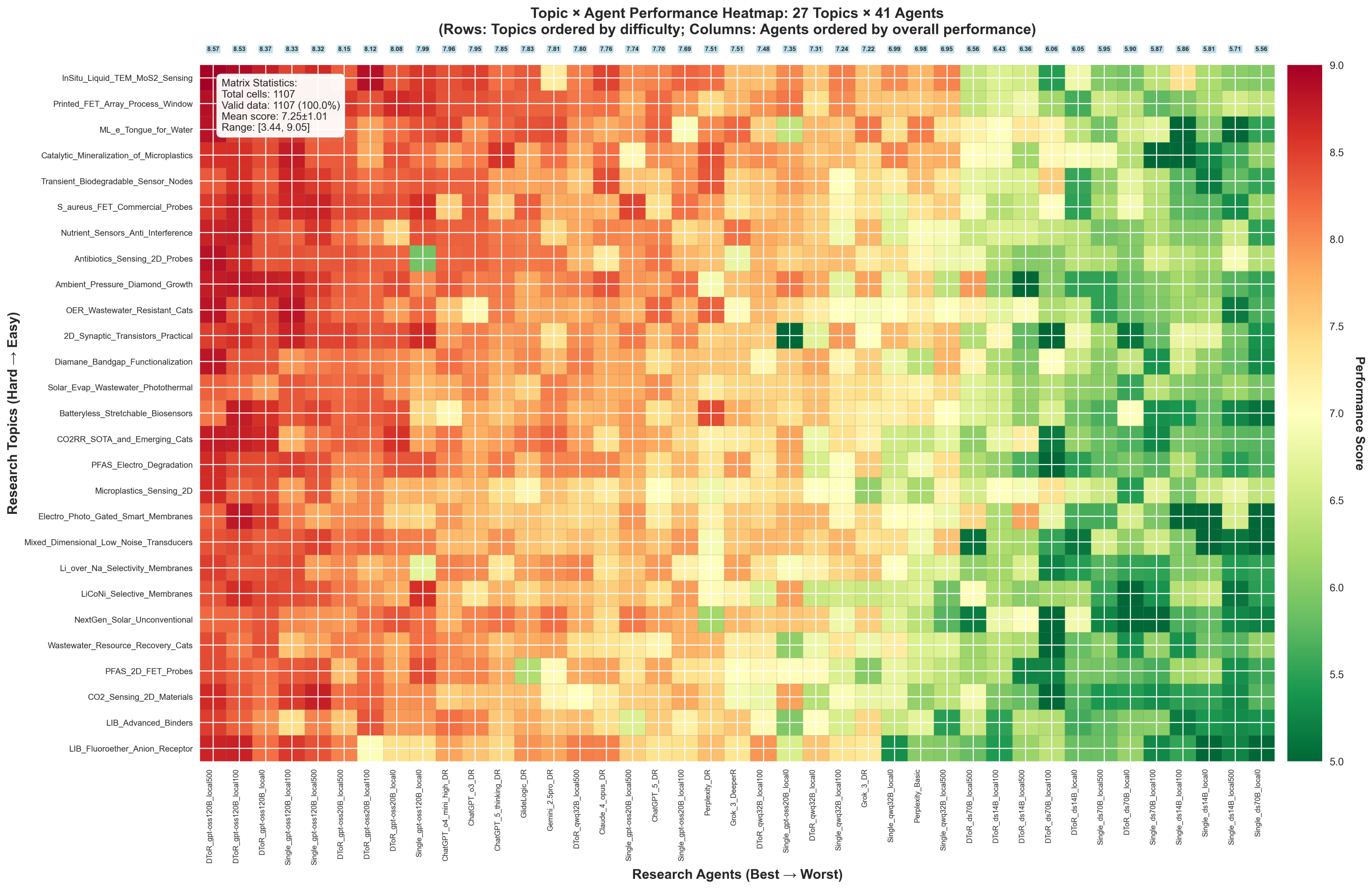}
  \caption{Performance matrix showing 27 research topics (rows, ordered by difficulty: easy\texorpdfstring{$\rightarrow$}{->}hard) and 41 agents (columns, ordered by overall performance: strong\texorpdfstring{$\rightarrow$}{->}weak). Because the colormap is reversed (\texttt{RdYlGn\_r}), deeper reds indicate higher scores while greens mark lower scores. The layout reveals topic-specific challenges and agent-specific strengths directly on the matrix.}  
  \label{fig:topic_anlaysis_heatmap} 
\end{figure}

\begin{figure}[htbp]
  \centering
  \includegraphics[width=\textwidth]{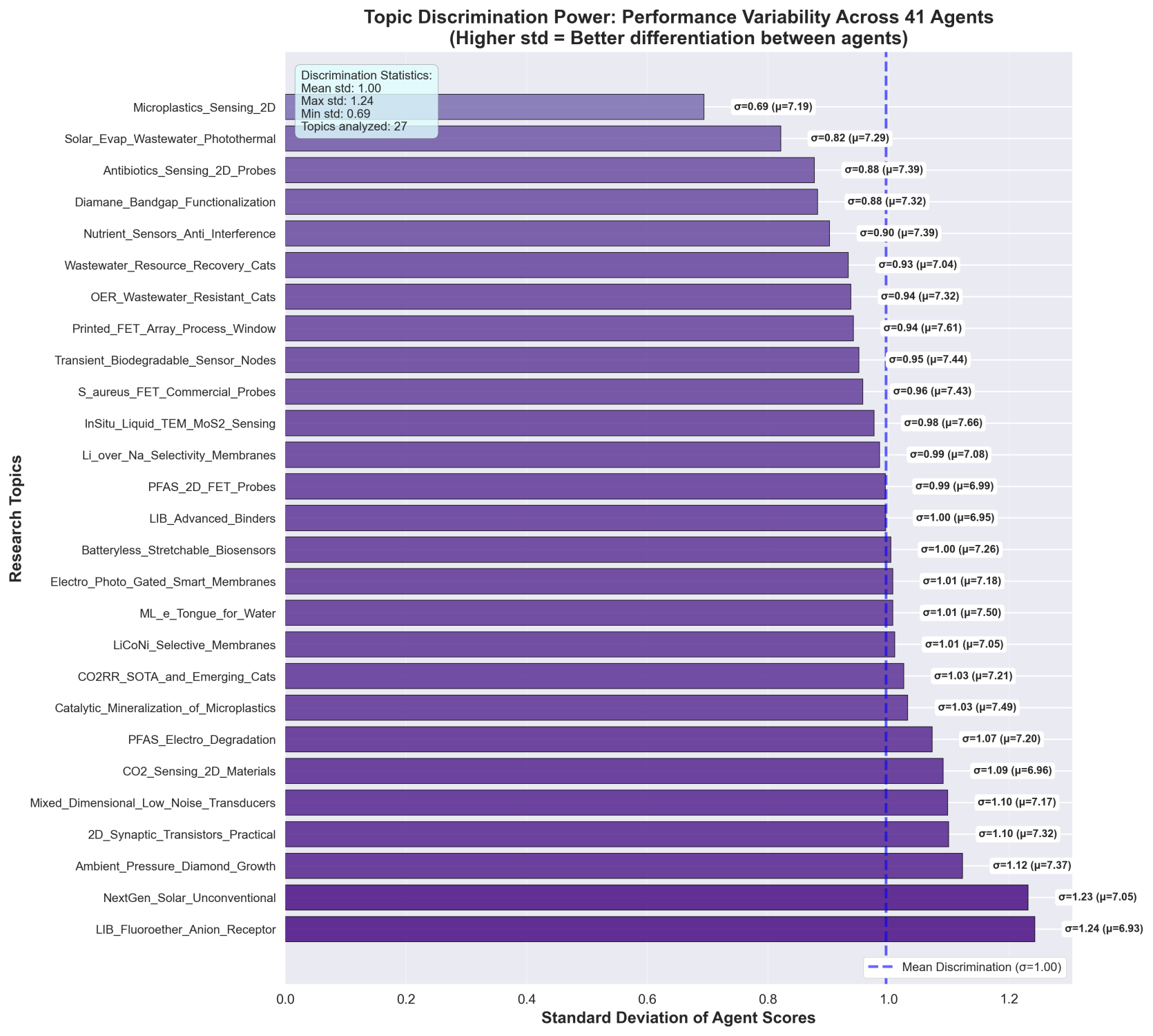}
  \caption{Standard deviation of agent scores for each topic, measuring how effectively each topic differentiates between strong and weak agents. Higher values indicate topics that create larger performance gaps across the agent spectrum, making them more suitable for benchmark evaluation.}  
  \label{fig:topic_anlaysis_discrimination} 
\end{figure}

\begin{figure}[htbp]
  \centering
  \includegraphics[width=\textwidth]{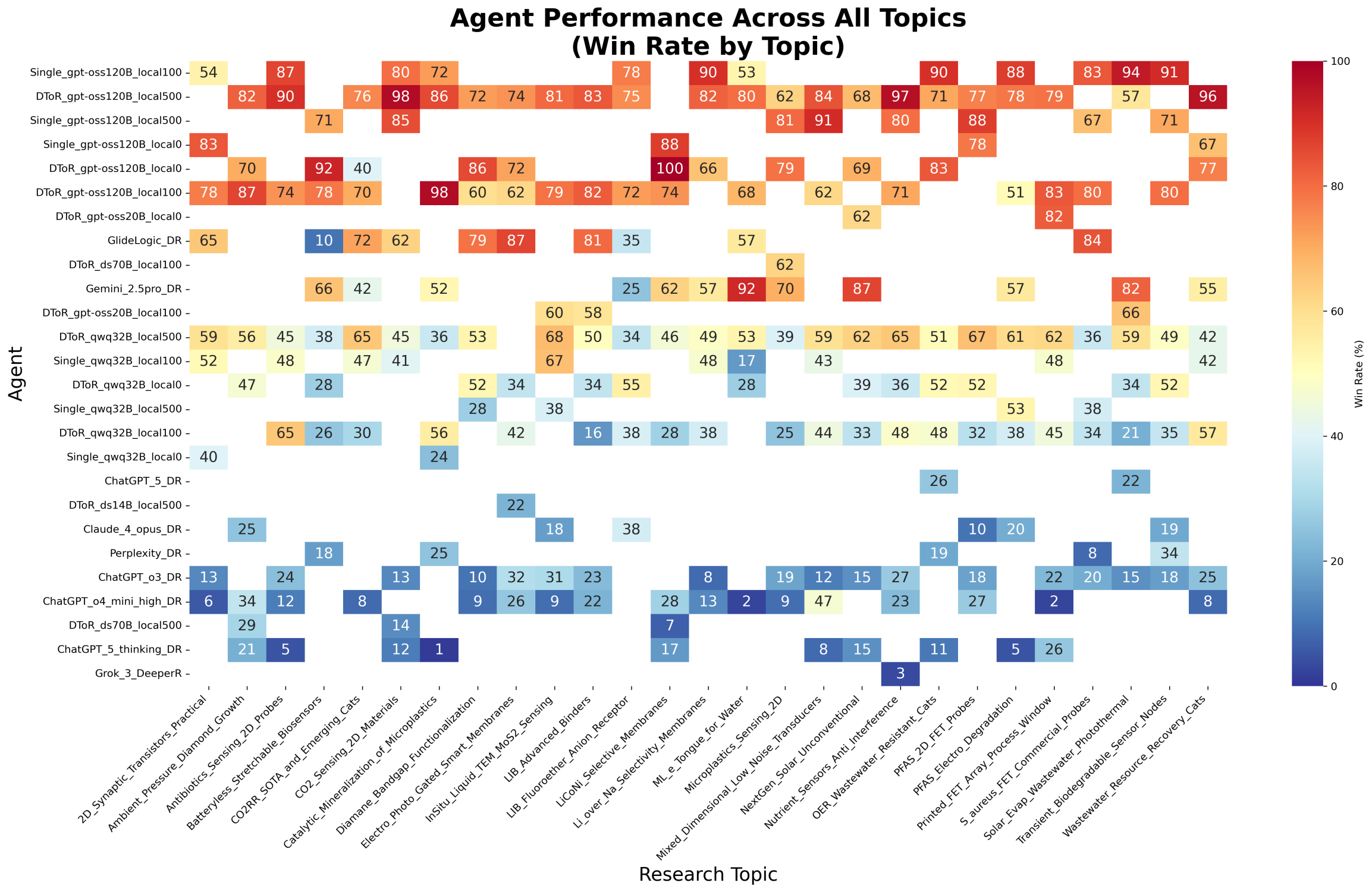}
  \caption{Win-rate heatmap across 27 research topics (columns) for every evaluated agent (rows); DToR variants dominate the warm/red cells, while weaker single and commercial baselines cluster in the cool/blue range, revealing topic-specific strengths and gaps}  
  \label{fig:topic_anlaysis_duel_heatmap_comparison} 
\end{figure}

\textbf{Result observations:} The analysis reveals a clear difficulty spectrum (Fig. \ref{fig:topic_ranking_bar}): the hard end is anchored by LIB\_Fluoroether\_Anion\_Receptor ($\mu \approx 6.9$), LIB\_Advanced\_Binders ($\mu \approx 6.9$), and CO2\_Sensing\_2D\_Materials ($\mu \approx 7.0$), while the easy end features InSitu\_Liquid\_TEM\_MoS2\_Sensing ($\mu \approx 7.7$), Printed\_FET\_Array\_Process\_Window ($\mu \approx 7.6$), and ML\_e\_Tongue\_for\_Water ($\mu \approx 7.5$); this ranking aligns with the row ordering in the heatmap (Fig. \ref{fig:topic_anlaysis_heatmap}). Importantly, discrimination power (Fig. \ref{fig:topic_anlaysis_discrimination}) is decoupled from difficulty: LIB\_Fluoroether\_Anion\_Receptor exhibits the largest spread ($\sigma \approx 1.24$), sharply separating agents despite its high difficulty, whereas Microplastics\_Sensing\_2D has low $\sigma$ ($\approx 0.69$) with mid-pack difficulty, indicating limited benchmarking sensitivity; easy topics such as InSitu\_Liquid\_TEM\_MoS2\_Sensing still show moderate $\sigma$, underscoring the independence. For benchmark design, topics with mid-range difficulty ($\mu \approx 7.2$--7.4) and elevated $\sigma$ give the best sensitivity---e.g., PFAS\_Electro\_Degradation ($\sigma \approx 1.07$), Ambient\_Pressure\_Diamond\_Growth ($\sigma \approx 1.12$), and Mixed\_Dimensional\_Low\_Noise\_Transducers ($\sigma \approx 1.10$)---while low-$\sigma$ topics (e.g., Microplastics\_Sensing\_2D, Solar\_Evap\_Wastewater\_Photothermal) add little discriminatory signal. Agent capability patterns follow the column ordering: stronger agents (left) show consistently higher bands, DToR configurations dominate the high-score region yet still dip on the hardest LIB chemistry topics, and commercial baselines (ChatGPT\_o3, ChatGPT\_o4\_mini\_high) sit mid-pack---no agent family is universally strong. However, the advantages of local DToR agents over commercial solutions are still obvious based on the statistics by in-pair dueling (Fig. \ref{fig:topic_anlaysis_duel_heatmap_comparison}). Domain-specific challenges cluster in battery/electrolyte design, pointing to persistent electrochemistry gaps; sensing/detection topics occupy the middle with modest discrimination (shared failure modes), and materials processing topics like Ambient\_Pressure\_Diamond\_Growth and NextGen\_Solar\_Unconventional couple mid difficulty with high $\sigma$, suggesting nuanced trade-offs that expose architectural differences. Collectively, we recommend selecting topics with $\sigma > 1.0$ and mean 6.5--7.5 for future benchmarks; we interpret low-$\sigma$ topics as signaling systematic weaknesses shared across approaches, and the difficulty--discrimination decoupling as evidence that agents face both ``universally hard'' challenges and ``high-variance'' generalization tests.

Taken together, the topic analysis indicates that our evaluation is both reliable and discriminative. The difficulty spectrum aligns with domain intuition (electrochemistry hardest; sensing mid-band), while discrimination ($\sigma$) is decoupled from difficulty and isolates topics that genuinely separate systems rather than amplifying noise. The resulting agent orderings are stable across topics
and mirror our minimal-sensitivity ablations (LLM\_only weakest, weak\_ToT degraded, local\_only below gap-triggered fusion), suggesting that the benchmark captures real capability differences instead of prompt or dataset artifacts. Practically, this lens identifies where progress is most needed and confirms that our orchestration-centric design delivers benefits that persist across model scales
and RAG choices. We therefore regard the evaluation as a fair, actionable basis for future comparisons.

\clearpage
\section{Power Consumption Estimation}\label{app:power_estimation}
\subsection{Power estimation and LLM/RAG time decomposition.}
We convert runtime to energy using a calibrated two–level power model on our
workstation (\textbf{4$\times$NVIDIA RTX A5500}, Ubuntu).
System power was monitored continuously during experiments using \texttt{nvidia-smi}
for GPU power draw and \texttt{powertop} for system–level consumption.
We acknowledge that our power measurement methodology is relatively basic;
however, the core contribution of this work lies in demonstrating the feasibility of
edge–device deployment, which already showcases significant democratization and
decentralization advantages over commercial subscription–based solutions.
The quantitative power estimates provided here serve primarily as reference points
for relative comparisons rather than definitive consumption benchmarks. 
Across runs we \emph{observed} a typical baseline draw during retrieval/IO of
\(P_{\text{base}}\approx 76\pm 3~\mathrm{W}\) (CPU, memory, disks, and idle GPUs).
When the LLM generates tokens the GPU adds a model–dependent draw reported by \texttt{nvidia-smi}:
\begin{align*}
&\Delta P_{\text{GPU}}(\text{ds14B})=282\pm 4~\mathrm{W},\quad
\Delta P_{\text{GPU}}(\text{ds70B})=310\pm 5~\mathrm{W}, \\
&\Delta P_{\text{GPU}}(\text{qwq32B})=301\pm 4~\mathrm{W}, 
\Delta P_{\text{GPU}}(\text{gpt-oss20B})=320\pm 5~\mathrm{W},\\
&\Delta P_{\text{GPU}}(\text{gpt-oss120B})=360\pm 6~\mathrm{W}.
\end{align*}
Hence \(P_{\text{LLM}}(\text{model}) = P_{\text{base}}+\Delta P_{\text{GPU}}(\text{model})\)
is \(358~\mathrm{W}\) (ds14B), \(386~\mathrm{W}\) (ds70B), \(377~\mathrm{W}\) (qwq32B), \(396~\mathrm{W}\) (gpt-oss20B), and \(436~\mathrm{W}\) (gpt-oss120B). 
During RAG (vector search over the local store, parsing, and file IO) the GPU utilization remained near–idle, so we bill RAG at \(P_{\text{RAG}}=P_{\text{base}}\approx 76~\mathrm{W}\).\footnote{Short web fetches last seconds relative to multi–hour runs and are neglected.}

\medskip\noindent
\subsection{Empirical split of runtime into LLM vs.\ RAG.}
Let \(\bar t_{\text{tot}}(s,m,k)\) be the mean wall–clock time for strategy \(s\in\{\text{Single},\text{DToR}\}\),
model \(m\in\{\text{ds14B},\text{ds70B},\text{qwq32B},\text{gpt-oss20B},\text{gpt-oss120B}\}\), and local–RAG size \(k\in\{0,100,500\}\).
For \(k=0\) (no local vector store) we roughly estimated that all time is attributable to LLM generation:
\[
t_{\text{LLM}}(s,m,0)=\bar t_{\text{tot}}(s,m,0),\qquad t_{\text{RAG}}(s,m,0)=0.
\]
For \(k>0\) we roughly split \(\bar t_{\text{tot}}(s,m,k)\) into LLM and RAG by scaling the \(k=0\) LLM time to reflect
the additional \emph{LLM} passes used in local–knowledge summarization and complementary–query generation.
The scale factors were estimated from repeated single–node micro–benchmarks
(profiling isolated summarization and query–generation calls with coarse-grained timing)
and calibrated against observed aggregate runtime patterns across full runs.
Given measurement noise and workload variability, we report representative values that
\emph{differed by both strategy and model size}:

{\small
\begin{align*}
\text{Single:}\quad &\alpha^{\text{Single}}_{\text{ds14B}}\approx 1.05,\quad
\alpha^{\text{Single}}_{\text{ds70B}}\approx 1.07,\quad
\alpha^{\text{Single}}_{\text{qwq32B}}\approx 1.09,\quad
\alpha^{\text{Single}}_{\text{gpt-oss20B}}\approx 1.11,\quad
\alpha^{\text{Single}}_{\text{gpt-oss120B}}\approx 1.13; \\
\text{DToR:}\quad &\alpha^{\text{DToR}}_{\text{ds14B}}\approx 1.10,\quad
\alpha^{\text{DToR}}_{\text{ds70B}}\approx 1.20,\quad
\alpha^{\text{DToR}}_{\text{qwq32B}}\approx 1.35,\quad
\alpha^{\text{DToR}}_{\text{gpt-oss20B}}\approx 1.25,\quad
\alpha^{\text{DToR}}_{\text{gpt-oss120B}}\approx 1.35.
\end{align*}
}

\noindent
We then compute
{\small
\[
t_{\text{LLM}}(s,m,k)=\alpha^{s}_{m}\,\bar t_{\text{tot}}(s,m,0),\qquad
t_{\text{RAG}}(s,m,k)=\bar t_{\text{tot}}(s,m,k)-t_{\text{LLM}}(s,m,k)\ \ (\text{clipped at }\ge 0).
\]
}
Two empirical patterns are worth noting.
(i) In \textbf{Single} mode the increase in \(t_{\text{LLM}}\) from \(k=0\) to \(k>0\) is modest
(ranging from \(+5\%\) to \(+13\%\) across models, with larger models showing slightly higher overheads),
and \(t_{\text{RAG}}\) exhibits a smaller difference than theoretically expected between \(k=100\) and \(k=500\) within measurement tolerance.
(ii) In \textbf{DToR} mode, multi–hour runs accumulate more local–knowledge summarization
and complementary–query steps; correspondingly \(t_{\text{LLM}}\) shows larger scaling
(ranging from \(1.10\times\) to \(1.35\times\) as above, with overhead correlating with model complexity),
and \(t_{\text{RAG}}\) increases with \(k\) due to heavier vector–store traffic and post–processing over longer context.
When attribution is ambiguous we conservatively charge time to the LLM bucket.
We emphasize that this decomposition serves as a \emph{coarse approximation} for comparative analysis
rather than exact per-component accounting.

\medskip\noindent
\subsection{Energy calculation.}
Per–report energy is the sum of LLM and RAG contributions:
\begin{align*}
E_{\text{LLM}}(s,m,k)&=\frac{P_{\text{LLM}}(m)\,t_{\text{LLM}}(s,m,k)}{1000}\ \ (\mathrm{kWh}), \\
E_{\text{RAG}}(s,m,k)&=\frac{P_{\text{RAG}}\,t_{\text{RAG}}(s,m,k)}{1000}\ \ (\mathrm{kWh}), \\
E_{\text{tot}}(s,m,k)&=E_{\text{LLM}}(s,m,k)+E_{\text{RAG}}(s,m,k).
\end{align*}
All results in Fig.~\ref{fig:main_energy_cost} use
\(P_{\text{base}}\approx 76~\mathrm{W}\),
\(\Delta P_{\text{GPU}}\approx\{282,310,301,320,360\}~\mathrm{W}\) (ds14B, ds70B, qwq32B, gpt-oss20B, gpt-oss120B),
and the strategy– and model–specific \(\alpha\) values above
(measurement uncertainties for power values range \(\pm 3\)\,--\,\(\pm 6~\mathrm{W}\)).
For example, substituting the estimated \(\{t_{\text{LLM}},t_{\text{RAG}}\}\) of DToR\_gpt-oss120B\_local500
yields \(E_{\text{tot}}\approx 4.37~\mathrm{kWh}\) per completed report\footnote{For context, 4.37 kWh/report is comparable to running a household refrigerator for 1.5 days or training a small CNN on ImageNet once—orders of magnitude below the carbon footprint of cloud-based DR APIs over repeated use.};
configuration–wise LLM/RAG splits are provided alongside runtime heatmaps in the Appendix. 

\end{appendices}

\end{document}